\documentclass{article}

\usepackage{graphicx}
\usepackage{booktabs} 
\usepackage{scalerel}
\usepackage{float}

\usepackage{subcaption}
\usepackage{caption}

\usepackage[toc,page]{appendix}

\usepackage{url}            
\usepackage{amsfonts}       
\usepackage{nicefrac}       
\usepackage{microtype}      
\usepackage{cancel}

\usepackage[dvipsnames]{xcolor}
\usepackage{amsmath, latexsym}
\usepackage[most]{tcolorbox}
\usepackage{placeins}
\usepackage{enumitem}
\usepackage{bold-extra}
\usepackage[T1]{fontenc}
\usepackage{standalone}
\usepackage{tikz}
\usepackage{tikzscale}
\usepackage{pgfplots}
\usepackage{pgf}
\usetikzlibrary{calc}
\usetikzlibrary{positioning}
\usetikzlibrary{angles,quotes}
\usetikzlibrary{backgrounds}
\usetikzlibrary{external}
\usetikzlibrary{fit}
\usetikzlibrary{arrows}
\usetikzlibrary{arrows.meta}
\usetikzlibrary{shapes.symbols}
\usetikzlibrary{shadings}
\usetikzlibrary{shapes}
\usetikzlibrary{fadings}
\usetikzlibrary{bayesnet}
\usetikzlibrary{matrix}
\usetikzlibrary{snakes}
\usetikzlibrary{decorations.pathmorphing, patterns}
\pgfplotsset{compat=1.15}

\newcommand{\Rect}[5]{
    \draw[#1] (#2,#3) rectangle(#2+#4,#3-#5);
}

\definecolor{C1}{HTML}{1F77B4}
\definecolor{C2}{HTML}{FF7F0E}
\definecolor{C3}{HTML}{2CA02C}
\definecolor{C4}{HTML}{D62728}
\definecolor{C5}{HTML}{9467BD}
\colorlet{C1light}{C1!70!white}
\colorlet{C2light}{C2!70!white}
\colorlet{C3light}{C3!70!white}
\colorlet{C4light}{C4!70!white}
\colorlet{C5light}{C5!70!white}
\colorlet{C1vlight}{C1!20!white}
\colorlet{C2vlight}{C2!20!white}
\colorlet{C3vlight}{C3!20!white}
\colorlet{C4vlight}{C4!20!white}
\colorlet{C5vlight}{C5!20!white}
\colorlet{linkcolor}{violet}
\colorlet{citecolor}{RedOrange}  %
\colorlet{urlcolor}{Aquamarine}
\definecolor{ultrapink}{rgb}{1.0, 0.44, 1.0}
\RequirePackage{contour}
\RequirePackage{xspace}

\tikzset{
    /pgf/decoration/amplitude = 0.1em,
    /pgf/decoration/segment length = 0.5em}
\usepackage{cuted}
\usepackage{flushend}

\usepackage{hyperref}

\usepackage{nccmath}
\usepackage[accepted]{icml2021}

\usepackage{xargs}                      
\usepackage[colorinlistoftodos,prependcaption,textsize=tiny]{todonotes}
\newcommandx{\unsure}[2][1=]{\todo[linecolor=red,backgroundcolor=red!25,bordercolor=red,#1]{#2}}
\newcommandx{\change}[2][1=]{\todo[linecolor=blue,backgroundcolor=blue!25,bordercolor=blue,#1]{#2}}
\newcommandx{\info}[2][1=]{\todo[linecolor=OliveGreen,backgroundcolor=OliveGreen!25,bordercolor=OliveGreen,#1]{#2}}
\newcommandx{\improvement}[2][1=]{\todo[linecolor=Plum,backgroundcolor=Plum!25,bordercolor=Plum,#1]{#2}}
\newcommandx{\thiswillnotshow}[2][1=]{\todo[disable,#1]{#2}}

\usepackage{mdframed}
\newmdenv[
  topline=false,
  bottomline=false,
  rightline=false,
  skipabove=\topsep,
  ]{siderules}
  
\newcommand{\sgparam}{\tau}
\newcommand{\slparam}{\eta}
\newcommand{\saparam}{\lambda}

\newcommand{\gparam}{\boldsymbol{\tau}}
\newcommand{\lparam}{\boldsymbol{\eta}}
\newcommand{\aparam}{\boldsymbol{\lambda}}
\newcommand{\gspace}{\Omega_\tau}
\newcommand{\lspace}{\Omega_\eta}
\newcommand{\aspace}{\Omega_\lambda}
\newcommand{\mixCompA}{B}

\newcommand{\ngrad}{\mbox{$\hat{g}$}}
\newcommand{\vngrad}{\mbox{$\myvecsym{\ngrad}$}}

\newcommand{\nGrad}{\mbox{$\hat{G}$}}
\newcommand{\vnGrad}{\mbox{$\myvecsym{\nGrad}$}}
\newcommand{\vlat}{\vw}
\newcommand{\lat}{w}
\newcommand{\vLat}{\vW}
\newcommand{\Lat}{W}
\newcommand{\varpar}{\lambda}
\newcommand{\vvarpar}{\vlambda}

\newcommand{\mix}{z}

\newcommand{\vfim}{\vF}

\newcommand{\entropy}{\mathcal{H}}

\newcommand\cut[1]{}

\newcommand{\squishlist}{
   \begin{list}{$\bullet$}
    { \setlength{\itemsep}{0pt}      \setlength{\parsep}{3pt}
      \setlength{\topsep}{3pt}       \setlength{\partopsep}{0pt}
      \setlength{\leftmargin}{1.5em} \setlength{\labelwidth}{1em}
      \setlength{\labelsep}{0.5em} } }

\newcommand{\squishlisttwo}{
   \begin{list}{$\bullet$}
    { \setlength{\itemsep}{0pt}    \setlength{\parsep}{0pt}
      \setlength{\topsep}{0pt}     \setlength{\partopsep}{0pt}
      \setlength{\leftmargin}{2em} \setlength{\labelwidth}{1.5em}
      \setlength{\labelsep}{0.5em} } }

\newcommand{\squishend}{
    \end{list}  }

{}
\newtheorem{thm}{Theorem}{}
{}
{}
\newtheorem{lemma}{Lemma}{}
{}
{}

\newenvironment{myproof}{{\bf Proof}}{}

\newcommand{\half}{\mbox{$\frac{1}{2}$}}

\newcommand{\real}{\mbox{$\mathbb{R}$}}

\newcommand{\rnd}[1]{\left(#1\right)}
\newcommand{\sqr}[1]{\left[#1\right]}
\newcommand{\crl}[1]{\left\{#1\right\}}
\newcommand{\myang}[1]{\langle#1\rangle}
\newcommand{\myexpect}{\mathbb{E}}
\newcommand{\Unmyexpect}[1]{\mathbb{E}_{\scaleto{#1\mathstrut}{6pt}}}

\newcommand{\gauss}{\mbox{${\cal N}$}}
\newcommand{\mgauss}{\mbox{${\cal MN}$}}

\newcommand{\Student}{\mbox{${\cal T}$}}

\newcommand{\myvec}[1]{\mbox{$\mathbf{#1}$}}
\newcommand{\myvecsym}[1]{\mbox{$\boldsymbol{#1}$}}

\newcommand{\valpha}{\mbox{$\myvecsym{\alpha}$}}

\newcommand{\vdelta}{\mbox{$\myvecsym{\delta}$}}
\newcommand{\vDelta}{\mbox{$\myvecsym{\Delta}$}}
\newcommand{\vepsilon}{\mbox{$\myvecsym{\epsilon}$}}
\newcommand{\veta}{\mbox{$\myvecsym{\eta}$}}

\newcommand{\vmu}{\mbox{$\myvecsym{\mu}$}}
\newcommand{\vlambda}{\mbox{$\myvecsym{\lambda}$}}

\newcommand{\vphi}{\mbox{$\myvecsym{\phi}$}}

\newcommand{\vpsi}{\myvecsym{\psi}}

\newcommand{\vSigma}{\mbox{$\myvecsym{\Sigma}$}}
\newcommand{\vOmega}{\mbox{$\myvecsym{\Omega}$}}
\newcommand{\vtau}{\mbox{$\myvecsym{\tau}$}}

\newcommand{\va}{\mbox{$\myvec{a}$}}
\newcommand{\vb}{\mbox{$\myvec{b}$}}

\newcommand{\vd}{\mbox{$\myvec{d}$}}

\newcommand{\vf}{\mbox{$\myvec{f}$}}
\newcommand{\vg}{\mbox{$\myvec{g}$}}
\newcommand{\vh}{\mbox{$\myvec{h}$}}

\newcommand{\vm}{\mbox{$\myvec{m}$}}

\newcommand{\vs}{\mbox{$\myvec{s}$}}

\newcommand{\vu}{\mbox{$\myvec{u}$}}
\newcommand{\vv}{\mbox{$\myvec{v}$}}
\newcommand{\vw}{\mbox{$\myvec{w}$}}

\newcommand{\vx}{\mbox{$\myvec{x}$}}

\newcommand{\vz}{\mbox{$\myvec{z}$}}
\newcommand{\vA}{\mbox{$\myvec{A}$}}
\newcommand{\vB}{\mbox{$\myvec{B}$}}
\newcommand{\vC}{\mbox{$\myvec{C}$}}
\newcommand{\vD}{\mbox{$\myvec{D}$}}
\newcommand{\vE}{\mbox{$\myvec{E}$}}
\newcommand{\vF}{\mbox{$\myvec{F}$}}
\newcommand{\vG}{\mbox{$\myvec{G}$}}
\newcommand{\vH}{\mbox{$\myvec{H}$}}
\newcommand{\vI}{\mbox{$\myvec{I}$}}
\newcommand{\vJ}{\mbox{$\myvec{J}$}}
\newcommand{\vK}{\mbox{$\myvec{K}$}}
\newcommand{\vL}{\mbox{$\myvec{L}$}}
\newcommand{\vM}{\mbox{$\myvec{M}$}}
\newcommand{\vN}{\mbox{$\myvec{N}$}}

\newcommand{\vP}{\mbox{$\myvec{P}$}}
\newcommand{\vQ}{\mbox{$\myvec{Q}$}}

\newcommand{\vS}{\mbox{$\myvec{S}$}}
\newcommand{\vT}{\mbox{$\myvec{T}$}}
\newcommand{\vU}{\mbox{$\myvec{U}$}}
\newcommand{\vV}{\mbox{$\myvec{V}$}}
\newcommand{\vW}{\mbox{$\myvec{W}$}}
\newcommand{\vX}{\mbox{$\myvec{X}$}}

\newcommand{\vZ}{\mbox{$\myvec{Z}$}}

\newcommand{\be}{\begin{equation}}
\newcommand{\ee}{\end{equation}}
\newcommand{\bea}{\begin{eqnarray}}
\newcommand{\eea}{\end{eqnarray}}
\newcommand{\beaa}{\begin{eqnarray*}}
\newcommand{\eeaa}{\end{eqnarray*}}

\begin{document}

\newcommand{\ourtitle}{Tractable Structured Natural-Gradient Descent Using Local Parameterizations}
\icmltitlerunning{\ourtitle}

\twocolumn[

\icmltitle{\ourtitle}




\vspace{-0.35cm}
\begin{icmlauthorlist}
\icmlauthor{Wu Lin}{ubc}
\icmlauthor{Frank Nielsen}{sony}
\icmlauthor{Mohammad Emtiyaz Khan}{riken}
\icmlauthor{Mark Schmidt}{ubc,amii}
\vspace{-0.1cm}
\end{icmlauthorlist}

\icmlaffiliation{ubc}{University of British Columbia.}
\icmlaffiliation{riken}{RIKEN Center for Advanced Intelligence Project.}
\icmlaffiliation{sony}{Sony Computer Science Laboratories Inc.}
\icmlaffiliation{amii}{CIFAR AI Chair, Alberta Machine Intelligence Institute}

\icmlcorrespondingauthor{Wu Lin}{yorker.lin@gmail.com
\vspace{-0.2cm}
}

\icmlkeywords{Natural Gradient Descent, Information Geometry, Variational Inference, Optimization, Search, Deep Learning}

\vskip 0.3in
]



\printAffiliationsAndNotice{}  

\begin{abstract}
Natural-gradient descent (NGD) on structured parameter spaces (e.g., low-rank covariances) is computationally challenging due to difficult Fisher-matrix computations.
We address this issue by using \emph{local-parameter coordinates} to obtain a flexible and efficient NGD method that works well for a wide-variety of structured parameterizations. 
We show four applications where our method (1) generalizes the exponential natural evolutionary strategy, (2) recovers existing Newton-like algorithms, (3) yields new structured second-order algorithms via matrix groups, and (4) gives new algorithms to learn covariances of Gaussian and Wishart-based distributions. 
We show results on a range of problems from deep learning, variational inference, and evolution strategies. Our work opens a new direction for scalable structured geometric methods.



\end{abstract}

\section{Introduction}
A wide-variety of problems that arise in the field of optimization, inference, and search can be expressed as 

\vspace{-0.4cm}
\begin{align}
   \min_{q(\text{\vlat})\in\mathcal{Q}} \Unmyexpect{q(\text{\vlat})} \sqr{ \ell(\vlat) } - \gamma\entropy (q(\vlat)) \label{eq:problem},
\end{align}

\vspace{-0.4cm}
where $\vlat$ is the parameter of interest, $q(\vlat)\in\mathcal{Q}$ is a distribution, $\entropy(q(\vlat))$ is Shannon's entropy, $\ell(\vlat)$ is a loss function, and $\gamma\ge 0$.
For example, in problems involving random search \citep{baba1981convergence}, stochastic optimization \cite{spall2005introduction}, and evolutionary strategies \cite{beyer2001theory}, $q(\vlat)$ is the so-called `search' distribution 
used to find a global minimum of a black-box function $\ell(\vlat)$. In
reinforcement learning, it can be the policy distribution which minimizes the 
expected value-function $\ell(\vlat)$ \citep{sutton1998introduction}, sometimes with entropy regularization \citep{williams1991function, teboulle1992, mnih2016asynchronous}.  
For Bayesian problems, $q(\vlat)$ is the posterior distribution or its approximation and the $\ell(\vlat)$ is the log of the joint distribution \citep{zellner1986bayesian} ($\gamma$ set to 1). Finally, many robust or global optimization techniques employ $q(\vlat)$ to smooth out local minima \citep{mobahi2015theoretical, leordeanu2008smoothing, hazan2016graduated}, where often $\gamma=0$. Developing fast and scalable algorithms for solving \eqref{eq:problem} potentially impacts all these fields.

Natural-gradient descent (NGD) is an attractive algorithm to solve \eqref{eq:problem} and can speed up the optimization by exploiting the information geometry of $q(\vlat)$ \cite{wierstra2008natural,sun2009efficient, hoffman2013stochastic, khan2017conjugate, salimbeni2018natural}. It also unifies a wide-variety of learning algorithms, which can be seen as its instances with a specific $q(\vlat)$ \cite{emti2020bayesprinciple}. This includes deep learning \citep{khan18a},
approximate inference \citep{khan2017conjugate}, and optimization \citep{emti2020bayesprinciple, khan2017variational}.
NGD also has better convergence properties compared to methods that ignore the geometry, for example, \citet{ranganath2014black, lezcano2019trivializations}.
\begin{figure*}[t]
   \center
   \includegraphics[width=6.4in]{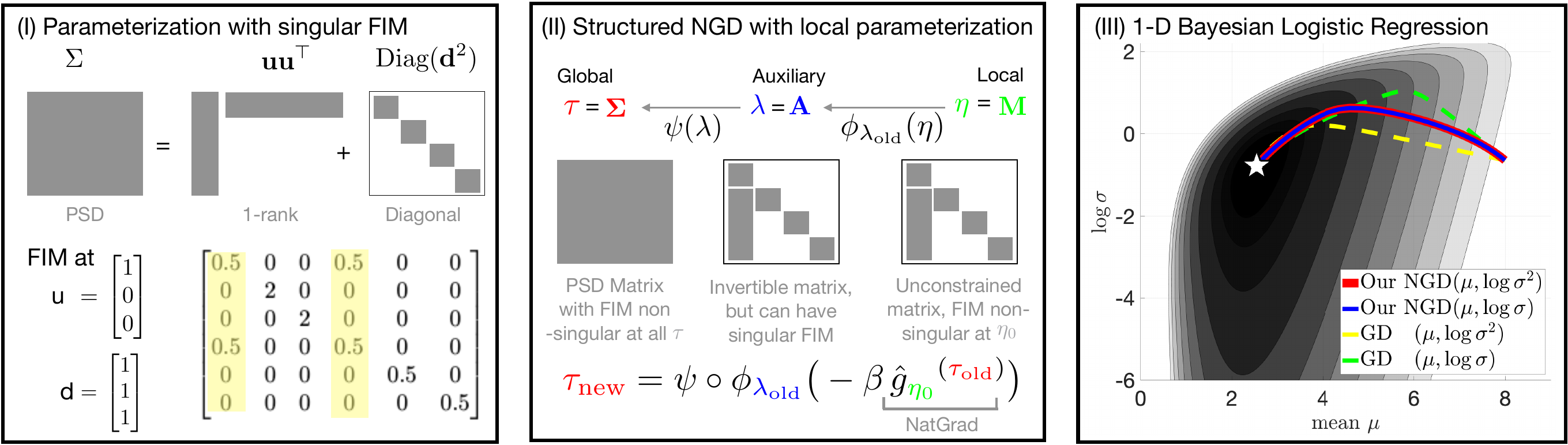}
   \caption{(I) The FIM can be singular, for example, when the covariance $\vSigma$ has a low rank structure (more details in Appx.~\ref{app:diag_zero}). The two identical columns of FIM are shown in yellow. (II) We fix such issues by using a local parameterization $\veta$ (here $\vM$, an unconstrained structured matrix) which is related to the global variable $\vtau$ ($=\vSigma$ for the low-rank example) through an auxiliary parameter $\vlambda$
   ($=\vA$, an invertible matrix with a specific structure to get a low-rank $\vSigma = \vA\vA^\top$).
   The three parameter-spaces are related through maps $\vtau = \vpsi(\vlambda) = \vA\vA^\top$ and $\vlambda = \vphi_{\text{\vlambda}_{\text{old}}}(\veta)= \vA = \vA_{\text{old}} \text{Exp}(\vM)$, and need to satisfy Assumptions 1 and 2~given in Section \ref{sec:local_params}.
   This results in a valid NGD step (shown at the bottom) in the local-parameter space (defined at $\veta_0 =0$ with learning rate $\beta)$. 
   (III) For a 1-D Bayesian logistic-regression, our NGD  is invariant to two different parameterizations, which is not the case for GD (details in Appx.~\ref{app:blr_gauss}).}
   \label{fig:fig1}
\end{figure*}

We consider NGD where parameters of $q(\vlat)$ assume special structures, for example, low-rank or sparse Gaussian covariances.
For such cases, NGD is often intractable and/or costly due to difficult Fisher Information Matrix (FIM) computations. First, the FIM can be singular for restricted parametrizations (see Fig. \ref{fig:fig1}(I)), which is often addressed with ad-hoc structural approximations, derived on a case-by-case basis
~\cite{sun2013linear,akimoto2016projection,li2017simple,mishkin2018slang,tran2020bayesian} (also see
Appx.~\ref{app:diff_st_gauss}).
Second, while we can switch parameterizations, the computation could be inefficient because the structure might be lost, for example, when switching from sparse precision to covariances. Using automatic differentiation could make the situation worse because such tools are often unaware of the structure \citep{salimbeni2018natural} (also see Appx.~\ref{app:indirect_limit}).
Finally, the choice of parameterizations and approximations themselves involve delicate choices to get a desired computation-accuracy trade-off. For example, for neural networks layer-wise approximations \cite{sun2017relative,zhang2018noisy,lin2019fast} might be better than low-rank/diagonal structures \cite{mishkin2018slang, tran2020bayesian,ros2008simple,khan18a}, but may also involve more computations.
Our goal is to address these difficulties and design a flexible method that works well for a variety of structured parameterizations.

We present \emph{local-parameter coordinates} to design flexible and tractable NGD for a variety of structured-parameter spaces.
The method is summarized in Fig. \ref{fig:fig1}(II), and involves specifying (i) a `local parameter coordinate' that satisfies the structural constraints of the original (global) parameters, (ii) a map to convert back to the global parameters via `auxiliary' parameters, and finally (iii) a tractable natural-gradient computation in the local-parameter space.
This construction ensures  a valid NGD update in local parameter spaces, while maintaining structures (often via matrix groups) in the auxiliary parameters. 
This decoupling enables a tractable NGD that exploits the structure, when these parameters and the map are chosen carefully.


%
We show four applications of our method.

\vspace{-0.3cm}
\begin{enumerate}
   \item We generalize \citet{glasmachers2010exponential}'s method to more general distributions and structures (Section \ref{sec:GaussSquareRoot}).
   \item In Section \ref{sec:gauss_newton_connect}, we recover Newton-like methods derived by \citet{lin2020handling} using Riemannian-gradients and by \citet{khan18a} using the standard NGD.
   \item Our approach is easily generalizable to other non-Gaussian cases; see Section \ref{sec:wishart} and \ref{sec:wishart_rgd_connect}.
   \item In Section \ref{sec:tri_group}, we derive new $2^{\text{nd}}$-order methods for low-rank, diagonal, and sparse covariances. The methods are only slightly more costly than diagonal-covariance methods. Moreover, they can be used as structured $2^{\text{nd}}$-order methods for unconstrained optimization.
\end{enumerate}

\vspace{-0.3cm}
We show applications to various problems for search, variational inference, and deep learning, obtaining much faster convergence than methods that ignore geometry. An example for 1-D logistic regression is shown in \ref{fig:fig1}(III).
Overall, our work opens a new direction to design efficient and structured geometric methods via local parameterizations.

\section{Structured NGD and its Challenges}

%
The distributions $q(\vlat)\in\mathcal{Q}$ are often parameterized, say using parameters $\gparam\in\gspace$, for which we write $q(\vlat|\gparam)$. The problem can then be conveniently expressed as an optimization problem in the space $\gspace$,

\vspace{-0.32cm}
\begin{equation}
    \gparam^* = \arg\min_{\gparam\in\gspace} \myexpect_{q(\text{\vlat}|\gparam)} \sqr{ \ell(\vlat)}, 
    \label{eq:problem_tau}
\end{equation}

\vspace{-0.36cm}
where we assume $\gamma =0$ for simplicity (general case is in Lemma \ref{lemma:eq4} of Appx.~\ref{app:FIM}).  
The NGD step is $\gparam_{t+1} \leftarrow \gparam_t  - \beta \vngrad_{\gparam_t}$ where $\beta>0$ is the step size and natural gradients are as

\vspace{-0.32cm}
\begin{equation}
    \vngrad_{\gparam_t} := \vF_{\gparam}(\gparam_t)^{-1} \nabla_{\gparam} \myexpect_{q(\text{\vlat}|\gparam)} \sqr{ \ell(\vlat)},
    \label{eq:ng_global}
\end{equation}

\vspace{-0.36cm}
where  
$\vF_{\gparam}(\gparam)\mspace{-6mu}:=\mathbb{E}_{q}[\nabla_\tau\log q(\vlat|\gparam)(\nabla_\tau^\top\log q(\vlat|\gparam))] $ is an invertible and well-defined FIM following the regularity condition (see Appx.~\ref{app:FIM}).
The iterates $\gparam_{t+1}$ may not always lie inside $\gspace$ and a projection step might be required.

In some cases, the NGD computation may not require an explicit FIM inversion. For example, when $q(\vlat|\gparam)$ is a minimal exponential-family (EF) distribution, FIM is always invertible, and natural gradients are equal to vanilla gradients with respect to the `expectation parameter'~\citep{malago2011towards, khan2018fast}.
By appropriately choosing $\mathcal{Q}$, the NGD then takes forms adapted by popular algorithms \citep{emti2020bayesprinciple}, for example, for Gaussians $q(\vlat|\gparam)=\gauss(\vlat|\vmu,\vS^{-1})$ where $\vS$ denotes the precision, it reduces to a Newton-like update \citep{khan18a},

\vspace{-0.32cm}
\begin{align}
\vmu_{t+1} & \leftarrow \vmu_t - \beta \vS_{t+1}^{-1} \myexpect_{q(\text{\vlat}|\gparam_t)}{ \sqr{ \nabla_\lat \ell( \vlat) }}, \nonumber \\ 
\vS_{t+1} & \leftarrow \vS_t + \beta  \Unmyexpect{q(\text{\vlat}|\gparam_t)}{ \sqr{ \nabla_\lat^2 \ell(\vlat) } }. \label{eq:newton_update}
\end{align}

\vspace{-0.36cm}
The standard Newton update for optimization is recovered by approximating the expectation at the mean and using a step-size of 1 with $\gamma =1$~\citep{emti2020bayesprinciple}. Several connections and extensions have been derived in the recent years establishing NGD as an important algorithm for optimization, search, and inference \citep{khan2017conjugate, khan2018fast, lin2019fast, osawa2019large}. 

This simplification of NGD breaks down when \eqref{eq:problem_tau} involves structured-parameter spaces $\gspace$, for example, spaces with constrains such as low-rank or sparse structures. Even for the simplest Gaussian case, where covariances lie in the positive-definite space, the update \eqref{eq:newton_update} may violate the constraint~\citep{khan18a}. Extensions have been derived using Riemannian gradient descent (RGD) to fix this issue \cite{lin2020handling}. Other solutions based on
Cholesky \citep{sun2009efficient,salimbeni2018natural} or square-root parameterization \citep{glasmachers2010exponential} have also been considered, where the problem is converted to another parameter space.
For example, \citet{glasmachers2010exponential} use a square-root parameterization  $q(\vlat)=\gauss(\vlat|\vmu,\vA\vA^T)$, where $\vA$ is the square-root of $\vS^{-1}$, to get the update,
%

\vspace{-0.36cm}
\begin{align} 
 \vmu_{t+1} &\leftarrow \vmu_t - \beta   \Unmyexpect{q(\text{\vlat}|\gparam_t)}{ \sqr{\big(\vA_t \vz_t \big) \ell( \vlat) }}, \nonumber \\
 \vA_{t+1} &\leftarrow \vA_t \mathrm{Exp} \rnd{ -\frac{\beta}{2}   \Unmyexpect{q(\text{\vlat}|\gparam_t)}{ \sqr{ \big( \vz_t \vz_t^T -\vI \big) \ell( \vlat) }}   }, \label{eq:xnes}
\end{align}

\vspace{-0.45cm}
where $\vz_t=\vA_t^{-1} (\vlat-\vmu_t)$ and $\mathrm{Exp}(\vX)=\vI + \sum_{k=1}^{\infty}\frac{\vX^k}{k!}$ is the matrix exponential function. 
These solutions however do not easily generalize. For example, it is not obvious how to apply these updates to cases where the covariance is 
low-rank~\citep{mishkin2018slang, tran2020bayesian}, Kronecker structured~\citep{zhang2018noisy, lin2019fast}, or to cases involving non-Gaussian distributions such as the Wishart, univariate exponential family distributions~\citep{lin2020handling} and Gaussian mixtures~\citep{lin2019fast}.

In fact, the issue with the structure and its effect on parameterization is a bit more involved than it might appear at first.
Certain choices of the structure/parameterization can make the Fisher matrix singular which can make NGD invalid, for example, for low-rank Gaussians as shown in Fig. \ref{fig:fig1}(I) where it requires new tricks such as auxiliary parameterization \citep{lin2019fast}, block approximations \citep{tran2020bayesian}, algorithmic approximations \citep{mishkin2018slang}, or damping \citep{zhang2018noisy}.
%
The computational cost depends on the parameterization, the choice of which is often not obvious. Some methods exploit structure in the covariances~\citep{glasmachers2010exponential} while the others work with its inverse such as~\eqref{eq:newton_update}. 
Customized structures, such as layer-wise and Kronecker-factored covariances in deep neural nets, may work well in one parameterization but not in the other.  
Thus, it is essential to have a flexible method that works well for a variety of structured-parameterizations and distributions. Our goal is to propose such a method.

\vspace{-0.2cm}
\section{Local Parameter Coordinates}
\label{sec:local_params}

\begin{figure}
   \center
     \vspace{-0.27cm}   
   \includegraphics[width=0.95\linewidth]{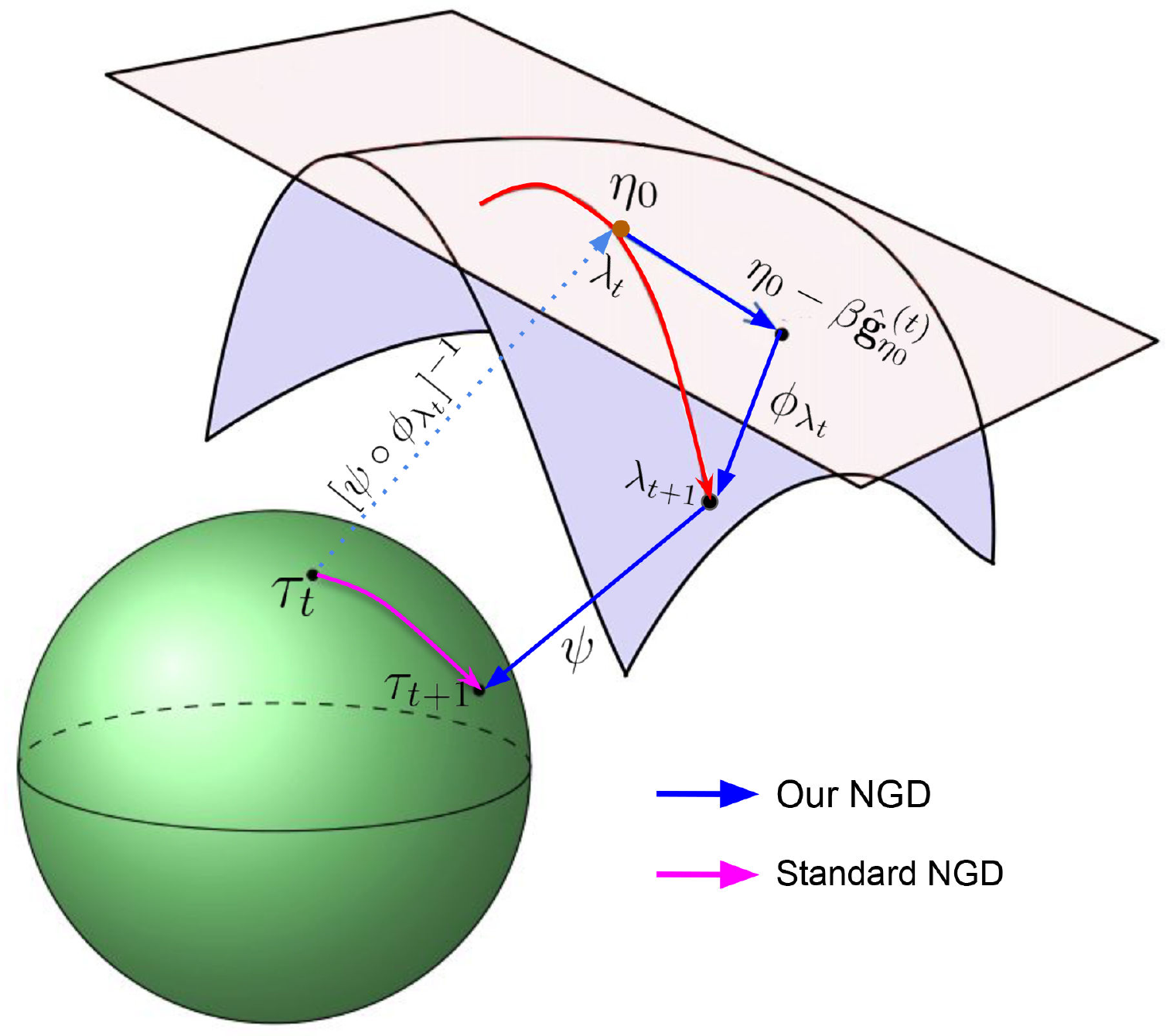}
     \vspace{-0.26cm}   
   \caption{
  The global, auxiliary, and local parameter spaces are highlighted in green, blue, and pink, respectively. 
 We perform NGD in the local space and preserve the structure in the auxiliary space.
  The proposed update~\eqref{eq:ngd_local_param} is denoted by blue solid lines.
  The standard NGD update is denoted by a purple line.
  In the local  space, the  brown dot is represented by $\lparam_0$. In the auxiliary  space, the brown  dot is represented by $\aparam_t=\vphi_{\aparam_t}(\lparam_0)$. Even when a standard NGD update (``purple NGD'') is intractable,
  our approach could still gives an efficient and tractable update (``blue NGD'').
   }
   \label{fig:fig2}
\vspace{-0.22cm}
\end{figure}

We present local-parameter coordinates to obtain a flexible and efficient NGD method that works well for a wide-variety of structured parameterizations. Table \ref{tab:updates} in Appx.~\ref{app:summary} summarizes the examples and extensions we consider. We describe the method in three steps (see Fig.~\ref{fig:fig2} for an illustration).

\textbf{Step 1.} The first step involves specifying a `local' parameterization, denoted by $\lparam\in \lspace$, so that the following assumption is satisfied (throughout, we set $\lparam_0 =\mathbf{0}$).

{\bf Assumption 1:} \emph{The Fisher matrix $\vF_{\lparam}(\lparam_0)$ is non-singular.}


\textbf{Step 2.} The second step involves specifying two maps shown below to connect to the original `global' parameters $\gparam$ via an `auxiliary' parameter $\aparam\in\aspace$,
%
%

\vspace{-0.35cm}
\begin{align}
\begin{split}
   \gparam= \vpsi(\aparam)\, \text{ and }\, \aparam = \vphi_{\aparam_t} (\lparam),
 \end{split}
    \label{eq:aux_update}
\end{align}

\vspace{-0.35cm}
where the first map is surjective and the second map is defined such that $\aparam_t = \vphi_{\aparam_t} (\lparam_0)$, i.e., the function is tight at $\lparam_0$ to match the current $\aparam_t$.
The local parameter $\lparam_0$ can be seen as a \emph{relative origin} tied to $\aparam_t$.
The overall map is $\gparam = \vpsi \circ \vphi_{\aparam_t}(\lparam)$ (the map could change with iterations).
Notice that we make no assumption about the non-singularity of the FIM in the auxiliary space $\aspace$. 
The FIM in the auxiliary space $\aspace$ can be singular (see Sec.~\ref{sec:GaussSquareRoot}).
The only restriction is a mild \emph{coordinate compatibility} assumption.

{\bf Assumption 2
:} \emph{
$\forall \aparam_t \in\aspace$, the map $\lparam \mapsto \vpsi \circ \vphi_{\aparam_t} (\lparam)$ is locally $C^1$-diffeomorphic  at an open neighborhood of $\lparam_0$. 
}

Assumption 2 implies that the local {$\lparam$} has the same degrees of freedom as $\gparam$, but the auxiliary {$\aparam$} can have a different one (an example is in Sec.~\ref{sec:GaussSquareRoot}).
Assumption 1-2, together with surjective $\vpsi(\cdot)$, imply a non-singular FIM  in the global space  $\gspace$, so there is no need to check the FIM $\vF_{\gparam}(\gparam)$ for specific cases.
On the other hand, if we know  the non-singularity of $\vF_{\gparam}(\gparam)$ beforehand, 
Assumption 2 together with surjective $\vpsi(\cdot)$ imply that Assumption 1 is satisfied.

\textbf{Step 3.} The final step is to compute the natural gradient at $\lparam_0$ in the local-parameter space to update the global $\gparam$, which can be done by using the chain rule,
%

\vspace{-0.32cm}
\begin{align}
        \vngrad_{\lparam_0}^{(t)} &= \vF_{\lparam}(\lparam_0)^{-1} \,\, \nabla_{\lparam_0} \sqr{ \vpsi \circ \vphi_{\aparam_t} (\lparam) } \,\, \vg_{\gparam_t}, \label{eq:ngrad_local} 
\end{align}

\vspace{-0.32cm}
where $\vg_{\gparam} := \nabla_{\gparam} \myexpect_{q(\text{\lat}|\gparam)}[ \ell(\vlat)]$ is the vanilla gradient.
An indirect computation is given in \eqref{eq:uni_ef_nd} in Appx.~\ref{app:FIM}.
The above computation is most useful when the computation of $\vngrad_{\lparam_0}^{(t)}$ is tractable, which ultimately depends on the choice of $\vpsi \circ \vphi_{\aparam_t}$ which in turn depends on the form of $q(\vlat)$.
Then, by using an NGD step $\lparam_0 - \beta \vngrad_{\lparam_0}^{(t)}$ in the local-parameter space, we get the following overall update for $\gparam$,

\vspace{0.15cm}
\begin{tcolorbox}[enhanced,colback=white,%
    colframe=red!75!black, attach boxed title to top right={yshift=-\tcboxedtitleheight/2, xshift=-1.25cm}, title= structured NGD using local parameters, coltitle=red!75!black, boxed title style={size=small,colback=white,opacityback=1, opacityframe=0}, size=title, enlarge top initially by=-\tcboxedtitleheight/2]
\begin{equation}
\aparam_{t+1} \leftarrow  \vphi_{\aparam_t} \rnd{ -\beta \vngrad_{\lparam_0}^{(t)} },\,\,\,\,\,
    \gparam_{t+1} \leftarrow \vpsi\rnd{\aparam_{t+1}} 
    \label{eq:ngd_local_param}
\end{equation}
\end{tcolorbox}

\vspace{-0.15cm}
since we assume $\lparam_0 = \mathbf{0}$.  
In summary, given an auxiliary parameter $\aparam_t$, we can use the natural gradient $\vngrad_{\lparam_0}^{(t)}$  to update $\gparam$ according to \eqref{eq:ngd_local_param}.
The NGD step using \eqref{eq:ng_global} is a special case of the above NGD step (see details in Appx.~\ref{app:general}).

Finally, we require the following Assumption to be satisfied to ensure that the NGD step $\lparam_0-\beta \vngrad_{\lparam_0}^{(t)} \in \lspace$ in  \eqref{eq:ngd_local_param} (this assumption is satisfied for all examples we discuss).

{\bf Assumption 3
:} \emph{
$\lspace$ has a vector-space structure so that the vector addition and the real-scalar product  are valid.
}

We will now discuss three applications of our method where we derive existing NGD strategies as special cases.

\subsection{Gaussian with square-root covariance structure}
\label{sec:GaussSquareRoot}
For a Gaussian case $\gauss(\vlat|\vmu,\vSigma)$, the covariance $\vSigma$ is positive definite. Standard NGD such as \eqref{eq:newton_update}, may violate the constraint~\cite{khan18a}. \citet{glasmachers2010exponential} use $\vSigma = \vA\vA^\top\mspace{-8mu}$ where $\vA$ is an invertible matrix (not  a Cholesky), and derive an update using a specific local parameterization.
We now show that their update is a special case of ours.

Following \citet{glasmachers2010exponential}, we use  the following parameterizations, where 
$\mathcal{S}_{++}^{p\times p}$, $\mathcal{S}^{p\times p}$, and $\mathrm{GL}^{p\times p}$  denote
the set of symmetric positive definite matrices, symmetric matrices, and invertible  matrices, respectively,

\vspace{-0.35cm}
\begin{equation}
    \begin{split}
        \gparam &:= \crl{\vmu \in\real^p, \,\,\, \vSigma \in \mathcal{S}_{++}^{p\times p} }, \\ 
        \aparam &:= \crl{ \vmu \in\real^p, \,\,\, \vA \in\mathrm{GL}^{p\times p} }, \\
        \lparam &:= \crl{ \vdelta\in\real^p, \,\,\, \vM \in\mathcal{S}^{p\times p}  },
    \end{split}
    \label{eq:squareroot_param}
\end{equation}

\vspace{-0.35cm}
where $\vdelta$ and $\vM$ are the local parameters.
The map $\vpsi \circ \vphi_{\aparam_t}(\lparam)$ at $\aparam_t := \{\vmu_t, \vA_t\}$ is chosen to be\footnote{We use  the 1/2 shown in red in \eqref{eq:gauss_ex_map} to match the parameterizations in \citet{glasmachers2010exponential}, but the update in \eqref{eq:ngd_aux_param} remains unchanged even when without it. }

\vspace{-0.4cm}
\begin{equation}
    \begin{split}
        \crl{ \begin{array}{c} \vmu \\ \vSigma \end{array} } &= \vpsi(\aparam) :=  \crl{ \begin{array}{c} \vmu \\ \vA\vA^\top \end{array} } \\
        \crl{ \begin{array}{c} \vmu \\ \vA \end{array} } &= \vphi_{\aparam_t}(\lparam) :=  \crl{ \begin{array}{c} \vmu_t + \vA_t \vdelta \\ \vA_t \mathrm{Exp} \rnd{ {\color{red} \half } \vM} \end{array} }.
    \end{split}
    \label{eq:gauss_ex_map}
\end{equation}

\vspace{-0.35cm}
Finally, we obtain the natural gradients \eqref{eq:ngrad_local} by using the {\color{red}exact} Fisher matrix $\vF_{\lparam}(\lparam_0)$ (see Appx.~\ref{sec:gauss_cov} 
for a derivation),

\vspace{-0.35cm}
\begin{equation}
    \Big( \mspace{-6.5mu}\begin{array}{c} \vngrad_{\text{\vdelta}_0}^{(t)}\\ \mathrm{vec}( \vngrad_{\text{\vM}_0}^{(t)})\end{array} \mspace{-6.5mu}\Big) \mspace{-2mu} = \mspace{-2mu}
    \Big(\mspace{-7mu}\begin{array}{cc} \vI_p & 0 \\ 0 & \half \vI_{p^2} \end{array}\mspace{-7mu} \Big)^{-1} \mspace{-3mu}\Big( \mspace{-6.5mu}\begin{array}{c}  \vA_t^\top \vg_{\text{\vmu}_t} \\ \mathrm{vec}( \vA_t^\top \vg_{\text{\vSigma}_t} \vA_t) \end{array} \mspace{-6.5mu}\Big)
    \label{eq:gauss_ex_ng}
\end{equation}

\vspace{-0.35cm}
By plugging \eqref{eq:gauss_ex_map} and \eqref{eq:gauss_ex_ng} in \eqref{eq:ngrad_local},  our update can be written in the space of $\aparam$ as below, where $\vS_t^{-1}=\vSigma_t$.

\vspace{-0.4cm}
\begin{equation}
    \begin{split}
    \vmu_{t+1} &\leftarrow \vmu_t - \beta \vS_{t}^{-1} \vg_{\mu_t} \\
    \vA_{t+1} &\leftarrow \vA_t \mathrm{Exp}\big(-\beta \vA_t^T\vg_{\Sigma_t} \vA_t \big)
    \end{split}
     \label{eq:ngd_aux_param}
 \end{equation}

\vspace{-0.35cm}
By the {REINFORCE trick} \citep{williams1992simple}, the gradients with respect to global parameters are 

\vspace{-0.35cm}
\begin{equation}
    \begin{split}
    \vg_\mu  
    &= \Unmyexpect{q(\text{\vlat}|\gparam)}{ \sqr{\big(\vA^{-T} \vz \big) \ell( \vlat) }}\\
    \vg_\Sigma 
    &= \half \Unmyexpect{q(\text{\vlat}|\gparam)}{ \sqr{ \vA^{-T} \big(  \vz \vz^T -\vI \big) \vA^{-1} \ell( \vlat) }}   
    \end{split}
    \label{eq:gauss_ex_grad}
\end{equation}

\vspace{-0.35cm}
where $\vz=\vA^{-1}(\vlat-\vmu)$.
By plugging in \eqref{eq:gauss_ex_grad} into  \eqref{eq:ngd_aux_param}, we recover the update \eqref{eq:xnes} used in  \citet{glasmachers2010exponential}.
Appx.~\ref{sec:gauss_cov}  shows that Assumptions 1-2 are satisfied. 

Parameterizations $\lparam=\{\vdelta,\vM\}$ and $\aparam=\{\vmu,\vA\}$ play distinct roles.
Local parameter $\vM$ is chosen to be symmetric with $p(p+1)/2$ degrees of freedom
so that Assumption 1 holds (also see Appx.~\ref{app:asym_M}).
Auxiliary parameter $\vA$ can be an invertible matrix with $p^2$ degrees of freedom and 
the Fisher matrix $\vF_{\aparam}(\aparam)$ is  singular.
Note that we perform natural-gradient descent in $\lparam$ instead of $\aparam$.
This is in contrast with the other works \citep{sun2009efficient,salimbeni2018natural} that require a Cholesky structure in $\vA$ with $p(p+1)/2$ degrees of freedom to ensure that $\vF_{\aparam}(\aparam)$ is non-singular.

\citet{glasmachers2010exponential} only demonstrated their method in the Gaussian case without complete derivations\footnote{There are a few typos in their paper. The matrix $\vA$ is missing in their Eq 8 and a factor $2$ is missing in Eq 11.} and a formal formulation.
It is difficult to generalize their method without explicitly knowing the distinct roles of parameterizations $\lparam$ and $\aparam$. Moreover, their approach only applied to a square-root structure of the covariance and it is unclear how to generalize it to other structures (e.g., low-rank  structures). 
Our method fixes these issues of their approach.

\subsection{Connection to Newton's method}
\label{sec:gauss_newton_connect}
We now show that the update \eqref{eq:xnes} derived using local parameterization is in fact closely related to a Newton-like algorithm.
Specifically, we will convert the update of $\vA_{t+1}$ in \eqref{eq:xnes} to the update over $\vS_{t+1}$, as in \eqref{eq:newton_update}, and recover the Newton's update derived by \citet{lin2020handling}.
To do so, we need to make two changes. First, we will expand $\mathrm{Exp}\big(\beta \vM \big) = $ 

\vspace{-0.35cm}
\begin{equation}
    \vI+\sum_{k=1}^{\infty}\frac{(\beta \vM)^k}{k!} = \vI + \beta \vM + \half (\beta\vM)^2 +O(\beta^3).
    \label{eq:matexp}
\end{equation}

\vspace{-0.35cm}
Second, instead of using \eqref{eq:gauss_ex_grad}, we will use {Stein's identity} \citep{opper2009variational,wu-report}:

\vspace{-0.35cm}
\begin{equation}
    \begin{split}
\vg_\mu = \Unmyexpect{q}{ \sqr{ \nabla_\lat \ell( \vlat) } }, \,\,\,\,\,  
\vg_{\Sigma}
 = \half \Unmyexpect{q}{ \sqr{ \nabla_\lat^2 \ell( \vlat) } } 
 \end{split}
 \label{eq:gauss_stein_2nd}
 \end{equation}
Using these changes, the update over $\vS_{t+1}$ can be rewritten as a modified Newton's update proposed by \citet{lin2020handling},

\vspace{-0.35cm}
\begin{align}
&\vS_{t+1}  = \big(\vA_{t+1} \vA_{t+1}^T \big)^{-1} =  \vA_t^{-T} \mathrm{Exp}\big(2\beta  \vA_t^T\vg_{\Sigma} \vA_t \big) \vA_t^{-1} \nonumber \\
&\quad = \vS_t + \beta \Unmyexpect{q}\sqr{\nabla_\lat^2 \ell( \vlat)} {+ \color{red} \frac{\beta^2}{2}  \vG \vS_{t}^{-1} \vG } + O(\beta^3) \label{eq:gauss_newton_exp}
\end{align}

\vspace{-0.35cm}
where $\vG = \Unmyexpect{q}\sqr{\nabla_\lat^2 \ell( \vlat)}$.
Ignoring the red term  gives us the update \eqref{eq:newton_update} derived by \citet{khan18a}. 
The term is added by \citet{lin2020handling} to fix the positive-definite constraint violation, by Riemannian gradient descent. 
Thus, these methods can be seen as special cases of ours with an approximation of the exponential map.

\citet{lin2020handling} show  NGD is a first-order approximation of a geodesic. 
Our NGD, which has first-order of accuracy, includes a second-order term  to handle the positive-definite constraint.
As we will discuss in Sec.~\ref{sec:tri_group}, a higher-order term $O(\beta^3)$ is introduced for structured updates.

\subsection{Wishart with square-root precision structure}
\label{sec:wishart}

We will now show an example that goes beyond Gaussians. We consider a Wishart distribution which is a distribution over $p$-by-$p$ positive-definite matrices,

\vspace{-0.35cm}
\[
{\cal W}_p(\vLat|\vS,n) =\frac{|\text{\vLat}|^{(n-p-1)/2} |\vS|^{n/2} }{ \Gamma_p(\frac{n}{2} ) 2^{np/2} } e^{ -\text{\half} \mathrm{Tr}(\text{\vS}\text{\vLat})},
\]

\vspace{-0.35cm}
where $\Gamma_p(\cdot)$ is the multivariate gamma function. Here, the global parameters are based on the precision matrix $\vS$, unlike the example in Sec.~\ref{sec:GaussSquareRoot}. We will see that  our update will automatically take care of this difference and report a  similar update to the one obtained using $\vSigma$ in \eqref{eq:ngd_aux_param}.
%

We start by specifying the parameterization,

\vspace{-0.3cm}
\begin{equation*}
    \begin{split}
        \gparam &:= \crl{n\in \real, \,\,\, \vS \in \mathcal{S}_{++}^{p\times p} \,\,\,|\,\,\, n>p-1 }, \,\,\,\\
        \aparam &:= \crl{ b \in\real, \,\,\, \vB \in\mathrm{GL}^{p\times p} }, \\
        \lparam &:= \crl{ \delta\in\real, \,\,\, \vM \in\mathcal{S}^{p\times p}  },
    \end{split}
\end{equation*}

\vspace{-0.35cm}
and their respective maps defined at $\aparam_t := \{b_t, \vB_t\}$

\vspace{-0.35cm}
\begin{equation*}
    \begin{split}
        \crl{ \begin{array}{c} n \\ \vS \end{array} } &= \vpsi(\aparam) :=  \crl{ \begin{array}{c} 2f(b)+p-1 \\ (2f(b)+p-1 ) \vB\vB^\top \end{array} }, \\
        \crl{ \begin{array}{c} b \\ \vB \end{array} } &= \vphi_{\aparam_t}(\lparam) :=  \crl{ \begin{array}{c} b_t + \delta \\ \vB_t \mathrm{Exp} \rnd{\vM} \end{array} }.
    \end{split}
\end{equation*} where $f(b)=\log(1+\exp(b))$ is the soft-plus function\footnote{We use the soft-plus function instead of the scalar exponential map for numerical stability.}. The auxiliary parameter $\vB$ here is defined as the square-root of the \emph{precision} matrix $\vS$, unlike in the previous examples.
%

Denoting the gradients by  

\vspace{-0.3cm}
\begin{align}
 \vG_{\text{\vS}^{-1}} := \nabla_{\text{\vs}^{-1}} \Unmyexpect{q} \sqr{{ \ell(\vLat) }},\,\,\,\,\,\,
 g_n :=\nabla_n \Unmyexpect{q} \sqr{{ \ell(\vLat) }},  \label{eq:wishart_egrad}
\end{align}
we can write the updates as (derivation in Appx.~\ref{app:wishart}):
\begin{align}
     \vB_{t+1} &  \leftarrow \vB_t \mathrm{Exp} \rnd{  \frac{\beta}{n_t^2} \vB_t^{-1} \vG_{\text{\vS}_t^{-1}} \vB_t^{-T} } \label{eq:wishart_rgd_A}\\
     b_{t+1} &  \leftarrow b_t - 
     \beta c_t 
     \sqr{g_n - \frac{1}{n_t} \mathrm{Tr}\rnd{  \vG_{\text{\vS}_t^{-1}} \vS_t^{-1} } }
     \label{eq:wishart_rgd_b}
\end{align}
where $c_t=\frac{ 2(1+\exp( b_t))}{ \exp( b_t)}\big( - \frac{2p }{n_t} +   D_{\psi,p} (\frac{n_t}{2})\big)^{-1}$ and  $D_{\psi,p}(x) $ is  the multivariate trigamma function.
Moreover, we can use \emph{re-parameterizable} gradients \citep{figurnov2018implicit,wu-report} for $\vG_{\text{\vS}_t^{-1}}$ and  $ g_n$  due to the Bartlett decomposition \citep{smith1972wishart} (see Appx.~\ref{app:rep_g_wishart} for details).

The update \eqref{eq:wishart_rgd_A} for $\vB$ (square-root of the precision matrix) is very similar to the update for $\vA$ (square-root for covariance) in \eqref{eq:ngd_aux_param}. The change from covariance to precision parameterization changes the sign of the update. The step size is modified using the parameter $n_t$. The local parameterization can automatically adjust to such changes in the parameter specification, giving rise to intuitive updates.


\subsection{Connection to Riemannian Gradient Descent}
\label{sec:wishart_rgd_connect}

We will show that the updates on the Wishart distribution is a generalization of Riemannian Gradient Descent (RGD) over the space of positive-definite matrices. Given an optimization problem
\vspace{-0.2cm}
\[
\min_{Z \in {\cal S}_{++}^{p \times p}} \ell (\vZ)
\]
over the space of symmetric positive-definite matrices, the RGD update with retraction can be written in terms of the inverse $\vU = \vZ^{-1}$  (see Appx.~\ref{app:rgd_at_u} for the details),

\vspace{-0.35cm}
\begin{align*}
\vU_{t+1} \leftarrow \vU_t + \beta_1 \nabla \ell(\vZ_t) + \frac{\beta_1^2}{2} \big[\nabla \ell(\vZ_t)] \vU_t^{-1} \big[\nabla \ell(\vZ_t)]
\end{align*}

\vspace{-0.35cm}
where $\nabla$ is taken with respect to $\vZ$, and $\beta_1$ is the step size. We now show that this is a special case of \eqref{eq:wishart_rgd_A} where gradients \eqref{eq:wishart_egrad} are approximated at the mean of the Wishart distribution as $\Unmyexpect{q} \sqr{{ \vLat }}=n \vS^{-1}$. Denoting the mean by $\vZ_t$, the approximation is (see the derivation in Appx.~\ref{app:gd_wishart_at_mean}), 

\vspace{-0.35cm}
\begin{align}
    \vG_{\text{\vS}^{-1}_t} \approx n_t \nabla { \ell(\vZ_t) }, \quad
    g_{n_t} \approx \mathrm{Tr} \sqr{ \nabla \ell(\vZ_t) \vS_t^{-1} } \label{eq:rgd_mean} 
\end{align}

\vspace{-0.35cm}
Plugging \eqref{eq:rgd_mean} into \eqref{eq:wishart_rgd_b}, $b$ remains constant after the update, 

\vspace{-0.35cm}
\begin{align*}
 b_{t+1} \leftarrow b_t -  \beta c_t 
 \sqr{  \cancel{ \mathrm{Tr} \sqr{ \nabla \ell(\vZ_t) \vS_t^{-1} }} -\cancel{ \mathrm{Tr} \sqr{ \nabla \ell(\vZ_t) \vS_t^{-1} }}  }
\end{align*} 

\vspace{-0.35cm}
so that $b_{t+1} \leftarrow b_t$ and $n_t$ is constant since $n=2f(b)+p-1$.
Resetting the step-size to be $\beta=\half \beta_1 n$,\footnote{Since $n$ remains constant, $\beta=\half\beta_1 n$ is a constant step-size.}
\eqref{eq:wishart_rgd_A} becomes

\vspace{-0.35cm}
\begin{align}
 \vB_{t+1} &  
 \leftarrow \vB_t \mathrm{Exp} \rnd{  \frac{\beta_1}{2} \vB_t^{-1} \big[   \nabla { \ell(\vZ_t) } \big] \vB_t^{-T} } \label{eq:wishart_update_A_at_mean}
\end{align} 

\vspace{-0.35cm}
Finally, we express the update in terms of $\vU_t := \vZ_t^{-1}= \vB_t\vB_t^T$ to rewrite  \eqref{eq:wishart_update_A_at_mean} as by using the second-order terms in the matrix exponential \eqref{eq:matexp},

\vspace{-0.35cm}
\begin{align*}
    &\vU_{t+1} \leftarrow \vB_t \mathrm{Exp}(  \beta_1 \vB_t^{-1} \big[   \nabla { \ell (\vZ_t) } \big] \vB_t^{-T} ) \vB_t^T \\
    &\leftarrow \vU_t + \beta_1 \nabla \ell(\vZ_t) + \frac{\beta_1^2}{2} \big[\nabla \ell(\vZ_t)] \vU_t^{-1} \big[\nabla \ell(\vZ_t)] + O(\beta_1^3)
\end{align*} 

\vspace{-0.35cm}
recovering the RGD update. Thus, the RGD update is a special case of our update, where the expectation is approximated at the mean. This is a \emph{local} approximation to avoid sampling from $q(\vLat)$.
This derivation is another instance of reduction to a \emph{local} method using NGD over distributions, similar to the ones obtained by \citet{emti2020bayesprinciple}.

\subsection{Generalizations and Extensions}
\label{sec:non_exp_map}
In previous sections, we use the matrix exponential map to define $\vphi_{\aparam_t}(\lparam)$, but other maps can be used.
This is convenient since the map can be difficult to compute and  numerically unstable.
We propose to use another map: 

\vspace{-0.35cm}
\[ \vh(\vM):=\vI+\vM+\half \vM^2.\]

\vspace{-0.25cm}
Map $\vh(\cdot)$ plays a key role for complexity reduction in Sec.~\ref{sec:tri_group},
since it simplifies the natural-gradient computation in Gaussian and Wishart cases without changing the form of the updates (due to Lemma \ref{lemma:eq1}-\ref{lemma:eq3} in Appx.~\ref{app:FIM}).
For example, consider the Gaussian case in Sec.~\ref{sec:GaussSquareRoot} where covariance $\vSigma$ is used. Using our approach, we could easily change the parameterization to the precision  $\vS=\vSigma^{-1}$ instead, by changing the parameters in \eqref{eq:squareroot_param} to

\vspace{-0.25cm}
\begin{equation}
    \begin{split}
        \gparam &:= \crl{\vmu \in\real^p, \,\,\, \vS \in \mathcal{S}_{++}^{p\times p} }\,\,\,\\
        \aparam &:= \crl{ \vmu \in\real^p, \,\,\, \vB \in\mathrm{GL}^{p\times p} } \\
        \lparam &:= \crl{ \vdelta\in\real^p, \,\,\, \vM \in\mathcal{S}^{p\times p}  }.
    \end{split}
    \label{eq:gauss_prec_params}
\end{equation}

\vspace{-0.25cm}
We can  use map $\vh(\cdot)$ in the following transformations:

\vspace{-0.3cm}
\begin{equation}
    \begin{split}
        \crl{ \begin{array}{c} \vmu \\ \vS \end{array} } &= \vpsi(\aparam) :=  \crl{ \begin{array}{c} \vmu \\ \vB\vB^\top \end{array} } \\
        \crl{ \begin{array}{c} \vmu \\ \vB \end{array} } &= \vphi_{\aparam_t}(\lparam) :=  \crl{ \begin{array}{c} \vmu_t + \vB_t^{-T} \vdelta \\ \vB_t \vh (\vM) \end{array} }.
    \end{split}
    \label{eq:gauss_xnes_prec}
\end{equation}

\vspace{-0.3cm}
An update (see \eqref{eq:gauss_prec_at_S} in Appx.~\ref{sec:gauss_prec}) almost identical to \eqref{eq:gauss_newton_exp} is obtained with this parameterization and map. The difference only appears in a $O(\beta^3)$ term.
Unlike the method originally described by \citet{glasmachers2010exponential}, our formulation makes it easy for a variety of parameterizations and maps, while keeping the natural-gradient computation tractable. %

To avoid computing $\nabla_\lat^2 \ell( \vlat)$ in Gaussians, we could use  
the {re-parameterizable} trick\footnote{$\nabla_\lat \ell( \vlat)$ is only required to exist almost surely.} for the covariance  \citep{wu-report,lin2020handling} in \eqref{eq:gauss_newton_exp}, where $ \vK(\vlat):=\vS (\vlat-\vmu)  \nabla_\lat^T \ell( \vlat)$.
\vspace{-0.25cm}
\begin{equation}
    \begin{split}
 \vg_\Sigma = \half \Unmyexpect{q}{ \sqr{ \vK(\vlat)  } }
 = \frac{1}{4} \Unmyexpect{q}{ \sqr{ \vK(\vlat) +  \vK^T(\vlat)} }
 \label{eq:rep_grad_cov}
    \end{split}
\end{equation} 
 
By the identities in
(\ref{eq:rep_grad_cov},~\ref{eq:gauss_ex_grad},~\ref{eq:gauss_stein_2nd}),
we establish the connection of our Gaussian update 
to variational inference by the re-parameterizable trick, 
to numerical optimization by Stein's identity, and
to black-box search by  the REINFORCE trick.

Our approach also gives NGD updates for common univariate exponential family (EF) distributions via Auto-Differentiation (see Appx.~\ref{app:uni_ef} for the detail). 

In practice, the FIM under  global parameter $\gparam$ or  local parameter $\lparam$ can be singular.
For example,
the FIM of curved EFs \citep{lin2019fast} and 
MLPs \citep{amari2018dynamics} can be singular.
The FIM of the low-rank structured Gaussian \citep{tran2020bayesian,mishkin2018slang} has the same issue. (see Fig.~\ref{fig:fig1} and Appx.~\ref{app:diag_zero} for a discussion).
We extend our approach to the following two kinds of curved EFs, where we relax Assumption 1 for local parameterizations. 

In Appx.~\ref{app:mat_gauss}, we adapt our local parameterization approach to a block approximation for matrix Gaussian cases, where
cross-block terms in the FIM are set to zeros (see \eqref{eq:block_approx_fim} in Appx.~\ref{app:mat_gauss}).
Our approximated FIM is guaranteed to be non-singular since matrix Gaussian is a \emph{minimal multi-linear} EF \citep{lin2019fast}.
Our approach is very different from noisy-KFAC \citep{zhang2018noisy}. In noisy-KFAC,
KFAC approximation along with a block-approximation is used, where the approximated FIM
can be singular without damping. Damping introduces an extra tuning hyper-parameter.

In Appx.~\ref{app:mog}, we extend our approach to mixtures such as finite  mixtures of Gaussians using the FIM defined by  the joint distribution of a mixture.
Another case is the rank-one Gaussian in Fig.~\ref{fig:fig1}, which is a mixture distribution discussed in \citet{lin2019fast}. 
The FIM of the marginal is singular (see Fig.~\ref{fig:fig1}(I)) while the FIM of the joint is not.
\citet{lin2019fast} show  the FIM of the \emph{joint distribution} of a \emph{minimal conditional} mixture is guaranteed to be  non-singular.

\vspace{-0.1cm}
\section{NGD for Structured Matrix Groups}
\label{sec:tri_group}

\begin{figure*}[t]
\captionsetup[subfigure]{aboveskip=-1pt,belowskip=-1pt}
        \begin{subfigure}[b]{0.1\textwidth}
        \includegraphics[width=1\linewidth]{upper_triangular.tikz}
        \caption{}
	     \label{fig:group1}
	      \end{subfigure}
	      \hspace{0.16cm}
        \begin{subfigure}[b]{0.1\textwidth}
        \includegraphics[width=1\linewidth]{lower_triangular.tikz}
        \caption{}
	      \label{fig:group2}
	      \end{subfigure}
	      \hspace{0.16cm}
        \begin{subfigure}[b]{0.1\textwidth}
        \includegraphics[width=1\linewidth]{Heisenberg.tikz}
        \caption{}
	      \label{fig:group3}
	\end{subfigure}
	      \hspace{0.16cm}
        \begin{subfigure}[b]{0.1\textwidth}
        \includegraphics[width=1\linewidth]{conjugation.tikz}
              \caption{}
	\label{fig:group4}
	 \end{subfigure}
	     \hspace{0.16cm}
        \begin{subfigure}[b]{0.1\textwidth}
        \includegraphics[width=1\linewidth]{sparsecholesky.tikz}
              \caption{}
	\label{fig:group6}
	 \end{subfigure} 
	 	     \hspace{0.16cm}
        \begin{subfigure}[b]{0.1\textwidth}
        \includegraphics[width=1\linewidth]{TriToeplitz.tikz}
              \caption{}
	\label{fig:group7}
	 \end{subfigure}
	      \hspace{0.16cm}
        \begin{subfigure}[b]{0.28\textwidth}
	 \includegraphics[width=0.98\linewidth]{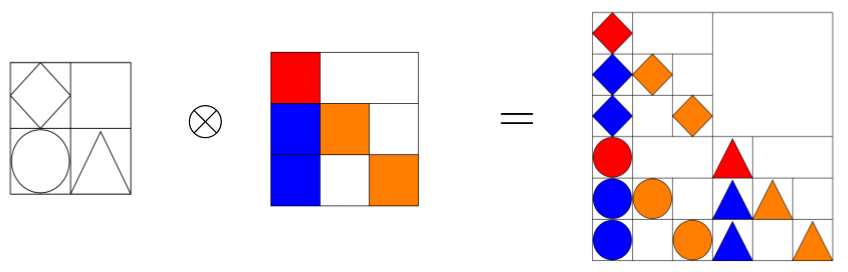} 
              \caption{}
	\label{fig:group5}
	 \end{subfigure}
	 \vspace{-0.2cm}
           \caption{ 
           Visualization of
         some useful  group structures.
         Figure~\ref{fig:group1} is a block upper-triangular group with $k=2$.
         Figure~\ref{fig:group2} is a block lower-triangular group with $k=2$.
         Figure~\ref{fig:group3} is a block Heisenberg group with $k_1=2$ and $k_2=4$, which is a hierarchical extension of Figure~\ref{fig:group2}.
         Figure~\ref{fig:group4} is a group conjugation of Figure~\ref{fig:group2} by a permutation matrix.
         Figure~\ref{fig:group6} is a sparse Cholesky group.
         Figure~\ref{fig:group7} is a triangular-Toeplitz group.
         Figure~\ref{fig:group5} is a Kronecker product group, which is a Kronecker product of two (block) lower-triangular groups.
         }
	\label{fig:groups}
\end{figure*}

We now show applications to NGD on matrices with special structures. The key idea is to use the fact that the auxiliary-parameter space $\mathrm{GL}^{p\times p}$ used in Sec.~\ref{sec:local_params} is a \emph{general linear group} (GL group) \citep{Group-notes}, and structured restrictions give us its subgroups.
For example, a Cholesky factor is a dense triangular group.
We can specify local parameterizations for the subgroups to get a tractable NGD. We will use the Gaussian example considered in Sec.~\ref{sec:non_exp_map} to illustrate this idea. A similar technique could be applied to the Wishart example.
We will discuss block triangular groups, and then discuss an extension inspired by the Heisenberg group.
Some useful  groups are illustrated in Fig.~\ref{fig:groups}.



\vspace{-0.1cm}
We denote ${\cal{B}_\text{up}}(k)$ 
the set of following block upper-triangular $p$-by-$p$ matrices as an auxiliary parameter space, where $k$ is the block size with $0\leq  k \leq  p$ and $d_0=p-k$, and ${\cal D}^{d_0 \times d_0}_{++}$ is the space of diagonal and invertible matrices.

\vspace{-0.3cm}
\begin{align*}
{\cal{B}_{\text{up}}}(k)  = \Big\{ 
\begin{bmatrix}
\vB_A &  \vB_B  \\
 \mathbf{0} & \vB_D
      \end{bmatrix} \Big| & \vB_A \in \mathrm{GL}^{k \times k},\,\,
 \vB_D  \in{\cal D}^{d_0 \times d_0}_{++}  \Big\}
\end{align*}

\vspace{-0.3cm}
When $k=0$, ${\cal{B}_{\text{up}}}(k)= {\cal D}^{p \times p}_{++}$ becomes a diagonal auxiliary space.
When $k=p$, ${\cal{B}_{\text{up}}}(k) = \mathrm{GL}^{p\times p}$ becomes a full space.
The following lemma shows ${\cal{B}_{\text{up}}}(k)$ is a \emph{matrix group}. 

\vspace{-0.15cm}
\begin{lemma}
\label{lemma:block_tri_gp}
${\cal{B}_{\text{up}}}(k)  $ is a matrix group that is closed under matrix multiplication.
\end{lemma}

\vspace{-0.1cm}
A local parameter space for ${\cal{B}_{\text{up}}}(k)$ is defined below with less degrees of freedom than the local space ${\cal S}^{p \times p}$ in \eqref{eq:gauss_prec_params}. 

\vspace{-0.25cm}
\begin{align*}
{\cal{M}_{\text{up}}}(k)  = \Big\{ 
\begin{bmatrix}
\vM_A &  \vM_B  \\
 \mathbf{0} & \vM_D
      \end{bmatrix} \Big| &  \vM_A \in{\cal S}^{k \times k}, \,\,
 \vM_D  \in{\cal D}^{d_0 \times d_0} \Big\}
\end{align*}

\vspace{-0.25cm}
where ${\cal D}^{d_0 \times d_0}$ denotes the space of diagonal matrices.
Lemma \ref{lemma:block_tri_assp1} shows that $\vh(\cdot)$ defined in Sec.~\ref{sec:non_exp_map} is essential. 

\vspace{-0.2cm}
\begin{lemma}
\label{lemma:block_tri_assp1}
For any $\vM \in {\cal{M}_{\text{up}}}(k)$,
 $\vh(\vM) \in {\cal{B}_{\text{up}}}(k) $.
\end{lemma}
\vspace{-0.1cm}

Using these spaces, we specify the parametrization for the Gaussian $\gauss(\vlat|\vmu,\vS^{-1})$, where the precision $\vS$ lives in  a sub-manifold\footnote{$\lparam$ locally gives a \href{https://en.wikipedia.org/wiki/Parametric_equation}{{parametric representation}} of the submanifold.
See \eqref{eq:global_low} in Appx.~\ref{app:proof_up_group_lemma} for an equivalent global parameterization of this submanifold using a sparse Cholesky factor.
} of $\mathcal{S}_{++}^{p\times p}$, 

\vspace{-0.25cm}
\begin{equation*}
    \begin{split}
       \gparam &:= \crl{\vmu \in\real^p, \,\,\, \vS=\vB \vB^T \in \mathcal{S}_{++}^{p\times p} \,\,\,|\,\,\, \vB \in {\cal{B}_{\text{up}}}(k) }, \,\,\,\\
        \aparam &:= \crl{ \vmu \in\real^p, \,\,\, \vB \in {\cal{B}_{\text{up}}}(k) }, \\
        \lparam &:= \crl{ \vdelta\in\real^p, \,\,\, \vM \in {\cal{M}_{\text{up}}}(k)   }.
    \end{split}
\end{equation*}

\vspace{-0.25cm}
The map $\vpsi \circ \vphi_{\aparam_t}(\lparam)$ at $\aparam_t := \{\vmu_t, \vB_t\}$ is chosen to be the same as \eqref{eq:gauss_xnes_prec} due to Lemma\ref{lemma:block_tri_gp} and Lemma \ref{lemma:block_tri_assp1}.
Lemma \ref{lemma:block_tri_assp2} below shows that this local parameterization is valid.

\vspace{-0.2cm}
\begin{lemma}
\label{lemma:block_tri_assp2}
Assumption 1-2 are satisfied in this case.
\end{lemma}
\vspace{-0.1cm}

The natural-gradients (see Appx.~\ref{app:ng_block_triangular_gauss}) computed using the {\color{red}exact} FIM are

\vspace{-0.25cm}
\begin{align*}
 \vngrad_{\delta_0}^{(t)}    = \vB_t^{-1} \vg_{\mu_t}; \,\,\,\,\, 
 \vngrad_{M_0}^{(t)} = \vC_{\text{up}} \odot \kappa_{\text{up}}\big( -2  \vB_t^{-1} \vg_{\Sigma_t} \vB_t^{-T} \big)
\end{align*} where $\odot$ is the element-wise product, $\kappa_{\text{up}}(\vX)$ extracts non-zero entries of   ${\cal{M}_{\text{up}}}(k)$ from $\vX$ so that $\kappa_{\text{up}}(\vX) \in {\cal{M}_{\text{up}}}(k)$,
 $\vJ $ is a matrix of ones,
 $\vC_{\text{up}}$ is a constant matrix defined as below,
where factor $\half$ appears in the symmetric part of ${\vC}_{\text{up}}$. 

\vspace{-0.25cm}
\begin{align*}
 \vC_{\text{up}} = 
 \begin{bmatrix}
\half \vJ_A & \vJ_B   \\
 \mathbf{0} & \half \vI_D
      \end{bmatrix}  \in {\cal{M}_{\text{up}}}(k)
\end{align*}
\vspace{-0.5cm}

The NGD update over the auxiliary parameters is
\vspace{0.15cm}
\begin{tcolorbox}[enhanced,colback=white,%
    colframe=red!75!black, attach boxed title to top right={yshift=-\tcboxedtitleheight/2, xshift=-1.25cm}, title=structured update, coltitle=red!75!black, boxed title style={size=small,colback=white,opacityback=1, opacityframe=0}, size=title, enlarge top initially by=-\tcboxedtitleheight/2]
\begin{align}
\vmu_{t+1} & \leftarrow \vmu_{t} - \beta \vS_t^{-1} \vg_{\mu_t} \nonumber \\
\vB_{t+1} & \leftarrow   \vB_t \vh \rnd{ \beta \vC_{\text{up}} \odot \kappa_{\text{up}}\big( 2 \vB_t^{-1} \vg_{\Sigma_t} \vB_t^{-T} \big) }
\label{eq:eq_group_ngd}
\end{align}
\end{tcolorbox}

\vspace{-0.25cm}
where 
\eqref{eq:eq_group_ngd} preserves the structure: $\vB_{t+1} \in {\cal{B}_{\text{up}}}(k)$ if $\vB_t \in {\cal{B}_{\text{up}}}(k)$.
When $k=p$,
update \eqref{eq:eq_group_ngd} recovers  update  \eqref{eq:gauss_prec_exp_updates} of the example in Sec.~\ref{sec:non_exp_map} and connects to  Newton's method in \eqref{eq:gauss_newton_exp} (see  \eqref{eq:gauss_prec_at_S} in Appx.~\ref{sec:gauss_prec}).
When $k<p$, \eqref{eq:eq_group_ngd} becomes a structured update preserved the group structure.

If we evaluate gradients using \eqref{eq:gauss_stein_2nd} at $\vmu_t$:  $\vg_{\mu_t} \approx \nabla_\mu \ell(\vmu_t) $ and $\vg_{\Sigma_t} \approx \half \nabla_\mu^2 \ell(\vmu_t) $,
\eqref{eq:eq_group_ngd} becomes a structured 2nd-order update with group structural invariance~\citep{lin2021snd}.

By exploiting the structure of $\vB$ (shown in Appx.~\ref{app:complexiity}), the update enjoys low time complexity $O(k^2 p)$. The product $\vS^{-1} \vg_{\mu}$ can be computed in $O(k^2 p)$. We can compute $\vB \vh(\vM)$ in $O(k^2 p)$ when  $\vB$ and $\vh(\vM)$ are block upper triangular matrices.
The gradient $\vg_{\Sigma}$ is obtained using Hessian where we only compute/approximate diagonal entries of the Hessian and use $O(k)$ Hessian-vector-products for non-zero entries of $\kappa_{\text{up}}\big( 2 \vB^{-1} \vg_{\Sigma} \vB^{-T} \big)$ (see \eqref{eq:upper_hvp} in Appx.~\ref{app:complexiity}).
We store the non-zero entries of $\vB$ with space complexity $O((k+1)p)$.
Map $\vh(\cdot)$  simplifies the computation and
reduces the time complexity, whereas the exponential map suggested by  \citet{glasmachers2010exponential} does not.

\begin{figure*}[t]
	\vspace{-0.1cm}
	\centering
	\hspace*{-1.2cm}
	\includegraphics[width=0.22\linewidth]{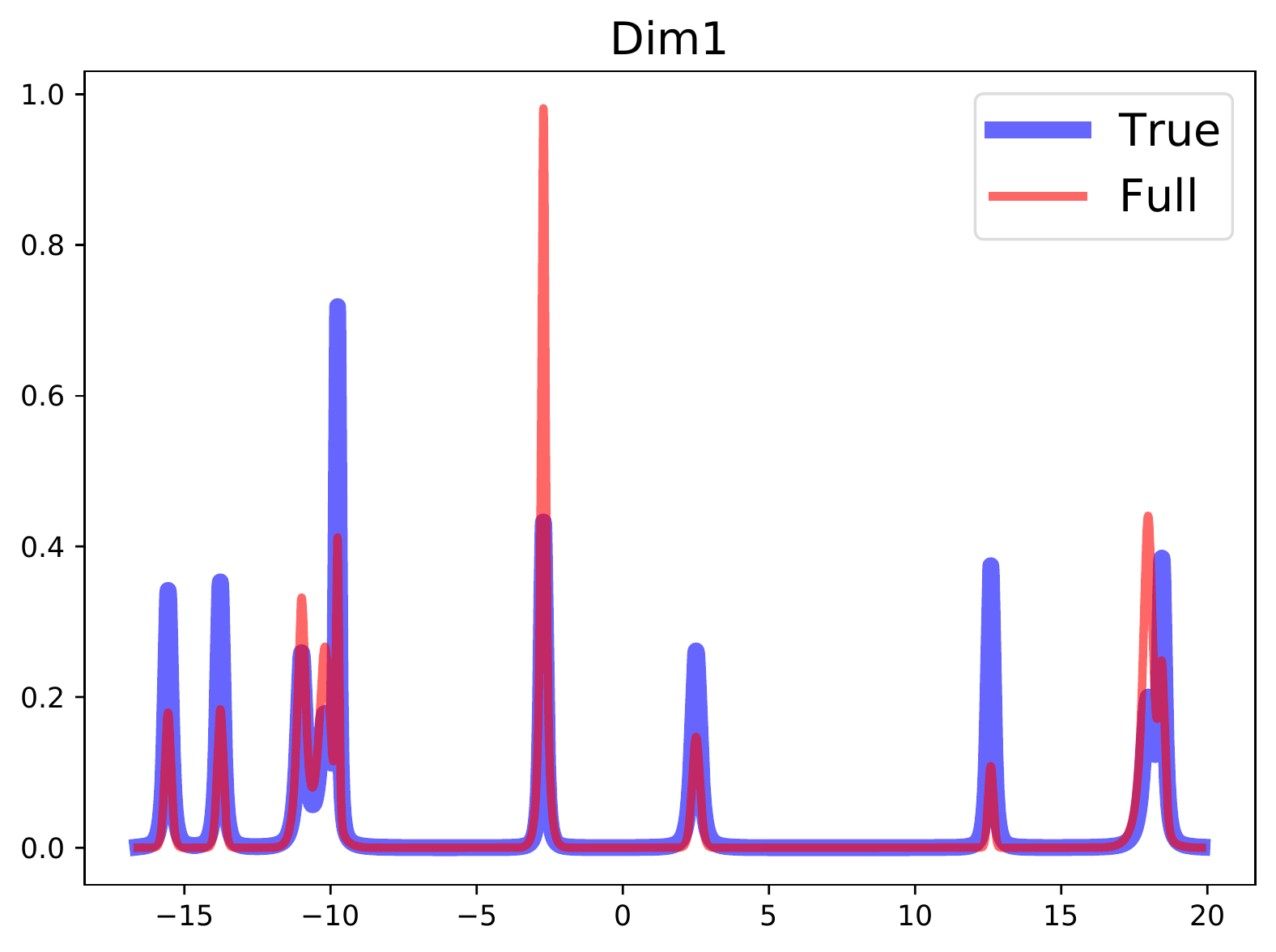}
	\includegraphics[width=0.22\linewidth]{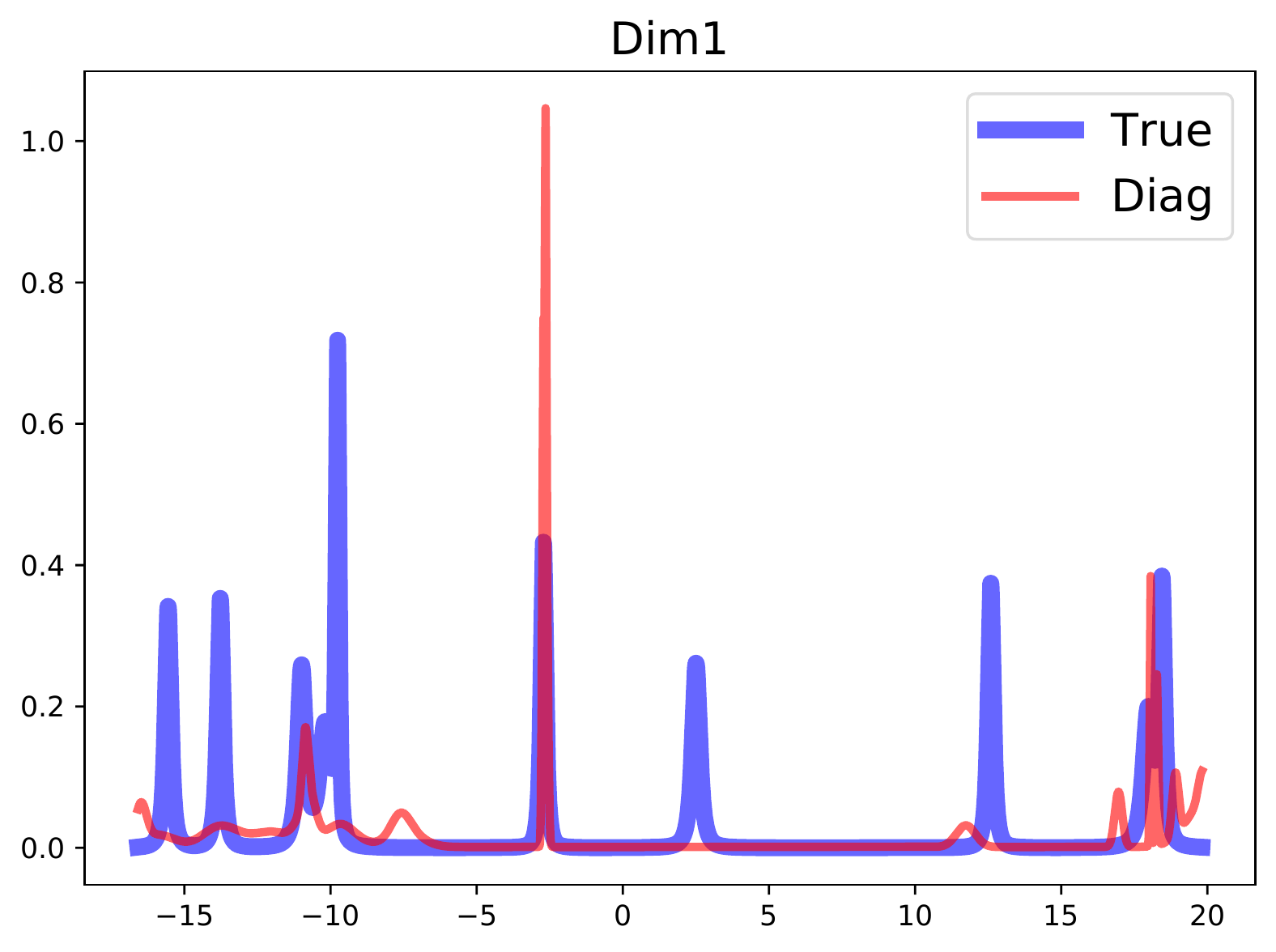} 
	\includegraphics[width=0.22\linewidth]{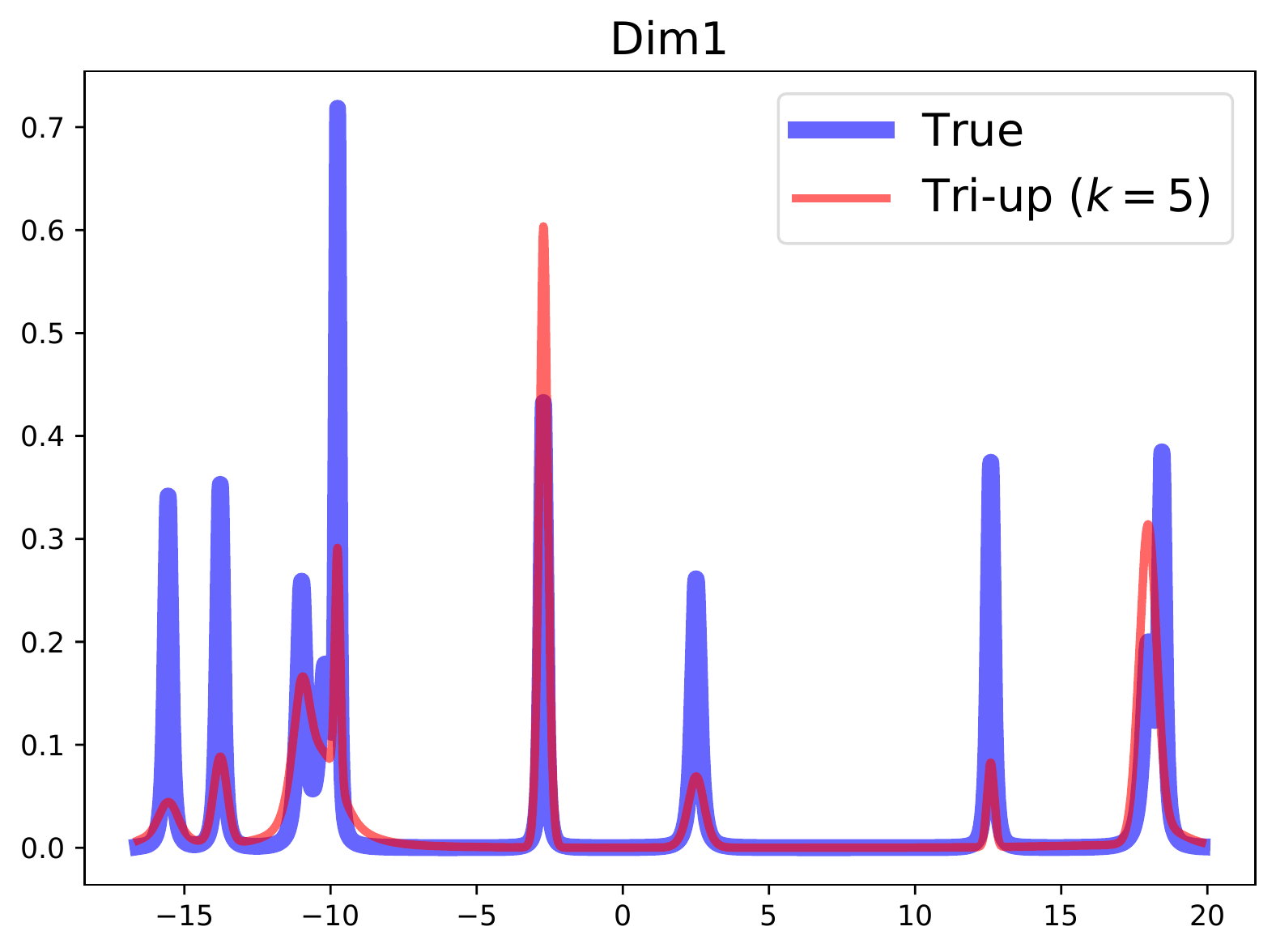}
	\includegraphics[width=0.22\linewidth]{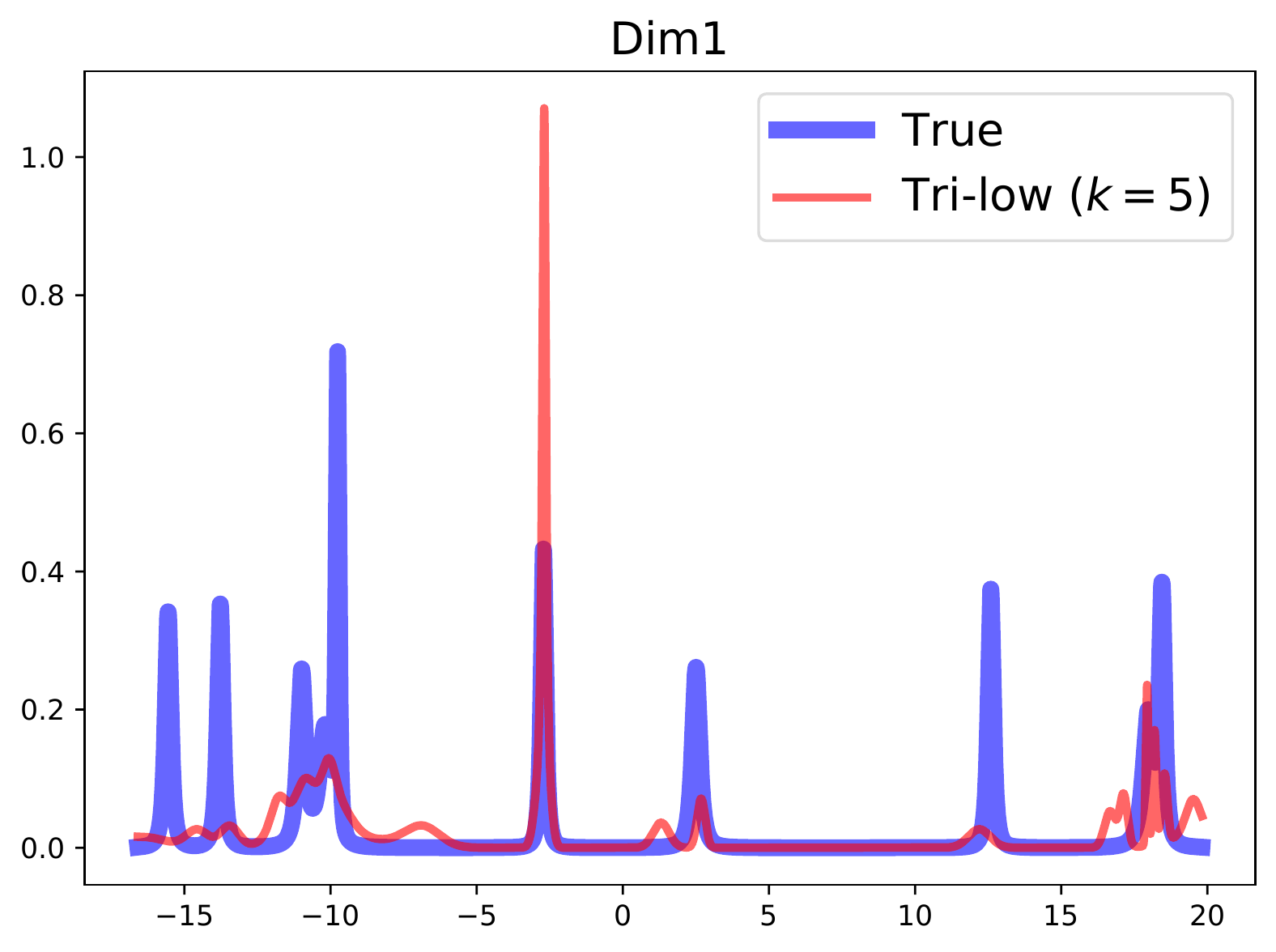}
	\hspace*{-1.2cm}
	\vspace{-0.35cm}
   \caption{
   Comparison results of structured Gaussian mixtures to fit a 80-Dim mixture of Student's t distributions with 10 components.
 The first marginal dimension obtained by our updates is shown in the figure, where an upper triangular structure in the precision form achieves better approximation than a lower triangular structure and a diagonal structure. The upper triangular structure performs comparably to the full covariance structure with lower computational cost. 
 Figure~\ref{fig:mixap1}-\ref{fig:mixap3} in~Appx.~\ref{app:more_results} show more dimensions and  results on other structures.
   }
   	\label{figure:mog}
\end{figure*}

As shown in Appx.~\ref{app:matrix_structure},
this  parameterization induces a special structure over $\vS_{\text{up}}=\vB\vB^T$, which is a block arrowhead matrix \citep{o1990computing}: 

\vspace{-0.35cm}
\begin{align*}
\vS_{\text{up}}
&=  \begin{bmatrix}
\vB_A  \vB_A^T + \vB_B \vB_B^T & \vB_B \vB_D \\
\vB_D \vB_B^T & \vB_D^2
      \end{bmatrix}
\end{align*}

\vspace{-0.35cm}
and over $\vSigma_{\text{up}}=\vS_{\text{up}}^{-1}$, which has a low-rank structure\footnote{  
The zero block  highlighted in red in the expression of $\vSigma_{\text{up}}$ guarantees the FIM to be non-singular (see Appx.~\ref{app:diag_zero}).}:

\vspace{-0.35cm}
\begin{align*}
\vSigma_{\text{up}} &= \vU_k \vU_k^T +  
      \begin{bmatrix}
     {\color{red} \mathbf{0} } & \mathbf{0} \\
 \mathbf{0}      &  \vB_D^{-2}
      \end{bmatrix};\,\,\,\, 
      \vU_k =  \begin{bmatrix}
-\vB_A^{-T} \\
\vB_D^{-1} \vB_B^T \vB_A^{-T}
      \end{bmatrix} 
\end{align*}

\vspace{-0.35cm}
where $\vU_k$ is a rank-$k$ matrix since $\vB_A^{-T}$ is invertible.

%
As shown in Appx.~\ref{app:block_lower_tri},
we obtain a similar update for a block lower-triangular group ${\cal{B}_{\text{low}}}(k)$ (see Fig.~\ref{fig:group2}).

\vspace{-0.35cm}
\begin{align*}
{\cal{B}_{\text{low}}}(k)  = \Big\{ 
\begin{bmatrix}
\vB_A & \mathbf{0}   \\
 \vB_C & \vB_D
      \end{bmatrix} \Big| & \vB_A \in \mathrm{GL}^{k \times k},\,\,
 \vB_D  \in{\cal D}^{d_0 \times d_0}_{++}  \Big\}
\end{align*}

\vspace{-0.3cm}
Our update with a structure $\vB \in {\cal{B}_{\text{low}}}(k)$ has a low-rank structure in precision $\vS_{\text{low}}=\vB\vB^T$.
Likewise, our update  with a structure  $\vB \in {\cal{B}_{\text{up}}}(k)$ has a low-rank structure in covariance\footnote{For the example in Sec.~\ref{sec:GaussSquareRoot}, our update with $\vA \in  {\cal{B}_{\text{low}}}(k)$ has a low-rank structure in covariance $\vSigma=\vA\vA^T$ (also see Figure~\ref{fig:fig1}).}
$\vS_{\text{up}}^{-1}=(\vB\vB^T )^{-1}$.
They are `structured second-order updates' where the precision can be seen as approximations of Hessians in Newton's method (see Sec.~\ref{sec:gauss_newton_connect}). 

An extension is to construct a \emph{hierarchical structure} inspired by the Heisenberg group~\cite{HeisenbergdDim-2018} by replacing a diagonal group\footnote{${\cal D}^{d_0 \times d_0}_{++}$ is indeed a diagonal matrix group.} in $\vB_D$ with a block triangular group, where $0 \leq k_1+k_2\leq p$ and $d_0=p-k_1-k_2$

\vspace{-0.35cm}
\begin{align*}
 {\cal B}_{\text{up}} (k_1,k_2)
  = \Big\{  \begin{bmatrix}
\vB_A & \vB_B \\
\mathbf{0} & \vB_D
      \end{bmatrix} \Big| 
      \vB_D = \begin{bmatrix}
      \vB_{D_1} & \vB_{D_2}\\
      \mathbf{0} & \vB_{D_4}
     \end{bmatrix}
 \Big\}
\end{align*}

 \vspace{-0.35cm}
where 
$\vB_A \in \mathrm{GL}^{k_1 \times k_1}$,
$\vB_{D_1}  \in{\cal D}_{++}^{d_0 \times d_0}$,
$\vB_{D_4} \in \mathrm{GL}^{k_2 \times k_2}$.

This  group has a flexible structure and recovers the block triangular group as a special case when $k_2=0$.
We can also define a lower Heisenberg group ${\cal B}_{\text{low}} (k_1,k_2)$ (see Fig.~\ref{fig:group3}).
In Appx.~\ref{app:Heisenberg_group}, we show that these groups can be used as structured parameter spaces for NGD, which could be useful for problems of interest in optimization, inference, and search.

If the Hessian $\nabla^2 \ell(\vlat)$ has a model-specific structure, 
we could design a customized group to capture such a structure in the precision.
For example, the Hessian of layer-wise matrix weights of a NN admits a Kronecker form (see Appx.~\ref{app:hessian_dnn}).
We can use a \emph{Kronecker product group} (see Fig.~\ref{fig:group5}) so that the precision can capture such structure of the Hessian.
This  group structure can reduce the time complexity from the quadratic complexity to a linear complexity in $k$ (see Appx.~\ref{app:com_red} and Fig.~\ref{figure:dnn}).
Even
when we employ the Gauss-Newton approximation to avoid computing the Hessian, this structure still preserves
a  Kronecker structure for each layer-wise matrix weight and leads us to a \emph{structured adaptive method} for the NN (see Sec.~\ref{sec:opt_dl}).

Many subgroups (e.g.,  invertible (block) circulant matrix groups,  invertible (block) triangular-Toeplitz matrix groups, sparse triangular groups)
of the GL group $\mathrm{GL}^{p \times p}$ and groups constructed from existing groups via \emph{the group conjugation} (the matrix similarity transform) by an element of the orthogonal group (see Fig.~\ref{fig:group4}-\ref{fig:group7})  can be used as structured auxiliary parameter spaces ${\cal B}$.
Our approach to construct a structured Gaussian-precision is valid
if there exists a local parameter space ${\cal M}$ so that $\vh(\vM) \in {\cal B}$ for any $\vM \in {\cal M}$ and Assumptions 1-3 are satisfied.
If these conditions hold, the inverse of FIM $\vF^{-1}_{\lparam}(\lparam_0)$ using ${\cal M}$ will be easy to compute due to Lemma
\ref{lemma:fim_M_gauss_prec} in Appx.~\ref{sec:gauss_prec}.
We can even weaken Assumption 1 as discussed in Sec.~\ref{sec:non_exp_map}. The computational requirements are (1) the group product and inverse can be efficiently  implemented and (2) 
$\kappa \big( 2 \vB^{-1} \vg_{\Sigma} \vB^{-T} \big) \in  {\cal M}$ can be implemented without computing the whole Hessian in \eqref{eq:gauss_stein_2nd}, where $\vB \in {\cal B}$ and $\kappa(\cdot)$ converts $\real^{p\times p}$ to ${\cal M}$.

Auxiliary parameter spaces for the matrix parameter  are indeed (closed) matrix Lie groups.
The corresponding local parameter space is a Lie sub-algebra\footnote{We use a Lie sub-algebra instead of its Lie algebra since the degrees of freedom for the auxiliary parameter could be greater  than the one for the local parameter.} with the matrix commutator as its Lie bracket.
The matrix exponential map is indeed the Lie-group exponential map.
Although
map $\vh(\cdot)$ is not the  exponential map, map $\vh(\cdot)$
simplifies the computation and satisfies the conditions of our NGD. 

\section{Numerical Results}
\label{sec:results}
\begin{figure*}[t]
\vspace{-0.1cm}
\captionsetup[subfigure]{aboveskip=-1pt,belowskip=-1pt}
        \centering
\hspace*{-2.5cm}
        \begin{subfigure}[b]{0.31\textwidth}
	\includegraphics[width=\textwidth]{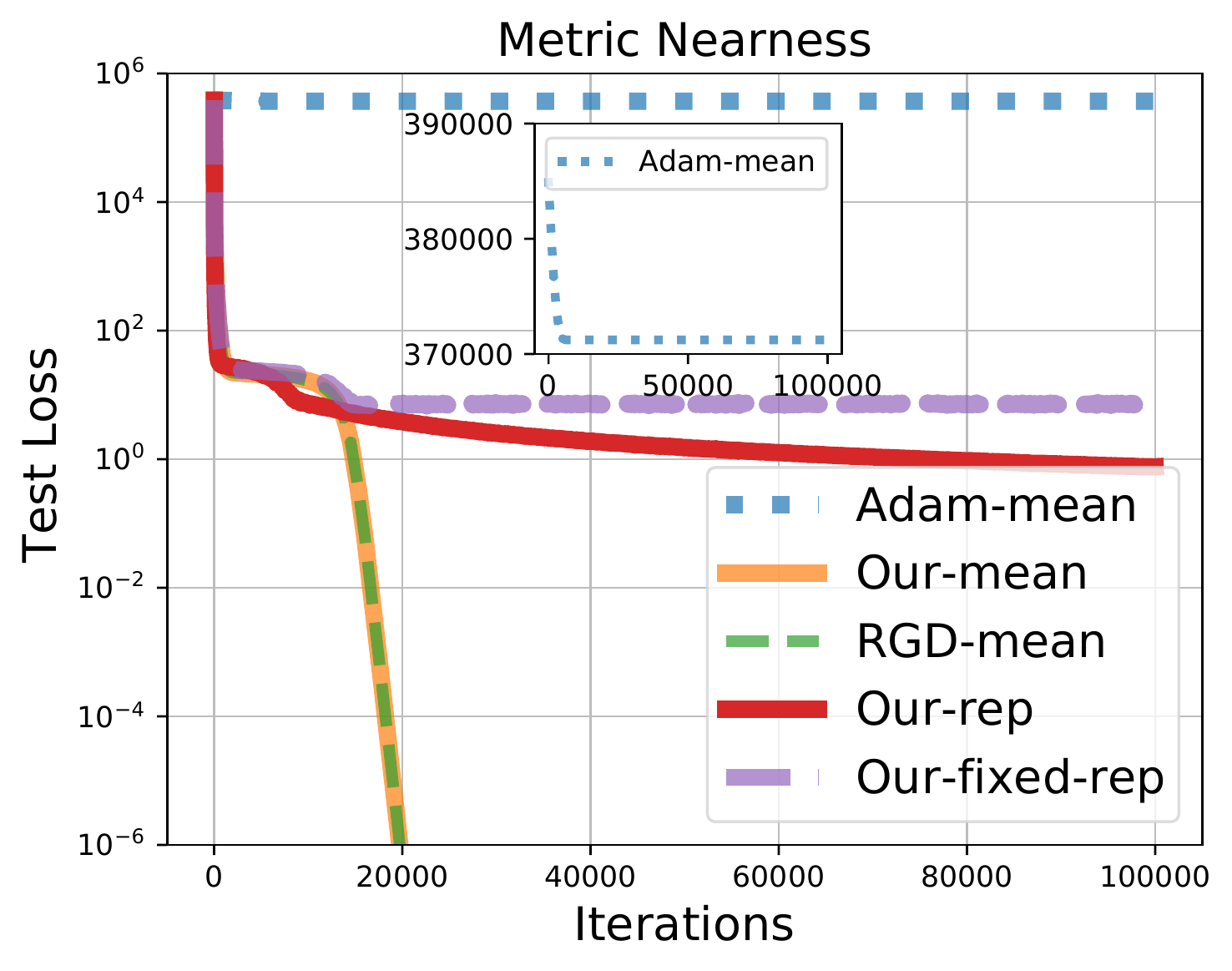}
                \caption{}
	      \label{fig:a}
        \end{subfigure}       
	\hspace*{-0.3cm}
        \begin{subfigure}[b]{0.31\textwidth}
	\includegraphics[width=\textwidth]{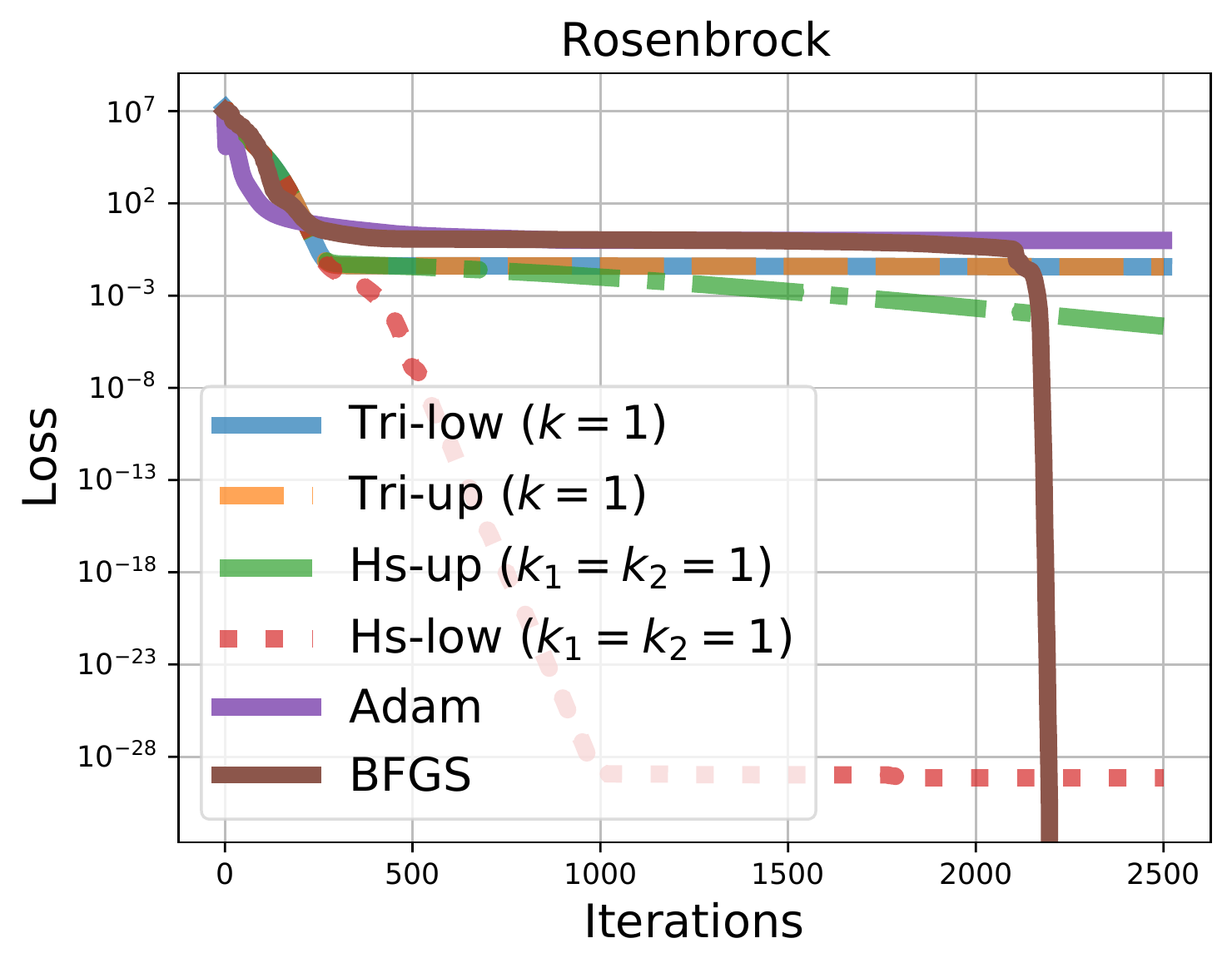}
                \caption{}
	      \label{fig:b}
        \end{subfigure}
	\hspace*{-0.3cm}
         \begin{subfigure}[b]{0.31\textwidth}
	\includegraphics[width=\textwidth]{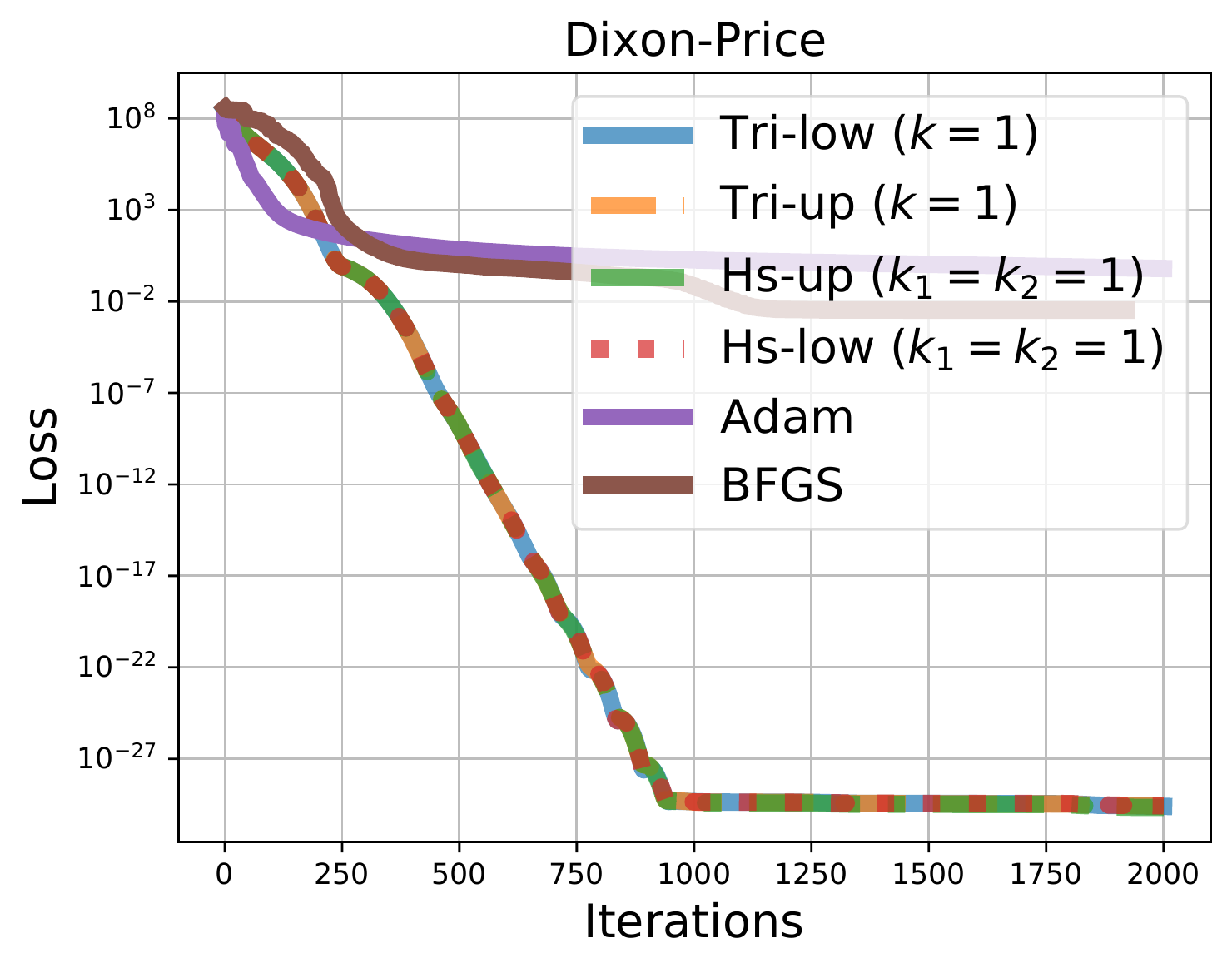}
                \caption{}
	      \label{fig:c}
        \end{subfigure}
     \vspace{-0.45cm}   
\hspace*{-1.2cm}
           \caption{ 
   The performances of our updates for search and optimization problems.
 Figure \ref{fig:a} shows the performances using a Wishart distribution to search the optimal solution of 
 a metric nearness task where our method evaluated at the mean behaves like RGD and converges faster than the Riemannian trivialization \citep{lezcano2019trivializations} with Adam. Our updates with re-parameterizable gradients also can find a solution near the optimal solution. 
Figure \ref{fig:b} and \ref{fig:c} show the performances using structured Newton's updates to optimize non-separable, valley-shaped, 200-dimensional functions, where our updates only require to compute diagonal entries of Hessian and Hessian-vector products. Our updates with a lower Heisenberg structure in the precision form converge faster than BFGS and Adam.
 }
\vspace{-0.3cm}
\end{figure*}

\begin{figure*}[t]
	\centering
	\hspace*{-1.2cm}
	\includegraphics[width=0.25\linewidth]{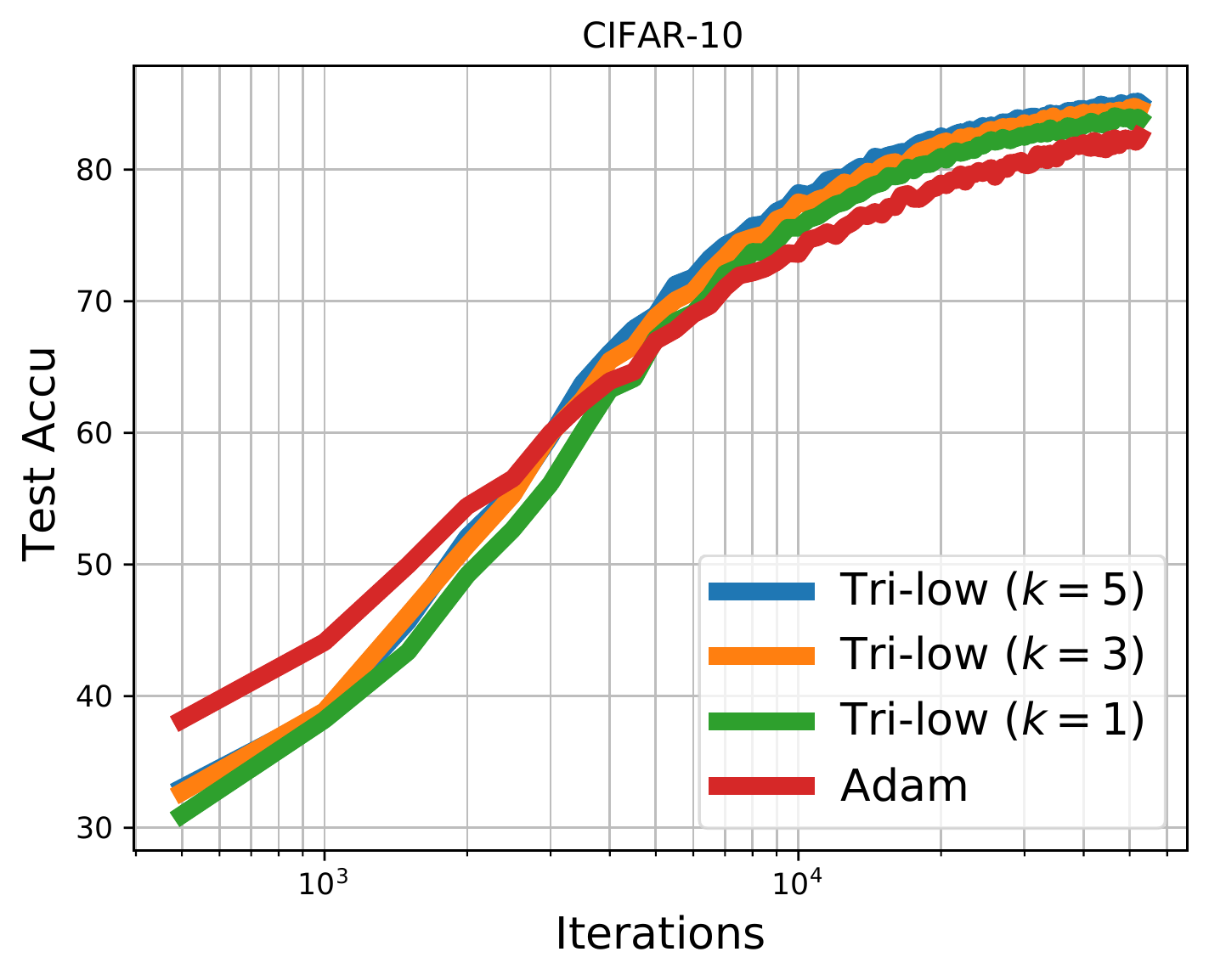}
	\includegraphics[width=0.25\linewidth]{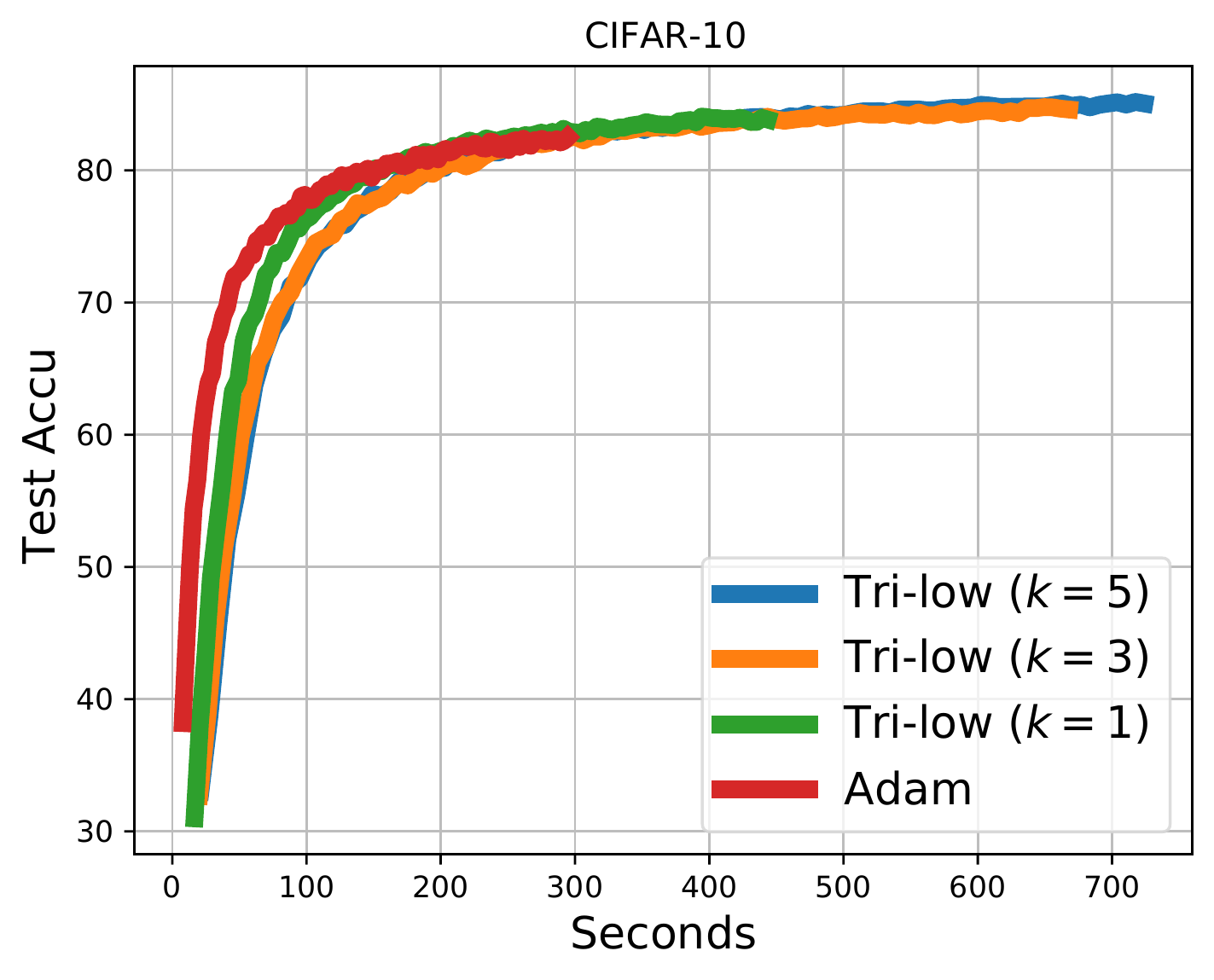}
	\includegraphics[width=0.25\linewidth]{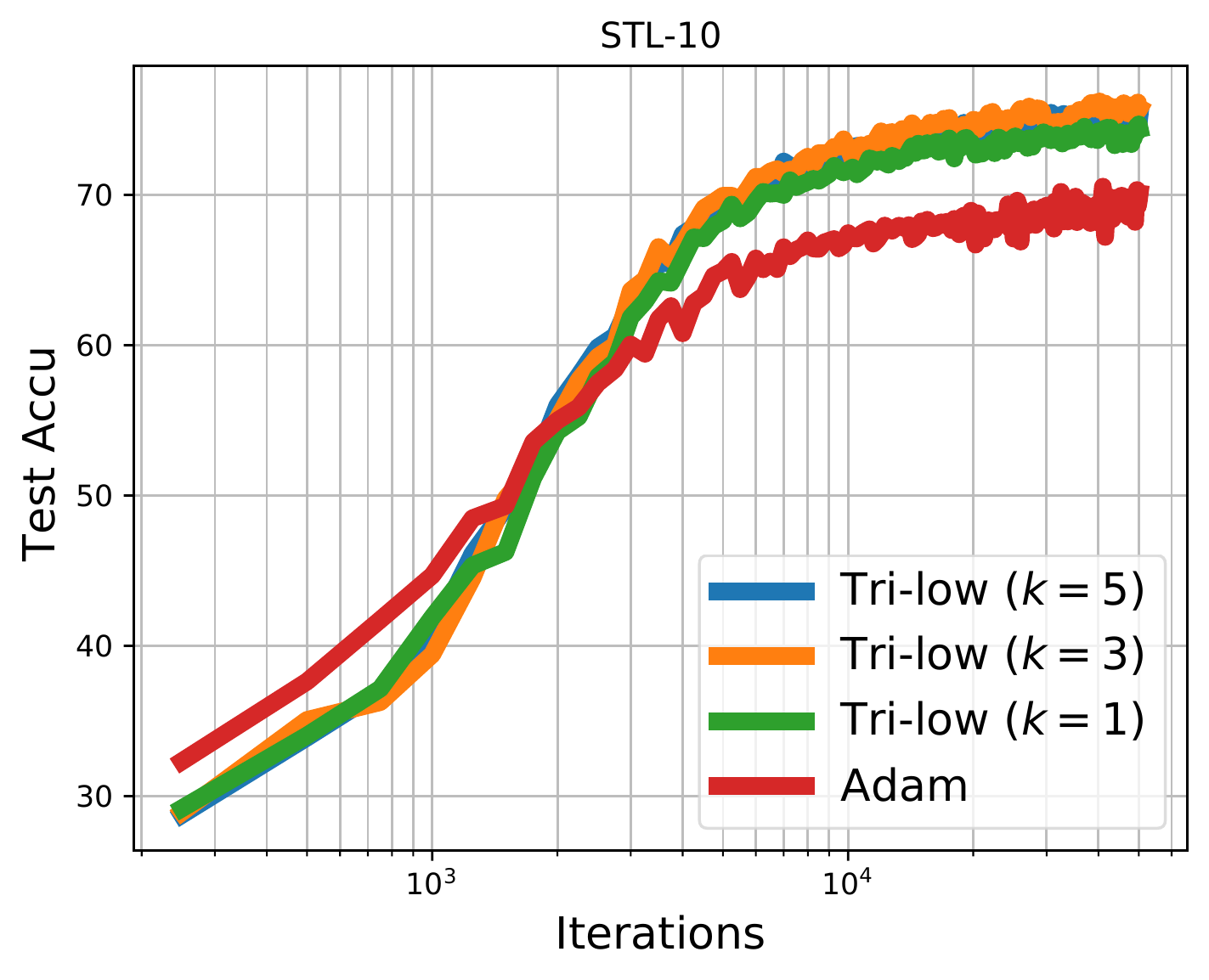}
	\includegraphics[width=0.25\linewidth]{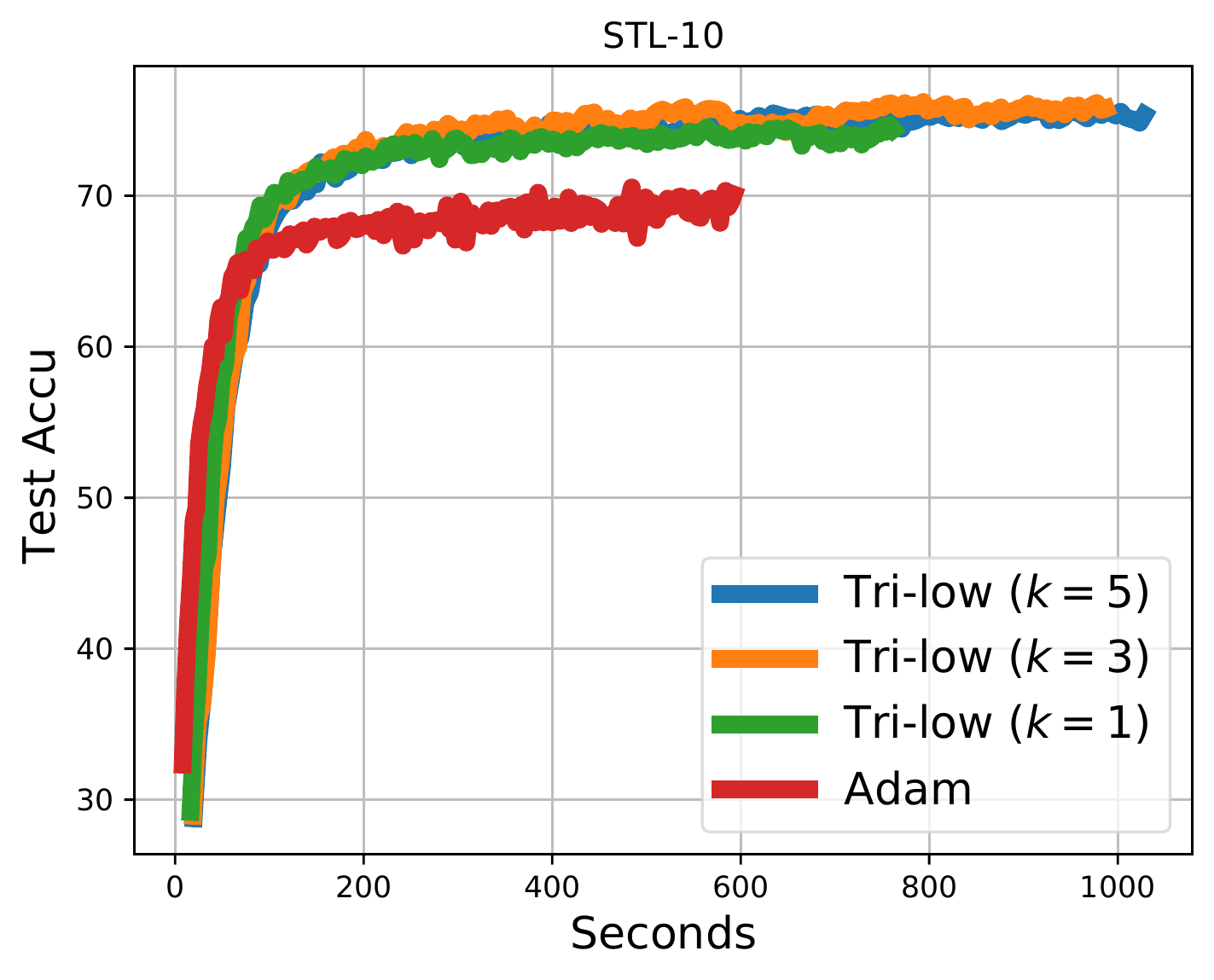}
	\hspace*{-1.2cm}
	\vspace{-0.45cm}
	\caption{
	   The performances for optimization of a CNN using matrix Gaussian with low-rank in a Kronecker precision form, where our updates ($O(k|\vlat|)$) have a linear iteration cost like Adam ($O(|\vlat|)$) and are automatically parallelized by Auto-Diff. Our updates achieve higher test accuracy ($75.8\%$ on ``STL-10'' and $85.0\%$ on ``CIFAR-10'') than Adam ($69.5\%$ on ``STL-10'' and $82.3\%$ on ``CIFAR-10'').
	}
	\label{figure:dnn}
\vspace{-0.2cm}
\end{figure*}

We present results on problems involving search, inference, optimization, and deep learning, where 
Table \ref{tab:updates} in Appx.~\ref{app:summary} summarizes our updates.
We use $\vh(\cdot)$ defined in Sec.~\ref{sec:non_exp_map} to replace the matrix exponential map in our proposed updates.

\vspace{-0.1cm}
\subsection{Search with Re-parameterizable Gradients}
We validate our update in the metric nearness task \cite{brickell2008metric} using a Wishart distribution as a search distribution $q$ with $\gamma=0$ in  \eqref{eq:problem}.
The objective function is $\ell(\vLat)= \frac{1}{2N} \sum_{i=1}^{N} \| \vLat \vQ \vx_i -\vx_i  \|_2^2 $, where $\vx_i \in {\cal R}^d$, $\vQ \in {\cal S}_{++}^{p \times p}$ and $\vLat \in {\cal S}_{++}^{p \times p}$.
The optimal is $\vQ^{-1}$. We randomly generate $\vx_i$ and $\vQ$ with $p=50$, $N_\text{train}=125,000$ for training and $N_{\text{test}}=25,000$ for testing.
All methods are trained using mini-batches, where the size of mini-batch is $100$.
We use re-parameterizable gradients with 1 Monte Carlo (MC) sample in our update (referred to as ``our-rep''), where we update $\vB$ and $b$. we also consider to only update $\vB$ with re-parameterizable gradients (referred to as ``our-fixed-rep'').
To numerically show the similarity between RGD and our update, we consider a case where gradients are evaluated at the mean (referred to as ``-mean'').
We consider these baselines: the  RGD update for positive-definite manifolds and the Riemannian trivialization\footnote{In variational inference (VI), trivializing a parametric distribution  is a special case of  black-box VI \citep{ranganath2014black}.} \citep{lezcano2019trivializations}, where gradients are evaluated at the mean. For the trivialization,
\citet{lezcano2019trivializations} suggests using Adam to perform updates in a trivialized (Euclidean) space.
We consider trivializations for the positive-definite manifold: a Cholesky factor and the matrix logarithmic function. 
We report the best result of the trivializations denoted by ``Adam''.
From Fig. \ref{fig:a}, we can see our update performs similarly to RGD if gradients are evaluated at the mean while the trivialization method is trapped in a local mode.
If we use re-parameterizable gradients, jointly updating both parameters is better than only updating $\vB$.

\subsection{Variational Inference with Gaussian Mixtures}
We consider the Gaussian mixture approximation problem \cite{lin2020handling}, where we use a Gaussian mixture
with $K$ components $q(\vlat)=\frac{1}{K}\sum_{k=1}^{K} \gauss(\vlat|\vmu_k,\vS_k^{-1})$ as a variational distribution $q$ with $\gamma=1$ in  \eqref{eq:problem}.
The goal of the problem is to approximate a mixture of $p$-dimensional Student's t distributions $\exp( - \ell(\vlat) ) =\frac{1}{C}\sum_{c=1}^{C} \Student(\vlat|\vu_c,\vV_c,\alpha)$ with $\alpha=2$.
We consider six kinds of structures of each Gaussian component: full precision (referred to as ``full''),  
diagonal precision (referred to as ``diag''),
precision with the  block upper triangular structure  (referred to as ``Tri-up''),
precision with the  block lower triangular structure (referred to as ``Tri-low''),
precision with the  block upper Heisenberg structure (referred to as ``Hs-up''),
precision with the  block lower Heisenberg structure (referred to as ``Hs-low'').
Each entry of  $\vu_c$ is generated uniformly in an interval $(-s,s)$. Each matrix $\vV_c$ is generated as suggested by \citet{lin2020handling}.
We consider a case with $K = 40, C = 10, p = 80, s=20$.
We update each  component  during training, where 10 MC samples are used to compute gradients. We compute gradients as suggested by \citet{lin2020handling}, where second-order information is used.
For structured updates, we  compute Hessian-vector products and diagonal entries of the Hessian without
directly computing the Hessian $\nabla_\lat^2 \ell(\vlat)$.
From Figure \ref{figure:mog}, we can see an \emph{upper structure} is better for inference problems\footnote{For
{variational inference}, an
{\emph{upper structure}} in the precision is better than a lower structure to capture off-diagonal correlations.}. 
Figure~\ref{fig:mixap1}-\ref{fig:mixap3} in~Appx.~\ref{app:more_results} show more results on dimensions and structures such as Heisenberg structures. 

\subsection{Structured Second-order Optimization}
We consider non-separable  valley-shaped test functions for optimization:
Rosenbrock: 
$ \ell_{\text{rb}}(\vlat) =\frac{1}{p} \sum_{i=1}^{p-1}\big[100(\lat_{i+1}-\lat_i)^2 + (\lat_i-1)^2 \big]$,
and Dixon-Price:
$ \ell_{\text{dp}}(\vlat) =\frac{1}{p}\big[ (\lat_i-1)^2 + \sum_{i=2}^{p} i(2\lat_i^2 - \lat_{i-1})^2 \big]$.
We test our structured Newton's updates, where we set $p=200$ and $\gamma =1$ in \eqref{eq:problem}.
We consider these  structures in the precision:
 the upper triangular structure (denoted by ``Tri-up''),
 the lower triangular structure (denoted by ``Tri-low''),
 the upper Heisenberg structure (denoted by ``Hs-up''),
and the lower Heisenberg structure (denoted by ``Hs-low''), where 
second-order information is used.
For our updates, we compute Hessian-vector products and diagonal entries of the Hessian without
directly computing the Hessian.
We consider baseline methods: the BFGS method provided by SciPy and the Adam optimizer, where the step-size is tuned for Adam.
We evaluate gradients at the mean for all methods.
Figure \ref{fig:b}-\ref{fig:c} show the performances of all methods\footnote{Empirically, we find out that a {\emph{lower structure}} in the precision  performs better than an upper structure for {optimization} tasks including optimization for neural networks.},
where our updates with a lower Heisenberg structure converge faster than BFGS and Adam.

\subsection{Optimization for Deep Learning}
\label{sec:opt_dl}
We consider a CNN model with 9 hidden layers, where 6 layers are convolution layers.
For a smooth objective, we use average pooling and GELU~\citep{hendrycks2016gaussian} as activation functions.
We employ $L_2$ regularization with weight $10^{-2}$.
We  set $\gamma =1$ in \eqref{eq:problem} in our updates.
We train the model with our updates derived from  matrix Gaussian (see Appx.~\ref{app:mat_gauss}) for each layer-wise matrix weight\footnote{
$\vLat \in \real^{c_{\text{out}}\times c_{\text{in}}p^2}$ is a weight matrix, where $p$, $c_{\text{in}}$,  $c_{\text{out}}$  are  the kernel size, the number of input, output channels, respectively.
}  on
datasets ``CIFAR-10'', ``STL-10''. 
Each Gaussian-precision has  a Kronecker product group structure of two lower-triangular groups (referred to as ``Tri-low'') for computational complexity reduction (see Appx.~\ref{app:com_red}).
For ``CIFAR-10'' and ``STL-10'', we train the model with mini-batch size 20.
Additional results on ``CIFAR-100'' can be found at Figure~\ref{figure:dnn2} in~Appx.~\ref{app:more_results}.
We evaluate gradients at the mean and approximate the Hessian by the Gauss-Newton approximation.
We compare our updates to Adam, where the step-size for each method is tuned by grid search.
We use the same initialization and hyper-parameters in all methods.
We report results in terms of test accuracy, where we average the results over 5 runs with distinct random seeds.
From Figure \ref{figure:dnn}, we can see our structured updates have a linear iteration cost like Adam while achieve higher test accuracy.

\section{Conclusion}
We propose a systematic approach  for NGD to 
incorporate group structures in parameter spaces.
Compared to existing NGD methods,
our method enables more flexible covariance structures with lower complexity
while keeping the update simple.
Moreover,
our approach gives
structured second-order methods for unconstrained optimization and
structured adaptive algorithms for NNs. 
An interesting direction is to evaluate our methods in  large-scale settings.

\section*{Acknowledgements}
WL is supported by a UBC International Doctoral Fellowship.
This research was partially supported by the Canada CIFAR AI Chair Program.

\bibliography{refs}
\bibliographystyle{icml2021}

\clearpage

\newpage
\clearpage
\pagebreak
\onecolumn
\begin{appendices}

Outline of the Appendix:
\begin{itemize}
 \item 
Appendix \ref{app:summary} summarizes parameterizations and updates used in this work, which gives a road-map of the appendix. 
 \item 
Appendix \ref{app:more_results} contains more experimental results.
\item
Appendix \ref{app:FIM} contains some useful results used in the remaining sections of the appendix.
\item
The rest of the appendix contains proofs of the claims and derivations of our update for examples summarized in Table  \ref{tab:examples} and Table \ref{tab:updates}.
\end{itemize}

\section{Summary of Parameterizations Used in This Work}
\label{app:summary}

\begin{table*}[ht]
\begin{minipage}{\columnwidth}
\begin{tabular}{l|l|l|l}
   \hline
$q(\vlat)$
& Name
   & Our update in auxiliary space $\aparam$
    \\
    \hline
$\gauss(\vlat|\vmu,\vSigma)$ (App. \ref{sec:gauss_cov} )  
& Gaussian with covariance
& See Eq \eqref{eq:ngd_aux_param_app}
    \\

    \hline

$\gauss(\vlat|\vmu,\vS^{-1})$  (App. \ref{sec:gauss_prec} ) 
& Gaussian with precision
  & See Eq \eqref{eq:gauss_prec_exp_updates} for a full structure;
    \\
  & &
  See Eq \eqref{eq:gauss_prec_triaup_ngd_aux} and \eqref{eq:low_sym_gauss_prec_ng} for a block triangular structure

  \\
  & &
See Eq \eqref{eq:gauss_prec_heisen_ngd_aux} for a block Heisenberg structure
\\
    \hline
    
            ${\cal W}_p(\vLat|\vS,n)$   (App. \ref{app:wishart} )
& Wishart with precision
  & See Eq \eqref{eq:wishar_ngd_aux_app}
    \\
    \hline

   $ \mgauss(\vLat|\vE,\vS_U^{-1},\vS_V^{-1})$ (App. \ref{app:mat_gauss} )
& Matrix Gaussian with Kronecker 
  & See Eq \eqref{eq:mat_gauss_exp_ngd_aux}
    \\   
    & structure in precision form
    &
    \\
 
    \hline
$\frac{1}{K} \sum_{k=1}^{K} \gauss(\vlat|\vmu_k, \vS_k^{-1})$    (App. \ref{app:mog} )
& Gaussian Mixture with precision
      &
      See Eq \eqref{eq:mog_prec_exp_updates}
    \\
    \hline
    
$B(\lat) \exp\big(\myang{\vT(\lat), \gparam } - A(\gparam)\big)$    (App. \ref{app:uni_ef} )
& Univariate Exponential Family
      &
      See Eq \eqref{eq:uef_ngd}
    \\
    \hline

\end{tabular}
\end{minipage}
  \caption{ Summary of our updates. See Table \ref{tab:examples} for the parameterizations used in our updates. }
  \vspace{-0.2cm}
 \label{tab:updates}
\end{table*}

\begin{table*}[ht]
\begin{minipage}{\columnwidth}
\begin{tabular}{l|l|l|l}
   \hline
$q(\vlat)$
   & global $\gparam$
   & auxiliary  $\aparam$
   & local $\lparam$
    \\
    \hline
$\gauss(\vlat|\vmu,\vSigma)$ (App. \ref{sec:gauss_prec} ) 
  &$\begin{bmatrix} \vmu \\ \vSigma \end{bmatrix}=\psi(\aparam)=\begin{bmatrix} \vmu \\ \vA\vA^T \end{bmatrix}$
   &$\begin{bmatrix} \vmu \\ \vA \end{bmatrix} = \phi_{\aparam_t}(\lparam)=\begin{bmatrix} \vmu_t +\vA_t \vdelta  \\ \vA_t\mathrm{Exp}(\half \vM) \end{bmatrix}$
   &
   $\begin{bmatrix} \vdelta \\ \vM \end{bmatrix}$

    \\

    \hline

$\gauss(\vlat|\vmu,\vS^{-1})$ (App. \ref{sec:gauss_cov} )  
  &$\begin{bmatrix} \vmu \\ \vS \end{bmatrix}=\psi(\aparam)=\begin{bmatrix} \vmu \\ \vB\vB^T \end{bmatrix}$
   &$\begin{bmatrix} \vmu \\ \vB \end{bmatrix} = \phi_{\aparam_t}(\lparam)=\begin{bmatrix} \vmu_t +\vB_t^{-T} \vdelta  \\ \vB_t \vh(\vM) \end{bmatrix}$
   &
   $\begin{bmatrix} \vdelta \\ \vM \end{bmatrix}$

    \\
    \hline
    
            ${\cal W}_p(\vLat|\vS,n)$   (App. \ref{app:wishart} )
  &$\begin{bmatrix} n \\ \vS \end{bmatrix}=\psi(\aparam)=\begin{bmatrix} 2(f(b)+c) \\ 2(f(b)+c)\vB\vB^T   \end{bmatrix}$
   &$\begin{bmatrix} b \\ \vB  \end{bmatrix} = \phi_{\aparam_t}(\lparam)=\begin{bmatrix}b_t+\delta  \\ \vB_t \mathrm{Exp}(\vM)   \end{bmatrix}$
   &
   $\begin{bmatrix} \delta \\ \vM   \end{bmatrix}$

    \\
    & $c=\frac{p-1}{2}, \,\,\, f(b)=\log(1+\exp(b))$
    &&
    \\
    
    \hline
    
      general $q(\vlat|\gparam)$  (App.  \ref{app:general} )
  &$\gparam=\psi(\aparam)=\aparam$
   &$\aparam = \phi_{\aparam_t}(\lparam)=\aparam_t+\lparam$
   &
   $\lparam$

    \\   
    \hline
    
   $ \mgauss(\vLat|\vE,\vS_U^{-1},\vS_V^{-1}) =$
  &$\begin{bmatrix} \vE \\ \vS_V \\ \vS_U \end{bmatrix}=\psi(\aparam)=\begin{bmatrix} \vE \\ \vA\vA^T \\ \vB\vB^T \end{bmatrix}$
   &$\begin{bmatrix} \vE \\ \vA \\ \vB \end{bmatrix} = \phi_{\aparam_t}(\lparam)=\begin{bmatrix}  \vE_t + \vB_t^{-T} \vDelta \vA_t^{-1} \\ \vA_t \vh(\vM) \\ \vB_t \vh(\vN) \end{bmatrix}$
   &
   $\begin{bmatrix} \vDelta \\ \vM \\ \vN \end{bmatrix}$

    \\   
    $\gauss(\textrm{vec}(\vLat)| \textrm{vec}(\vE), \vS_V^{-1} \otimes \vS_U^{-1})$  
&&&
    \\
   Kronecker structure
(App. \ref{app:mat_gauss} ) 
&&&
\\
    \hline
    
$\frac{1}{K} \sum_{k=1}^{K} \gauss(\vlat|\vmu_k, \vS_k^{-1})$  
    & $\gparam= \begin{bmatrix} \vmu_k \\ \vS_k \end{bmatrix}_{k=1}^{K}$, $\psi(\aparam) = \{\psi_k(\aparam_k)\}_{k=1}^{K}$
    & $\aparam= \begin{bmatrix} \vmu_k \\ \vB_k \end{bmatrix}_{k=1}^{K}$ , $\phi_{\aparam_t}(\lparam) = \{\phi_{k,\aparam_t}(\lparam_k)\}_{k=1}^{K}$
    &  $\begin{bmatrix} \vdelta_k \\ \vM_k \end{bmatrix}_{k=1}^{K}$
    \\
    (App. \ref{app:mog} )
      &$\begin{bmatrix} \vmu_k \\ \vS_k \end{bmatrix}= \psi_k(\aparam_k)  =\begin{bmatrix} \vmu_k \\ \vB_k\vB_k^T \end{bmatrix}$
   &$\begin{bmatrix} \vmu_k \\ \vB_k \end{bmatrix} = \phi_{k,\aparam_t}(\lparam_k)=\begin{bmatrix} \vmu_{k,t} +\vB_{k,t}^{-T} \vdelta_k  \\ \vB_{k,t}\vh( \vM_k) \end{bmatrix}$
   &

    \\
    \hline

      univariate EF $q(\lat|\gparam)$  (App.  \ref{app:uni_ef} )
  &$\gparam=\psi(\aparam)=f(\aparam)$
   &$\aparam = \phi_{\aparam_t}(\lparam)=\aparam_t+\lparam$
   &
   $\lparam$
 \\
   
   $B(\lat) \exp\big(\myang{\vT(\lat), \gparam } - A(\gparam)\big)$
   &$f(\aparam) = \log(1+\exp(\aparam))$ 
   &&

    \\   
    \hline

\end{tabular}
\end{minipage}
  \caption{ Summary of the parameterizations  }
  \vspace{-0.2cm}
 \label{tab:examples}
\end{table*}

\section{More Results}
\label{app:more_results}

\begin{figure*}[t]
	\centering
	\hspace*{-1.2cm}
	\includegraphics[width=0.5\linewidth]{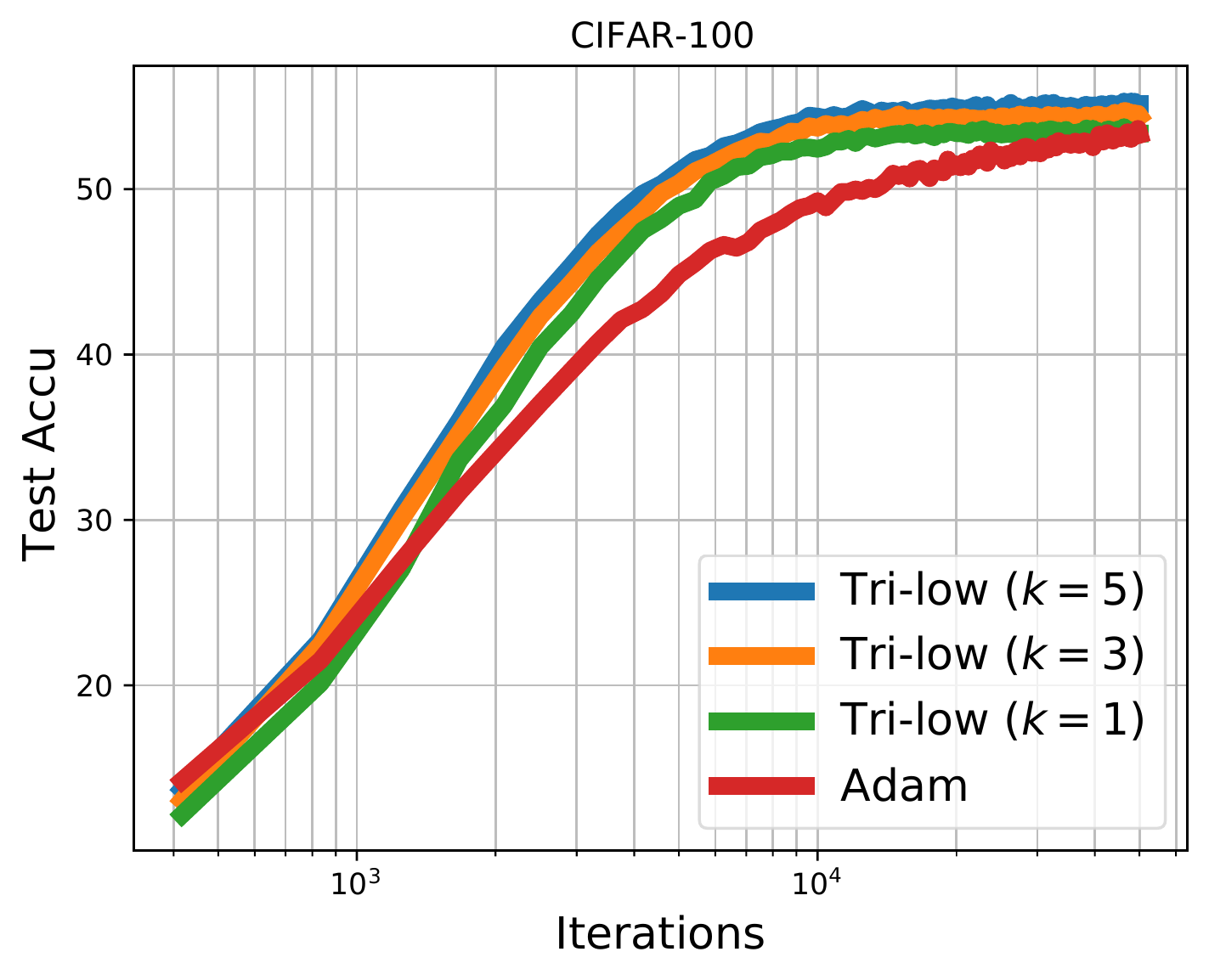}
	\includegraphics[width=0.5\linewidth]{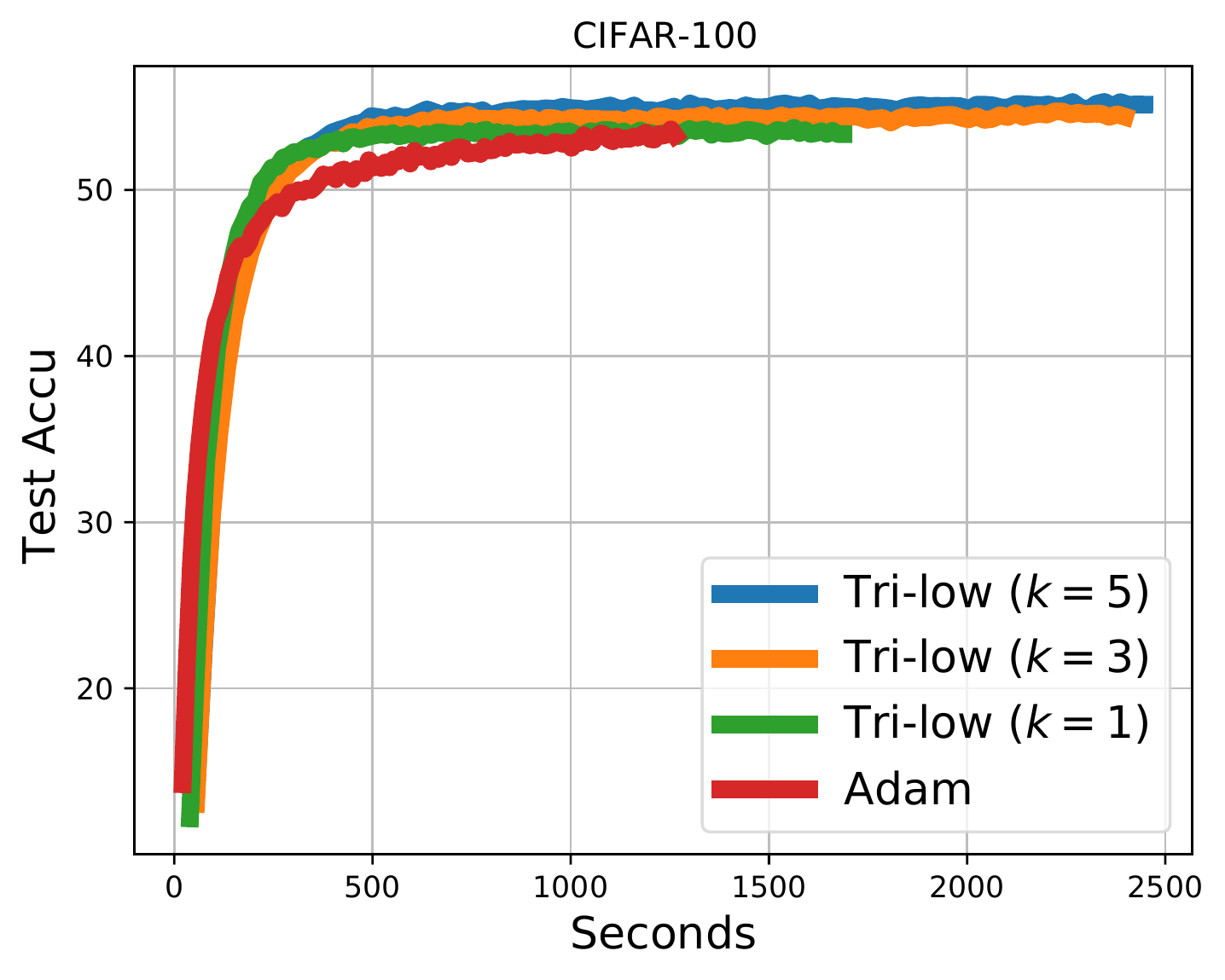}
	\hspace*{-1.2cm}
	\vspace{-0.2cm}
	\caption{
	   The performances of our updates for optimization of a CNN model on CIFAR-100 using layer-wise matrix Gaussian with low-rank structures in  a Kronecker-precision form, where our updates ($O(k |\vlat|)$) have a linear iteration cost like Adam ($O(|\vlat|)$) in terms of time.
For dataset ``CIFAR-100'', we train the model with mini-batch size 120.
Our updates achieve higher test accuracy ($55.2\%$ on ``CIFAR-100'') than Adam ($53.3\%$ on ``CIFAR-100'').
	}
	\label{figure:dnn2}
\end{figure*}

\begin{figure}[H]
\captionsetup[subfigure]{aboveskip=-1pt,belowskip=-1pt}
        \centering
\hspace*{-1.5cm}
        \begin{subfigure}[b]{0.51\textwidth}
  \includegraphics[width=\textwidth]{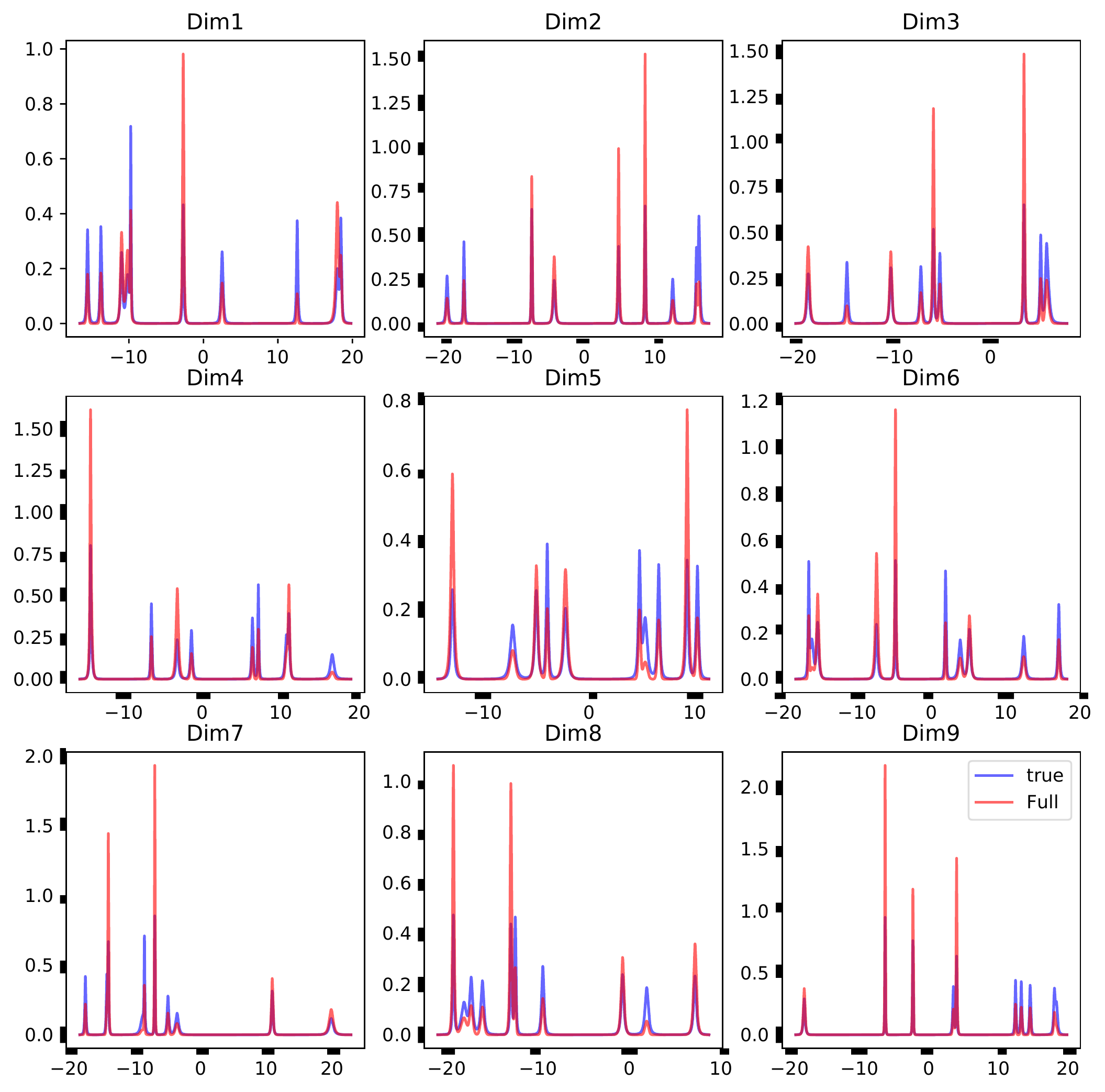}
                \caption{}
\label{fig:mix_a}
        \end{subfigure}       
	\hspace*{-0.4cm}
        \begin{subfigure}[b]{0.51\textwidth}
  \includegraphics[width=\textwidth]{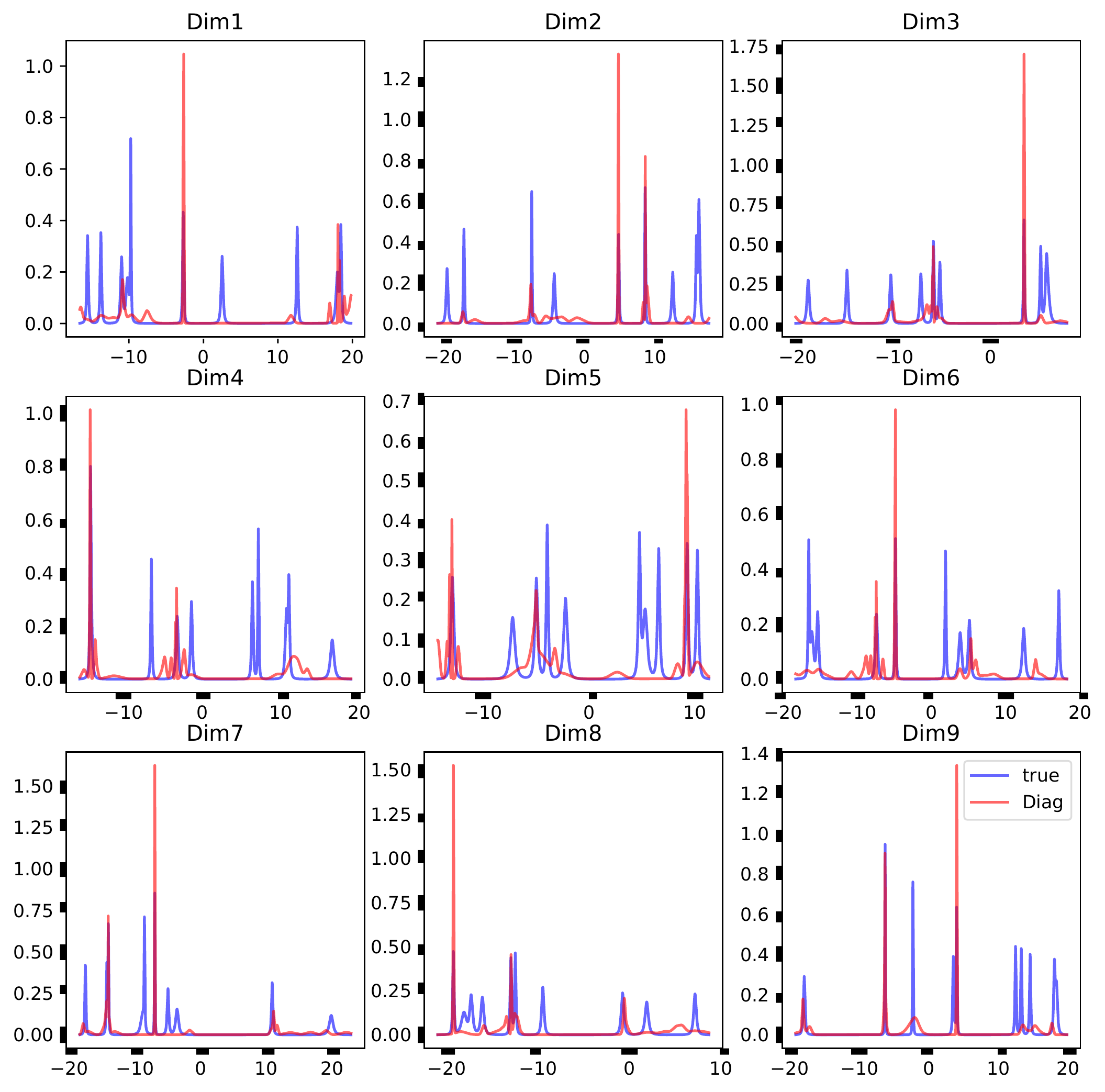}
                \caption{}
\label{fig:mix_b}
        \end{subfigure}
        	\hspace*{-0.4cm}
\vspace{-0.4cm}
   \caption{ 
      Comparison results of structured Gaussian mixtures to fit a 80-Dim mixture of Student's t distributions with 10 components.
 The first 9 marginal dimensions obtained by our updates is shown in the figure, where
 we consider the full covariance structure and the diagonal structure.
 }
 \label{fig:mixap1}
\end{figure}

\begin{figure}[H]
\captionsetup[subfigure]{aboveskip=-1pt,belowskip=-1pt}
        \centering
\hspace*{-1.5cm}
        \begin{subfigure}[b]{0.51\textwidth}
  \includegraphics[width=\textwidth]{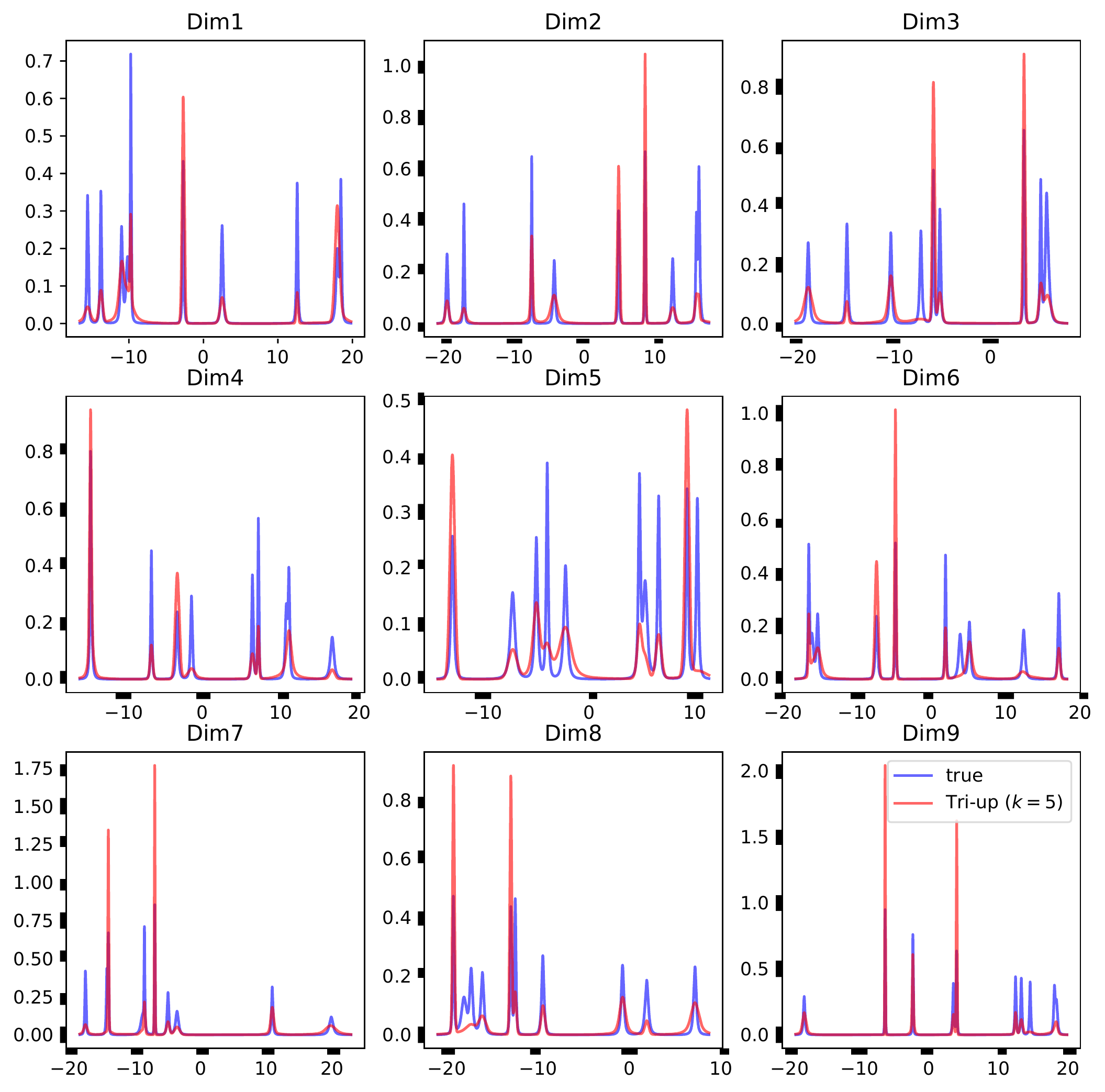}
                \caption{}
\label{fig:mix_c}
        \end{subfigure}       
	\hspace*{-0.4cm}
        \begin{subfigure}[b]{0.51\textwidth}
  \includegraphics[width=\textwidth]{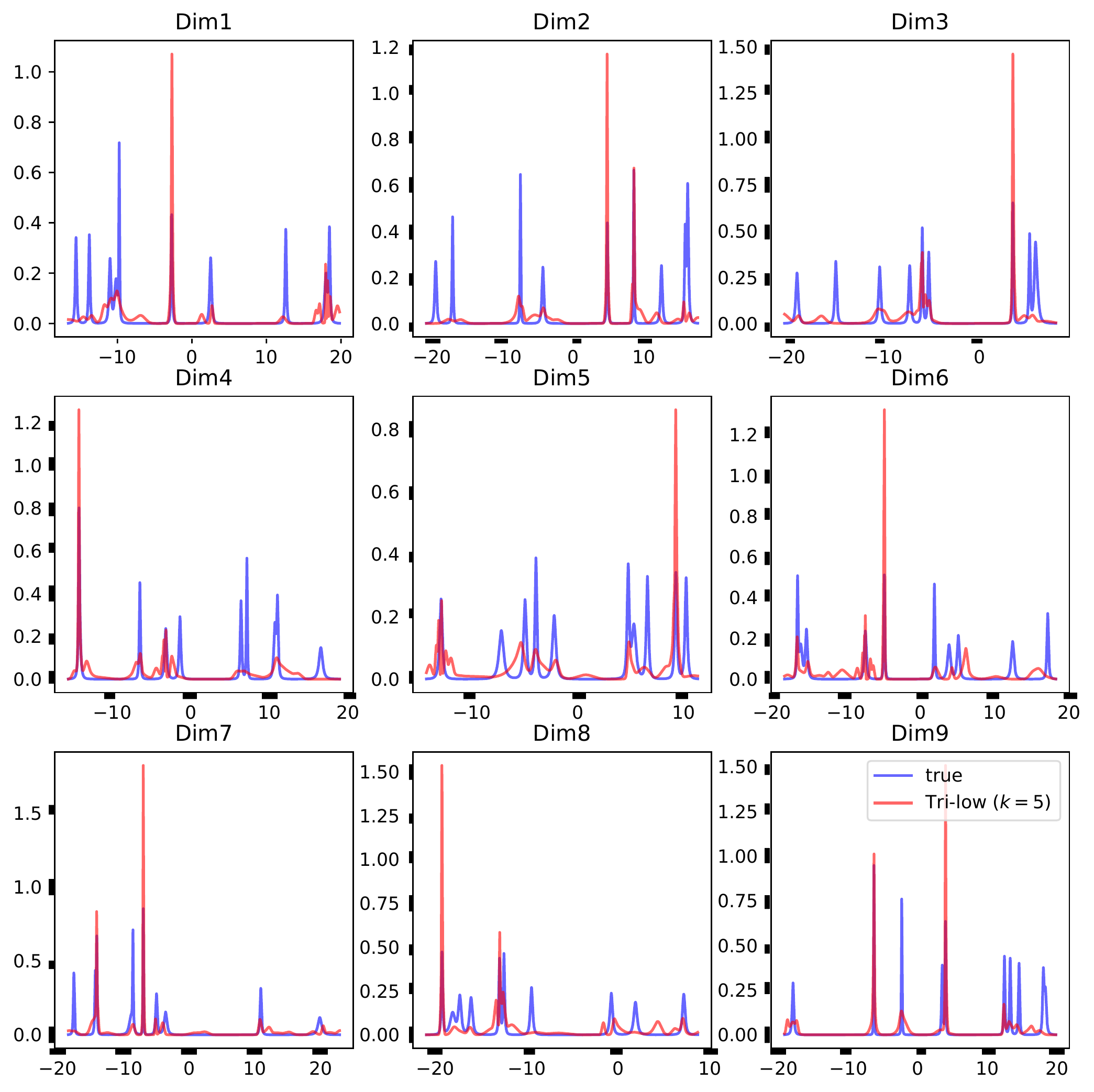}
                \caption{}
\label{fig:mix_d}
        \end{subfigure}
        	\hspace*{-0.4cm}
\vspace{-0.4cm}
    \caption{ 
      Comparison results of structured Gaussian mixtures to fit a 80-Dim mixture of Student's t distributions with 10 components.
 The first 9 marginal dimensions obtained by our updates is shown in the figure, where
 we consider the upper triangular structure  and the lower triangular structure in the precision form.
 The upper triangular structure performs comparably to the full covariance structure with lower computational cost. 
 }
 \label{fig:mixap2}
\end{figure}

\begin{figure}[H]
\captionsetup[subfigure]{aboveskip=-1pt,belowskip=-1pt}
        \centering
\hspace*{-1.5cm}
        \begin{subfigure}[b]{0.51\textwidth}
  \includegraphics[width=\textwidth]{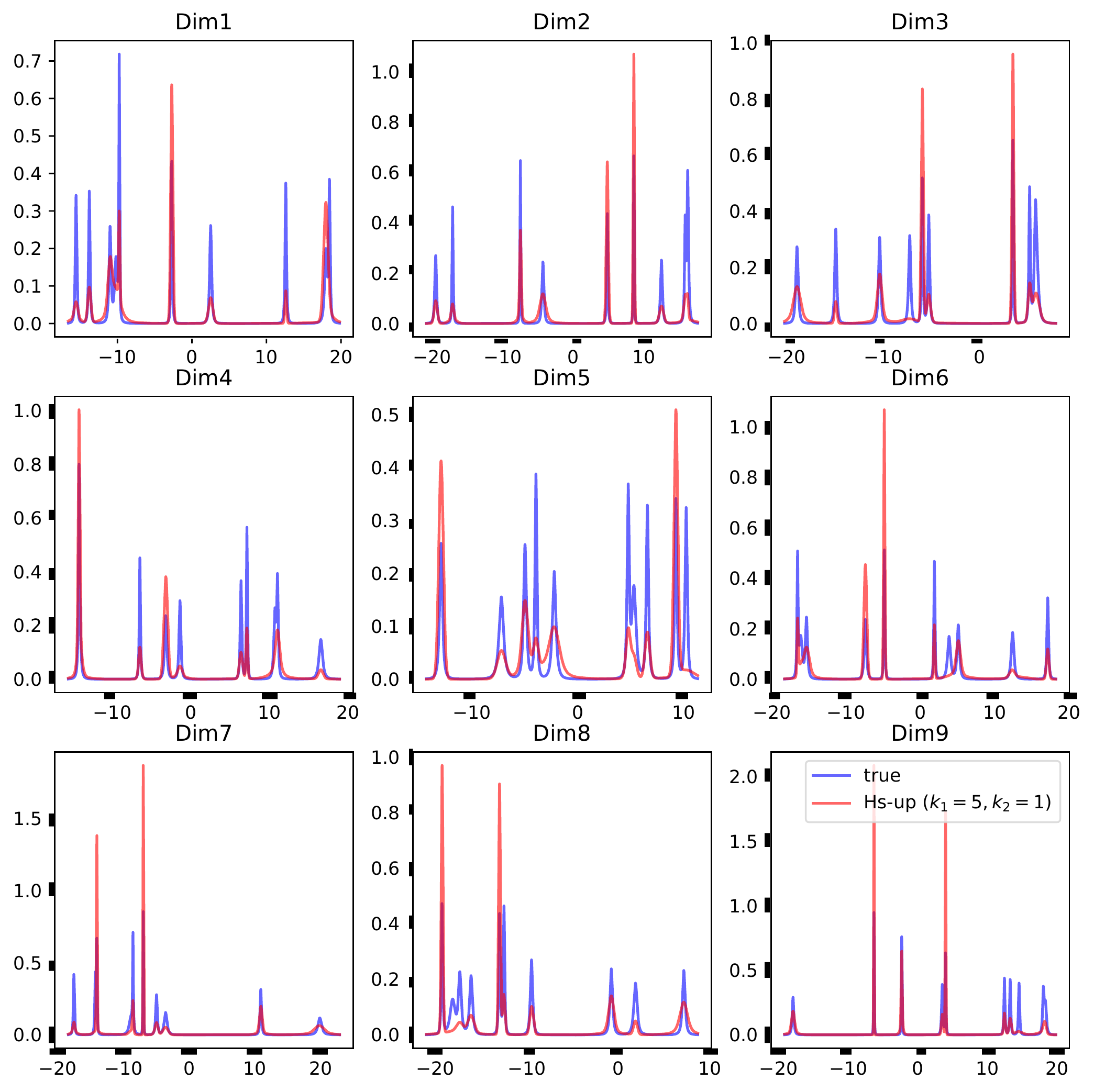}
                \caption{}
\label{fig:mix_e}
        \end{subfigure}       
	\hspace*{-0.4cm}
        \begin{subfigure}[b]{0.51\textwidth}
  \includegraphics[width=\textwidth]{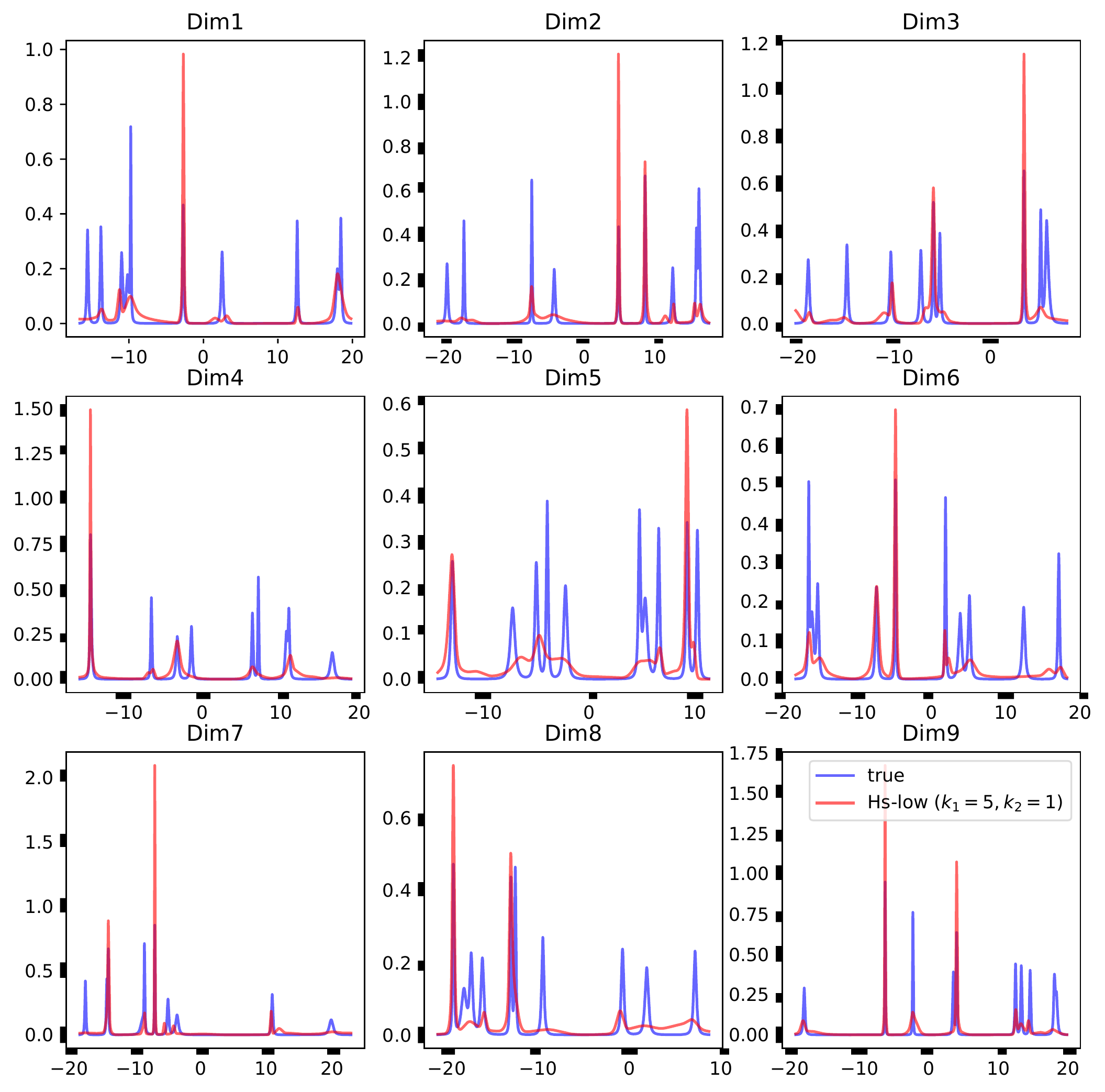}
                \caption{}
\label{fig:mix_f}
        \end{subfigure}
        	\hspace*{-0.4cm}
\vspace{-0.4cm}
   \caption{ 
      Comparison results of structured Gaussian mixtures to fit a 80-Dim mixture of Student's t distributions with 10 components.
 The first 9 marginal dimensions obtained by our updates is shown in the figure, where
 we consider the upper   Heisenberg structure  and the  lower Heisenberg structure in the precision form.
 The upper triangular structure performs comparably to the full covariance structure with lower computational cost. 
 }
 \label{fig:mixap3}
\end{figure}

\section{Fisher information matrix and Some Useful Lemmas}
\label{app:FIM}

The Fisher information matrix (FIM) $\vF_{\gparam}(\gparam)$ of a parametric family of probability distributions $\{q_{\gparam}\}$ is expressed by $\vF_{\gparam}(\gparam)=\mathrm{Cov}_{q_{\gparam}}(\nabla_{\gparam}\log q_{\gparam}(\vlat),\nabla_{\gparam}\log q_{\gparam}(\vlat))$.
Under mild regularity conditions (i.e., expectation of the score is zero and interchange of integrals with gradient operators), we have $\vF_{\gparam}(\gparam)=\Unmyexpect{q_{\gparam}}\sqr{\nabla_{\gparam}\log q_{\gparam}(\vlat) (\nabla_{\gparam}\log q_{\gparam}(\vlat))^\top} =-\Unmyexpect{q_{\gparam}}[\nabla_{\gparam}^2\log q_{\gparam}(\vlat)]$. 

\begin{lemma}
\label{lemma:eq4}
In a general case, Eq \eqref{eq:problem} can be expressed as:
\begin{align*}
 {\cal L}(\gparam) :=\myexpect_{q(\text{\vlat}|\gparam)} \sqr{ \ell(\vlat)} - \gamma\entropy (q(\vlat|\gparam))
\end{align*}
We have the following result:
\begin{align*}
\vg_{\gparam_t}:=
\nabla_{\gparam} {\cal L}(\gparam)  \Big|_{\gparam=\gparam_t}= 
 \nabla_{\gparam} \myexpect_{q(\text{\vlat}|\gparam)} \sqr{ \ell(\vlat) + \gamma \log q(\vlat| {\color{red} \gparam_t })  }  \Big|_{\gparam=\gparam_t}
\end{align*}
Therefore, we could re-define $\ell(\vlat)$ to include $\gamma \log q(\vlat|\gparam_t)$ when we compute gradient $\nabla_{\gparam} {\cal L}(\gparam) \Big|_{\gparam=\gparam_t}$, where $\gparam_t$ highlighted in red is considered as a constant.
\end{lemma}

The following lemma gives us an indirect approach to compute natural gradients. See Appendix  
\ref{app:uni_ef} for the indirect approach and Appendix \ref{app:indirect_limit} for its limitation.

\begin{lemma}
\label{lemma:uef_scalar} 
(Indirect Natural-gradient Computation)
If $\gparam=\vpsi \circ \vphi_{\aparam_t}(\lparam)$ is $C^1$-smooth w.r.t. $\lparam$, we have the following (covariant) transformation\footnote{This is the component transform for a type $(0,2)$-tensor in Riemannian geometry.}.
\begin{align*}
\vF_{\lparam}(\lparam_0) = \big[ \nabla_{\lparam} \gparam   \big] \big[ \vF_{\gparam}(\gparam_t) \big]  \big[ \nabla_{\lparam} \gparam  \big]^T \Big|_{\lparam=\lparam_0}
\end{align*} where we use a layout so that $\nabla_{\slparam_i} \gparam$ and $\nabla_{\lparam} \sgparam_j$ are a row vector and a column vector\footnote{We assume $\lparam$ and $\gparam$ are vectors. For a matrix parameter, we could use the vector representation of the matrix via $\text{vec}(\cdot)$. }, respectively.

If $\vngrad_{\gparam_t}$  is easy to compute\footnote{$\gparam_t$ may stay in a constrained parameter space}, 
the natural gradient $\vngrad_{\lparam_0}$ can be computed via the following (contravariant) transformation\footnote{This is the component transform for a type $(1,0)$-tensor in Riemannian geometry.}, where we assume 
$\vF_{\gparam}(\gparam_t)$  and 
the Jacobian $\big[ \nabla_{\lparam} \gparam  \big]\Big|_{\lparam=\lparam_0}$ are both  non-singular
\begin{align}
\vngrad_{\lparam_0} = \big[ \nabla_{\gparam} \lparam \big]^T  \vngrad_{\gparam_t} \Big|_{\gparam=\gparam_t}  
=\big[ \nabla_{\lparam} \gparam \big]^{-T}  \vngrad_{\gparam_t} \Big|_{\lparam=\lparam_0}  \label{eq:uni_ef_nd}
\end{align} 
where the $j$-th entry of the natural gradient $\vngrad_{{\slparam_0}}$ can be re-expressed as  $\ngrad_{{\slparam_0}_j} = \sum_i  \big[\nabla_{\sgparam_i} \slparam_j \big] \ngrad_{{\sgparam_t}_i}$ when the Jacobian is invertible.

Therefore, $\vngrad_{\lparam_0}$ can be  computed via a Jacobian-vector product used in forward-mode differentiation if
$\vngrad_{\gparam_t}$ is computed beforehand and the Jacobian is invertible.

\end{lemma}

We will use
the following lemmas to show that $\vh(\cdot)$ can replace the matrix exponential map used in the main text while still keeping the natural-gradient computation tractable.

\begin{lemma}
\label{lemma:eq1}
Let $\vh(\vM) = \vI + \vM + \half \vM^2 $.
If the matrix determinant  $|\vh (\vM) | >0$, 
we have the identity:   
\begin{align*}
 \nabla_{M}  \log |\vh (\vM) |  = \vI  + C(\vM), 
\end{align*}
where  $\nabla_{M_{ij}} C(\vM)\Big|_{M=0}=\mathbf{0}$ and $M_{ij}$ is the entry of $\vM$ at position $(i,j)$.
\end{lemma}

\begin{lemma}
\label{lemma:eq2}
Let $ \mathrm{Exp}(\vM) := \vI + \sum_{k=1}^{\infty} \frac{\vM^k}{k!} $.
We have a similar identity as  Lemma \ref{lemma:eq1}:
\begin{align*}
 \nabla_{M}  \log |\mathrm{Exp} (\vM) |  = \vI  + C(\vM),
\end{align*} where
$\nabla_{M_{ij}} C(\vM)\Big|_{M=0}=\mathbf{0}$ and $M_{ij}$ is the entry of $\vM$ at position $(i,j)$.
\end{lemma}
\begin{lemma}
\label{lemma:eq3}
Let $\vf(\vM)=\vh(\vM) $ or $\vf(\vM)=\mathrm{Exp}(\vM)$. 
We have the following expressions:
 \begin{align*}
& \big[ \nabla_{M_{ij}} \vf(\vM) \big] \vf(\vM)^T 
= \big[ (\nabla_{M_{ij}} \vM ) + \half \vM (\nabla_{M_{ij}}\vM) + \half  (\nabla_{M_{ij}} \vM)\vM + (\nabla_{M_{ij}} \vM ) \vM^T   \big] + O(\vM^2) (\nabla_{M_{ij}} \vM ) \\
& \vf(\vM) \big[ \nabla_{M_{ij}} \vf(\vM)^T \big]  
= \big[ (\nabla_{M_{ij}} \vM^T ) + \half \vM^T (\nabla_{M_{ij}}\vM^T) + \half  (\nabla_{M_{ij}} \vM^T)\vM^T + \vM(\nabla_{M_{ij}} \vM^T )    \big] +  O(\vM^2) (\nabla_{M_{ij}} \vM )
\end{align*}

Moreover,
it is obvious that
$\nabla_{M_{kl}} \big[ O(\vM^2) (\nabla_{M_{ij}} \vM )\big]=\mathbf{0}$, where  $M_{kl}$ is the entry of $\vM$ at position $(k,l)$.
\end{lemma}

\subsection{Proof of Lemma \ref{lemma:eq4}}
\begin{myproof}
Since $\entropy (q(\vlat|\gparam)) = -\Unmyexpect{q(\text{\vlat}|\gparam)}\sqr{ \log q(\vlat|\gparam) } $,
we can re-express $\nabla_{\gparam} {\cal L}(\gparam)  \Big|_{\gparam=\gparam_t}$ as 
\begin{align*}
\nabla_{\gparam} {\cal L}(\gparam)  \Big|_{\gparam=\gparam_t}  & =
\nabla_{\gparam} \myexpect_{q(\text{\vlat}|\gparam)} \sqr{ \ell(\vlat) + \gamma \log q(\vlat|\gparam)  }  \Big|_{\gparam=\gparam_t} \\
& = \nabla_{\gparam} \myexpect_{q(\text{\vlat}|\gparam)} \sqr{ \ell(\vlat) + \gamma \log q(\vlat|\gparam_t)  } + \gamma \myexpect_{q(\text{\vlat}|\gparam)} \sqr{ \nabla_{\gparam}  \log q(\vlat|\gparam)  } \Big|_{\gparam=\gparam_t} \,\,\,\, (\text{By the chain rule})
\end{align*}
Note that
\begin{align}
& \myexpect_{q(\text{\vlat}|\gparam)} \sqr{ \nabla_{\gparam}  \log q(\vlat|\gparam)  } \Big|_{\gparam=\gparam_t} \nonumber \\
=& \myexpect_{q(\text{\vlat}|\gparam)} \sqr{ \frac{\nabla_{\gparam}   q(\vlat|\gparam)}{ q(\vlat|\gparam) }  } \Big|_{\gparam=\gparam_t} \nonumber  \\
=& \nabla_{\gparam} \myexpect_{q(\text{\vlat}|\gparam)} \sqr{  1 } \Big|_{\gparam=\gparam_t}  \nonumber\\
=& \mathbf{0} \label{eq:score_fun_grad}
\end{align}
Therefore,
\begin{align*}
\nabla_{\gparam} {\cal L}(\gparam)  \Big|_{\gparam=\gparam_t} & =
 \nabla_{\gparam} \myexpect_{q(\text{\vlat}|\gparam)} \Big[ \ell(\vlat) + \gamma \log q(\vlat| \overbrace{\gparam_t}^{\text{Constant}})  \Big]  \Big|_{\gparam=\gparam_t}
\end{align*}

\end{myproof}

\subsection{Proof of Lemma \ref{lemma:uef_scalar} }
\label{app:proof_lemma_indirect}
\begin{myproof}
 Let's consider an entry of the FIM $\vF_{\lparam}(\lparam_0)$ at position $(j,i)$.
\begin{align*}
\underbrace{F_{\slparam_{ji}} (\lparam_0)}_{\text{scalar}} &= 
 \Unmyexpect{q(\text{\vlat}|\lparam)}\big[
 \big[ \nabla_{\slparam_j}  \log q(\vlat| \lparam ) \big] \big[\nabla_{\slparam_i} \log q(\vlat|\lparam)\big] \big] \big|_{\lparam=\lparam_0} \\
 &=
 \Unmyexpect{q(\text{\vlat}|\lparam)}\big[
 \big[   \underbrace{ \nabla_{\slparam_j} \gparam }_{\text{ row vector } }  \underbrace{  \nabla_{\gparam}  \log q(\vlat| \gparam ) }_{\text{column vector }} \big] 
  \big[    \nabla_{\slparam_i} \gparam \nabla_{\gparam}  \log q(\vlat| \gparam ) \big] \big|_{\lparam=\lparam_0} \\
  &=
\underbrace{ \big[ \nabla_{\slparam_j} \gparam \big]}_{\text{row vector}}
 \Unmyexpect{q(\text{\vlat}|\lparam)}\big[
 \big[   \nabla_{\gparam}  \log q(\vlat| \gparam ) \big] \big[\nabla_{\gparam}  \log q(\vlat| \gparam ) \big]^T \big]  \underbrace{ \big[ \nabla_{\slparam_i} \gparam \big]^T }_{\text{column vector}} \big|_{\lparam=\lparam_0} \\
   &=
 \big[ \nabla_{\slparam_j} \gparam \big]
 \Unmyexpect{q(\text{\vlat}|\gparam_t)}\big[
 \big[    \nabla_{\gparam}  \log q(\vlat| \gparam ) \big] \big[\nabla_{\gparam}  \log q(\vlat| \gparam ) \big]^T \big]   \big[ \nabla_{\slparam_i} \gparam \big]^T \big|_{\lparam=\lparam_0} \\
&= \big[ \nabla_{\slparam_j} \gparam \big]  \vF_{\gparam}(\gparam_t)
\big[\nabla_{\slparam_i} \gparam\big]^T \Big|_{\lparam=\lparam_0}
\end{align*}

Therefore, we have
$\vF_{\lparam}(\lparam_0) =  \big[ \nabla_{\lparam} \gparam \big]  \vF_{\gparam}(\gparam_t)
\big[\nabla_{\lparam} \gparam\big]^T \Big|_{\lparam=\lparam_0}$.

The natural gradient $\vngrad_{\lparam_0}$  can be computed as follows.
\begin{align*}
 \vngrad_{\lparam_0} & = \big( \vF_{\lparam}(\lparam_0)\big)^{-1} \vg_{\lparam_0} \Big|_{\lparam=\lparam_0} \\
 &=
 \big[\nabla_{\lparam} \gparam\big]^{-T}
  \big(\vF_{\gparam}(\gparam_t)\big)^{-1}
 \big[ \nabla_{\lparam} \gparam \big]^{-1} \vg_{\lparam_0} \Big|_{\lparam=\lparam_0}  \\
 &=
 \big[\nabla_{\gparam} \lparam\big]^{T}
  \big(\vF_{\gparam}(\gparam_t)\big)^{-1}
 \big[ \nabla_{\gparam} \lparam \big] \vg_{\lparam_0} \Big|_{\lparam=\lparam_0}  \\
 &=
 \big[\nabla_{\gparam} \lparam\big]^{T}
  \big(\vF_{\gparam}(\gparam_t)\big)^{-1}
  \vg_{\gparam_t}\Big|_{\lparam=\lparam_0} \\
 &=
 \big[\nabla_{\gparam} \lparam\big]^{T} \vngrad_{\gparam_t}\Big|_{\gparam=\gparam_t}
\end{align*} where
$\vF_{\gparam}(\gparam_t)$ and
$\nabla_{\lparam} \gparam$ are invertible by the assumption, and
$\gparam_t = \vpsi \circ \vphi_{\aparam_t}(\lparam_0)$.

\end{myproof}

\subsection{Proof of Lemma \ref{lemma:eq1}}
\begin{myproof}
We first consider the entry $M_{ij}$ of $\vM$.
By matrix calculus, 
we have the following expression.
\begin{align*}
& \nabla_{M_{ij}}  \log | \vh(\vM) | \\
=&  \mathrm{Tr} \big( (\vh(\vM))^{-1} \nabla_{M_{ij}} \vh(\vM) \big)  \\
=&  \mathrm{Tr} \big( (\vh(\vM))^{-1} \big[ (\nabla_{M_{ij}}\vM) + \half \vM (\nabla_{M_{ij}}\vM) + \half  (\nabla_{M_{ij}} \vM ) \vM   \big] \big)  \\
=&  \mathrm{Tr} \big( (\vh(\vM))^{-1} \big[ \half (\vI +\vM ) (\nabla_{M_{ij}}\vM) + \half  (\nabla_{M_{ij}}\vM) (\vI+\vM)  \big] \big)  \\
=&  \mathrm{Tr} \big( (\vh(\vM))^{-1} \big[ \half \vh(\vM) (\nabla_{M_{ij}}\vM) + \half  (\nabla_{M_{ij}}\vM) \vh(\vM) - \frac{1}{4}\big[ \vM^2 (\nabla_{M_{ij}}\vM)  + (\nabla_{M_{ij}}\vM) \vM^2\big] \big)  \\
=&  \mathrm{Tr} \big( (\nabla_{M_{ij}}\vM)\big) - \frac{1}{4}\mathrm{Tr} \big( (\vh(\vM))^{-1} \big[ \vM^2 (\nabla_{M_{ij}}\vM)  + (\nabla_{M_{ij}}\vM) \vM^2\big]\big)
\end{align*}
Therefore, we can express the gradient in a matrix form.
\begin{align*}
 \nabla_{M}  \log | \vh(\vM) | = \vI -\frac{1}{4} \big( \vM^{2} \big)^{T} \vh(\vM)^{-T}-\frac{1}{4} \vh(\vM)^{-T} \big( \vM^{2} \big)^{T} 
\end{align*}

We will show $-\frac{1}{4} \big( \vM^{2} \big)^{T} \vh(\vM)^{-T}-\frac{1}{4} \vh(\vM)^{-T} \big( \vM^{2} \big)^{T}$ is a $ C(\vM) $ function defined in our claim.
We first show that
\begin{align*}
 \nabla_{M_{ij}} \big[  \vM^{2}  \vh(\vM)^{-1} \big] \Big|_{M=0} =\mathbf{0}
\end{align*}

By the product rule, we have
\begin{align*}
& \nabla_{M_{ij}} \big[  \vM^{2}  \vh(\vM)^{-1} \big] \Big|_{M=0} \\
=&  \big[\nabla_{M_{ij}}  \vM\big] \underbrace{ \vM }_{=\mathbf{0}}  \vh(\vM)^{-1}  \Big|_{M=0} + \underbrace{ \vM}_{=\mathbf{0}} \big[\nabla_{M_{ij}}  \vM\big]   \vh(\vM)^{-1} \Big|_{M=0} + \underbrace{\vM^2}_{=\mathbf{0}}\big[\nabla_{M_{ij}}  \vh(\vM)^{-1} \big]  \Big|_{M=0} = \mathbf{0}
\end{align*}

Similarly, we can show

\begin{align*}
  \nabla_{M_{ij}} \big[ \big( \vM^{2} \big)^{T} \vh(\vM)^{-T} \big] \Big|_{M=0} = \mathbf{0}; \,\,\, \,\,\, \nabla_{M_{ij}} \big[ \vh(\vM)^{-T} \big( \vM^{2} \big)^{T} \big] \Big|_{M=0} = \mathbf{0}
\end{align*}

Finally, we obtain the result as
$ \nabla_{M}  \log |\vh (\vM) |  = \vI  + C(\vM)$, where $ C(\vM) = -\frac{1}{4} \big( \vM^{2} \big)^{T} \vh(\vM)^{-T}-\frac{1}{4} \vh(\vM)^{-T} \big( \vM^{2} \big)^{T}$
\end{myproof}

\subsection{Proof of Lemma \ref{lemma:eq2}}
\begin{myproof}
First of all, 
$ |\mathrm{Exp}(\vM)|>0$
and $(\mathrm{Exp}(\vM))^{-1} =\mathrm{Exp}(-\vM) $.
We consider the following expressions.
\begin{align*}
\mathrm{Exp}(-\vM) &= \vI -\vM + \underbrace{ O(\vM^2)}_{\text{remaining higher-order terms}} \\
\mathrm{Exp}(\vM) &= \vI +\vM + \half \vM^2 + \underbrace{ O(\vM^3)}_{\text{remaining higher-order terms}} \\
\nabla_{M_{ij}} \mathrm{Exp}(\vM) &= (\nabla_{M_{ij}}\vM) + \half \vM (\nabla_{M_{ij}}\vM) + \half  (\nabla_{M_{ij}} \vM ) \vM + \underbrace{ O(\vM^2)(\nabla_{M_{ij}} \vM )}_{\text{remaining higher-order terms}} 
\end{align*}

By matrix calculus, 
we have the following expression.
\begin{align*}
& \nabla_{M_{ij}}  \log | \mathrm{Exp}(\vM) | \\
=&  \mathrm{Tr} \big(  \mathrm{Exp}(-\vM) \nabla_{M_{ij}} \mathrm{Exp}(\vM) \big)  \\
=&  \mathrm{Tr} \big(  \mathrm{Exp}(-\vM) \big[ (\nabla_{M_{ij}}\vM) + \half \vM (\nabla_{M_{ij}}\vM) + \half  (\nabla_{M_{ij}} \vM ) \vM + O(\vM^2)(\nabla_{M_{ij}} \vM ) \big] \big)  \\
=&  \mathrm{Tr} \big( \mathrm{Exp}(-\vM) \big[ (\nabla_{M_{ij}}\vM) + \half \vM (\nabla_{M_{ij}}\vM) + \half  (\nabla_{M_{ij}} \vM ) \vM  + O(\vM^2)(\nabla_{M_{ij}} \vM )  \big] \big)  \\
=&  \mathrm{Tr} \big( \big( \vI - \vM + O(\vM^2 ) \big) \big[ (\nabla_{M_{ij}}\vM) + \half \vM (\nabla_{M_{ij}}\vM) + \half  (\nabla_{M_{ij}} \vM ) \vM   + O(\vM^2)(\nabla_{M_{ij}} \vM ) \big] \big)  \\
=&  \mathrm{Tr} \big( (\nabla_{M_{ij}}\vM) \big) + \mathrm{Tr} \big(-\half \vM (\nabla_{M_{ij}}\vM) + \half  (\nabla_{M_{ij}}\vM) \vM + O(\vM^2 )(\nabla_{M_{ij}}\vM) \big)
\end{align*}

Therefore, we have
\begin{align*}
 \nabla_{M}  \log | \mathrm{Exp}(\vM) | = \vI -\half \vM^T + \half \vM^T + O(\vM^2) = \vI + O(\vM^2)
\end{align*}
Now, we show that the remaining $O(\vM^2)$ term is a $C(\vM)$ function defined in our claim.
Note that
\begin{align*}
\nabla_{M_{ij}} O(\vM^2)\Big|_{M=0} = \mathrm{Tr} ( \underbrace{ O(\vM)}_{=\mathbf{0}} \big[ \nabla_{M_{ij}} \vM \big] ) \Big|_{M=0}= \mathbf{0}
\end{align*} where 
$O(\vM)$ contains at least the first order term of $\vM$.

Therefore, the remaining $O(\vM^2)$ term is a $C(\vM)$ function.

\end{myproof}

\subsection{Proof of Lemma \ref{lemma:eq3}}
\begin{myproof}

First note that
\begin{align*}
\vf(\vM)^T &= \vI + \vM^T + O(\vM^2) \\
\vf(\vM) &= \vI + \vM + \half \vM^2 + D(\vM^3) \\
\nabla_{M_{ij}} \vf(\vM) &= (\nabla_{M_{ij}} \vM ) + \half \vM (\nabla_{M_{ij}}\vM) + \half  (\nabla_{M_{ij}} \vM)\vM + D(\vM^2) (\nabla_{M_{ij}} \vM)
\end{align*} where $D(\vM^3) = O(\vM^3)$ and $D(\vM^2)= O(\vM^2)$ when $\vf(\vM)=\mathrm{Exp}(\vM)$ while
 $D(\vM^3) = \mathbf{0}$ and $D(\vM^2)= \mathbf{0}$ when $\vf(\vM)=\vh(\vM)$.

 We will show the first identity.
 \begin{align*}
& \big[ \nabla_{M_{ij}} \vf(\vM) \big] \vf(\vM)^T \\
=& \big[ (\nabla_{M_{ij}} \vM ) + \half \vM (\nabla_{M_{ij}}\vM) + \half  (\nabla_{M_{ij}} \vM)\vM + D(\vM^2) (\nabla_{M_{ij}} \vM) \big] \vf(\vM)^T \\
=& \big[ (\nabla_{M_{ij}} \vM ) + \half \vM (\nabla_{M_{ij}}\vM) + \half  (\nabla_{M_{ij}} \vM)\vM + D(\vM^2) (\nabla_{M_{ij}} \vM) \big] \big( \vI + \vM^T + O(\vM^2) \big) \\
=& \big[ (\nabla_{M_{ij}} \vM ) + \half \vM (\nabla_{M_{ij}}\vM) + \half  (\nabla_{M_{ij}} \vM)\vM + (\nabla_{M_{ij}} \vM ) \vM^T   \big] + O(\vM^2)(\nabla_{M_{ij}} \vM),
\end{align*} where  $ \vM (\nabla_{M_{ij}}\vM)  \vM^T, (\nabla_{M_{ij}}\vM) \vM \vM^T \in O(\vM^2)(\nabla_{M_{ij}} \vM)$.

Similarly, we can show the second expression holds.
\end{myproof}

\section{Gaussian Distribution}
\label{app:vec_gauss}

\subsection{Gaussian with square-root precision structure}
\label{sec:gauss_prec}
Let's consider a global parameterization $\gparam=\{\vmu, \vS\}$, where $\vS$ is the precision and $\vmu$ is the mean.
We use the following parameterizations:
\begin{equation*}
    \begin{split}
        \gparam &:= \crl{\vmu \in\real^p, \,\,\, \vS \in \mathcal{S}_{++}^{p\times p} }\,\,\,\\
        \aparam &:= \crl{ \vmu \in\real^p, \,\,\, \vB \in\mathrm{Gl}^{p\times p} } \\
        \lparam &:= \crl{ \vdelta\in\real^p, \,\,\, \vM \in\mathcal{S}^{p\times p}  }.
    \end{split}
\end{equation*} and maps:
\begin{equation*}
    \begin{split}
        \crl{ \begin{array}{c} \vmu \\ \vS \end{array} } &= \vpsi(\aparam) :=  \crl{ \begin{array}{c} \vmu \\ \vB\vB^\top \end{array} } \\
        \crl{ \begin{array}{c} \vmu \\ \vB \end{array} } &= \vphi_{\aparam_t}(\lparam) :=  \crl{ \begin{array}{c} \vmu_t + \vB_t^{-T} \vdelta \\ \vB_t \vh (\vM) \end{array} }.
    \end{split}
\end{equation*}

Under this local parametrization, we can re-expressed the negative logarithm of the Gaussian P.D.F. as below.
\begin{align*}
-\log q(\vlat|\lparam) &=   - \log | \vB_t \vh(\vM) | + \half (   \vmu_t + \vB_t^{-T} \vdelta  - \vlat )^T \vB_t \vh(\vM) \vh(\vM)^T \vB_t^T ( \vmu_t + \vB_t^{-T} \vdelta  - \vlat ) + C
\end{align*} where $C$ is a constant number and $\aparam_t=\{\vmu_t, \vB_t\}$ is the auxiliary parameterization evaluated at iteration $t$.

\begin{lemma}
\label{lemma:fim_bdiag_gauss_prec}
Under this local parametrization $\lparam$,  $\vF_{\lparam}$ is block diagonal with two blocks--the $\vdelta$ block and the $\vM$ block. 
The claim holds even when  $\vM$ is not symmetric.
\end{lemma}

\begin{myproof}
Any cross term of $\vF_{\lparam}$  between these two blocks is zero as shown below.
\begin{align*}
 &-\Unmyexpect{q(\lat|\lparam)}\sqr{  \nabla_{M_{ij}} \nabla_{\delta} \log q(\vlat|\lparam) }   \\
=& \Unmyexpect{q(\lat|\lparam)}\Big[ \nabla_{M_{ij}} \Big(  \vh(\vM)  \vh(\vM)^T \vB_t^T ( \vmu_t + \vB_t^{-T} \vdelta  - \vlat )  \Big)  \Big] \\
=&  \nabla_{M_{ij}}\big(\vh(\vM)\vh(\vM)^T\big) \Big(    \vB_t^T \Unmyexpect{q(\lat|\lparam)}\Big[  \underbrace{( \vmu_t + \vB_t^{-T} \vdelta  - \vlat )}_{=\mathbf{0}}    \Big]   \Big) \\
=& \mathbf{0}
\end{align*} where $\Unmyexpect{q(\lat|\lparam)}\big[  \vlat \big]= \vmu_t + \vB_t^{-T} \vdelta$ and $M_{ij}$ denotes the  element of the matrix $\vM$ at $(i,j)$.
\end{myproof}

\begin{lemma}
 The FIM  w.r.t. block $\vdelta$ denoted by $\vfim_\delta$ is $\vI_\delta$ when we evaluate it at $\lparam_0=\{\vdelta_0, \vM_0\}=\mathbf{0}$.
The claim holds even when  $\vM$ is not symmetric.
\end{lemma}

\begin{myproof}
\begin{align*}
\vfim_\delta(\lparam_0) & = - \Unmyexpect{q(\lat|\lparam)}\sqr{  \nabla_\delta^2  \log q(\vlat|\lparam)  }\Big|_{\lparam=\mathbf{0}} \\
 &=  \Unmyexpect{q(\lat|\lparam)}\sqr{  \nabla_\delta  \Big( \vh(\vM) \vh(\vM)^T \vB_t^T ( \vmu_t + \vB_t^{-T} \vdelta  - \vlat ) \Big)  }\Big|_{\lparam=\mathbf{0}} \\
 &=  \Unmyexpect{q(\lat|\lparam)}\sqr{  \nabla_\delta  \Big(   \vdelta + \vB_t^T ( \vmu_t  - \vlat )    \Big)  } \Big|_{\lparam=\mathbf{0}}\\
 &= \vI_\delta
\end{align*} where we use the fact that $\vh(\vM)=\vI$ when $\vM=\mathbf{0}$ to move from step 2 to step 3.
\end{myproof}

Now, we discuss how to compute the FIM w.r.t. $\vM$, where
the following expressions hold even when  $\vM$ is not symmetric since we deliberately do not make use the symmetric constraint.
The only requirement for $\vM$ is $|\vh(\vM)|>0$  due to Lemma  \ref{lemma:eq1}.

Let $\vZ = \vB_t^T ( \vmu_t + \vB_t^{-T} \vdelta  - \vlat ) ( \vmu_t + \vB_t^{-T} \vdelta  - \vlat )^T \vB_t$. By matrix calculus, we have the following expression.
\begin{align*}
& \half \nabla_{M_{ij}} \big[ ( \vmu_t + \vB_t^{-T} \vdelta  - \vlat )^T \vB_t \vh(\vM) \vh(\vM)^T \vB_t^T ( \vmu_t + \vB_t^{-T} \vdelta  - \vlat ) \big] \\
= & \half \nabla_{M_{ij}} \mathrm{Tr}\big(\vZ \vh(\vM) \vh(\vM)^T \big) \\
=& \half \mathrm{Tr}\big(\vZ  \big[ \nabla_{M_{ij}} \vh(\vM) \big] \vh(\vM)^T + \vZ \vh(\vM) \nabla_{M_{ij}} \big[ \vh(\vM)^T \big] \big) 
\end{align*}

By Lemma  \ref{lemma:eq3},  we obtain a simplified expression.
\begin{align*}
& \half \nabla_{M} \big[ ( \vmu_t + \vB_t^{-T} \vdelta  - \vlat )^T \vB_t \vh(\vM) \vh(\vM)^T \vB_t^T ( \vmu_t + \vB_t^{-T} \vdelta  - \vlat ) \big] \\
=& \half \big[ 2 \vQ +  \vQ \vM^T + \vM^T \vQ  + 2 \vQ \vM \big] +O(\vM^2)  \vZ \\
=&  \vZ +  (\vZ \vM^T + \vM^T \vZ)/2  +  \vZ \vM  +  O(\vM^2) \vZ 
\end{align*} where $\vQ = ( \vZ^T + \vZ )/2 = \vZ$

By Lemma  \ref{lemma:eq1}, 
we can re-express the gradient w.r.t. $\vM$ as
\begin{align}
&- \nabla_{M}  \log q(\vlat|\lparam) = \underbrace{ -\vI - C(\vM) }_{ - \nabla_{M}  \log | \vh(\vM) |}+ \vZ   + (\vZ \vM^T + \vM^T \vZ)/2 + \vZ \vM + O(\vM^2)\vZ
\label{eq:gauss_first_M}
\end{align}

Finally, we have the following lemma to compute the FIM w.r.t. $\vM$ (denoted by $\vfim_M$) evaluated at $\lparam_0=\mathbf{0}$.
\begin{lemma}
\label{lemma:fim_M_gauss_prec}
 $- \Unmyexpect{q(\lat|\lparam)}\sqr{  \nabla_{M_{ij}} \nabla_M \log q(\vlat|\lparam)  }\Big|_{\lparam=\mathbf{0}} 
 =\nabla_{M_{ij}} \Big(  \vM  +  \vM^T   \Big)
 $.
The claim holds even when  $\vM$ is not symmetric as long as $|\vh(\vM)|>0$. 
\end{lemma}
\begin{myproof}
\begin{align}
 &- \Unmyexpect{q(\lat|\lparam)}\sqr{  \nabla_{M_{ij}} \nabla_M \log q(\vlat|\lparam)  }\Big|_{\lparam=\mathbf{0}} \nonumber \\
 =& \Unmyexpect{q(\lat|\lparam)}\sqr{  \nabla_{M_{ij}} \Big( -\vI - C(\vM) + \vZ  + (\vZ \vM^T + \vM^T \vZ)/2 + \vZ \vM  + O(\vM^2)\vZ \Big)} \Big|_{\lparam=\mathbf{0}}\,\,\, (\text{by Eq \ref{eq:gauss_first_M}}) \nonumber \\
 =& \sqr{  \nabla_{M_{ij}} \Big(    ( \vM^T + \vM^T )/2 +  \vM + O(\vM^2)  \Big)} \Big|_{\lparam=\mathbf{0}} - \underbrace{ \nabla_{M_{ij}} C(\vM) \Big|_{\lparam=\mathbf{0}}}_{=\mathbf{0}} \nonumber \\
 =&  \nabla_{M_{ij}} \Big(  \vM  +  \vM^T   \Big) + O(\vM) \Big|_{\lparam=\mathbf{0}} \nonumber \\
 =&  \nabla_{M_{ij}} \Big(  \vM  +  \vM^T   \Big) \label{eq:gauss_fim}
\end{align} where we use the fact that
$ \Unmyexpect{q(\lat|\lparam)}\sqr{  \vZ } = \Unmyexpect{q(\lat|\lparam)}\sqr{ \vB_t^T ( \vmu_t + \vB_t^{-T} \vdelta  - \vlat ) ( \vmu_t + \vB_t^{-T} \vdelta  - \vlat )^T \vB_t  } = \vI$ evaluated at $\lparam=\mathbf{0}$ to move from step 2 to step 3.
\end{myproof}

Now, we discuss the symmetric constraint in $\vM \in {\cal S}^{p \times p}$. The constraint is essential since the FIM can be singular without a proper constraint. 

\subsubsection{Symmetric Constraint ${\cal S}^{p \times p}$ in $\vM$}
\label{app:sym_m}

Instead of directly using the symmetric property of $\vM$ to simplify Eq \eqref{eq:gauss_fim}, we present a general approach so that we can deal with asymmetric $\vM$ discussed in Appendix  \ref{app:group}. The key idea is to decomposition $\vM$ as a sum of special matrices so that the FIM computation is simple. We also numerically verify the following computation of FIM by Auto-Diff.

First of all, we consider a symmetric constraint in $\vM$. We will show that this constraint ensures the FIM is non-singular, which implies that we can use Lemma \ref{lemma:fim_M_gauss_prec} in this case.

\begin{lemma}
When $\vM$ is symmetric, $|\vh(\vM)|>0$.
\end{lemma}

\begin{myproof}
 \begin{align*}
\vh(\vM)  &= \vI + \vM + \half \vM^2 \\
 &= \half ( \vI + (\vI+\vM) (\vI+\vM) ) \\
 &= \half ( \vI + (\vI+\vM) (\vI+\vM)^T ) \,\,\, (\text{ since $\vM$ is symmetric }) \\
 & \succ \mathbf{0}\,\,\, (\text{ positive-definite })
\end{align*}

Therefore, $|\vh(\vM)|>0$.
\end{myproof}

Since $\vM$ is symmetric, we can re-express the matrix $\vM$ as follows.
\begin{align*}
 \vM = \vM_{\text{low}} + \vM_{\text{low}}^T + \vM_{\text{diag}},
\end{align*} where $\vM_{\text{low}}$ contains the lower-triangular half of $\vM$ excluding the diagonal elements, and $\vM_{\text{diag}}$ contains the diagonal entries of $\vM$.
\begin{align*}
 \vM_{\text{low}} =\begin{bmatrix}
                   0 & 0  & \cdots & 0 \\
                   {\color{blue} M_{21} } & 0 & \cdots & 0  \\
                   \cdots & \cdots & \cdots & 0 \\
                   {\color{blue}M_{d1}} & {\color{blue}M_{d2}} & \cdots & 0
                   \end{bmatrix}\,\,\,\,\,
 \vM_{\text{diag}} =\begin{bmatrix}
                   {\color{blue}  M_{11}} & 0  & \cdots & 0 \\
                   0 & {\color{blue}  M_{22}} & \cdots & 0  \\
                   \cdots & \cdots & \cdots & 0 \\
                   0& 0 & \cdots & { \color{blue}  M_{dd}  }
                   \end{bmatrix}
\end{align*}

By Eq. \ref{eq:gauss_first_M} and the chain rule, we have the following expressions, where $i>  j$.
\begin{align*}
&- \nabla_{{M_{\text{low}}}_{ij}}  \log q(\vlat|\lparam) = - \mathrm{Tr} \big( \underbrace{ \big[ \nabla_{{M_{\text{low}}}_{ij}} \vM \big] }_{\vI_{ij} + \vI_{ji}}  \big[ \nabla_{M}  \log q(\vlat|\lparam) \big] \big) \\
&- \nabla_{{M_{\text{diag}}}_{ii}}  \log q(\vlat|\lparam) = - \mathrm{Tr} \big( \underbrace{ \big[ \nabla_{{M_{\text{diag}}}_{ii}} \vM \big] }_{\vI_{ii}}  \big[ \nabla_{M}  \log q(\vlat|\lparam) \big] \big)
\end{align*}

Therefore, we have
\begin{align}
 &- \nabla_{{M_{\text{low}}}}  \log q(\vlat|\lparam) = - \mathrm{Low} \big( \nabla_{M}  \log q(\vlat|\lparam) + \nabla_{M}^T  \log q(\vlat|\lparam)  \big) \label{eq:low_sym_M_gauss_prec} \\
 &- \nabla_{{M_{\text{diag}}}}  \log q(\vlat|\lparam) = - \half \mathrm{Diag} \big( \nabla_{M}  \log q(\vlat|\lparam) + \nabla_{M}^T  \log q(\vlat|\lparam)  \big) 
 = - \mathrm{Diag} \big( \nabla_{M}  \log q(\vlat|\lparam)  \big) \label{eq:diag_sym_M_gauss_prec}
\end{align} where we define the $\mathrm{Diag}(\cdot)$ function that returns a diagonal matrix with the same structure as $\vM_{\text{diag}}$ and 
the $\mathrm{Low}(\cdot)$ function that returns a lower-triangular matrix with the same structure as $\vM_{\text{low}}$.

By Lemma  \ref{lemma:fim_bdiag_gauss_prec}, the FIM $\vF_{\lparam}$ is block-diagonal with two blocks---the $\vdelta$ block and the $\vM$ block.
We have the following lemma for $\vF_{M}$

\begin{lemma}
\label{eq:gauss_fim_sym_prec}
The $\vM$ block of the FIM denoted by $\vF_{M}$ is also block-diagonal with two block--- the diagonal block denoted by non-zero entries in $\vM_{\text{diag}}$, and the lower-triangular block denoted by non-zero entries in $\vM_{\text{low}}$.
\end{lemma}
\begin{myproof}
We will prove this lemma by showing any cross term of the FIM  between the non-zero entries in $\vM_{\text{low}}$ and the non-zero entries in $\vM_{\text{diag}}$ is also zero.

Notice that we only consider non-zero entries in $\vM_{\text{low}}$, which implies that $i> j$ in the following expression. Therefore, any cross term can be expressed as below.
\begin{align*}
 &- \Unmyexpect{q(\lat|\lparam)}\sqr{  \nabla_{{M_{\text{low}}}_{ij}} \nabla_{M_\text{diag}} \log q(\vlat|\lparam)  }\Big|_{\lparam=\mathbf{0}} 
 =- \Unmyexpect{q(\lat|\lparam)}\big[  \nabla_{{M_{\text{low}}}_{ij}}  \mathrm{Diag} \big( \nabla_{M}  \log q(\vlat|\lparam)  \big) \big]\Big|_{\lparam=\mathbf{0}} \,\,\, (\text{   by Eq. \ref{eq:diag_sym_M_gauss_prec} } ) \\
 =&- \Unmyexpect{q(\lat|\lparam)}\big[ \sum_{k,l} \big[ \nabla_{{M_{\text{low}}}_{ij}} M_{kl} \big] \nabla_{M_{kl}}  \mathrm{Diag} \big( \nabla_{M}  \log q(\vlat|\lparam)  \big) \big] \Big|_{\lparam=\mathbf{0}}  \\
 =&- \Unmyexpect{q(\lat|\lparam)}\big[ \underbrace{ \big[ \nabla_{{M_{\text{low}}}_{ij}} M_{ij} \big]}_{=1} \nabla_{M_{ij}}  \mathrm{Diag} \big( \nabla_{M}  \log q(\vlat|\lparam)  \big)
 + \underbrace{ \big[ \nabla_{{M_{\text{low}}}_{ij}} M_{ji} \big]}_{=1} \nabla_{M_{ji}}  \mathrm{Diag} \big( \nabla_{M}  \log q(\vlat|\lparam)  \big)
\big]\Big|_{\lparam=\mathbf{0}}   \\
 =&- \Unmyexpect{q(\lat|\lparam)}\sqr{ \nabla_{M_{ij}}  \mathrm{Diag} \big( \nabla_{M}  \log q(\vlat|\lparam)  \big)
 +  \nabla_{M_{ji}}  \mathrm{Diag} \big( \nabla_{M}  \log q(\vlat|\lparam)  \big)
}\Big|_{\lparam=\mathbf{0}}  \\
 =&-  \mathrm{Diag} \big(\Unmyexpect{q(\lat|\lparam)}\sqr{ \nabla_{M_{ij}}  \nabla_{M}  \log q(\vlat|\lparam)   +  \nabla_{M_{ji}}   \nabla_{M}  \log q(\vlat|\lparam)   } \big)\Big|_{\lparam=\mathbf{0}}   \\
 =&  \mathrm{Diag} \big( \underbrace{ \nabla_{M_{ij}}   (\vM+\vM^T)}_{\vI_{ij} + \vI_{ji}} + \underbrace{ \nabla_{M_{ji}}   (\vM+\vM^T) }_{\vI_{ij} + \vI_{ji}}   \big)  =\mathbf{0}\,\,\, (\text{ by Lemma \ref{lemma:fim_M_gauss_prec}})
\end{align*}  where
${M_{\text{low}}}_{ij}$ denotes the entry of $M_{\text{low}}$ at position $(i,j)$, 
we use $ \vM = \vM_{\text{low}} + \vM_{\text{low}}^T + \vM_{\text{diag}}$ to move from step 2 to step 3,
and  obtain the last step since $i > j$ and $\mathrm{Diag}(\vI_{ij})=\mathbf{0}$

\end{myproof}

To compute the FIM w.r.t a symmetric $\vM$, we can consider the FIM w.r.t. the non-zero entries in both $\vM_{\text{low}}$ and $\vM_{\text{diag}}$ separately due to the block-diagonal structure of the FIM.
Now, we compute the FIM w.r.t. $\vM_{\text{diag}}$ and $\vM_{\text{low}}$.

By the chain rule, we have
\begin{align*}
 &- \Unmyexpect{q(\lat|\lparam)}\sqr{  \nabla_{{M_{\text{diag}}}_{ii}} \nabla_{M_\text{diag}} \log q(\vlat|\lparam)  }\Big|_{\lparam=\mathbf{0}} \\
 =&- \Unmyexpect{q(\lat|\lparam)}\sqr{  \nabla_{{M_{\text{diag}}}_{ii}}  \mathrm{Diag} \big( \nabla_{M}  \log q(\vlat|\lparam)  \big) }\Big|_{\lparam=\mathbf{0}} \,\,\, (\text{   by Eq. \ref{eq:diag_sym_M_gauss_prec} } )  \\
 =&- \Unmyexpect{q(\lat|\lparam)}\sqr{ \sum_{j,k} \big[ \nabla_{{M_{\text{diag}}}_{ii}} M_{jk} \big] \nabla_{M_{jk}}  \mathrm{Diag} \big( \nabla_{M}  \log q(\vlat|\lparam)  \big) }\Big|_{\lparam=\mathbf{0}}  \\
 =&- \Unmyexpect{q(\lat|\lparam)}\sqr{  \big[ \underbrace{ \nabla_{{M_{\text{diag}}}_{ii}} M_{ii} \big] }_{=1} \nabla_{M_{ii}}  \mathrm{Diag} \big( \nabla_{M}  \log q(\vlat|\lparam)  \big) }\Big|_{\lparam=\mathbf{0}}  \\
 =&- \mathrm{Diag}  \big( \Unmyexpect{q(\lat|\lparam)}\sqr{  \nabla_{M_{ii}}  \nabla_{M}  \log q(\vlat|\lparam) } \big)  \Big|_{\lparam=\mathbf{0}} 
\end{align*}

By Lemma \ref{lemma:fim_M_gauss_prec},  the FIM w.r.t. $\vM_{\text{low}}$ is
\begin{align}
 &- \Unmyexpect{q(\lat|\lparam)}\sqr{  \nabla_{{M_{\text{diag}}}_{ii}} \nabla_{M_\text{diag}} \log q(\vlat|\lparam)  }\Big|_{\lparam=\mathbf{0}} 
 =- \mathrm{Diag}  \big( \Unmyexpect{q(\lat|\lparam)}\sqr{  \nabla_{M_{ii}}  \nabla_{M}  \log q(\vlat|\lparam) } \big)  \Big|_{\lparam=\mathbf{0}}  
 = \mathrm{Diag}  \big( \nabla_{M_{ii}} \Big(  \vM  +  \vM^T   \Big) \big) 
 = 2 \mathrm{Diag} ( \vI_{ii} ) \label{eq:gauss_prec_diag_M}
\end{align}

Now, we compute the FIM w.r.t. $\vM_{\text{low}}$.
By the chain rule, we have
\begin{align*}
 &- \Unmyexpect{q(\lat|\lparam)}\sqr{  \nabla_{{M_{\text{low}}}_{ij}} \nabla_{M_\text{low}} \log q(\vlat|\lparam)  }\Big|_{\lparam=\mathbf{0}} \\
 =&- \Unmyexpect{q(\lat|\lparam)}\sqr{  \nabla_{{M_{\text{low}}}_{ij}}  \mathrm{Low} \big( \nabla_{M}  \log q(\vlat|\lparam) +  \nabla_{M}^T  \log q(\vlat|\lparam) \big) }\Big|_{\lparam=\mathbf{0}} \,\,\, (\text{   by Eq. \ref{eq:low_sym_M_gauss_prec} } )
\end{align*}
We will first consider the following term.
\begin{align*}
&- \Unmyexpect{q(\lat|\lparam)}\sqr{  \nabla_{{M_{\text{low}}}_{ij}}  \mathrm{Low} \big( \nabla_{M}  \log q(\vlat|\lparam) \big) }\Big|_{\lparam=\mathbf{0}}  \\
=&- \Unmyexpect{q(\lat|\lparam)}\sqr{ \sum_{k,l} \big[ \nabla_{{M_{\text{low}}}_{ij}} M_{kl} \big] \nabla_{M_{kl}} \mathrm{Low} \big( \nabla_{M}  \log q(\vlat|\lparam) \big) }\Big|_{\lparam=\mathbf{0}} \\
=&- \Unmyexpect{q(\lat|\lparam)}\sqr{  \underbrace{\big[ \nabla_{{M_{\text{low}}}_{ij}} M_{ji} \big]}_{=1} \nabla_{M_{ji}} \mathrm{Low} \big( \nabla_{M}  \log q(\vlat|\lparam) \big) + \underbrace{\big[ \nabla_{{M_{\text{low}}}_{ij}} M_{ij} \big]}_{=1} \nabla_{M_{ij}} \mathrm{Low} \big( \nabla_{M}  \log q(\vlat|\lparam) \big) }\Big|_{\lparam=\mathbf{0}} \\
=&- \mathrm{Low} \big(\Unmyexpect{q(\lat|\lparam)}\sqr{   \nabla_{M_{ji}}  \nabla_{M}  \log q(\vlat|\lparam)  + \nabla_{M_{ij}}  \nabla_{M}  \log q(\vlat|\lparam)  } \big) \Big|_{\lparam=\mathbf{0}} \\
=& \mathrm{Low} \big( \underbrace{ \nabla_{M_{ji}}\big[ \vM + \vM^T \big]}_{=\vI_{ji}+\vI_{ij}} + \underbrace{ \nabla_{M_{ij}}\big[ \vM + \vM^T \big]}_{=\vI_{ij}+\vI_{ji}}  \big) = 2 \vI_{ij}\,\,\, (\text{By Lemma \ref{lemma:fim_M_gauss_prec}} )
\end{align*} where we obtain the last step by Eq 
\ref{eq:gauss_fim} and the fact that $\vM$ is symmetric.

Similarly, we can show 
\begin{align}
- \Unmyexpect{q(\lat|\lparam)}\sqr{  \nabla_{{M_{\text{low}}}_{ij}}  \mathrm{Low} \big( \nabla_{M}^T  \log q(\vlat|\lparam) \big) }\Big|_{\lparam=\mathbf{0}}  
= 2 \vI_{ij} \label{eq:gauss_prec_diag_Mv2}
\end{align}
Therefore,  the FIM w.r.t. $\vM_{\text{low}}$ is
\begin{align}
 - \Unmyexpect{q(\lat|\lparam)}\sqr{  \nabla_{{M_{\text{low}}}_{ij}} \nabla_{M_\text{low}} \log q(\vlat|\lparam)  }\Big|_{\lparam=\mathbf{0}} 
 = 4 \vI_{ij} \label{eq:gauss_prec_low_M}
\end{align}.

Now, we discuss how to compute the Euclidean gradients.
Recall that
\begin{align*}
\vmu & = \vmu_t + \vB_t^{-T} \vdelta \\
\vS &= \vB_t \vh(\vM) \vh(\vM)^T \vB_t^T
\end{align*}

Let ${\cal L}:=  \Unmyexpect{q(\text{\vlat})} \sqr{ \ell(\vlat) } - \gamma\entropy (q(\vlat)) $.
By the chain rule, we have
\begin{align*}
 \nabla_{\delta_i} {\cal L} &= \big[ \nabla_{\delta_i} \vmu \big]^T \nabla_{\mu} {\cal L } + \mathrm{Tr}\big(\overbrace{\big[ \nabla_{\delta_i} \vS \big]}^{=0} \nabla_{S} { \cal L}) \\
 &= \big[ \nabla_{\delta_i} \vdelta \big]^T \vB^{-1} \nabla_{\mu} {\cal L } \\
  \nabla_{M_{ij}} {\cal L} &= \underbrace{\big[ \nabla_{M_{ij}} \vmu \big]^T}_{=0} \nabla_{\mu} {\cal L } + \mathrm{Tr}\big(\big[ \nabla_{M_{ij}} \vS \big] \nabla_{S} {\cal L }\big)  \\
  &= \mathrm{Tr}\big(\big[ \nabla_{M_{ij}} \vS \big] \nabla_{S} {\cal L }\big) \\
  &=- \mathrm{Tr}\big(\big[ \nabla_{M_{ij}} \vS \big] \vSigma \big[\nabla_{\Sigma} {\cal L }\big] \vSigma \big) \\
  &=- \mathrm{Tr}\big(\big[  \vB \big\{ \big[\nabla_{M_{ij}}\vh(\vM) \big] \vh(\vM)^T + \vh(\vM) \big[\nabla_{M_{ij}}\vh(\vM)^T \big] \big\} \vB^T \big] \vSigma \big[\nabla_{\Sigma} {\cal L }\big] \vSigma \big) 
\end{align*} where $\vSigma=\vS^{-1}$ and we use the gradient identity $\nabla_S {\cal L} = -\vSigma \big[ \nabla_\Sigma {\cal L}\big] \vSigma$.

Therefore, when we  evaluate the gradient at $\lparam_0=\{\vdelta_0,\vM_0\} =\mathbf{0}$, we have
\begin{align}
 \nabla_{\delta_i} {\cal L} \big|_{\lparam=0} &= \big[ \nabla_{\delta_i} \vdelta \big]^T \vB_t^{-1} \nabla_{\mu} {\cal L } \nonumber\\
 \nabla_{M_{ij}} {\cal L}\big|_{\lparam=0} &= 
- \mathrm{Tr}\big(\big[  \vB_t \big( \big[\nabla_{M_{ij}}\vh(\vM) \big] \underbrace{\vh(\mathbf{0})^T }_{=\vI}+ \underbrace{ \vh(\mathbf{0})}_{=\vI} \big[\nabla_{M_{ij}}\vh(\vM)^T \big] \big) \vB_t^T \big] \underbrace{\vSigma_t}_{\vB_t^{-T} \vB_t^{-1}} \big[\nabla_{\Sigma} {\cal L }\big] \vSigma_t \big) \nonumber \\
&= - \mathrm{Tr}\big(\big[  \vB_t \big( \big[\nabla_{M_{ij}}\vh(\vM) \big] + \big[\nabla_{M_{ij}}\vh(\vM)^T \big] \big) \vB_t^T \big] \vB_t^{-T} \vB_t^{-1} \big[\nabla_{\Sigma} {\cal L}\big] \vB_t^{-T} \vB_t^{-1}  \big) \nonumber \\
&= - \mathrm{Tr}\big( \big( \big[\nabla_{M_{ij}}\vM  \big] + \big[\nabla_{M_{ij}}\vM^T \big] \big)  \vB_t^{-1} \big[\nabla_{\Sigma} {\cal L }\big] \vB_t^{-T}   \big) \nonumber \\
&= - \mathrm{Tr}\big(  \big[\nabla_{M_{ij}} \big(\vM + \vM^T\big) \big]   \vB_t^{-1} \big[\nabla_{\Sigma} {\cal L }\big] \vB_t^{-T}   \big)
\label{eq:gauss_sym_prec_ng}
\end{align} where note that $\vh(\vM)=\vI+\vM+O(\vM^2)$ and its gradient evaluated at $\lparam=\mathbf{0}$ can be simplified as
\begin{align*}
 \nabla_{M_{ij}}\vh(\vM) \big|_{\lparam=0} = \nabla_{M_{ij}}\vM + \underbrace{ O(\vM)}_{=\mathbf{0}} \big[\nabla_{M_{ij}}\vM\big] \big|_{\lparam=0}
 = \nabla_{M_{ij}}\vM
\end{align*}

Let's denote
$\vG_M =-2\vB_t^{-1} \big[\nabla_{\Sigma} {\cal L}\big] \vB_t^{-T} $.
Therefore, we can show that
\begin{align*}
  \nabla_{M_{\text{diag}}} {\cal L}\big|_{\lparam=0} = \mathrm{Diag} (\vG_M); \,\,\,\,\,\,
   \nabla_{M_{\text{low}}} {\cal L}\big|_{\lparam=0} = \mathrm{Low} \big ( \vG_M + \vG_M^T \big) = 2 \mathrm{Low} (\vG_M)
\end{align*}

The FIM is block-diagonal w.r.t. three blocks, the $\vdelta$ block, the $\vM_{\text{diag}}$ block, and the $\vM_{\text{low}}$ block

Recall that the FIM w.r.t. $\vdelta$, $\vM_{\text{diag}}$ and $\vM_{\text{low}}$ are $\vI$, $2\vI$, $4\vI$, respectively.
The above statement implies that Assumption 1 is satisfied.

The natural gradients w.r.t. $\vM_{\text{diag}}$ and $\vM_{\text{low}}$ are $\half \mathrm{Diag} (\vG_{M})$ and $\half \mathrm{Low} (\vG_M)$.

Therefore, the natural gradients w.r.t. $\vdelta$ and  w.r.t. $\vM$ are
\begin{align}
\vngrad_\delta =  \vB_t^{-1} \nabla_{\mu} {\cal L},\,\,\,\,
\vngrad_M = \half  \vG_M= -\vB_t^{-1} \big[\nabla_{\Sigma} {\cal L}\big] \vB_t^{-T}
\label{eq:gauss_prec_ng_sym}
\end{align}

Now, we show that Assumption 2 is also satisfied.
We will use the inverse function theorem to show this.

Recall that we have shown that Assumption 1 is satisfied by using the lower-triangular half (i.e., $\vM_{\text{low}}$ and $\vM_{\text{diag}}$) of $\vM$ since $\vM$ is symmetric.
Let's consider the vector representation of the non-zero entries of the \emph{lower-triangular} part of $\vM$ denoted by $\vm_{\text{vec}}$.
We consider the following function denoted by $\mathrm{Mat}(\vm_{\text{vec}})$ to obtain $\vM$ given the vector.
It is easy to see that this function is linear and therefore it is $C^1$-smooth w.r.t. $\vm_{\text{vec}}$.
Consider the vector representation of the local parameter $\lparam_{\text{vec}}= \{\vmu, \vm_{\text{vec}}\}$.
Assumption 1 implies that the FIM $\vF_{\lparam_{\text{vec}}} (\mathbf{0})$ is non-singular at $\lparam_0=\mathbf{0}$.

Note that $\vS$ is a symmetric positive-definite matrix and it can be represented by using a (lower-triangular) Cholesky factor $\vL$ such as $\vS=\vL\vL^T$.
We denote the vector representation of the non-zero entries of $\vL$ denoted by $\mathrm{vec}(\vL)$. 
Moreover, the length of $\vm_{\text{vec}}$ is the same as the length $\mathrm{vec}(\vL)$. Indeed, this length is the (effective) degrees of freedom of the local parameter. 

Now, consider a new global parameterization $\gparam_{\text{new}} = \{\vmu, \mathrm{vec}(\vL)\}$ and the new map $\gparam_{\text{new}} = \vpsi_{\text{new}} \circ \vphi_{\aparam_t}(\lparam_{\text{vec}})$.
\begin{align}
\begin{bmatrix}
 \vmu \\
\mathrm{vec}(\vL) 
\end{bmatrix}=
\vpsi_{\text{new}} \circ \vphi_{\aparam_t}
 \big(
\begin{bmatrix}
 \vdelta \\
 \vm_{\text{vec}}
\end{bmatrix}
\big) = 
\begin{bmatrix}
\vmu_t + \vB_t^{-T} \vdelta \\
\mathrm{vec} ( 
\mathrm{Chol}( \vB_t  \vh(\vM) \vh(\vM)^T \vB_t^T ) )
\end{bmatrix}
\label{eq:chol_map}
\end{align} where $\vM=\mathrm{Mat}( \vm_{\text{vec}} )$.

It is obvious that  Jacobian matrix $\nabla_{\lparam_{\text{vec}}} \gparam_{\text{new}}$ is a square matrix.
Moreover, since $\vS=\vL\vL^T$,
this new FIM under this parameterization remains the same, denoted by $\vF_{\lparam_{\text{new}}}(\mathbf{0})$. It is non-singular at $\lparam_{\text{vec}}=\mathbf{0}$ due to Assumption 1.

By Lemma \ref{lemma:uef_scalar}, we know that
\begin{align*}
\vF_{\lparam_{\text{new}}}(\mathbf{0}) = \big[ \nabla_{\lparam_{\text{vec}}} \gparam_{\text{new}}   \big] \big[ \vF_{\gparam_{\text{new}}}(\gparam_{{\text{new}}_t}) \big]  \big[ \nabla_{\lparam_{\text{vec}}} \gparam_{\text{new}}  \big]^T \Big|_{\lparam_{\text{new}}=\mathbf{0}}
\end{align*}

Since $\vF_{\lparam_{\text{new}}}(\mathbf{0})$ is non-singular and the Jacobian matrix $\nabla_{\lparam_{\text{vec}}} \gparam_{\text{new}}$ is a square matrix, 
 the Jacobian matrix is  non-singular at $\lparam_{\text{vec}}=\mathbf{0}$.

Notice that the Cholesky decomposition $\mathrm{Chol}(\vX)$ is $C^1$-smooth w.r.t. $\vX$.
The smoothness of the Cholesky decomposition is used  by \citet{sun2009efficient,salimbeni2018natural}.
We can see that this map  $\gparam_{\text{new}} = \vpsi_{\text{new}} \circ \vphi_{\aparam_t}(\lparam_{\text{vec}})$ is $C^1$-smooth w.r.t. $\lparam_{\text{vec}}$.

By the inverse function theorem, we know that
 there exist a (local) inverse function of
$ \{\vmu, \mathrm{vec}(\vL) \}= \vpsi_{\text{new}} \circ \vphi_{\aparam_t}(\lparam_{\text{vec}})$
at an open neighborhood of $\lparam_{\text{vec}}=\mathbf{0}$, which is also  $C^1$-smooth.

Since $\vS = \vL\vL^T$, we know that $\gparam=\{\vmu,\vS\}$  and $\lparam=\{\vdelta,\vM\}$ are  
locally $C^1$-diffeomorphic  at an open neighborhood of $\lparam_0$.

\subsubsection{Connection to Newton's Method}
\label{app:gauss_newton}

In Eq \eqref{eq:problem}, we consider the following problem.
\begin{align*}
   \min_{q(\text{\vlat})\in\mathcal{Q}} \Unmyexpect{q(\text{\vlat})} \sqr{ \ell(\vlat) } - \gamma\entropy (q(\vlat)) 
\end{align*} 
Note that we assume $\gamma=0$ in Eq \eqref{eq:problem_tau} for simplicity.

By Eq \eqref{eq:gauss_prec_ng_sym},
our update in the auxiliary parameter space with step-size $\beta$ is
\begin{align}
\vmu_{t+1} & \leftarrow \vmu_{t} + \vB_t^{-T} (-\beta) \vB_t^{-1} \vg_\mu  = \vmu_t -\beta \overbrace{\vB_t^{-T} \vB_t^{-1}}^{\vS_t^{-1}} \vg_\mu \nonumber \\
\vB_{t+1} & \leftarrow \vB_t \vh( \beta \vB_t^{-1} \big[\vg_\Sigma  \big] \vB_t^{-T} ) \label{eq:gauss_prec_exp_updates}
\end{align}

When $\gamma\geq 0$, due to Stein's identities,  we have
\begin{align*}
\vg_\mu  = \Unmyexpect{q(\text{\vlat}|\mu,\Sigma)} \sqr{ \nabla_\lat \ell(\vlat) }, \,\,\,\,\,
\vg_\Sigma  = \half\big( \Unmyexpect{q(\text{\vlat}|\mu,\Sigma)} \sqr{ \nabla_\lat^2 \ell(\vlat) } -\gamma \vSigma^{-1} \big)
\end{align*}

Let $\vG_t =\Unmyexpect{q}{ \sqr{ \nabla_\lat^2 \ell( \vlat) } } - \gamma \vSigma_t^{-1} = \Unmyexpect{q}{ \sqr{ \nabla_\lat^2 \ell( \vlat) } } - \gamma \vS_t$

Therefore, our update in $\vS$ is
\begin{align}
 \vS_{t+1} &= \vB_{t+1} \vB_{t+1}^T = \vB_t \vh( \beta \vB_t^{-1} \big[\vg_{\Sigma_t} \big] \vB_t^{-T} ) \vh( \beta \vB_t^{-1} \big[\vg_{\Sigma_t} \big] \vB_t^{-T} )^T \vB_t^T \nonumber \\
 &= \vB_t \big[ \vI + 2 \big( \beta \vB_t^{-1} \big[\vg_{\Sigma_t} \big] \vB_t^{-T} \big) + 2 \big( \beta \vB_t^{-1} \big[\vg_{\Sigma_t} \vB_t^{-T} \big)^2 + O(\beta^3) \big] \vB_t^T \nonumber \\
 &=\vB_t \big[ \vI + \beta \vB_t^{-1} \vG_t \vB_t^{-T} + \frac{\beta^2}{2}\vB_t^{-1} \vG_t \vB_t^{-T} \vB_t^{-1} \vG_t \vB_t^{-T} + O(\beta^3)  \big] \vB_t^T \nonumber \\
 &= \vS_t + \beta \vG_t + \frac{\beta^2}{2}\vG_t \vS_t^{-1} \vG_t  + O(\beta^3) \label{eq:gauss_prec_at_S}
\end{align} where we use the following result when $\vX$ is symmetric
\begin{align*}
 \vh(\vX)\vh(\vX)^T = \vh(\vX)\vh(\vX)  = (\vI+\vX+\half \vX^2)(\vI+\vX+\half \vX^2)
 = \vI+2\vX+ 2 \vX^2 + O(\vX^3)
\end{align*}

When $\gamma=1$, we obtain the update proposed by \citet{lin2020handling} if we  ignore the $O(\beta^3)$ term.
\begin{align*}
 \vS_{t+1} 
 &= \vS_t + \beta \vG_t + \frac{\beta^2}{2}\vG_t \vS_t^{-1} \vG_t  + O(\beta^3) \\
 &= (1-\beta) \vS_t + \beta \Unmyexpect{q}{ \sqr{ \nabla_\lat^2 \ell( \vlat) } }+ \frac{\beta^2}{2}\vG \vS_t^{-1} \vG_t  + O(\beta^3) 
\end{align*} where  $\vG_t = \Unmyexpect{q}{ \sqr{ \nabla_\lat^2 \ell( \vlat) } } - \vS_t$

\subsubsection{Unconstrained $\vM$ }
\label{app:asym_M}

In Appendix \ref{app:sym_m}, we show that if $\vM$ is symmetric, the FIM $\vF_{\lparam}(\lparam_0)$ is non-singular.
Unfortunately, if $\vM \in \real^{p \times p}$ is unconstrained, the FIM is indeed singular. In this appendix, we consider the square-root case for the precision. It is easy to show that the following result is also true for the square-root case of the covariance discussed in Appendix
\ref{sec:gauss_cov}.

To see why the FIM is indeed singular, we will use the vector representation of $\vM$ as $\vv = \mathrm{vec}(\vM)$.
Let's consider these two entries $M_{ij}$ and $M_{ji}$, where $i\neq j$.
Unlike the symmetric case, $M_{ij}$ and $M_{ji}$ are \emph{distinct} parameters in the unconstrained case.
In our vector representation, we use $v_{k_1}$ and $v_{k_2}$ to uniquely represent
$M_{ij}$ and  $M_{ji}$, respectively, where $k_1 \neq k_2$ since $i \neq j$.

First of all, since $\vv = \mathrm{vec}(\vM)$, we have the following identity.
\begin{align*}
  - \nabla_{\text{\vv}}  \log q(\vlat|\lparam)  = \mathrm{vec} ( - \nabla_{M}  \log q(\vlat|\lparam) )
\end{align*}

Recall that FIM is block-diagonal with two blocks---the $\vdelta$ block and the $\vM$ block.
To show that the FIM is singular, we will show that  the $\vM$ block contains two identical columns/rows.
For simplicity, we will instead show that the FIM w.r.t. $\vv$ contains two identical columns/rows, where $\vv$ is the vector representation of $\vM$.

Let's consider the following row/column of the FIM for the $\vM$ block.
\begin{align*}
&- \Unmyexpect{q(\lat|\lparam)}\sqr{  \nabla_{v_{k_1}}  \big( \nabla_{\text{\vv}}  \log q(\vlat|\lparam) \big) }\Big|_{\lparam=\mathbf{0}}  \\
=&- \Unmyexpect{q(\lat|\lparam)}\sqr{  \nabla_{v_{k_1}}  \mathrm{vec}\big( \nabla_{M}  \log q(\vlat|\lparam) \big) }\Big|_{\lparam=\mathbf{0}}  \\
=&- \Unmyexpect{q(\lat|\lparam)}\sqr{ \sum_{l,m} \big[ \nabla_{v_{k_1}} M_{lm} \big] \nabla_{M_{lm}} \mathrm{vec}\big( \nabla_{M}  \log q(\vlat|\lparam) \big) }\Big|_{\lparam=\mathbf{0}} \\  
=&- \Unmyexpect{q(\lat|\lparam)}\sqr{  \underbrace{\big[ \nabla_{v_{k_1}} M_{ij} \big]}_{=1} \nabla_{M_{ij}} \mathrm{vec}\big( \nabla_{M}  \log q(\vlat|\lparam) \big) }\Big|_{\lparam=\mathbf{0}}   
\end{align*} we obtain the last step since 
$v_{k_1}$  uniquely represents $M_{ij}$.

Similarly, we can show
\begin{align*}
- \Unmyexpect{q(\lat|\lparam)}\sqr{  \nabla_{v_{k_1}}  \big( \nabla_{\text{\vv}}  \log q(\vlat|\lparam) \big) }\Big|_{\lparam=\mathbf{0}}  
=- \Unmyexpect{q(\lat|\lparam)}\sqr{  \underbrace{\big[ \nabla_{v_{k_2}} M_{ji} \big]}_{=1} \nabla_{M_{ji}} \mathrm{vec}\big( \nabla_{M}  \log q(\vlat|\lparam) \big) }\Big|_{\lparam=\mathbf{0}}   
\end{align*}

According to Eq \ref{eq:gauss_fim}, we have
\begin{align*}
&- \Unmyexpect{q(\lat|\lparam)}\sqr{  \nabla_{v_{k_1}}  \big( \nabla_{\text{\vv}}  \log q(\vlat|\lparam) \big) }\Big|_{\lparam=\mathbf{0}}  \\
=&- \Unmyexpect{q(\lat|\lparam)}\sqr{  \underbrace{\big[ \nabla_{v_{k_1}} M_{ij} \big]}_{=1} \nabla_{M_{ij}} \mathrm{vec}\big( \nabla_{M}  \log q(\vlat|\lparam) \big) }\Big|_{\lparam=\mathbf{0}}   \\
=&- \mathrm{vec}\big(\Unmyexpect{q(\lat|\lparam)}\sqr{  \nabla_{M_{ij}}  \nabla_{M}  \log q(\vlat|\lparam)  } \big)\Big|_{\lparam=\mathbf{0}}   \\
=& \mathrm{vec}\big( \nabla_{M_{ij}} \Big(  \vM  +  \vM^T   \Big) \big)  \\
=& \mathrm{vec}\big( \vI_{ij} + \vI_{ji} \big)
\end{align*}
Similarly, we have
\begin{align*}
- \Unmyexpect{q(\lat|\lparam)}\sqr{  \nabla_{v_{k_2}}  \big( \nabla_{v}  \log q(\vlat|\lparam) \big) }\Big|_{\lparam=\mathbf{0}}  
= \mathrm{vec}\big( \nabla_{M_{ji}} \Big(  \vM  +  \vM^T   \Big) \big)  
= \mathrm{vec}\big( \vI_{ji} + \vI_{ij} \big)
\end{align*}

Therefore, the FIM of the $\vM$ block contains two identical columns/rows and it must be singular.

\subsection{Gaussian with square-root covariance structure}
\label{sec:gauss_cov}  
Let's consider a global parameterization $\gparam=\{\vmu, \vSigma\}$, where $\vSigma$ is the covariance and $\vmu$ is the mean.
We use the following Parameterizations:
\begin{equation*}
    \begin{split}
        \gparam &:= \crl{\vmu \in\real^p, \,\,\, \vSigma \in \mathcal{S}_{++}^{p\times p} }\,\,\,\\
        \aparam &:= \crl{ \vmu \in\real^p, \,\,\, \vA \in\mathrm{GL}^{p\times p} } \\
        \lparam &:= \crl{ \vdelta\in\real^p, \,\,\, \vM \in\mathcal{S}^{p\times p}  }.
    \end{split}
\end{equation*} and maps:
\begin{equation*}
    \begin{split}
        \crl{ \begin{array}{c} \vmu \\ \vSigma \end{array} } &= \vpsi(\aparam) :=  \crl{ \begin{array}{c} \vmu \\ \vA\vA^\top \end{array} } \\
        \crl{ \begin{array}{c} \vmu \\ \vA \end{array} } &= \vphi_{\aparam_t}(\lparam) :=  \crl{ \begin{array}{c} \vmu_t + \vA_t \vdelta \\ \vA_t  \mathrm{Exp}(\half \vM) \end{array} }.
    \end{split}
\end{equation*}

Now, we will use the fact that that $\vM$ is symmetric.
Under this local parametrization, we can re-expressed the negative logarithm of the Gaussian P.D.F. as below.
\begin{align*}
-\log q(\vlat|\lparam) &=    \log | \vA_t \mathrm{Exp}(\half \vM) | + \half (   \vmu_t + \vA_t \vdelta  - \vlat )^T \vA_t^{-T}\mathrm{Exp}(-\vM) \vA_t^{-1} ( \vmu_t + \vA_t \vdelta  - \vlat ) + C
\end{align*} where $C$ is a constant number and $\aparam_t=\{\vmu_t, \vA_t\}$ is the auxiliary parameterization evaluated at iteration $t$.

Like Sec \ref{sec:gauss_prec}, we can show the FIM w.r.t. $\lparam$ is block-diagonal w.r.t. two blocks--- the $\vdelta$ block and the $\vM$ block.

Now, we show that the FIM  w.r.t. block $\vdelta$ denoted by $\vfim_\delta$ is $\vI_\delta$ when we evaluate it at $\lparam_0=\{\vdelta_0, \vM_0\}=\mathbf{0}$.
\begin{align*}
\vfim_\delta(\lparam_0) & = - \Unmyexpect{q(\lat|\lparam)}\sqr{  \nabla_\delta^2  \log q(\vlat|\lparam)  }\Big|_{\lparam=\mathbf{0}} \\
 &=  \Unmyexpect{q(\lat|\lparam)}\sqr{  \nabla_\delta  \Big( \mathrm{Exp}(-\vM)  \vA_t^{-1} ( \vmu_t + \vA_t \vdelta  - \vlat ) \Big)  }\Big|_{\lparam=\mathbf{0}} \\
 &=  \Unmyexpect{q(\lat|\lparam)}\sqr{  \nabla_\delta  \Big(   \vdelta + \vA_t^{-1} ( \vmu_t  - \vlat )    \Big)  } \Big|_{\lparam=\mathbf{0}}\\
 &= \vI_\delta
\end{align*} where we use the fact that $\mathrm{Exp}(-\vM)=\vI$ when $\vM=\mathbf{0}$ to move from step 2 to step 3.

Now, we discuss how to compute the FIM w.r.t. $\vM$, where we explicitly use the fact that $\vM$ is symmetric.

Let $\vZ = \vA_t^{-1} ( \vmu_t + \vA_t \vdelta  - \vlat ) ( \vmu_t + \vA_t \vdelta  - \vlat )^T \vA_t^{-T}$. By matrix calculus, we have the following expression.
\begin{align*}
& \half \nabla_{M_{ij}} \big[ ( \vmu_t + \vA_t \vdelta  - \vlat )^T \vA_t^{-T} \mathrm{Exp}(-\vM) \vA_t^{-1} ( \vmu_t + \vA_t \vdelta  - \vlat ) \big] \\
= & \half \nabla_{M_{ij}} \mathrm{Tr}\big(\vZ \mathrm{Exp}(-\vM) \big) \\
=& \half \mathrm{Tr}\big(\vZ \nabla_{M_{ij}} (- \vM + \half \vM^2 + O(\vM^3) ) \big) 
\end{align*}

Therefore, we have
\begin{align*}
 \half \nabla_{M} \big[ ( \vmu_t + \vA_t \vdelta  - \vlat )^T \vA_t^{-T} \mathrm{Exp}(-\vM) \vA_t^{-1} ( \vmu_t + \vA_t \vdelta  - \vlat ) \big]
=- \half \vZ +  \frac{1}{4}(\vZ \vM + \vM\vZ) + O(\vM^2)\vZ
\end{align*}

By Lemma  \ref{lemma:eq2}, 
we can re-express the gradient w.r.t. $\vM$ as
\begin{align}
&- \nabla_{M}  \log q(\vlat|\lparam) = \underbrace{ \half (\vI + C(\vM)) }_{  \nabla_{M}  \log |\half \mathrm{Exp}(\vM) |}
- \half \vZ +  \frac{1}{4}(\vZ \vM + \vM\vZ) + O(\vM^2)\vZ
\label{eq:gauss_cov_first_M}
\end{align}

Finally, we have the following lemma to compute the FIM w.r.t. $\vM$ (denoted by $\vfim_M$) evaluated at $\lparam_0=\mathbf{0}$.
\begin{lemma}
 $- \Unmyexpect{q(\lat|\lparam)}\sqr{  \nabla_{M_{ij}} \nabla_M \log q(\vlat|\lparam)  }\Big|_{\lparam=\mathbf{0}} 
 =\frac{1}{2} \nabla_{M_{ij}}   \vM 
 $.
The claim assumes  $\vM$ is symmetric.
\end{lemma}
\begin{myproof}
\begin{align}
 &- \Unmyexpect{q(\lat|\lparam)}\sqr{  \nabla_{M_{ij}} \nabla_M \log q(\vlat|\lparam)  }\Big|_{\lparam=\mathbf{0}} \nonumber \\
 =& \Unmyexpect{q(\lat|\lparam)}\sqr{  \nabla_{M_{ij}} \Big( \half (\vI + C(\vM)) - \half \vZ +  \frac{1}{4}(\vZ \vM + \vM\vZ) + O(\vM^2)\vZ  \Big)} \Big|_{\lparam=\mathbf{0}}\,\,\, (\text{by Eq \ref{eq:gauss_cov_first_M}}) \nonumber \\
 =& \sqr{  \nabla_{M_{ij}} \Big(    \half \vM + O(\vM^2)  \Big)} \Big|_{\lparam=\mathbf{0}} +\half \underbrace{ \nabla_{M_{ij}} C(\vM) \Big|_{\lparam=\mathbf{0}}}_{=\mathbf{0}} \nonumber \\
 =&  \half \nabla_{M_{ij}} \Big(  \vM     \Big) + O(\vM) \Big|_{\lparam=\mathbf{0}} \nonumber \\
 =& \half \nabla_{M_{ij}} \Big(  \vM    \Big) 
\end{align} where we use the fact that
$ \Unmyexpect{q(\lat|\lparam)}\sqr{  \vZ } = \Unmyexpect{q(\lat|\lparam)}\sqr{ \vA_t^{-1} ( \vmu_t + \vA_t \vdelta  - \vlat ) ( \vmu_t + \vA_t \vdelta  - \vlat )^T \vA_t^{-T}  } = \vI$ evaluated at $\lparam=\mathbf{0}$ to move from step 2 to step 3.
\end{myproof}

Therefore, $\vF_{M}(\lparam_0)=\half \vI_{M}$.

Now, we discuss how to compute the Euclidean gradients.
Recall that
\begin{align*}
\vmu & = \vmu_t + \vA_t \vdelta \\
\vSigma &= \vA_t  \mathrm{Exp}(\vM) \vA_t^T
\end{align*}

Let ${\cal L}:=  \Unmyexpect{q(\text{\vlat})} \sqr{ \ell(\vlat) } - \gamma\entropy (q(\vlat)) $.
By the chain rule, we have
\begin{align*}
 \nabla_{\delta_i} {\cal L} &= \big[ \nabla_{\delta_i} \vmu \big]^T \nabla_{\mu} {\cal L } + \mathrm{Tr}\big(\overbrace{\big[ \nabla_{\delta_i} \vSigma \big]}^{=0} \nabla_{\Sigma} { \cal L}) \\
 &= \big[ \nabla_{\delta_i} \vdelta \big]^T \vA_t^T \nabla_{\mu} {\cal L } \\
  \nabla_{M_{ij}} {\cal L} &= \underbrace{\big[ \nabla_{M_{ij}} \vmu \big]^T}_{=0} \nabla_{\mu} {\cal L } + \mathrm{Tr}\big(\big[ \nabla_{M_{ij}} \vSigma \big] \nabla_{\Sigma} {\cal L }\big)  \\
  &= \mathrm{Tr}\big(\big[ \nabla_{M_{ij}} \vSigma \big] \nabla_{\Sigma} {\cal L }\big) \\
  &= \mathrm{Tr}\big(\vA_t\big[  \nabla_{M_{ij}} \mathrm{Exp}(\vM) \big] \vA_t^T \nabla_{\Sigma} {\cal L }\big)
\end{align*}

Therefore, when we  evaluate the gradient at $\lparam_0=\{\vdelta_0,\vM_0\} =\mathbf{0}$, we have
\begin{align*}
 \nabla_{\delta_i} {\cal L} \big|_{\lparam=0} &= \big[ \nabla_{\delta_i} \vdelta \big]^T \vA_t^{T} \nabla_{\mu} {\cal L }\\
 \nabla_{M_{ij}} {\cal L}\big|_{\lparam=0} &= \mathrm{Tr}\big(\vA_t\big[  \nabla_{M_{ij}} \mathrm{Exp}(\vM) \big] \vA_t^T \nabla_{\Sigma} {\cal L }\big)\big|_{\lparam=0}\\
&= \mathrm{Tr}\big(\vA_t\big[  \nabla_{M_{ij}} \vM \big] \vA_t^T \nabla_{\Sigma} {\cal L }\big)
\end{align*} where note that $\mathrm{Exp}(\vM)=\vI+\vM+O(\vM^2)$ and its gradient evaluated at $\lparam=\mathbf{0}$ can be simplified as
\begin{align*}
 \nabla_{M_{ij}}\mathrm{Exp}(\vM) \big|_{\lparam=0} = \nabla_{M_{ij}}\vM + \underbrace{ O(\vM)}_{=\mathbf{0}} \big[\nabla_{M_{ij}}\vM\big] \big|_{\lparam=0}
 = \nabla_{M_{ij}}\vM
\end{align*}

Therefore,
\begin{align*}
 \nabla_{\delta} {\cal L} \big|_{\lparam=0} &=  \vA_t^{T} \nabla_{\mu} {\cal L }\\
 \nabla_{M_{ij}} {\cal L}\big|_{\lparam=0} &=  \vA_t^T \big[\nabla_{\Sigma} {\cal L} \big]\vA_t
\end{align*}

Recall that the FIM w.r.t. $\vdelta$ and $\vM$ are $\vI$ and $\half\vI$, respectively. In other words,
\begin{align*}
 \vF_{\lparam}(\lparam_0) = \begin{bmatrix}
    \vI_{\delta} & \mathbf{0} \\
     \mathbf{0} &  \half \vI_{M}\\
                            \end{bmatrix},
\end{align*} which implies that Assumption 1 is satisfied.

Therefore, the natural gradient w.r.t. $\vdelta$ is $ \vngrad_\delta = \vA_t^{T} \nabla_{\mu} {\cal L}$.
The natural-gradient w.r.t. $\vM$ as $\vngrad_M = 2 \vA_t^{T} \big[\nabla_{\Sigma} {\cal L}\big] \vA_t$.

Therefore, our update in the auxiliary parameter space is
\begin{align}
    \vmu_{t+1} &\leftarrow \vmu_t - \beta \vS_{t}^{-1} \vg_{\mu}\nonumber  \\
    \vA_{t+1} &\leftarrow \vA_t \mathrm{Exp}\big(- \beta \vA_t^T\vg_{\Sigma} \vA_t \big)
     \label{eq:ngd_aux_param_app}
\end{align} recall that $\vA = \vA_t \mathrm{Exp} \big(-\beta \half \vngrad_M  \big)$.

Now, we show that Assumption 2 is also satisfied.
Since $\{\vmu,\vSigma\}=\gparam = \vpsi \circ \vphi_{\aparam_t}(\{ \vdelta, \vM \}) $, where $\aparam_t=\{\vmu_t, \vA_t\}$, we have
\begin{align*}
\begin{bmatrix}
 \vmu \\
\vSigma 
\end{bmatrix}=
\vpsi \circ \vphi_{\aparam_t}\big(
\begin{bmatrix}
 \vdelta \\
\vM 
\end{bmatrix}
\big) = 
\begin{bmatrix}
\vmu_t + \vA_t \vdelta \\
\vA_t  \mathrm{Exp}(\vM) \vA_t^T
\end{bmatrix}
\end{align*}

It is easy to see that
$\vpsi \circ \vphi_{\aparam_t}(\lparam)$ is $C^1$-smooth  w.r.t. $\lparam$.

Since we have shown Assumption 1 is satisfied, we have $\vF_{\lparam}(\lparam_0)$ is non-singular.
By Lemma \ref{lemma:uef_scalar}, we know that both 
$\vF_{\gparam}(\gparam_t)$ and the Jacobian matrix
$\nabla_{\lparam} \gparam$ evaluated at $\lparam_0$  are non-singular.
By the inverse function theorem, we know that
 there exist a (local) inverse function of $\vpsi \circ \vphi_{\aparam_t}(\lparam)$ at an open neighborhood of $\lparam_0$, which is also  
 $C^1$-smooth.

Therefore,
we know that $\{\vmu,\vSigma\}=\gparam = \vpsi \circ \vphi_{\aparam_t}(\{ \vdelta, \vM \}) $ is 
locally $C^1$-diffeomorphic  at an open neighborhood of $\lparam_0$.

\subsection{Our NG Updates for the 1-Dim Bayesian Logistic Regression}
\label{app:blr_gauss}
Now, we consider the following parameterization
$ \gparam =\{\mu \in \real, \log\sigma \in \real \}$ for a Gaussian distribution $q$, where $\sigma^2$ is the variance and $\sigma>0$.
The FIM under this parameterization is
\begin{align*}
 \vF_{\gparam}(\gparam) = \begin{bmatrix}
 \sigma^{-2} &0\\
 0 & 2
 \end{bmatrix}
\end{align*}

The standard NGD using this (global) parameterization $\gparam$ with step-size $\beta>0$ is
\begin{align*}
 \mu & \leftarrow \mu - \beta \sigma^2 g_{\mu} \\
 \log\sigma & \leftarrow \log\sigma - \beta \half g_{\log\sigma} = \log\sigma - \beta \half \overbrace{( 2 \sigma^2 g_{\sigma^2} )}^{g_{\log \sigma}} = 
  \log\sigma - \beta   \sigma^2  g_{\sigma^2}
\end{align*}
Recall that our local-parameter approach also includes the standard NGD as a special case shown in 
Appendix \ref{app:general}.
We can also similarly show that the standard NGD on parameterization $\gparam=\{\mu,\log\sigma^2\}$ obtain an equivalent update.

For our local-parameter approach,
consider the following parameterizations:
\begin{align*}
\gparam &= \{\mu \in \real, \sigma^{-2}>0\}\\
 \aparam &= \{ \mu \in \real, b \in \real \setminus \{0\} \} \\
 \lparam & =\{\delta \in \real, m\in \real \} \\
\begin{bmatrix}
\mu \\ b 
\end{bmatrix}
& =  \vphi_{\aparam_t}(\lparam)=
\begin{bmatrix}
 \mu_t + b_t^{-1} \delta \\
 b_t \exp(m)
\end{bmatrix}
\end{align*} where $\sigma^2 = b^{-2}$  is the variance.

Our NGD update (see \eqref{eq:gauss_prec_exp_updates}) under these parameterizations is
\begin{align*}
 \mu &\leftarrow \mu-\beta b^{-2} g_{\mu} = \mu -\beta \sigma^2 g_{\mu} \\
 b & \leftarrow b \exp( \beta b^{-2} g_{\sigma^2} ) \iff \underbrace{ \log b}_{-\log \sigma} \leftarrow \log b + \beta \sigma^2 g_{\sigma^2} ,\,\,\,\,\,\, \text{we assume } b>0 \text{ for } \log(b) \text{ otherwise we  use } \log(-b)
\end{align*} where we use  the exponential map.

Consider another set of parameterizations for our approach:
\begin{align*}
\gparam &= \{\mu \in \real, \sigma^2 >0\}\\
 \aparam &= \{ \mu \in \real, a \in \real \setminus \{0\} \} \\
 \lparam & =\{\delta \in \real, m\in \real \} \\
\begin{bmatrix}
\mu \\ a 
\end{bmatrix}
& =  \vphi_{\aparam_t}(\lparam)=
\begin{bmatrix}
 \mu_t + a_t \delta \\
 a_t \exp({\color{red} \half} m)
\end{bmatrix}
\end{align*} where $\sigma^2 = a^2$ and the red term $\half$ appears  since we use the same  parameterizations as
 \citet{glasmachers2010exponential}.

Our NGD update (see \eqref{eq:ngd_aux_param}) under these parameterizations is 
\begin{align*}
 \mu &\leftarrow \mu-\beta a^{2} g_{\mu} = \mu -\beta \sigma^2 g_{\mu} \\
 a & \leftarrow a \exp( - \beta a^{2} g_{\sigma^2} ) \iff \underbrace{ \log (a)}_{\log \sigma} \leftarrow \log a - \beta \sigma^2 g_{\sigma^2} ,\,\,\,\,\,\, \text{we assume } a>0 \text{ for } \log(a) \text{ otherwise we  use } \log(-a)
\end{align*}

Therefore, we can see our NG updates including standard NGD in global parameterization $\gparam=\{\mu,\log\sigma^2\}$ in this univariate case are all equivalent under these parameterizations and maps.
We could also use map $h(\cdot)$ defined in
Sec.\ref{sec:non_exp_map}.
As shown in \eqref{eq:gauss_newton_exp}, this  map matches the first two order and in practice, 
there is no difference between these two maps in terms of performance.

For (Euclidean) gradient descent (GD), it is not invariant to these parameterizations.
Let's consider a unconstrained parameterization $\{\mu,\log\sigma^2\}$.
The GD update under  parameterization $\{\mu,\log\sigma^2\}$ with step size $\beta>0$ is
\begin{align*}
\mu &\leftarrow \mu - {\color{red} \beta} g_{\mu} \\
\log\sigma^2 &\leftarrow \log\sigma^2 - \beta g_{\log\sigma^2} = \log\sigma^2 - {\color{red} \beta } (\sigma^2 g_{\sigma^2}) 
\end{align*}

Now, we consider another unconstrained parameterization $\{\mu,\log\sigma\}$.
The GD update with  parameterization $\{\mu,\log\sigma\}$  step size $\beta>0$ is
\begin{align*}
\mu &\leftarrow \mu - {\color{red} \beta} g_{\mu} \\
\log\sigma &\leftarrow \log\sigma - \beta g_{\log\sigma} = \log\sigma - \beta (2\sigma^2 g_{\sigma^2}) 
\iff \log\sigma^2 \leftarrow \log\sigma^2 - {\color{red} 4 \beta } (\sigma^2 g_{\sigma^2})
\end{align*}

Clearly, GD is not invariant to the change of parameterizations and its performance depends on the parameterization  even in this simple case.

\subsection{Difficulties of  the standard NGD involving structured covariance/precision}
\label{app:diff_st_gauss}

Before we discuss issues in structured cases, we first revisit cases with full covariance, where we have a Kronecker structure.
This Kronecker structure plays a key role for computational reduction.
Unfortunately, this structure could be  missing in structured covariance/precision cases.

\subsubsection{Cases with full covariance}
Let's consider the following parameterization  $\gparam=\{\vmu, \mathrm{vec}(\vSigma)\}$, where $\vSigma$ is the covariance and $\vmu$ is the mean
The negative-log Gaussian distribution is  $-\log q(\vlat|\vmu,\mathrm{vec}(\vSigma) )= \half \sqr{ \log |\vSigma| +\mathrm{Tr} (\vSigma^{-1} (\vlat-\vmu) (\vlat-\vmu)^T ) } $.
The FIM  under this parameterization is
\begin{align*}
 \vF_{\gparam}(\gparam) &= 
- \Unmyexpect{q}  \sqr{\nabla_{\gparam}^2 \log q(\vlat|\gparam) }  \\
&= \Unmyexpect{q}  \begin{bmatrix}
\vSigma^{-1} & \nabla_{\mathrm{vec}(\Sigma)} \vSigma^{-1} (\vlat-\vmu)  \\
\nabla_{\mathrm{vec}(\Sigma)}^T \vSigma^{-1} (\vlat-\vmu) &
 \half \nabla_{ \mathrm{vec}(\Sigma)}^2 \sqr{ \log |\vSigma| +\mathrm{Tr} (\vSigma^{-1} (\vlat-\vmu) (\vlat-\vmu)^T ) }
   \end{bmatrix}
\\
&=  \begin{bmatrix}
\vSigma^{-1} &\sqr{ \nabla_{\mathrm{vec}(\Sigma)} \vSigma^{-1}} \Unmyexpect{q} \sqr{(\vlat-\vmu)}  \\
\sqr{ \nabla_{\mathrm{vec}(\Sigma)}^T \vSigma^{-1}} \Unmyexpect{q} \sqr{(\vlat-\vmu)} &
 \half  \rnd{ \sqr{ \nabla_{ \mathrm{vec}(\Sigma)}^2 \log |\vSigma|} + \mathrm{Tr}( \sqr{ \nabla_{ \mathrm{vec}(\Sigma)}^2 \vSigma^{-1} } \Unmyexpect{q}\sqr{ (\vlat-\vmu) (\vlat-\vmu)^T }) }
   \end{bmatrix}
\\
&=
\begin{bmatrix}
                          \vSigma^{-1}  & \mathbf{0} \\
                          \mathbf{0} &  \mathrm{Hess} ( f(\vSigma) )
                          \end{bmatrix}  \\
                           &= \begin{bmatrix}
                          \vF_{\mu}( \gparam) & \mathbf{0} \\
                          \mathbf{0} &  \vF_{\mathrm{vec}(\Sigma)}(\gparam)
                          \end{bmatrix}  
\end{align*} where $\vV_0= \Unmyexpect{q}\sqr{ (\vlat-\vmu) (\vlat-\vmu)^T } =\vSigma$ is considered as a constant,
\begin{align*}
f(\vX) &:=\half \sqr{\log |\vX| +\mathrm{Tr}(\vX^{-1} \vV_0)} \\
 \mathrm{Hess} ( f(\vSigma) ) &:= \nabla_{\mathrm{vec}(\Sigma)}^2 f(\vSigma) 
\end{align*}

Similarly, 
let's consider another
parameterization $\gparam=\{\vmu, \mathrm{vec}(\vS)\}$, where $\vS$ is the precision.
The FIM under this parameterization is
\begin{align*}
 \vF_{\gparam}(\gparam) 
 &= \begin{bmatrix}
                          \vF_{\mu}( \gparam) & \mathbf{0} \\
                          \mathbf{0} &  \vF_{\mathrm{vec}(S)}(\gparam)
                          \end{bmatrix}  
\\
&=
\begin{bmatrix}
                          \vP  & \mathbf{0} \\
                          \mathbf{0} & \nabla_{\mathrm{vec}(S)}^2 f(\vS^{-1})
                          \end{bmatrix}  
\end{align*} where $\vV_0$ is a constant used in function $f(\cdot)$ defined above and the value of $\vV_0=\vS^{-1}$.

Let's denote a Euclidean gradient of $\myexpect_{q}\sqr{\ell(\vlat)}$ w.r.t. $\vSigma$ by $\vG_{\Sigma}$, where
$\ell(\vlat)$ is a model loss function and $q(\vlat):=\gauss(\vlat|\vmu,\vSigma)$.
We also denote the corresponding natural-gradient w.r.t. $\vSigma$ by $\hat{\vG}_{\Sigma}$.

Since the FIM is block-diagonal,
we see the FIM block for the vector form of this precision $\mathrm{vec}(\vS)$ is
\begin{align*}
 \vF_{\mathrm{vec}(S)}(\gparam):= \nabla_{\mathrm{vec}(S)}^2 f(\vS^{-1})
\end{align*} 
Note that this FIM block has a Kronecker form as $ \vF_{\mathrm{vec}(S)}(\gparam)= \half ( \vS^{-1} \otimes \vS^{-1} ) $ for  $\mathrm{vec}(\vS)$.
The natural gradient for $\mathrm{vec}(\vS)$ is
\begin{align*}
\mathrm{vec}(\hat{\vG}_{S}) =
\hat{\vg}_{\mathrm{vec}(S)} = \big(\vF_{\mathrm{vec}(S)}(\gparam)\big)^{-1} \mathrm{vec}(\vG_{S}) = 2\big( \vS \otimes \vS \big) \mathrm{vec}(\vG_{S})
\end{align*} where $\mathrm{vec}(\vG_{S})=\vg_{\mathrm{vec}(S)}$ is the Euclidean gradient w.r.t. $\mathrm{vec}(\vS)$.

Exploiting the Kronecker structure,
we can convert this vector form of natural-gradient in a matrix form as
\begin{align*}
\mathrm{Mat}(\hat{\vg}_{\mathrm{vec}(S)}) =
\mathrm{Mat}(2\big( \vS \otimes \vS \big) \mathrm{vec}(\vG_{S}) )
&= 2\vS  \big( \vG_{S} \big) \vS \,\,\,\,  \text{ { \color{red}(exploiting the Kronecker structure)} } \\
&= -2\vG_{S^{-1}} \,\,\,\,\text{(using matrix calculus)}\\
&= -2\vG_{\Sigma}\\
&= -  \myexpect_{q(\lat)}\sqr{ \nabla_\lat^2 \ell(\vlat) } \,\,\,\,\text{(using Stein's identity)},
\end{align*} which  is the natural-gradient for the precision matrix $\vS$.

\subsubsection{Issue involving structured cases}

In low-rank Gaussian cases, as an example, consider the following parameterization
$ \gparam = \{\vmu, \valpha\}$
where $\valpha:=\begin{bmatrix}\vv \\ \vd \end{bmatrix}$ and $\vSigma:= \vv\vv^T + \mathrm{Diag}(\vd^2)$.
The FIM under this parameterization is
\begin{align}
 \vF_{\gparam}(\gparam) = \begin{bmatrix}
                          \vSigma^{-1} & \mathbf{0} \\
                          \mathbf{0} & \mathrm{Hess} ( h(\valpha) )
                          \end{bmatrix} \label{eq:fim_singular_st_gauss}
\end{align}
where $\vV_0=\vSigma$ is considered as a constant, 
 $\vSigma$ is considered as a function of $\valpha$, and
\begin{align*}
h(\valpha) &:= f(\vSigma(\valpha) )\\
 \mathrm{Hess} ( h(\valpha) ) &:= \nabla_{\alpha}^2 f(\vSigma(\valpha))
\end{align*}

There are several issues about NGD for structured Gaussian cases, which lead to a case-by-case derivation for structures.
\begin{itemize}
 \item 
 One issue is that
$\vF_{\alpha}(\gparam)$
can be singular for an arbitrary structure as shown in Appendix \ref{app:diag_zero}.
\item
A critical issue in  that
$\vF_{\alpha}(\gparam) = \mathrm{Hess} ( h(\valpha) ) $ may not have a Kronecker form exploited in full Gaussian cases.
Without the Kronecker from, a computational challenge is
how to  efficiently compute
\begin{align*}
\hat{\vg}_{\alpha} = \big( \vF_{\alpha}(\gparam)\big)^{-1} \vg_{\alpha} =  
 \mathrm{Hess} ( h(\valpha) )^{-1} \vg_{\alpha}
\end{align*} 
\item
If we want to make use of second-order information via Stein's identity, another computational challenge is
about how to re-express 
$ \mathrm{Hess} ( h(\valpha) )^{-1} \vg_{\alpha}$
in terms of $\vG_{\Sigma} =\half \myexpect_{q}\sqr{\nabla_\lat^2 \ell(\vlat)}$ and how to efficiently compute  natural-gradients for $\valpha$ without computing the whole Hessian
$\nabla_{\lat}^2 \ell(\vlat)$.
Note that
$\vSigma= \vv\vv^T + \mathrm{Diag}(\vd^2)$ and 
$ \gparam = \{\vmu, \valpha\}$. Therefore, $\vg_{\alpha}$ could be re-expressed in terms of $\vG_{\Sigma}$ by the chain rule.

\end{itemize}

\section{Wishart distribution with square-root precision structure}
\label{app:wishart}
Let's consider a global parameterization $\gparam=\{\vS, n\}$.
The P.D.F. of a Wishart distribution under this parameterization is
\begin{align*}
 q(\vLat|\gparam) = \exp\{ -\half \mathrm{Tr}(\vS\vLat) + \frac{n-p-1}{2} \log |\vLat| - \frac{np}{2} \log 2 + \frac{n}{2} \log |\vS| - \log\Gamma_p(\frac{n}{2} ) \} 
\end{align*} where $\vLat$ is a $p$-by-$p$ positive-definite matrix.
The parameterization constraint for Wishart distribution is $n>p-1$ and $\vS \in {\cal S}^{p\times p}_{++}$, where 
${\cal S}^{p\times p}_{++}$ denotes the set of $p$-by-$p$ positive-definite matrices.

We start by specifying the parameterization,
\begin{equation*}
    \begin{split}
        \gparam &:= \crl{n\in \real, \,\,\, \vS \in \mathcal{S}_{++}^{p\times p} \,\,\,|\,\,\, n>p-1 }, \,\,\,\\
        \aparam &:= \crl{ b \in\real, \,\,\, \vB \in\mathrm{GL}^{p\times p} }, \\
        \lparam &:= \crl{ \delta\in\real, \,\,\, \vM \in\mathcal{S}^{p\times p}  },
    \end{split}
\end{equation*}
and their respective maps defined at $\aparam_t := \{b_t, \vB_t\}$
\begin{equation*}
    \begin{split}
        \crl{ \begin{array}{c} n \\ \vS \end{array} } &= \vpsi(\aparam) :=  \crl{ \begin{array}{c} 2f(b)+p-1 \\ (2f(b)+p-1 ) \vB\vB^\top \end{array} }, \\
        \crl{ \begin{array}{c} b \\ \vB \end{array} } &= \vphi_{\aparam_t}(\lparam) :=  \crl{ \begin{array}{c} b_t + \delta \\ \vB_t \mathrm{Exp} \rnd{\vM} \end{array} }.
    \end{split}
\end{equation*} where $f(b)=\log(1+\exp(b))$ is the soft-plus function.

For simplicity, we assume $\vM$ is symmetric. We can also exploit structures in the Wishart case.

Under this local parameterization, we have the following result.
\begin{align*}
-\log q(\vLat|\lparam) &=  
(f(b_t+\delta)+c) \mathrm{Tr}( \vB_t \mathrm{Exp}(\vM)  \mathrm{Exp}(\vM)^T \vB_t^T \vLat)
- (f(b_t+\delta)-1)\log |\vLat| \\
-&  (f(b_t+\delta)+c) p \log ( f(b_t+\delta) +c)
-2 (f(b_t+\delta)+c) (\log |\mathrm{Exp}(\vM)|+ \log |\vB_t|) \\
+& \log \Gamma_p( f(b_t+\delta)+c )
\end{align*} where $c=\frac{p-1}{2} $.

\begin{lemma}
\label{lemma:fim_bdiag_wishart_prec}
Under this local parametrization $\lparam$,  $\vF_{\lparam}(\lparam_0)$ is block diagonal with two blocks--the $\vdelta$ block and the $\vM$ block. 
\end{lemma}

\begin{myproof}
The cross term at $\lparam_0=\mathbf{0}$ is
\begin{align*}
 &- \Unmyexpect{q(\Lat|\lparam)} \big[ \nabla_\delta \nabla_M \log q(\vLat|\lparam) \big] \big|_{\lparam=0} \\
 = & \nabla_\delta 2(f(b_t+\delta)+c) \Unmyexpect{q(\lat|\lparam)}\big[ \vB_t^T \vLat\vB_t - \vI  \big]\big|_{\lparam=0} \\
 = & \nabla_\delta 2(f(b_t+\delta)+c) \underbrace{\big[ \vI - \vI  \big]}_{=\mathbf{0}}\big|_{\lparam=0} \\
 = & \mathbf{0}
\end{align*} where we have the fact that $\Unmyexpect{q(\Lat|\lparam)}\big[  \vLat \big]\big|_{\lparam=0} = \vB_t^{-T} \vB_t^{-1}$.
 
\end{myproof}

Let $\vZ = \vB_t^T \vLat \vB_t$
First, we consider the following result.
\begin{align*}
 \nabla_{M_{ij}}\mathrm{Tr}( \vB_t \mathrm{Exp}(\vM)  \mathrm{Exp}(\vM)^T \vB_t^T \vLat)
&= \nabla_{M_{ij}}\mathrm{Tr}( \vZ \mathrm{Exp}(\vM)  \mathrm{Exp}(\vM)^T )\\
&=  \mathrm{Tr}\big(\vZ  \big[ \nabla_{M_{ij}} \mathrm{Exp}(\vM) \big] \mathrm{Exp}(\vM)^T + \vZ \mathrm{Exp}(\vM) \nabla_{M_{ij}} \big[ \mathrm{Exp}(\vM)^T \big] \big) 
\end{align*}

By Lemma \ref{lemma:eq3},
we obtain a simplified expression.
\begin{align*}
  \nabla_{M} \big[  \vB_t \mathrm{Exp}(\vM) \mathrm{Exp}(\vM)^T \vB_t^T \vLat  \big] 
=  2\vZ +  (\vZ \vM^T + \vM^T \vZ)  +  2\vZ \vM  +  \vZ O(\vM^2)
\end{align*} 

By 
Lemma  \ref{lemma:eq2}, 
we have
\begin{align}
- \nabla_{M}  \log | \mathrm{Exp}(\vM) | = -\vI - C(\vM)  
\end{align}

Now, we can compute the FIM w.r.t. block $\vM$ as follows.
Note that we also numerically verify the following computation of FIM by Auto-Diff.
\begin{align*}
 &- \Unmyexpect{q(\Lat|\lparam)} \big[  \nabla_M^2 \log q(\vLat|\lparam) \big] \big|_{\lparam=0} \\
=&  \Unmyexpect{q(\Lat|\lparam)} \big[ (f(b_t+\delta)+c) \nabla_M \big[  2\vZ +  (\vZ \vM^T + \vM^T \vZ)  +  2\vZ \vM  + \vZ O(\vM^2) -2\vI - 2C(\vM) \big] \big] \big|_{\lparam=0} \\
=&   \big[ (f(b_t+\delta)+c) \nabla_M \big[  2\vI + 2 \vM^T  +  2 \vM    -2\vI + O(\vM^2)  \big] \big] \big|_{\lparam=0} 
 -  2\big[ (f(b_t+\delta)+c)  \big] \underbrace{ \nabla_M \big[  C(\vM) \big] \big|_{\lparam=0} }_{=\mathbf{0}} \\
=&   \big[ (f(b_t+\delta)+c) \nabla_M \big[    2 \vM^T  +  2 \vM   + O(\vM^2)   \big] \big] \big|_{\lparam=0} \\
=&    2(f(b_t)+c) \nabla_M \big(  \vM^T  +  \vM  \big) 
\end{align*}
where we use the fact that
$  \Unmyexpect{q(\Lat|\lparam)}\sqr{  \vZ } = \Unmyexpect{q(\Lat|\lparam)}\sqr{ \vB_t^T \vLat \vB_t  } = \vI$ evaluated at $\lparam=\mathbf{0}$ to move from step 2 to step 3.

When $\vM$ is symmetric, we have $\vF_M(\lparam_0) = 4(f(b_t)+c) \vI= 2n_t \vI$. 

Next, we discuss how to compute the FIM w.r.t. $\delta$.
Let $z(\delta):= \big[\mathrm{Tr}( \vB_t \mathrm{Exp}(\vM)  \mathrm{Exp}(\vM)^T \vB_t^T \vLat)
- \log |\vLat| 
-   p \log( f(b_t+\delta)+c)
- p
- 2 (\log |\mathrm{Exp}(\vM)|+ \log |\vB_t|)
+ \psi_p( f(b_t+\delta)+c) \big]$,
where  $\psi_p(x) :=\nabla_x \log \Gamma_p( x)$ is  the multivariate digamma function.

First, let's observe that
\begin{align*}
-   \nabla_\delta \log q(\vLat|\lparam) 
= z(\delta) \frac{\exp(b_t+\delta)}{1+\exp(b_t+\delta)} 
\end{align*}

Similarly, we have
\begin{align*}
 -   \nabla_\delta^2 \log q(\vLat|\lparam) 
= z(\delta) \big[\nabla_\delta \frac{\exp(b_t+\delta)}{1+\exp(b_t+\delta)}\big]  +
\big[\nabla_\delta z(\delta)\big]
\frac{\exp(b_t+\delta)}{1+\exp(b_t+\delta)} 
\end{align*}

Let's consider the first term in the above expression.
\begin{align*}
 z(\delta) \big[\nabla_\delta \frac{\exp(b_t+\delta)}{1+\exp(b_t+\delta)}\big]  = 
 - \big[ \nabla_\delta \log q(\vLat|\lparam) \big]  \frac{1+\exp(b_t+\delta)}{\exp(b_t+\delta)} \big[\nabla_\delta \frac{\exp(b_t+\delta)}{1+\exp(b_t+\delta)}\big]
\end{align*}

Note that $\lparam_0=\{\vM_0,\delta_0\}=\mathbf{0}$. We have the following result.
\begin{align*}
& \Unmyexpect{q(\Lat|\lparam)} \big[z(\delta) \big[\nabla_\delta \frac{\exp(b_t+\delta)}{1+\exp(b_t+\delta)}\big]\big|_{\lparam=0} \\
=& - \underbrace{\Unmyexpect{q(\Lat|\lparam)} \big[\nabla_\delta \log q(\vLat|\lparam) \big]\big|_{\lparam=0}}_{=0\, (\text{see Eq \eqref{eq:score_fun_grad}}) } \Big( \frac{1+\exp(b_t+\delta)}{\exp(b_t+\delta)} \big[\nabla_\delta \frac{\exp(b_t+\delta)}{1+\exp(b_t+\delta)}\big] \Big)\big|_{\lparam=0} \\
=& 0
\end{align*}

Now, we consider the second term.
Note that
\begin{align*}
 \big[\nabla_\delta z(\delta)\big] =  \frac{\exp(b_t+\delta)}{1+\exp(b_t+\delta)} \big(
 -\frac{p}{f(b_t+\delta)+c} + D_{\psi,p}\big( f(b_t+\delta)+c\big) \big)
\end{align*} where $D_{\psi,p}(x) = \nabla \psi_p(x)$ is  the multivariate trigamma function.

Therefore, we can compute the FIM w.r.t. $\delta$ as follows.
\begin{align*}
\vF_\delta(\lparam_0) &= 
 - \Unmyexpect{q(\Lat|\lparam)} \big[  \nabla_\delta^2 \log q(\vLat|\lparam) \big] \big|_{\lparam=0} 
 = \big( \frac{\exp(b_t)}{1+\exp(b_t)} \big)^2 \big( - \frac{2p }{2f(b_t)+p-1} + D_{\psi,p}\big( f(b_t)+\frac{p-1}{2}\big) \big)\\
&= \big( \frac{\exp(b_t)}{1+\exp(b_t)} \big)^2 \big( - \frac{2p }{n_t} + D_{\psi,p}\big( \frac{n_t}{2}\big) \big)
\end{align*}

Now, we discuss how to compute the Euclidean gradients.
First note that 
\begin{align*}
n &:=    2( f(b_t+\delta)+\frac{p-1}{2}) \\
\vV^{-1} & :=  \vS =  2( f(b_t+\delta)+\frac{p-1}{2}) \vB_t \mathrm{Exp}(\vM)\mathrm{Exp}(\vM)^T \vB_t^T
\end{align*} where we will evaluate $n$ and $\vV$ at $\delta=0$ and $\vM=0$.

Let ${\cal L}:=  \Unmyexpect{q(\text{\vlat})} \sqr{ \ell(\vlat) } - \gamma\entropy (q(\vlat)) $.
By the chain rule, we have
\begin{align*}
 \nabla_{\delta}{\cal L} & := \mathrm{Tr}( \big[\nabla_{V}{\cal L}\big]  \big[\nabla_{\delta} \vV \big]) +  \big[ \nabla_n {\cal L} \big] \big[ \nabla_{\delta} n \big] \\
 \nabla_{M_{ij}}{\cal L} & := \mathrm{Tr}( \big[\nabla_{V}{\cal L}\big]  \big[\nabla_{M_{ij}} \vV \big]) +  \big[ \nabla_n {\cal L} \big] \overbrace{\big[ \nabla_{M_{ij}} n \big]}^{=\mathbf{0}} \\
 &=  \mathrm{Tr}( \big[\nabla_{V}{\cal L}\big]  \big[\nabla_{M_{ij}} \vV \big]) \\
 &=  -\mathrm{Tr}( \big[\nabla_{V}{\cal L}\big] \vV \big[  \nabla_{M_{ij}} \vV^{-1}  \big] \vV) 
\end{align*}

Note that
\begin{align*}
 \nabla_{\delta}{\cal L}\big|_{\lparam=0} & :=\mathrm{Tr}( \big[\nabla_{V}{\cal L}\big] \big[  \nabla_{\delta} \vV  \big]) + \big[ \nabla_n {\cal L} \big] \big[ \nabla_{\delta} n \big]\big|_{\lparam=0}  \\
 &= \frac{-1}{ 2 (f(b_t)+\frac{p-1}{2})^2 } \frac{\exp(b_t)}{1+\exp(b_t)}\mathrm{Tr}( \big[\nabla_{V}{\cal L}\big] \vB_t^{-T} \vB_t^{-1} )
 +  \frac{2\exp(b_t)}{1+\exp(b_t)} \big[ \nabla_n {\cal L} \big]  \\
 &= \frac{2\exp(b_t)}{1+\exp(b_t)}\big( \frac{-1}{ 4 (f(b_t)+\frac{p-1}{2})^2 }\mathrm{Tr}( \big[\nabla_{V}{\cal L}\big] \vB_t^{-T} \vB_t^{-1} ) +   \big[ \nabla_n {\cal L} \big] \big) \\
 &= \frac{2\exp(b_t)}{1+\exp(b_t)}\big( \frac{-1}{ n_t^2 }\mathrm{Tr}( \big[\nabla_{V}{\cal L}\big] \underbrace{\vB_t^{-T} \vB_t^{-1}}_{=n_t \vV_t } ) +   \big[ \nabla_n {\cal L} \big] \big) \\
 &= \frac{2\exp(b_t)}{1+\exp(b_t)}\big( -\frac{\mathrm{Tr}( \big[\nabla_{V}{\cal L}\big] \vV_t )}{n_t} +   \big[ \nabla_n {\cal L} \big] \big) \\
\\
 \nabla_{M_{ij}}{\cal L}\big|_{\lparam=0} & :=-\mathrm{Tr}( \big[\nabla_{V}{\cal L}\big] \vV \big[  \nabla_{M_{ij}} \vV^{-1}  \big] \vV) \big|_{\lparam=0} \\
 &= - n_t\mathrm{Tr}( \big[\nabla_{V}{\cal L}\big] \vV_t \big[\vB_t \nabla_{M_{ij}} ( \vM +\vM^T)  \vB_t^T \big] \vV_t ) \\
 &= - n_t \mathrm{Tr}( \big[\nabla_{V}{\cal L}\big] \underbrace{ n_t^{-1}\vB_t^{-T}\vB_t^{-1} }_{=\vV_t}\big[\vB_t \nabla_{M_{ij}} ( \vM +\vM^T)  \vB_t^T \big] \underbrace{ n_t^{-1}\vB_t^{-T}\vB_t^{-1}}_{=\vV_t} ) \\
 &= - n_t^{-1}\mathrm{Tr}( \big[\nabla_{V}{\cal L}\big]  \vB_t^{-T}\big[ \nabla_{M_{ij}} ( \vM +\vM^T) \big] \vB_t^{-1} ) 
\end{align*} 

when $\vM$ is symmetric, we have
\begin{align*}
 \nabla_{M}{\cal L}\big|_{\lparam=0} & := -\frac{2}{n_t}\mathrm{Tr}( \vB_t^{-1}\big[\nabla_{V}{\cal L}\big]  \vB_t^{-T}  ) \\
 \nabla_{\delta}{\cal L}\big|_{\lparam=0} & :=
  \frac{2\exp(b_t)}{1+\exp(b_t)} \big[ \frac{ - \mathrm{Tr}( \big[\nabla_{V}{\cal L}\big] \vV_t )}{n_t} + \big[ \nabla_n {\cal L} \big] \big]
\end{align*}  where we use the fact that
$\big[\nabla_{V}{\cal L}\big]$ is symmetric.

In the symmetric case,  the FIM w.r.t. $\lparam$ at $\lparam_0$ is
\begin{align*}
\vF_{\lparam}(\lparam_0) = \begin{bmatrix}
                          2n_t \vI_M & \mathbf{0} \\
                          \mathbf{0} & \big( \frac{\exp(b_t)}{1+\exp(b_t)} \big)^2 \big( - \frac{2p }{n_t} + D_{\psi,p}\big( \frac{n_t}{2}\big) \big)
                         \end{bmatrix},
\end{align*} which implies that Assumption 1 is satisfied.

The natural gradients  are
\begin{align*}
\hat{\vg}_{M} &:= \frac{1}{2 n_t} \vG =  - \frac{1}{n_t^2} \vB_t^{-1} \big[\nabla_{V}{\cal L}\big] \vB_t^{-T}  \\ 
\hat{\vg}_{\delta} &:=
 \frac{2(1+\exp( b_t))}{ \exp( b_t)} \big( - \frac{2p }{n_t} +  D_{\psi,p}\big( \frac{n_t}{2}\big) \big)^{-1}
 \big[ \frac{ - \mathrm{Tr}( \big[\nabla_{V}{\cal L}\big] \vV_t )}{n_t} + \big[ \nabla_n {\cal L} \big] \big]
\end{align*}
where $\nabla_{V}{\cal L}$ and $\nabla_{n}{\cal L}$ can be computed by the implicit reparametrization trick in the following section.

Therefore, our update with step-size $\beta$ is
\begin{align}
 \vB_{t+1} & \leftarrow \vB_{t} \mathrm{Exp}( 0 - \beta \hat{\vg}_{M} ) = \vB_t \mathrm{Exp}(  \frac{\beta}{n_t^2} \vB_t^{-1} \big[\nabla_{V}{\cal L}\big] \vB_t^{-T} ) \nonumber \\
 b_{t+1} & \leftarrow b_t + (0-\beta \hat{\vg}_{\delta} ) = b_t - 
 \frac{2\beta (1+\exp( b_t))}{ \exp( b_t)} \big( - \frac{2p }{n_t} +  D_{\psi,p}\big( \frac{n_t}{2}\big) \big)^{-1}
 \big[ \frac{ - \mathrm{Tr}( \big[\nabla_{V}{\cal L}\big] \vV_t )}{n_t} + \big[ \nabla_n {\cal L} \big] \big] \label{eq:wishar_ngd_aux_app}
\end{align}

We can similarly show that Assumption 2 is also satisfied by the inverse function theorem as discussed in Gaussian cases (see
Appendix
\ref{sec:gauss_prec})
since the soft-plus function $f(b)$  and $\mathrm{Exp}(\vM)$ are both $C^1$-smooth.

\subsection{Reparametrizable Gradients}
\label{app:rep_g_wishart}

Recall that we can generate  a Wishart random variable $\vLat$ due to the Bartlett decomposition as shown below.
$\vLat = \vL \vOmega \vOmega^T \vL^T$, where $\vL$ is the lower-triangular Cholesky factor of $\vS^{-1}=\vV$ and $\vOmega$ is the random lower-triangular matrix defined according to the Bartlett decomposition as follows
\begin{align*}
 \vOmega =\begin{bmatrix}
 c_1 & 0 & 0 & \cdots & 0\\
n_{21} & c_2 &0 & \cdots& 0 \\
n_{31} & n_{32} & c_3 & \cdots & 0\\
\vdots & \vdots & \vdots &\ddots & \vdots \\
n_{d1} & n_{d2} & n_{d3} &\cdots & c_d
          \end{bmatrix}
\end{align*}
where the square of diagonal entry $c_i^2$ is independently generated from  Gamma distribution with shape $\frac{n-i+1}{2}$ and rate $\half$, and other non-zero entries $n_{ij}$ are independently drawn from standard normal distribution.  

Let ${\cal L}_1=\Unmyexpect{q} \sqr{{ \ell(\vLat) }}$.
According to this sampling scheme, we can clearly see that Wishart distribution is reparametrizable.
The gradient w.r.t. $\vV$ can be computed as
\begin{align*}
\nabla_V {\cal L}_1  =  \Unmyexpect{q(\Omega)} \sqr{ \nabla_\Lat { \ell(\vLat) } \nabla_V \big( \vL \vOmega \vOmega^T \vL^T\big) }
\end{align*}

Since Gamma distribution is implicitly re-parametrizable, we can also compute the gradient  $\nabla_n {\cal L}_1$ thanks to the implicit reparametrization trick \citep{figurnov2018implicit,wu-report} for Gamma distribution.

\subsection{Riemannian Gradient Descent at $\vU$}
\label{app:rgd_at_u}
\begin{align*}
\min_{Z \in {\cal S}_{++}^{p \times p}} \ell (\vZ)
\end{align*}

Instead of optimizing $\vZ$, we optimize $\vU=\vZ^{-1}$. 
A Riemannian gradient \citep{hosseini2015matrix,lin2020handling} in the manifold ${\cal S}_{++}^{p \times p}$ is $\vnGrad=\vU \big(\nabla_U \ell \big) \vU$.
The RGD update with retraction and step-size $\beta_1$ is
\begin{align*}
\vU \leftarrow \vU -  \beta_1 \vnGrad + \frac{\beta_1^2}{2} \vnGrad (\vU)^{-1} \vnGrad.
\end{align*}
Due to matrix calculus, we have $\nabla_Z \ell = -\vU \big(\nabla_U \ell \big) \vU$. 
We can re-express the RGD update as
\begin{align*}
\vU \leftarrow \vU+  \beta_1 \nabla_Z \ell + \frac{\beta_1^2}{2} \big[\nabla_Z \ell] \vU^{-1} \big[\nabla_Z \ell].
\end{align*}

\subsection{Gradients Evaluated at the Mean}
\label{app:gd_wishart_at_mean}
Recall that the mean of the Wishart distribution as $\vZ = \Unmyexpect{q} \sqr{{ \vLat }}=n \vS^{-1}=n\vV$.
We can approximate the Euclidean gradients as below.

\begin{align*}
    \nabla_{V_{ij}} \Unmyexpect{q(\Lat)} \sqr{{ \ell(\vLat) }}
\approx \mathrm{Tr}\big( \nabla_Z { \ell(\vZ) } \nabla_{V_{ij}} \big( n\vV \big) \big) = n \nabla_{Z_{ij}} { \ell(\vZ) } \\
    \nabla_{n} \Unmyexpect{q(\Lat)} \sqr{{ \ell(\vLat) }}
\approx \mathrm{Tr}\big( \nabla_Z { \ell(\vZ) } \nabla_{n} \big( n\vV \big) \big) = \mathrm{Tr}(\nabla_Z \ell(\vZ) \vV ) \\
\end{align*} where $\vZ=n\vV$.

Therefore,
\begin{align*}
    \vG_{\text{\vV}_t} \approx n_t \nabla { \ell(\vZ_t) }, \quad
    g_{n_t} \approx \mathrm{Tr} \sqr{ \nabla \ell(\vZ_t) \vV_t }
\end{align*}

\section{Standard NGD is a Special Case}
\label{app:general}
The standard NGD in a global parameter $\gparam$ is a special case of using a local parameter $\lparam$.
We assume $\gparam$ is unconstrained and the FIM is non-singular for $\gparam \in \gspace$.
Note that if $\gparam$ stays in a constraint set, the standard NGD is not well-defined since the update could violate the constraint.
In this case, we choose the auxiliary parameter $\aparam$ to be the same as $\gparam$. 
The map $\vpsi \circ \vphi_{\aparam_t}(\lparam)$ is chosen to be
\begin{align*}
\gparam = \vpsi(\aparam) := \aparam; \,\,\,\,\,
        \aparam = \vphi_{\aparam_t}(\lparam) :=   \aparam_t + \lparam.  
\end{align*}
        
\begin{thm}
\label{thm:global_fim}
Let $\vF_{\lparam}$ and $\vF_{\gparam}$ be the FIM under the local parameter $\lparam$ and the global parameter $\gparam$, respectively.
\begin{align*}
\vF_{\lparam}(\lparam_0) = \vF_{\lparam}(\mathbf{0}) = \vF_{\gparam}(\gparam_t)
\end{align*}
\end{thm}

It is obvious that Assumption 2 is satisfied since the map is linear.
Since $\vfim_{\gparam}( \gparam_t)$ is non-singular, we know that Assumption 1 is satisfied due to Theorem \ref{thm:global_fim}.
We can also verify that Assumption 3 is satisfied when $\gparam$ is unconstrained.
Since $\gparam = \vpsi \circ \vphi_{\aparam_t}(\lparam) = \aparam_t + \lparam$, by the chain rule, we have
 $\vg_{\lparam_0} = \big[\nabla_{ \lparam} \gparam \big] \vg_{\gparam_t } = \vg_{\gparam_t }$

Therefore, the NGD update with step-size $\beta$ in this local parameterization is
\begin{align*}
 \lparam^{\text{new}}  = \mathbf{0} - \beta \vF_{\lparam}(\mathbf{0})^{-1} \vg_{\lparam_0}
 =- \beta \vF_{\gparam}(\gparam_t)^{-1} \vg_{\gparam_t}
\end{align*}

Finally, we re-express the update in the global parameter as:
\begin{align*}
\gparam_{t+1}  =   \vpsi \circ \vphi_{\aparam_t}(\lparam^{\text{new}})
= \gparam_t +  \lparam^{\text{new}} 
= \gparam_t - \beta \vF_{\gparam}(\gparam_t)^{-1} \vg_{\gparam_t}
\end{align*} which is exactly the standard NGD update in $\gparam$.

\subsection{Proof of theorem \ref{thm:global_fim} }
\label{app:proof_thm_global}
Note that $\gparam = \psi \circ \phi_{\aparam_t}(\lparam) = \aparam_t + \lparam = \gparam_t + \lparam $.
Now, we will show that the FIM under the local parameter $\lparam$ can be computed as
\begin{align*}
\vF_{\lparam} (\mathbf{0}) &=
- \Unmyexpect{q(\text{\vlat}|\lparam)}\big[
 \nabla_{\lparam}^2 \log q(\vlat| \lparam ) \big] \big|_{\lparam=0} \\
&= - \Unmyexpect{q(\text{\vlat}|\lparam)}\big[  \nabla_{\lparam} \big[ \underbrace{\nabla_{\lparam} \gparam}_{\vI} \nabla_{\gparam}  \log q(\vlat| \gparam ) \big]  \big] \big|_{\lparam=0} \\
&= -  \Unmyexpect{q(\text{\vlat}|\lparam)}\big[ \nabla_{\lparam } \big[ \nabla_{ \gparam }  \log q(\vlat| \gparam ) \big] \big] \big|_{\lparam=0}\\
&= -  \Unmyexpect{q(\text{\vlat}|\lparam)}\big[ \big[ \nabla_{\lparam} \gparam \big]  \nabla_{\gparam } \big[ \nabla_{ \gparam }  \log q(\vlat| \gparam ) \big] \big] \big|_{\lparam=0}\\
&= - \Unmyexpect{q(\text{\vlat}|\lparam)}\big[ \nabla_{\gparam} \big[  \nabla_{\gparam}  \log q(\vlat| \gparam ) \big] \big] \big|_{\lparam=0} \\
&= - \Unmyexpect{q(\text{\vlat}|\gparam)}\big[ \nabla_{\gparam } \big[  \nabla_{\gparam}  \log q(\vlat| \gparam ) \big] \big] \big|_{\gparam=\gparam_t} \\
&= \vF_{\gparam} (\gparam_t)
\end{align*}

\section{Univariate Minimal Exponential Family Distributions}
\label{app:uni_ef}
Using Lemma 
\ref{lemma:uef_scalar}, 
we can generalize the indirect method of \citet{salimbeni2018natural} to compute natural-gradients for univariate \emph{minimal} EF distributions using a local parameterization.
\citet{salimbeni2018natural} only consider the method for multivariate Gaussian cases using a global parameterization.

Note that the main issue to perform the standard NGD update in the global parameter space is that the NGD update in $\gparam$ may violate a parameter constraint. However, we can perform a NGD update in an unconstrained space (e.g., the auxiliary space of $\aparam$ )
if the natural gradient computation  in the space of unconstrained space of $\aparam$ is simple.
\citet{salimbeni2018natural} suggest using the indirect method to compute natural gradients via Auto-Differentiation (Auto-Diff).

For univariate minimal EF distributions, we can also use this indirect method to compute natural gradients.
We consider a class of univariate EF distributions.
We make the following assumptions for the class of distributions:
(A) Each distribution in the class contains separable natural parameter blocks so that each  parameter constraint only appears once in a block and
each block only contains a scalar parameter.
(B) The natural gradient w.r.t. the natural parameterization is easy to compute.

We choose the natural parameterization as a global parameterization $\gparam$ with $K$ blocks:
$q(\lat|\gparam) = B(\lat) \exp( \myang{\vT(\lat), \gparam} - A(\gparam))$, where $B(\lat)$ is the base measure, $A(\gparam)$ is the log partition function\footnote{
$\exp(\cdot)$ is the scalar exponential function and do not confuse it with the matrix exponential function $\mathrm{Exp}(\cdot)$.
$A(\gparam)$ is $C^{2}$-smooth w.r.t. $\gparam$ as shown in \citet{johansen1979introduction}.
}, and $\vT(\lat)$ is the sufficient statistics.
A common parameter constraint in $\gparam$ is the scalar positivity constraint denoted by 
$\mathcal{S}_{++}^{1}$.
For simplicity, we assume $\mathcal{S}_{++}^{1}$ is the only parameter constraint.
Common univariate EF distributions such as Bernoulli, exponential,
Pareto, Weibull,  Laplace, Wald, univariate Gaussian, Beta, and Gamma distribution all satisfy Assumption A.
Assumption B is also valid for these univariate EF distributions since we can either compute the natural gradient $\vngrad_{\gparam_t}$ via the Euclidean gradient w.r.t. the expectation parameter \citep{khan2017conjugate} or use the direct natural gradient computation when $K$ is small ($K<3$ in common cases).

Given a distribution in the class, we consider the following parameterizations:
\begin{equation*}
    \begin{split}
        \gparam := \begin{bmatrix} \sgparam_1  \in \mathcal{S}_{++}^{1} \\
        \cdots \\
        \sgparam_K  \in \mathcal{S}_{++}^{1}
\end{bmatrix}, \,\,\,\,  
        \aparam := \begin{bmatrix} \saparam_1  \\
        \cdots \\
        \saparam_K  
\end{bmatrix} \in \real^{K},\,\,\,\,  
        \lparam := 
        \begin{bmatrix} \slparam_1  \\
        \cdots \\
        \slparam_K 
\end{bmatrix} \in \real^{K}
    \end{split}
\end{equation*}
and maps:
\begin{equation*}
    \begin{split}
 \gparam =     \vpsi(\aparam)   := \begin{bmatrix} f(\saparam_1 )  \\
        \cdots \\
        f(\saparam_K) 
\end{bmatrix},\,\,\,\, 
   \aparam =    \vphi_{\aparam_t}(\lparam)   :=  \aparam_{t} + \lparam = \begin{bmatrix} \saparam_{1,t} + \slparam_{1}   \\
        \cdots \\
        \saparam_{K,t} + \slparam_{K}
\end{bmatrix}
    \end{split}
\end{equation*}
where $f(b):=\log(1+\exp(b))$ is the soft-plus function\footnote{We use the soft-plus function instead of the scalar exponential map for numerical stability.} and $\gparam$ is the natural parameterization.

In this case,
we can easily compute the Jacobian, 
where $\nabla f(b):=\frac{\exp(b)}{1+\exp(b)}$.
\begin{align*}
  \nabla_{\lparam} \gparam \Big|_{\lparam=\lparam_0=\mathbf{0}} = \mathrm{Diag} \Big( \begin{bmatrix}
\nabla f(  \saparam_{1,t})   \\
        \cdots \\
\nabla f(  \saparam_{K,t})  
                                            \end{bmatrix} \Big)
\end{align*}

By Lemma \ref{lemma:uef_scalar}, we have
\begin{align*}
\vngrad_{\lparam_0} = \big[ \nabla_{\lparam} \gparam \big]^{-T}  \vngrad_{\gparam_t} \Big|_{\lparam=\mathbf{0}}  
\end{align*} where natural-gradient $\vngrad_{\gparam_t}$ can be computed via the Euclidean gradient w.r.t. its expectation parameter or via direct inverse FIM computation as below
\begin{align*}
 \vngrad_{\gparam_t} & =  \big( \vF_{\gparam} (\gparam_t)\big)^{-1} \vg_{\gparam_t} \\
& = \big( \nabla_{\gparam} \vm \big)^{-1}\vg_{\gparam_t} \\
&= \vg_{\text{\vm}}
\end{align*} where $\vm = \Unmyexpect{q}\sqr{ \vT(\lat) } = \nabla_{\gparam} A(\gparam) $ is the expectation parameter and $ \vF_{\gparam} (\gparam_t) = \nabla_{\gparam}^2 A(\gparam_t)$ is the FIM which is non-singular due to the minimality of the distribution.

Our update in the  auxiliary parameter space is
\begin{align}
 \aparam_{t+1} \leftarrow  \aparam_{t} + (-\beta \vngrad_{\lparam_0})  \label{eq:uef_ngd}
\end{align}

Since $\aparam =  \vphi_{\aparam_t}(\lparam)=\aparam_t+\lparam$, we can easily show that $\vngrad_{\lparam_0} = \vngrad_{\aparam_t}$.
In other words, our update recovers the standard NGD update in an unconstrained space of $\aparam$.
\begin{align*}
 \aparam_{t+1} \leftarrow  \aparam_{t} -\beta \vngrad_{\aparam_t},
\end{align*} which recovers the method proposed by 
\citet{salimbeni2018natural} in multivariate Gaussian cases.

Therefore, by choosing $\aparam = \vphi_{\aparam_t}(\lparam)= \aparam_t +\lparam$, Lemma \ref{lemma:uef_scalar} generalizes the indirect method proposed by \citet{salimbeni2018natural}.

\subsection{Discussion about the Indirect Method}
\label{app:indirect_limit}
\citet{salimbeni2018natural} propose an indirect method to compute natural-gradients via Auto-Differentiation (Auto-Diff) for multivariate Gaussian with full covariance structure via a unconstrained parameter transform.
We have shown that this method is a special case of our approach by using a particular local parameterization and have extended it to univariate \emph{minimal} EF distributions by using Lemma \ref{lemma:uef_scalar}.

 The indirect approach requires us to first define one parameterization $\gparam$ so that natural-gradient $\vngrad_{\gparam}$ is easy to compute under this parameterization.
To compute natural-gradient in another  parameterization $\lparam$,
the indirect method avoids computing the FIM $\vF_{\lparam}(\lparam)$ by computing the Jacobian $\big[ \nabla_{\gparam} \lparam \big]$ instead.
Unfortunately, the Jacobian matrix computation can be very complicated when it comes to a matrix parameter.
\citet{salimbeni2018natural} suggest using Auto-Diff to track non-zero terms in the Jacobian matrix $\big[ \nabla_{\gparam} \lparam \big]$
(e.g., $\lparam$ can be a Cholesky factor of $\vS$ and $\gparam=\vS$ is the precision matrix in Gaussian cases with a constant mean)
and to perform the Jacobian-vector product as shown in Lemma \ref{lemma:uef_scalar}.

However, this indirect method has several limitations when it comes to a structured matrix parameter $\lparam$ such as structured Gaussian and Wishart cases.
\begin{itemize}
 \item 
 The parameterization transform used in this indirect approach often requires the Jacobian matrix $\big[ \nabla_{\gparam} \lparam \big]$ to be square and invertible (see Lemma \ref{lemma:uef_scalar}).
 For a new structured  parameter $\lparam$,  the Jacobian between $\gparam$ and $\lparam$ can be a non-square matrix and therefore the classical parameter transform rule fails (e.g., Lemma \ref{lemma:uef_scalar}).
Furthermore, it is difficult to automatically verify whether the Jacobian is invertible or not even when the Jacobian is a square matrix.
\item The existing Auto-Diff implementation of the Jacobian-vector product requires us to compute a dense natural-gradient $\vngrad_{\gparam}$ (e.g., $\vg_{\Sigma}$ has to compute the Hessian matrix in Gaussian cases with a constant mean) beforehand, which is not efficient for a sparse structured parameter $\lparam$. 

 \item
 For a structured Gaussian NGD with second-order information, 
the Auto-Diff system has to first record non-zero entries  in the Jacobian matrix from a structured parameterization $\lparam$ to the precision $\gparam=\vS$ and then query 
the corresponding entries of natural gradient $\vngrad_{\gparam}$ for the precision (which can be expressed in terms of $\vG_{S^{-1}}=\half \myexpect_{q}\sqr{\nabla_\lat^2 \ell(\vlat)}$ via Stein's identity \citep{khan18a}).
Since Auto-Diff does not know how to organize the required entries in $\vG_{S^{-1}}$ in \emph{a compact and structural way}, Auto-Diff may  perform 
too many Hessian-vector products to obtain the entries in $\vG_{S^{-1}}$ even when we allow Auto-Diff to compute $\vngrad_{\gparam}$ on the fly.

 \item
 It is also unclear whether the Jacobian matrix $\big[ \nabla_{\gparam} \lparam \big]$ is sparse even when the parameter $\lparam$ is sparse. 

\item
 As demonstrated  by \citet{lin2020handling}, the indirect method via Auto-Diff could be inefficient and numerically unstable for matrix parameters such as multivariate Gaussian cases with full precision $\gparam=\vS$.
 
\end{itemize}

The flexibility of our approach allows us to freely use either the indirect method (see Eq  \eqref{eq:uni_ef_nd}) or the direct method (Eq \eqref{eq:ngrad_local}) to compute natural gradients.
By using a proper local parameterization,
we can directly compute the natural-gradient
$\vF_{\lparam}(\lparam_0)^{-1} \vngrad_{\lparam_0}$  without computing the Jacobian matrix.
As shown in the main text, our update recovers the direct method suggested by \citet{lin2020handling}.
Moreover, we can easily exploit a sparse structure in a matrix parameter as discussed in Sec. 
\ref{sec:tri_group} of the main text. Our structured updates also reduce the number of Hessian-vector products.

The indirect method is also related to the Riemannian trivialization method \citep{lezcano2019trivializations}, where
the unconstrained transform is considered as
a push-forward map.
In the  trivialization method, the authors suggest doing a unconstrained transform and then performing {\color{red}\emph{Euclidean gradient descent}} in the
trivialized (unconstrained) space.
Unfortunately, the update via a trivialization (e.g., Euclidean gradient descent in a unconstrained space) can converge very slowly as shown in our experiments (see Figure
\ref{fig:a} in the main text).
In variational inference, the Riemannian trivialization method is known as the black-box variational inference \citep{ranganath2014black}. \citet{khan2017conjugate,lin2019fast} demonstrate that natural-gradient variational inference  converges faster than block-box variational inference.

The Riemannian trivialization method is different from the natural-gradient transform method suggested by
\citet{salimbeni2018natural}. 
In the  method of \citet{salimbeni2018natural}, the authors suggest using a unconstrained global parameterization and then performing {\color{red} \emph{natural gradient descent}} in the unconstrained space. In other words, 
the  method of \citet{salimbeni2018natural} uses the Fisher-Rao metric while 
the Riemannian trivialization suggested by \citet{lezcano2019trivializations} does not.
As shown in Appx.~\ref{app:uni_ef},
our approach contains the method of \citet{salimbeni2018natural} as a special case.

\section{Finite Mixture of Gaussians}
\label{app:mog}
In this appendix, we consider the following Gaussian mixture  distribution $q$ with $K$ components.
\begin{align*}
q(\vlat|\gparam) = \frac{1}{K} \sum_{k=1}^{K} \gauss(\vlat|\vmu_k, \vS_k^{-1})
\end{align*} where $\gparam=\{\vmu_k, \vS_k\}_{k=1}^{K=1}$ and $\vS_k$ is the precision matrix of the $k$-th Gaussian component.

As discussed in \citet{lin2019fast}, the FIM of $q(\vlat|\lparam)$ can be singular. Therefore, Assumption 1 is not satisfied.

We define $\varpar_{\mix_k}=\log(\frac{\pi_k}{\pi_K})=0$, where $\pi_k = \frac{1}{K}$.
However, we can consider the Gaussian mixture as the marginal distribution of the following joint distribution such that $\int  q(\vlat,\mix | \gparam) d\mix = q(\vlat|\gparam) $.
\begin{align*}
q(\vlat,\mix|\gparam) &= q(\mix|\vvarpar_\mix) q(\vlat|\mix, \gparam) \\
q(\mix|\vvarpar_\mix) &=  \exp( \sum_{k=1}^{K-1} \mathbb{I}(\mix=k) \varpar_{\mix_k} - A_\mix(\vvarpar_\mix) ) \\
q(\vlat|\mix, \gparam) &=
\exp\Big( \sum_{k=1}^{K} \mathbb{I}(\mix=k)\sqr{ -\half  \vlat^T\vS_k\vlat + \vlat^T\vS_k\vmu_k} - A_\lat(\gparam,\mix)\Big) 
\end{align*} where 
$\mixCompA(\vmu_k, \vS_k)=\half\sqr{ \vmu_{k}^T\vS_{k}\vmu_{k} - \log \left|\vS_{k}/(2\pi)\right| }$,
$A_\lat(\gparam,\mix)= \sum_{k=1}^{K} \mathbb{I}(\mix=k) \mixCompA(\vmu_k, \vS_k)$,
$A_\mix(\vvarpar_\mix) = \log (1+ \sum_{k=1}^{K-1} \exp(\varpar_{\mix_k}) ) $.

As discussed in \citet{lin2019fast}, the FIM of the joint distribution $q(\vlat,\mix|\gparam)$ is not singular.
To solve a variational inference problem, \citet{lin2019fast} consider the following problem with $\gamma=1$ in Eq \eqref{eq:problem}.
\begin{align*}
   \min_{q(\text{\vlat,\mix})\in\mathcal{Q}} \Unmyexpect{q(\text{\vlat,\mix})} \sqr{ \ell(\vlat) } - \gamma\entropy (q(\vlat)) ,
  \end{align*} where we use the entropy of the marginal distribution $q(\vlat)$.
This approach has been  studied  by \citet{agakov2004auxiliary}.
  
This formalization allows us to relax Assumption 1 and use the joint FIM instead.
\citet{lin2019fast} further show that the joint FIM is block-diagonal for each component.

Therefore,
we use the following parameterizations:
\begin{equation*}
    \begin{split}
        \gparam &:= \crl{\vmu_k \in\real^p, \,\,\, \vS_k \in \mathcal{S}_{++}^{p\times p} }_{k=1}^{K} \,\,\,\\
        \aparam &:= \crl{ \vmu_k \in\real^p, \,\,\, \vB_k \in\mathrm{Gl}^{p\times p} }_{k=1}^{K} \\
        \lparam &:= \crl{ \vdelta_k \in\real^p, \,\,\, \vM_k \in\mathcal{S}^{p\times p}  }_{k=1}^{K}.
    \end{split}
\end{equation*} and maps are defined as
\begin{equation*}
    \begin{split}
      \vpsi(\aparam) &= \crl{\vpsi_k(\aparam_k)}^{K}_{k=1} \\
  \vphi_{\aparam_t}(\lparam) &=\crl{\vphi_{k,\aparam_t}(\lparam_k)}_{k=1}^{K} \\    
        \crl{ \begin{array}{c} \vmu_k \\ \vS_k \end{array} } &= \vpsi_k(\aparam_k) :=  \crl{ \begin{array}{c} \vmu_k \\ \vB_k\vB_k^\top \end{array} } \\
        \crl{ \begin{array}{c} \vmu_k \\ \vB_k \end{array} } &= \vphi_{k,\aparam_t}(\lparam_k) :=  \crl{ \begin{array}{c} \vmu_{k,t} + \vB_{k,t}^{-T} \vdelta_k \\ \vB_{k,t} \vh (\vM_k) \end{array} }.
    \end{split}
\end{equation*} where 
$\vB_{k,t}$ denotes the value of $\vB_k$ at iteration $t$ and 
$\aparam_t = \crl{\vmu_{k,t}, \vB_{k,t}}_{k=1}^{K}$.

We can show that Assumption 2 is also satisfied as discussed in Gaussian cases (see Appendix \ref{sec:gauss_prec}).

Natural gradients w.r.t. $\vdelta_k$ and   $\vM_k$ can be computed as below, which is similar to \eqref{eq:gauss_prec_ng_sym}.
\begin{align}
\vngrad_{\delta_k} = \frac{1}{\pi_k} \vB_{k,t}^{-1} \nabla_{\mu_k} {\cal L},\,\,\,\,
\vngrad_{M_k} = -\frac{1}{\pi_k}\vB_{k,t}^{-1} \big[\nabla_{\Sigma_k} {\cal L}\big] \vB_{k,t}^{-T}
\end{align}
where ${\cal L}:=  \Unmyexpect{q(\text{\vlat,\mix})} \sqr{ \ell(\vlat) } - \gamma\entropy (q(\vlat)) $ and $\pi_k = \frac{1}{K}$.

Therefore, our update for the $k$ Gaussian component is
\begin{align}
\vmu_{k,t+1} & \leftarrow \vmu_{k,t} - \frac{\beta}{\pi_k} \vB_{k,t}^{-T}  \vB_{k,t}^{-1} \nabla_{\mu_k} {\cal L} \nonumber \\
\vB_{k,t+1} & \leftarrow \vB_{k,t} \vh( \frac{\beta}{\pi_k}  \vB_{k,t}^{-1} \big[\nabla_{\Sigma_k} {\cal L}\big] \vB_{k,t}^{-T} ) \label{eq:mog_prec_exp_updates}
\end{align} where $\pi_k = \frac{1}{K}$.

Euclidean gradients $\nabla_{\mu_k} {\cal L}$ and $\nabla_{\Sigma_k} {\cal L}$ can be computed as suggested by \citet{lin2019fast}, where
we use second-order information to compute $\nabla_{\Sigma_k} {\cal L}$.
\citet{lin2020handling} also show that we can compute $\nabla_{\Sigma_k} {\cal L}$ by first-order information if  second-order information is not available.

\begin{align*}
 \nabla_{\mu_k} {\cal L} &= \Unmyexpect{q(\lat)}{\sqr{ \pi_k \delta_k \nabla_\lat b(\vlat) } }\\
\nabla_{\Sigma_k} {\cal L} &= \half \Unmyexpect{q(\lat)}{\sqr{ \pi_k \delta_k \nabla_\lat^2 b(\vlat) } }\\
&= \half \Unmyexpect{q(\lat)}{\sqr{ \pi_k \delta_k \vS_k (\vlat-\vmu_k) \nabla_\lat^T b(\vlat) } }
\end{align*}   where
$\delta_k:=\gauss(\vlat|\vmu_k,\vS_k)/ \sum_{c=1}^{K}\pi_c \gauss(\vlat|\vmu_c,\vS_c)$,
$b(\vlat) := \ell(\vlat) + \gamma \log q(\vlat|\gparam)$.

\section{Matrix Gaussian for Matrix Weights in Deep Learning}
\label{app:mat_gauss}
In this appendix, we consider a matrix Gaussian for layer-wise matrix weights in a neural network, where a precision form will be used.
\begin{align*}
   \mgauss(\vLat|\vE,\vS_U^{-1},\vS_V^{-1}) := \gauss(\textrm{vec}(\vLat)| \textrm{vec}(\vE), \vS^{-1})
\end{align*} where the precision $\vS=\vS_V \otimes \vS_U$ has a Kronecker form,  $\vLat \in {\real}^{d \times p}$ is a matrix,
$\vS_V \in {\cal S}_{++}^{p \times p}$, $\vS_U \in {\cal S}_{++}^{d \times d}$, and $\otimes$ denotes the Kronecker product.

In this case, Assumption 1 is not satisfied since the FIM of a matrix Gaussian is singular due to the cross terms between $\vS_U$ and $\vS_V$ in the FIM.
However, a block-diagonal approximation for the FIM is non-singular.
This approximation has been used in many works such as \citet{tran2020bayesian,glasmachers2010exponential,lin2019fast}.
Therefore, we relax Assumption 1 and use the block-diagonal approximation of the FIM instead. The update is known as simultaneous block coordinate (natural-gradient) descent in optimization.

We consider the following optimization problem for NNs with $L_2$ regularization.
\begin{align*}
    \min_{\gparam\in\gspace} \myexpect_{q(\text{\vLat}|\gparam)} \sqr{ \ell(\vLat) +  \frac{\alpha}{2} \mathrm{Tr}(\vLat^T \vLat) } - \gamma\entropy (q(\vLat|\gparam))
\end{align*}
where $q(\vLat) = \prod_l q(\vLat_l)$ and for each layer $l$, $q(\vLat_l)$ is a matrix Gaussian distribution with precision matrix  $\vS_l= \vS_{l,V} \otimes \vS_{l,U} $.

For simplicity, we only consider one layer and drop the layer index $l$.

Let's consider a global parameterization $\gparam=\{\vE, \vS_U, \vS_V \}$
We use the following parameterizations:
\begin{equation*}
    \begin{split}
        \gparam &:= \crl{\vE \in \real^{d \times p}, \,\,\, \vS_V \in \mathcal{S}_{++}^{p\times p}, \,\,\, \vS_U \in \mathcal{S}_{++}^{d\times d} }\,\,\,\\
        \aparam &:= \crl{ \vE \in \real^{d \times p}, \,\,\, \vA \in\mathrm{GL}^{p\times p}, \,\,\, \vB \in\mathrm{GL}^{d\times d} } \\
        \lparam &:= \crl{ \vDelta \in \real^{d \times p}, \,\,\, \vM \in\mathcal{S}^{p\times p}, \,\,\, \vN \in\mathcal{S}^{d\times d}  }.
    \end{split}
\end{equation*} and maps:
\begin{equation*}
    \begin{split}
        \crl{ \begin{array}{c} \vE \\ \vS_V \\ \vS_U \end{array} } &= \vpsi(\aparam) :=  \crl{ \begin{array}{c} \vE \\ \vA\vA^\top \\ \vB\vB^\top \end{array} } \\
        \crl{ \begin{array}{c} \vE \\ \vA \\ \vB \end{array} } &= \vphi_{\aparam_t}(\lparam) :=  \crl{ \begin{array}{c} \vE_t + \vB_t^{-T} \vDelta \vA_t^{-1} \\ \vA_t \vh (\vM) \\ \vB_t \vh (\vN) \end{array} }.
    \end{split}
\end{equation*}

Thanks to this parameterization, it is also easy to generate samples from a matrix Gaussian $\mgauss(\vLat|\vE,\vS_U^{-1},\vS_V^{-1})$ as
\begin{align*}
\vLat = \vE + \vB^{-T} \mathrm{Mat}(\vz) \vA^{-1} 
\end{align*} where $\vz \sim \gauss(\vz|\mathbf{0},\vI)$.

The block-diagonal approximation of the FIM under the local parameterization $\lparam$ is given below.
Note that we also numerically verify the following computation of FIM by Auto-Diff.
\begin{align}
 \vF_{\lparam}(\lparam_0) =\begin{bmatrix}
 \vI_{\Delta} & \mathbf{0} & \mathbf{0} \\
  \mathbf{0} & 2 d \vI_{M} & {\color{red} \mathbf{0}} \\
  \mathbf{0} & {\color{red} \mathbf{0}} & 2 p \vI_{N}  
                         \end{bmatrix}
                         \label{eq:block_approx_fim}
\end{align} where the  red terms are set to be zero due to the block-diagonal approximation while the black terms are obtained from the exact FIM.

Thanks to the block-diagonal approximation of the FIM,
we can show that Assumption 2 is satisfied for each parameter block by holding the remaining blocks fixed.

Now, we discuss how to compute Euclidean gradients w.r.t. local parameterization $\lparam$.
Since each matrix Gaussian $\mgauss(\vLat|\vE,\vS_U^{-1},\vS_V^{-1})$ can be re-expressed as a vector Gaussian $\gauss(\vlat| \vmu, \vS^{-1})$,
The Euclidean gradients w.r.t. global parameter $\gparam_{\text{vec}}=\{\vmu, \vS\}$ of the vector Gaussian are
\begin{align*}
\vg_{\mu} &= \alpha \vmu + \Unmyexpect{\gauss(\text{\vlat}|\gparam_{\text{vec}})} \sqr{ \nabla_\lat \ell(\vlat) } \\
\vg_{\Sigma} &= \half \big( \alpha \vI_\Sigma  +  \Unmyexpect{\gauss(\text{\vlat}|\gparam_{\text{vec}})} \sqr{ \nabla_\lat^2 \ell(\vlat) } -  \gamma \vS \big)
\end{align*} where $\vlat=\mathrm{vec}(\vLat)$, $\vmu=\mathrm{vec}(\vE)$, $\vSigma=\vS^{-1}=\vS_V^{-1} \otimes \vS_U^{-1}$.

To avoid computing the Hessian $ \nabla_\lat^2 \ell(\vlat)$, we use the per-example Gauss-Newton approximation \citep{graves2011practical,osawa2019practical} as
\begin{align*}
\vg_{\Sigma} \approx \half \big( \alpha \vI_\Sigma  +  \Unmyexpect{\gauss(\text{\vlat}|\gparam_{\text{vec}})} \sqr{ \nabla_\lat \ell(\vlat) \nabla_\lat^T \ell(\vlat) } - \gamma \vS \big)
\end{align*}

Recall that
\begin{align*}
\vE & = \vE_t + \vB_t^{-T} \vDelta \vA_t^{-1} \\
\vS_V &= \vA_t  \vh(\vM) \vh(\vM)^T \vA_t^T \\
\vS_U &= \vB_t  \vh(\vN) \vh(\vN)^T \vB_t^T 
\end{align*}

Let's denote $\vg= \nabla_\lat \ell(\vlat)$ and $\vG= \nabla_\Lat \ell(\vLat)$, where $\vlat=\mathrm{vec}(\vLat)$ and $\vg=\mathrm{vec}(\vG)$.
By matrix calculus, we have
\begin{align*}
 \vg_{\Delta} \Big|_{\lparam=\mathbf{0}} = \vB_t^{-1} \mathrm{Mat} (\vg_{\mu}) \vA_t^{-T} = 
 \vB_t^{-1} \big(\alpha \vE + \Unmyexpect{q(\text{\vLat}|\gparam)} \sqr{ \nabla_\Lat \ell(\vLat) } \big) \vA_t^{-T} 
 = \vB_t^{-1} \big(\alpha \vE + \Unmyexpect{q(\text{\vLat}|\gparam)} \sqr{ \vG } \big) \vA_t^{-T} 
\end{align*}

Now, we discuss how to compute a Euclidean gradient w.r.t. $\vM$.
By the chain rule, we have
\begin{align*}
\vg_{M_{ij}} \Big|_{\lparam=\mathbf{0}} & = \mathrm{Tr} \big( \big[\nabla_{M_{ij}} \vSigma\big] \vg_{\Sigma} \big) \\
&= -2 \mathrm{Tr} \big( \big[  (\vA_t^{-T} \big[ \nabla_{M_{ij}} \vM \big] \vA_t^{-1} ) \otimes (\vB_t^{-T}\vB_t^{-1}) \big]  \vg_{\Sigma} \big)
\end{align*} where $M_{ij}$ is the entry of $\vM$ at position $(i,j)$.

By the Gauss-Newton approximation of the Hessian, we have
\begin{align*}
\vg_{M_{ij}} \Big|_{\lparam=\mathbf{0}} & \approx 
 - \mathrm{Tr} \big( \big[  (\vA_t^{-T} \big[ \nabla_{M_{ij}} \vM \big] \vA_t^{-1} ) \otimes (\vB_t^{-T}\vB_t^{-1}) \big]  \big( \alpha \vI_\Sigma  +  \Unmyexpect{\gauss(\text{\vlat}|\gparam_{\text{vec}})} \sqr{ \nabla_\lat \ell(\vlat) \nabla_\lat^T \ell(\vlat) } - \gamma \vS_t \big) \big)
\end{align*}

Let's consider the first term in the approximated $\vg_\Sigma$.
\begin{align*}
 - \mathrm{Tr} \big( \big[  (\vA_t^{-T} \big[ \nabla_{M_{ij}} \vM \big] \vA_t^{-1} ) \otimes (\vB_t^{-T}\vB_t^{-1}) \big]   \alpha \vI_\Sigma  \big)
 = -\alpha \mathrm{Tr} (\vB_t^{-T}\vB_t^{-1}) \mathrm{Tr}(  \vA^{-1} \vA^{-T} \big[ \nabla_{M_{ij}} \vM \big])
\end{align*}

Now, we consider the second term in the approximated $\vg_\Sigma$.
\begin{align*}
 &- \mathrm{Tr} \big( \big[  (\vA_t^{-T} \big[ \nabla_{M_{ij}} \vM \big] \vA_t^{-1} ) \otimes (\vB_t^{-T}\vB_t^{-1}) \big]     \Unmyexpect{\gauss(\text{\vlat}|\gparam_{\text{vec}})} \sqr{ \nabla_\lat \ell(\vlat) \nabla_\lat^T \ell(\vlat) }  \big) \\
 = & - \Unmyexpect{\gauss(\text{\vlat}|\gparam_{\text{vec}})} \sqr{ \mathrm{Tr} \big( \vg^T  \big[  (\vA_t^{-T} \big[ \nabla_{M_{ij}} \vM \big] \vA_t^{-1} ) \otimes (\vB_t^{-T}\vB_t^{-1}) \big]      \vg    \big) } \\
 = & - \Unmyexpect{q(\text{\vlat}|\gparam)} \sqr{ \mathrm{Tr} \big( \mathrm{vec}(\vG )^{T} \big[  (\vA_t^{-T} \big[ \nabla_{M_{ij}} \vM \big] \vA_t^{-1} ) \otimes (\vB_t^{-T}\vB_t^{-1}) \big]  \mathrm{vec}(\vG ) \big) } 
\end{align*}

Using the identity $(\vB^T \otimes \vA) \mathrm{vec}(\vX) = \mathrm{vec}(\vA\vX\vB)$, we can simplify the above expression as
\begin{align*}
 &- \mathrm{Tr} \big( \big[  (\vA_t^{-T} \big[ \nabla_{M_{ij}} \vM \big] \vA_t^{-1} ) \otimes (\vB_t^{-T}\vB_t^{-1}) \big]     \Unmyexpect{\gauss(\text{\vlat}|\gparam_{\text{vec}})} \sqr{ \nabla_\lat \ell(\vlat) \nabla_\lat^T \ell(\vlat) }  \big) \\
 = & - \Unmyexpect{q(\text{\vlat}|\gparam)} \sqr{ \mathrm{Tr} \big( \mathrm{vec}(\vG )^{T} \big[  (\vA_t^{-T} \big[ \nabla_{M_{ij}} \vM \big] \vA_t^{-1} ) \otimes (\vB_t^{-T}\vB_t^{-1}) \big]  \mathrm{vec}(\vG ) \big) } \\
 = & - \Unmyexpect{q(\text{\vlat}|\gparam)} \sqr{ \mathrm{Tr} \big( \mathrm{vec}(\vG )^{T} \mathrm{vec}\big[ (\vB_t^{-T}\vB_t^{-1}) \vG (\vA_t^{-T} \big[ \nabla_{M_{ij}} \vM^T \big] \vA_t^{-1} )  \big]   \big) } \\
 = & - \Unmyexpect{q(\text{\vlat}|\gparam)} \sqr{ \mathrm{Tr} \big( \vG^{T}  (\vB_t^{-T}\vB_t^{-1}) \vG (\vA_t^{-T} \big[ \nabla_{M_{ij}} \vM^T \big] \vA_t^{-1} )     \big) } \\
 = & - \Unmyexpect{q(\text{\vlat}|\gparam)} \sqr{ \mathrm{Tr} \big( \vA_t^{-1} \vG^{T}  \vB_t^{-T}\vB_t^{-1} \vG \vA_t^{-T} \big[ \nabla_{M_{ij}} \vM^T \big]      \big) }  \\
 = & - \Unmyexpect{q(\text{\vlat}|\gparam)} \sqr{ \mathrm{Tr} \big( \vA_t^{-1} \vG^{T}  \vB_t^{-T}\vB_t^{-1} \vG \vA_t^{-T} \big[ \nabla_{M_{ij}} \vM \big]      \big) }  \,\,\, ( \text{ since } \mathrm{Tr}(\vC\vD) = \mathrm{Tr}(\vC^T \vD^T) )
\end{align*} where $\vC:=\vA_t^{-1} \vG^{T}  \vB_t^{-T}\vB_t^{-1} \vG \vA_t^{-T}$, $\vD:=\nabla_{M_{ij}} \vM^T$ and $\vC^T=\vC$.

Finally, we consider the last term in the approximated $\vg_\Sigma$.
\begin{align*}
 &- \mathrm{Tr} \big( \big[  (\vA_t^{-T} \big[ \nabla_{M_{ij}} \vM \big] \vA_t^{-1} ) \otimes (\vB_t^{-T}\vB_t^{-1}) \big]  (-\gamma \vS_t)  \big) \\
 =&\gamma \mathrm{Tr} \big( \big[  (\vA_t^{-T} \big[ \nabla_{M_{ij}} \vM \big] \vA_t^{-1} ) \otimes (\vB_t^{-T}\vB_t^{-1}) \big]  \vS_t  \big) \\
 =&\gamma \mathrm{Tr} \big( \big[  (\vA_t^{-T} \big[ \nabla_{M_{ij}} \vM \big] \vA_t^{-1} ) \otimes (\vB_t^{-T}\vB_t^{-1}) \big]  \big[ (\vA_t\vA_t^T) \otimes (\vB_t\vB_t^T) \big]  \big) \\
 =&\gamma \mathrm{Tr} \big( \big[  (\vA_t^{-T} \big[ \nabla_{M_{ij}} \vM \big] \vA_t^{-1} (\vA_t\vA_t^T)) \otimes (\vB_t^{-T}\vB_t^{-1}  (\vB_t\vB_t^T) \big]  \big) \\
 =&\gamma \mathrm{Tr} \big( \big[  (\vA_t^{-T} \big[ \nabla_{M_{ij}} \vM \big] \vA_t^T) \otimes \vI_B \big]  \big) \\
 =&\gamma \mathrm{Tr} \big( \big[  (\vA_t^{-T} \big[ \nabla_{M_{ij}} \vM \big] \vA_t^T) \big) \mathrm{Tr}\big( \vI_B \big)  \\
 =&\gamma d\mathrm{Tr} \big(  \big[ \nabla_{M_{ij}} \vM \big]  \big)  
\end{align*}

Therefore,  we have the following expression due to the Gauss-Newton approximation.
\begin{align*}
\vg_{M_{ij}} \Big|_{\lparam=\mathbf{0}} & \approx 
  -\alpha \mathrm{Tr} (\vB_t^{-T}\vB_t^{-1}) \mathrm{Tr}(  \vA^{-1} \vA^{-T} \big[ \nabla_{M_{ij}} \vM \big])
  - \Unmyexpect{q(\text{\vlat}|\gparam)} \sqr{ \mathrm{Tr} \big( \vA_t^{-1} \vG^{T}  \vB_t^{-T}\vB_t^{-1} \vG \vA_t^{-T} \big[ \nabla_{M_{ij}} \vM \big]      \big) } 
 + \gamma d\mathrm{Tr} \big(  \big[ \nabla_{M_{ij}} \vM \big]  \big)  
\end{align*}

We can re-express it in a matrix form as
\begin{align*}
\vg_{M} \Big|_{\lparam=\mathbf{0}} & \approx 
  -\alpha \mathrm{Tr} (\vB_t^{-T}\vB_t^{-1})  \vA_t^{-1} \vA_t^{-T} 
  - \Unmyexpect{q(\text{\vlat}|\gparam)} \sqr{  \vA_t^{-1} \vG^{T}  \vB_t^{-T}\vB_t^{-1} \vG \vA_t^{-T} } 
 + \gamma d \vI_M
\end{align*}
Similarly, we can show
\begin{align*}
\vg_{N} \Big|_{\lparam=\mathbf{0}} & \approx 
  -\alpha \mathrm{Tr} (\vA_t^{-T}\vA_t^{-1})  \vB^{-1} \vB^{-T} 
  - \Unmyexpect{q(\text{\vlat}|\gparam)} \sqr{  \vB_t^{-1} \vG  \vA_t^{-T}\vA_t^{-1} \vG^T \vB_t^{-T} } 
 + \gamma p \vI_N
\end{align*}

Our update in terms of the auxiliary parameterization is
\begin{align}
\vE_{t+1}  & \leftarrow \vE_t - \beta \overbrace{ \vB_t^{-T}  \vB_t^{-1}}^{\vS_U^{-1}} \big[ \alpha \vE_t + \Unmyexpect{q(\text{\vLat}|\gparam_t)} \sqr{ \vG } \big)   \big] \overbrace{ \vA_t^{-T} \vA_t^{-1}}^{ \vS_V^{-1} } \nonumber\\
\vA_{t+1} & \leftarrow \vA_t \vh\big[ \frac{\beta}{2d} \crl{ -d\gamma \vI_A + \alpha\mathrm{Tr}( (\vB_t\vB_t^T)^{-1} ) \vA_t^{-1}\vA_t^{-T} + \Unmyexpect{q(\text{\vlat}|\gparam_t)} \sqr{ \vA_t^{-1}  \vG^T  (\vB_t\vB_t^T)^{-1} \vG \vA_t^{-T}} } \big] \nonumber \\
\vB_{t+1} & \leftarrow \vB_t \vh\big[ \frac{\beta}{2p} \Big\{ \underbrace{ -p\gamma \vI_B}_{\text{from the entropy}}  + \underbrace{ \alpha\mathrm{Tr}( (\vA_t\vA_t^T)^{-1} ) \vB_t^{-1}\vB_t^{-T} }_{\text{from the regularization}}+ \underbrace{ \Unmyexpect{q(\text{\vlat}|\gparam_t)} \sqr{ \vB_t^{-1}\vG  (\vA_t\vA_t^T)^{-1} \vG^T \vB_t^{-T} } }_{\text{from the NN loss }} \Big\} \big]  \label{eq:mat_gauss_exp_ngd_aux}
\end{align}

By adding a natural momentum term $\vZ$ \citep{khan18a} and  an exponential weighted step-size $\beta_t=\frac{1-c_2^t}{1-c_1^t}$, we can obtain  the following  update for DNN with the Gauss-Newton approximation.
\begin{align}
\vZ_{t} &\leftarrow (1-c_1) \big[ \alpha \vE_t + \Unmyexpect{q(\text{\vLat}|\gparam_t)} \sqr{ \vG } \big)   \big] + c_1 \vZ_{t-1} \nonumber \\
\vE_{t+1}  & \leftarrow \vE_t - \beta_t  \vB_t^{-T}  \vB_t^{-1} \vZ_{t}  \vA_t^{-T} \vA_t^{-1} \nonumber \\
\vA_{t+1} & \leftarrow \vA_t \vh\big[ \frac{\beta_t}{2d} \crl{ -d\gamma \vI_A + \alpha\mathrm{Tr}( (\vB_t\vB_t^T)^{-1} ) \vA_t^{-1}\vA_t^{-T} + \Unmyexpect{q(\text{\vlat}|\gparam_t)} \sqr{ \vA_t^{-1}  \vG^T  (\vB_t\vB_t^T)^{-1} \vG \vA_t^{-T}} } \big] \nonumber \\
\vB_{t+1} & \leftarrow \vB_t \vh\big[ \frac{\beta_t}{2p} \crl{ -p\gamma \vI_B  + \alpha\mathrm{Tr}( (\vA_t\vA_t^T)^{-1} ) \vB_t^{-1}\vB_t^{-T} +  \Unmyexpect{q(\text{\vlat}|\gparam_t)} \sqr{ \vB_t^{-1}\vG  (\vA_t\vA_t^T)^{-1} \vG^T \vB_t^{-T} } } \big] 
\label{eq:gauss_newton_matgauss}
\end{align} where 
$\vG= \nabla_\Lat \ell(\vLat)$,
$c_1$ and $c_2$ are fixed to $0.9$ and $0.999$, respectively, as the same used in the Adam optimizer.

The time complexity for our  update above is $O(d^3+p^3)$, which is the same as noisy-KFAC \citep{zhang2018noisy}.
In our approach, the update for $\vA$ ($\vS_V$) and $\vB$ ($\vS_U$) blocks use the \emph{exact} FIM block.
It can be shown that the corresponding updates for
$\vS_V=\vA\vA^T$ and $\vS_U=\vB\vB^T$ blocks also use the \emph{exact} FIM block and our update ensures that $\vS_V$ and $\vS_U$ are always non-singular.
Our approach is different from noisy-KFAC \citep{zhang2018noisy}. In noisy-KFAC,  the FIM of $\vS_V$ and $\vS_U$ are approximated by KFAC. The authors have to use additional damping to ensure that  $\vS_V$ and $\vS_U$ are non-singular.

\subsection{Complexity Reduction}
\label{app:com_red}
A nice property of our update in \eqref{eq:gauss_newton_matgauss} is that we can easily incorporate extra structures to reduce the time and space complexity.
As shown in Appendix \ref{app:group}, we can further exploit group-structures both in $\vA$ and $\vB$ so that
the precision $\vS=\vS_V \otimes \vS_U= (\vA\vA^T) \otimes (\vB\vB^T)= (\vA \otimes \vB ) (\vA \otimes \vB)^T$ has a \emph{low-rank} Kronecker structure to further reduce the computational complexity. Note that 
the Kronecker product of two matrix groups such as $\vA \otimes \vB $ is also a matrix group closed under the matrix multiplication.
Therefore, $\vA \otimes \vB$  is a \emph{Kronecker product group} when $\vA$ and $\vB$ are  matrix groups.

Recall that the time complexity of Adam for a matrix weight $\vLat \in \real^{d  \times p}$  is linear $O(d p)$.
If a block triangular group structure (see Appendix \ref{app:tria_group})
is exploited in both $\vA$ and $\vB$, the time complexity of our update reduces to $O(k d p)$ from $O(d^3+p^3)$, where $0< k < \min(d,p)$ is a sparsity parameter for the group defined in Appendix \ref{app:group}. In this case, our update has a linear time complexity like Adam, which is much faster than noisy-KFAC. Although we present the update based on the Gauss-Newton approximation of the Hessian, 
our update with the triangular group structure can be easily applied to the case with Hessian information if each Hessian has a Kronecker form such as a example about layer-wise weight matrices in a NN discussed in the next section.

Notice that our update can be automatically parallelized by Auto-Diff since our update only use basic linear algebra operations (i.e., matrix multiplication, low-rank matrix solve, and the Einstein summation) , which is more efficient than Newton-CG type updates, where a sequential conjugate-gradient (CG) step is used at each iteration.

\subsection{A Layer-wise Hessian and its Approximation}
\label{app:hessian_dnn}

We consider the following loss function parameterized by a MLP/CNN evaluated at one data point.
We will show that a layer-wise Hessian of matrix weights has a Kronecker form. This result has been exploited in  \citet{dangel2020modular,chen2019ea}. 
For simplicity, we only consider the matrix weight $\vW$ at the input layer of a MLP. It is easy to extend this computation to other layers and CNN.
\begin{align*}
\ell (\vW) &= c(  f(\vW\vx))
\end{align*} where $x$ is a single data point with shape $p \times 1$, $c(\cdot)$ is a function that returns a scalar output, 
and $\vW$ is the matrix weight at the input layer with shape $d \times p$.

 We assume $f(\vz)$ is an element-wise $C^2$-smooth activation function (e.g., the $\text{tanh}$ function).
Let $\vu:=\vW \vx$ and $\vv :=f(\vu) = f(\vW\vx)$

By the chain rule, it is easy to check that
\begin{align*}
 \nabla_{W} \ell(\vW) &= \big[ \nabla_v \ell  \big] \big[ \nabla_W \vv ]\\
 &= \big[ \underbrace{\big[ \nabla_v \ell  \big]}_{d \times 1} \odot \underbrace{ f'(\vu)}_{ d \times 1} \big] \underbrace{ \vx^T}_{ 1 \times p}
\end{align*} where $\odot$ denotes the element-wise product.

Let $\vW_{i,:}$ denotes the $i$-th row of the matrix $\vW$. We know that the shape of $\vW_{i,:}$ is $1 \times p$.

Now, we can show that the Hessian is a Kronecker product.
\begin{align*}
 \nabla_{W_{i,:}} \nabla_{W_{k,:}} \ell(\vW) & = {\cal I}(i == k) \big[ \nabla_{v_i} \ell \big] f''(u_i) \vx \vx^T + \big[ \nabla_{v_i} \nabla_{v_j} \ell \big] f'(u_k) f'(u_i) \vx\vx^T \\
 &= \underbrace{\Big( {\cal I}(i == k) \big[ \nabla_{v_i} \ell \big] f''(u_i)  + \big[ \nabla_{v_i} \nabla_{v_k} \ell \big] f'(u_k) f'(u_i) \Big)}_{\text{a scalar}} \vx \vx^T
\end{align*}

We assume $\mathrm{vec}$ uses the row-major order.
Therefore, if we use $\vw = \mathrm{vec}(\vW)$ to denote a vector representation of $\vW$, the Hessian w.r.t. $\vw = \mathrm{vec}(\vW)$ with shape $dp \times 1$ is
\begin{align*}
\nabla_{w}^2 \ell = \underbrace{ \vA }_{ d \times d} \underbrace{ \otimes }_{\text{Kronecker Product}} \underbrace{(\vx\vx^T)}_{p \times p} 
\end{align*} where $\vA $ is a symmetric matrix with entry $A_{ik}={\cal I}(i == k) \big[ \nabla_{v_i} \ell \big] f''(u_i)  + \big[ \nabla_{v_i} \nabla_{v_k} \ell \big] f'(u_k) f'(u_i)$.

Now, we discuss the  Gauss-Newton approximation of the Hessian.
Note that
\begin{align*}
 \nabla_{W_{i,:}} \ell(\vW) &= \underbrace{\big[ \big[ \nabla_{v_i} \ell  \big]  f'(u_i) \big] }_{\text{a scalar}} \vx^T
\end{align*} where $\odot$ denotes the element-wise product.
\begin{align*}
  \nabla_{W_{k,:}}^T \ell(\vW) \big[  \nabla_{W_{i,:}} \ell(\vW) \big] = 
\underbrace{ \big[  \nabla_{v_i} \ell  \big]  f'(u_i) \big[ \nabla_{v_k} \ell  \big]  f'(u_k) }_{\text{a scalar}} \vx\vx^T
\end{align*}

Therefore, the Gauss-Newton approximation in term of $\vw$ can be re-expressed as
\begin{align*}
 \vB \otimes \big(\vx\vx^T\big)
\end{align*} where $\vB $ is a symmetric matrix with entry $B_{ik}= \big[\nabla_{v_i} \ell \nabla_{v_k} \ell  \big]  f'(u_k) f'(u_i)  $.

From the above expression, we can clearly see that the Gauss-Newton approximation ignores diagonal terms involving $ f''(u_i)$ and approximates 
$\big[ \nabla_{v_i} \nabla_{v_k} \ell \big] $ by $\big[\nabla_{v_i} \ell \nabla_{v_k} \ell  \big]$.

\section{Group Structures}
\label{app:group}

In this section, we use the Gaussian example with square-root precision form to illustrate group structures. 
\subsection{Block Triangular Group}
\label{app:tria_group}

\subsubsection{Proof of Lemma \ref{lemma:block_tri_gp}}
\begin{myproof}
Now, we show that ${\cal{B}_{\text{up}}}(k)$ is a matrix group.
\begin{align*}
{\cal{B}_{\text{up}}}(k)  = \Big\{ 
\begin{bmatrix}
\vB_A &  \vB_B  \\
 \mathbf{0} & \vB_D
      \end{bmatrix} \Big| & \vB_A \in \mathrm{GL}^{k \times k},\,\,
 \vB_D  \in{\cal D}^{d_0 \times d_0}_{++}  \Big\}
\end{align*} 
(0) It is clear that matrix multiplication is an associate  product.

(1) It is obvious that $\vI = \begin{bmatrix}
\vI_A &  \mathbf{0}  \\
 \mathbf{0} & \vI_D
      \end{bmatrix} \in {\cal{B}_{\text{up}}}(k)$ since $\vI_A \in \mathrm{GL}^{k \times k}$ and $\vI_D \in {\cal D}^{d_0 \times d_0}_{++}$.

(2) For any $\vB \in {\cal{B}_{\text{up}}}(k)$, we have
\begin{align*}
\vB^{-1} =  \begin{bmatrix}
\vB_A^{-1} & -\vB_A^{-1} \vB_B \vB_D^{-1}    \\          
\mathbf{0}& \vB_D^{-1}  
            \end{bmatrix} \in {\cal{B}_{\text{up}}}(k) 
\end{align*} since $\vB_A^{-1} \in \mathrm{GL}^{k \times k}$ and $\vB_D^{-1} \in {\cal D}^{d_0 \times d_0}_{++}$.

(3) For any $\vB,\vC \in {\cal{B}_{\text{up}}}(k)$, the matrix product is
\begin{align*}
\vB \vC =  
\begin{bmatrix}
\vB_A &  \vB_B  \\
 \mathbf{0} & \vB_D
      \end{bmatrix}
      \begin{bmatrix}
\vC_A &  \vC_B  \\
 \mathbf{0} & \vC_D
      \end{bmatrix}=
      \begin{bmatrix}
   \vB_A \vC_A   & \vB_A \vC_B + \vB_B \vC_D \\
   \mathbf{0}   & \vB_D \vC_D
      \end{bmatrix} \in  {\cal{B}_{\text{up}}}(k)
\end{align*} since $ \vB_A \vC_A \in \mathrm{GL}^{k \times k}$ and 
$\vB_D \vC_D \in  {\cal D}^{d_0 \times d_0}_{++} $.

\end{myproof}

\subsubsection{Proof of Lemma \ref{lemma:block_tri_assp1}}
\begin{myproof}
For any $\vM \in {\cal{M}_{\text{up}}}(k)$, we have
\begin{align*}
 \vM =\begin{bmatrix}
\vM_A &  \vM_B  \\
 \mathbf{0} & \vM_D
      \end{bmatrix},
\end{align*} where
$\vM_A$ is symmetric and $\vM_D$ is diagonal.
Therefore,
\begin{align*}
 \vh(\vM) &= \vI + \vM + \half \vM^2 \\
 &=\begin{bmatrix}
\vI_A + \vM_A &  \vM_B  \\
 \mathbf{0} & \vI_D + \vM_D
      \end{bmatrix} + \half 
      \begin{bmatrix}
 \vM_A &  \vM_B  \\
 \mathbf{0} &  \vM_D
      \end{bmatrix}
            \begin{bmatrix}
 \vM_A &  \vM_B  \\
 \mathbf{0} &  \vM_D
      \end{bmatrix}\\
     &=
     \begin{bmatrix}
\vI_A + \vM_A &  \vM_B  \\
 \mathbf{0} & \vI_D + \vM_D
      \end{bmatrix} + \half 
      \begin{bmatrix}
 \vM_A^2 &   \vM_A \vM_B + \vM_B \vM_D\\
 \mathbf{0} & \vM_D^2
      \end{bmatrix}\\
     &=
           \begin{bmatrix}
 \vI_A+\vM_A +\half \vM_A^2 & \vM_B+\half(  \vM_A \vM_B + \vM_B \vM_D )\\
 \mathbf{0} &\vI_D +\vM_D +\half \vM_D^2
      \end{bmatrix} \in {\cal{B}_{\text{up}}}(k)
\end{align*}

Since $\vM_A$ is symmetric, we have
$ \vI_A+\vM_A +\half \vM_A^2  = \half\big( \vI_A + (\vI_A+\vM_A) (\vI_A+\vM_A)^T \big) \succ \mathbf{0}$ is invertible and symmetric.
Similarly, $ \vI_D+\vM_D +\half \vM_D^2 $ is diagonal and invertible.

Thus, $ \vh(\vM)  \in {\cal{B}_{\text{up}}}(k)$. Moreover, the determinant $|\vh(\vM)|>0$
\end{myproof}

\subsubsection{Proof of Lemma \ref{lemma:block_tri_assp2}}
\label{app:proof_up_group_lemma}
\begin{myproof}
we consider the following parametrization for the Gaussian $\gauss(\vlat|\vmu,\vS^{-1})$, where the precision $\vS$ belongs to  a sub-manifold of $\mathcal{S}_{++}^{p\times p}$, auxiliary parameter $\vB$ belongs to ${\cal{B}_{\text{up}}}(k)$, and local parameter $\vM$ belongs to ${\cal{M}_{\text{up}}}(k)$,
\begin{equation*}
    \begin{split}
       \gparam &:= \crl{\vmu \in\real^p, \,\,\, \vS=\vB \vB^T \in \mathcal{S}_{++}^{p\times p} \,\,\,|\,\,\, \vB \in {\cal{B}_{\text{up}}}(k) }, \,\,\,\\
        \aparam &:= \crl{ \vmu \in\real^p, \,\,\, \vB \in {\cal{B}_{\text{up}}}(k) }, \\
        \lparam &:= \crl{ \vdelta\in\real^p, \,\,\, \vM \in {\cal{M}_{\text{up}}}(k)   }.
    \end{split}
\end{equation*}
The map $\vpsi \circ \vphi_{\aparam_t}(\lparam)$ at $\aparam_t := \{\vmu_t, \vB_t\}$ is chosen as below, which is the same as     \eqref{eq:gauss_xnes_prec}

\begin{align*}
\begin{bmatrix}
 \vmu \\
\vS 
\end{bmatrix}=
\vpsi \circ \vphi_{\aparam_t}\big(
\begin{bmatrix}
 \vdelta \\
\vM 
\end{bmatrix}
\big) = 
\begin{bmatrix}
\vmu_t + \vB_t^{-T} \vdelta \\
\vB_t  \vh(\vM) \vh(\vM)^T \vB_t^T
\end{bmatrix}
\end{align*}

As shown in Appendix \ref{app:ng_block_triangular_gauss}, the FIM is non-singular. Therefore, Assumption 1 is satisfied.

In Appendix \ref{app:ng_block_triangular_gauss}, we show that $\vM$ can be decomposed as
\begin{align*}
\vM =  \vM_{\text{diag}} + \vM_{\text{up}} + \vM_{\text{up}}^T + \vM_{\text{asym}}
\end{align*}
Let ${\cal I}_{\text{up}}$, ${\cal I}_{\text{diag}}$, ${\cal I}_{\text{asym}}$ be the index set of the non-zero entries of $\vM_{\text{up}}$, $\vM_{\text{diag}}$, and $\vM_{\text{asym}}$ respectively.

Now, we can  show that Assumption 2 is also satisfied.
This proof is similar to the one at \eqref{eq:chol_map}.
The key idea is to use an effective representation to represent $\gparam$ and $\lparam$.

Now, let's consider the global matrix parameter.
Let  ${\cal S}_1=\{\vB\vB^T | \vB \in {\cal B}_{\text{up}}(k) \}$, which represents the parameter space of the global matrix parameter.
Consider another set
\begin{align}
{\cal S}_2=\{ \vU \vU^T |  \vU = \begin{bmatrix}
        \vU_{A}  & \vU_B \\
      \mathbf{0}  & \vU_D
       \end{bmatrix} \} \label{eq:global_low},
\end{align}
where $\vU_A \in \real^{k \times k}$ is an \emph{upper-triangular} and invertible matrix, $\vU_D$ is an invertible and diagonal matrix and $\vU$ has \emph{positive} diagonal entries.
We will first show that ${\cal S}_1 = {\cal S}_2$ and therefore, ${\cal S}_2$ represents the sub-manifold.
The key reason is that $\vU$ can be used as a global parameter while $\vB$ does not. Recall that in $\vB$ is used as an auxiliary parameter, which could be over-parameterized. Note that a global parameter should have the same degree of freedoms as a local parameter.
It is easy to verify that ${\cal S}_2$ and ${\cal M}_{\text{up}}(k)$ both have $(k+1)k/2+(p-k)k+(p-k)=(k+1)(p-k/2)$ degrees of freedom.

      We will see that $\vU$ is indeed the output of the upper-triangular version of the Cholesky method \citep{4114067}, denoted by $\mathrm{CholUP}$. 
      In other words, if $\vS=\vU_1 \vU_1^T \in {\cal S}_2$ and $\vU_2 = \mathrm{CholUP}(\vS)$, we will show $\vU_1 = \vU_2$.
This Cholesky algorithm takes a positive-definite matrix $\vX$ as an input and returns an upper-triangular matrix $\vW$ with \emph{positive diagonal entries} so that  $\vX=\vW\vW^T$ (e.g., $\vW=\mathrm{CholUP}(\vX)$). Like the original Cholesky method, this method gives a unique decomposition and is $C^1$-smooth w.r.t. its input $\vX$ when $\vX$ is positive-definite.
      
Now, We  show that ${\cal S}_1 = {\cal S}_2$.
It is obvious that ${\cal S}_2 \subset {\cal S}_1$ since by construction $\vU \in {\cal B}_{up}(k)$.
Now, we show that ${\cal S}_1 \subset {\cal S}_2$.
Consider any $\vS \in {\cal S}_1$, it can be expressed as
\begin{align*}
 \vS &= 
  \begin{bmatrix}
        \vB_A & \vB_B\\
        \mathbf{0} & \vB_D
       \end{bmatrix}
        \begin{bmatrix}
        \vB_A & \vB_B\\
        \mathbf{0} & \vB_D
       \end{bmatrix}^T \\
      &= 
        \begin{bmatrix}
        \vB_A \vB_A^T +\vB_B \vB_B^T & \vB_B \vB_D \\
       \vB_D \vB_B^T & \vB_D^2
       \end{bmatrix}
\end{align*}

Since $\vB_D$ is an invertible and diagonal matrix,   $\vd := \mathrm{abs}(\mathrm{diag}(\vB_D)) \odot \mathrm{diag}^{-1}(\vB_D)$ is a vector with entries whose value is either 1 or -1.
Let  $\vU_A : = \mathrm{CholUP}(\vB_A\vB_A^T)$ be an upper-triangular matrix as an output by the upper-triangular version of the Cholesky  method.
Consider the following upper-triangular matrix $\vU$
\begin{align*}
 \vU = \begin{bmatrix}
        \vU_{A}  & \vB_B \mathrm{Diag}(\vd^{-1}) \\
        \mathbf{0} & \mathrm{Diag}(\vd) \vB_D
       \end{bmatrix}
\end{align*} 
We can show  that this $\vU$ has positive diagonal entries.
Moreover, $\vU \vU^T \in {\cal S}_2$.
Note that $\vB_D$ is a diagonal matrix. We can show $\vU\vU^T =\vS$ since
\begin{align*}
\vU\vU^T  = 
\begin{bmatrix}
     \overbrace{   \vU_{A}\vU_A^T}^{\vB_A \vB_A^T}  + \vB_B \overbrace{ \mathrm{Diag}(\vd^{-2})}^{\vI} \vB_B^T & \vB_B \vB_D \\
     \\
        \underbrace{\mathrm{Diag}(\vd) \vB_D \mathrm{Diag}(\vd^{-1})}_{\vB_D } \vB_B^T & \underbrace{ \mathrm{Diag}(\vd) \vB_D \mathrm{Diag}(\vd) \vB_D}_{\vB_D^2}
       \end{bmatrix} =\vS \in  {\cal S}_2
\end{align*}
Therefore, ${\cal S}_1 = {\cal S}_2$ and we now show that  $\vU$ can be used as a global parameterization to represent the sub-manifold.
Since ${\cal S}_1 = {\cal S}_2$, we can use ${\cal S}_2$ to denote the sub-manifold.
Furthermore, $\vU$ is indeed an upper-triangular and invertible matrix with positive diagonal entries, which implies that $\vU$ is a (upper-triangular) Cholesky factor of $\vS \in {\cal S}_2$. Note that the Cholesky decomposition gives a \emph{unique} representation.
Therefore, for any $\vS = \vU\vU^T \in {\cal S}_2$, we have $\vU_2 = \mathrm{CholUP}(\vS)$.

For the local parameter, 
since $\vM \in {\cal M}_{\text{up}}(k)$, we have
\begin{align*}
 \vM = \begin{bmatrix}
        \vM_A & \vM_B\\
        \mathbf{0} & \vM_D
       \end{bmatrix}
\end{align*}
Since $\vM_A$ is symmetric, we can consider the upper-triangular part of $\vM_A$, denoted by $\mathrm{triu}(\vM_A)$.
Therefore, the upper-triangular part of $\vM$ is
\begin{align*}
 \mathrm{triu}(\vM)= \begin{bmatrix}
        \mathrm{triu}(\vM_A) & \vM_B\\
        \mathbf{0} & \vM_D
       \end{bmatrix}
\end{align*}
Consider the vector representation of the non-zero entries of $\mathrm{triu}(\vM)$ denoted by $\vm_{\text{vec}}$.
Similarly, consider the vector representation of the non-zero entries of $\vU$ denoted by $\mathrm{vec}(\vU)$.
The length of $\vm_{\text{vec}}$ is the same as the length of $\mathrm{vec}(\vU)$. Therefore, we can use these two vector representations to represent the global parameter and the local parameter in the structured spaces. Moreover, they have the same degree of freedoms.
The remaining proof can be found at \eqref{eq:chol_map} by using the inverse function theorem and Assumption 1, where we need to use the result that
if $\vS=\vU\vU^T \in {\cal S}_2$ and $\vU_2 = \mathrm{CholUP}(\vS)$, then $\vU=\vU_2$ and $\vS \in {\cal S}_1$.
Moreover, for any positive-definite matrix $\vX$, $\mathrm{CholUP}(\vX)$ is $C^1$-smooth w.r.t. $\vX$, which is as smooth as the original Cholesky 
method.
\end{myproof}

\subsubsection{ Natural Gradient Computation for Structured $\vM$}
\label{app:ng_block_triangular_gauss}

we use a similar technique  discussed in  Appendix \ref{app:sym_m} to deal with the FIM computation w.r.t. an asymmetric $\vM$.
The main idea is to decomposition $\vM$ as a sum of special matrices so that the FIM computation is simple. We also numerically verify the following computation of FIM by Auto-Diff.

Since \begin{align*}
\vM = \begin{bmatrix}
\vM_A & \vM_B \\
\mathbf{0} & \vM_D
      \end{bmatrix} \in {\cal{M}_{\text{up}}}(k),
\end{align*} 
by Lemma \ref{lemma:block_tri_assp1}, $\vh(\vM)$ is invertible for any $\vM \in {\cal{M}_{\text{up}}}(k) $.
Moreover, by the structure of $\vM$,  $|\vh(\vM)|>0$.

Since $\vM_A$ is symmetric, we can re-express the matrix $\vM_A$ as follows. We use a similar decomposition  in Appendix \ref{app:sym_m}.
\begin{align*}
 \vM_A = \vM_{A_\text{up}} + \vM_{A_\text{up}}^T + \vM_{A_\text{diag}},
\end{align*} where $\vM_{A_\text{up}}$ contains the upper-triangular half of $\vM_A$ excluding the diagonal elements, and $\vM_{A_\text{diag}}$ contains the diagonal entries of $\vM_A$. 

We will decompose the $\vM$ as follows
\begin{align*}
\vM =  \vM_{\text{diag}} + \vM_{\text{up}} + \vM_{\text{up}}^T + \vM_{\text{asym}}
\end{align*}
where $\vM_{\text{diag}}$ is a diagonal matrix, $\vM_{\text{asym}}$ is an asymmetric matrix, and $\vM_{\text{low}}$ is a upper-triangular matrix with zero diagonal entries.
\begin{align*}
 \vM_{\text{diag}} = 
 \begin{bmatrix}
\vM_{A_\text{diag}} & \mathbf{0} \\
\mathbf{0} & \vM_D
      \end{bmatrix}\,\,\,\,\,
       \vM_{\text{asym}} = 
 \begin{bmatrix}
 \mathbf{0} & \vM_B \\
\mathbf{0} &\mathbf{0} 
      \end{bmatrix}\,\,\,\,\,
      \vM_{\text{up}} = 
 \begin{bmatrix}
 \vM_{A_\text{up}} & \mathbf{0} \\
\mathbf{0}  &\mathbf{0} 
      \end{bmatrix}\,\,\,\,\,
\end{align*}

Note that  $\vM_{\text{diag}}$, $\vM_{\text{asym}}$, and $\vM_{\text{low}}$ respectively contain the diagonal entries of $\vM$,
the asymmetric entries of $\vM$,
the upper-triangular half of the symmetric part of $\vM$ excluding the diagonal entries.

Recall that the FIM $\vF_{\lparam}(\lparam_0)$ is block-diagonal with two blocks---the $\vdelta$ block and the $\vM$ block.
We will can show that the $\vM$ block of the FIM is also block-diagonal with three blocks, where each block represents the non-zero entries in $\vM_{\text{up}}$, 
$\vM_{\text{diag}}$, and $\vM_{\text{asym}}$, respectively.

Now, we will show that any cross term of the FIM between any two of these blocks is zero. We have three cases.
Let ${\cal I}_{\text{up}}$, ${\cal I}_{\text{diag}}$, ${\cal I}_{\text{asym}}$ be the index set of the non-zero entries of $\vM_{\text{up}}$, $\vM_{\text{diag}}$, and $\vM_{\text{asym}}$ respectively.

Case 1: For a cross term of the FIM between $\vM_{\text{up}}$ and $\vM_{\text{diag}}$, it is zero since
this is the case shown in the symmetric case (see Lemma \ref{eq:gauss_fim_sym_prec} in Appendix \ref{app:sym_m} for details).

Case 2: For a cross term of the FIM between $\vM_{\text{asym}}$ and $\vM_{\text{diag}}$, we can compute it as follows.

By Eq. \ref{eq:gauss_first_M} and the chain rule, we have the following expressions, where $j > i$.
\begin{align*}
&- \nabla_{{M_{\text{asym}}}_{ij}}  \log q(\vlat|\lparam) = - \mathrm{Tr} \big( \underbrace{ \big[ \nabla_{{M_{\text{asym}}}_{ij}} \vM \big] }_{\vI_{ij} }  \big[ \nabla_{M}  \log q(\vlat|\lparam) \big] \big) \\
&- \nabla_{{M_{\text{diag}}}_{ii}}  \log q(\vlat|\lparam) = - \mathrm{Tr} \big( \underbrace{ \big[ \nabla_{{M_{\text{diag}}}_{ii}} \vM \big] }_{\vI_{ii}}  \big[ \nabla_{M}  \log q(\vlat|\lparam) \big] \big)
\end{align*}

Therefore, we have
\begin{align*}
 &- \nabla_{{M_{\text{asym}}}}  \log q(\vlat|\lparam) = - \mathrm{Asym} \big( \nabla_{M}  \log q(\vlat|\lparam)  \big) \\
 &- \nabla_{{M_{\text{diag}}}}  \log q(\vlat|\lparam) 
 = - \mathrm{Diag} \big( \nabla_{M}  \log q(\vlat|\lparam)  \big)
\end{align*} where we define the $\mathrm{Diag}(\cdot)$ function that returns a diagonal matrix with the same structure as $\vM_{\text{diag}}$ and 
the $\mathrm{asym}(\cdot)$ function that returns a (upper) triangular matrix with the same structure as $\vM_{\text{asym}}$.

Notice that we only consider non-zero entries in $\vM_{\text{asym}}$, which implies that $j>i$ and $(i,j) \in {\cal I}_{\text{asym}}$ in the following expression. Therefore, any cross term can be expressed as below.
\begin{align*}
 &- \Unmyexpect{q(\lat|\lparam)}\sqr{  \nabla_{{M_{\text{asym}}}_{ij}} \nabla_{M_\text{diag}} \log q(\vlat|\lparam)  }\Big|_{\lparam=\mathbf{0}} 
 =- \Unmyexpect{q(\lat|\lparam)}\sqr{  \nabla_{{M_{\text{asym}}}_{ij}}  \mathrm{Diag} \big( \nabla_{M}  \log q(\vlat|\lparam)  \big) }\Big|_{\lparam=\mathbf{0}}  \\
 =&- \Unmyexpect{q(\lat|\lparam)}\sqr{ \sum_{k,l} \big[ \nabla_{{M_{\text{asym}}}_{ij}} M_{kl} \big] \nabla_{M_{kl}}  \mathrm{Diag} \big( \nabla_{M}  \log q(\vlat|\lparam)  \big) }\Big|_{\lparam=\mathbf{0}}  \\
 =&- \Unmyexpect{q(\lat|\lparam)}\sqr{ \underbrace{ \big[ \nabla_{{M_{\text{asym}}}_{ij}} M_{ij} \big]}_{=1} \nabla_{M_{ij}}  \mathrm{Diag} \big( \nabla_{M}  \log q(\vlat|\lparam)  \big)
}\Big|_{\lparam=\mathbf{0}}   \\
 =&- \Unmyexpect{q(\lat|\lparam)}\sqr{ \nabla_{M_{ij}}  \mathrm{Diag} \big( \nabla_{M}  \log q(\vlat|\lparam)  \big)
}\Big|_{\lparam=\mathbf{0}}  \\
 =&-  \mathrm{Diag} \big(\Unmyexpect{q(\lat|\lparam)}\sqr{ \nabla_{M_{ij}}  \nabla_{M}  \log q(\vlat|\lparam)  } \big)\Big|_{\lparam=\mathbf{0}}   \\
 =&  \mathrm{Diag} \big( \underbrace{ \nabla_{M_{ij}}   (\vM+\vM^T)}_{\vI_{ij} + \vI_{ji}}  \big)  =\mathbf{0}
\end{align*}  where
we  obtain the last step since $j>i$ and $\mathrm{Diag}(\vI_{ij})=\mathbf{0}$ since $(i,j)\in {\cal I}_{\text{asym}}$ and $(i,j) \not \in {\cal I}_{\text{diag}}$.

Case 3: Now, we show that any cross term of the FIM between $\vM_{\text{asym}}$ and $\vM_{\text{up}}$ is zero.
Let's denote a $\mathrm{Up}(\cdot)$ function  that returns a upper-triangular part of an input matrix  with the same (non-zero) structure as $\vM_{\text{up}}$.
Similarly, we can define  a $\mathrm{Asym}(\cdot)$ function.

It is obvious see that the intersection between any two of these index sets are empty.

For any $i<j$, where $(i,j) \in {\cal I}_{\text{up}}$, we have 
$(i,j) \not\in {\cal I}_{\text{asym}}$ and $\mathrm{Asym} \big( \vI_{ij}\big)= \mathrm{Asym} \big( \vI_{ji}\big) = \mathbf{0}$.

In this case, let $(i,j) \in {\cal I}_{\text{up}}$. The cross term can be computed as follows.
\begin{align*}
 &- \Unmyexpect{q(\lat|\lparam)}\sqr{  \nabla_{{M_{\text{up}}}_{ij}} \nabla_{M_\text{asym}} \log q(\vlat|\lparam)  }\Big|_{\lparam=\mathbf{0}} 
 =- \Unmyexpect{q(\lat|\lparam)}\sqr{  \nabla_{{M_{\text{up}}}_{ij}}  \mathrm{Asym} \big( \nabla_{M}  \log q(\vlat|\lparam)  \big) }\Big|_{\lparam=\mathbf{0}}  \\
 =&- \Unmyexpect{q(\lat|\lparam)}\sqr{ \sum_{k,l} \big[ \nabla_{{M_{\text{up}}}_{ij}} M_{kl} \big] \nabla_{M_{kl}}  \mathrm{Asym} \big( \nabla_{M}  \log q(\vlat|\lparam)  \big) }\Big|_{\lparam=\mathbf{0}}  \\
 =&- \Unmyexpect{q(\lat|\lparam)}\sqr{ \underbrace{ \big[ \nabla_{{M_{\text{up}}}_{ij}} M_{ij} \big]}_{=1} \nabla_{M_{ij}}  \mathrm{Asym} \big( \nabla_{M}  \log q(\vlat|\lparam)  \big)
 + \underbrace{ \big[ \nabla_{{M_{\text{up}}}_{ij}} M_{ji} \big]}_{=1} \nabla_{M_{ji}}  \mathrm{Asym} \big( \nabla_{M}  \log q(\vlat|\lparam)  \big)
}\Big|_{\lparam=\mathbf{0}}   \\
 =&- \Unmyexpect{q(\lat|\lparam)}\sqr{ \nabla_{M_{ij}}  \mathrm{Asym} \big( \nabla_{M}  \log q(\vlat|\lparam)  \big)
 +  \nabla_{M_{ji}}  \mathrm{Asym} \big( \nabla_{M}  \log q(\vlat|\lparam)  \big)
}\Big|_{\lparam=\mathbf{0}}  \\
 =&-  \mathrm{Asym} \big(\Unmyexpect{q(\lat|\lparam)}\sqr{ \nabla_{M_{ij}}  \nabla_{M}  \log q(\vlat|\lparam)   +  \nabla_{M_{ji}}   \nabla_{M}  \log q(\vlat|\lparam)   } \big)\Big|_{\lparam=\mathbf{0}}   \\
 =&  \mathrm{Asym} \big( \underbrace{ \nabla_{M_{ij}}   (\vM+\vM^T)}_{\vI_{ij} + \vI_{ji}} + \underbrace{ \nabla_{M_{ji}}   (\vM+\vM^T) }_{\vI_{ij} + \vI_{ji}}   \big)  =\mathbf{0} 
\end{align*}  where
we use $ \vM =  \vM_{\text{diag}} + \vM_{\text{up}} + \vM_{\text{up}}^T + \vM_{\text{asym}}$ to move from step 2 to step 3,
and  obtain the last step since $\mathrm{Asym} \big( \vI_{ij}\big)= \mathrm{Asym} \big( \vI_{ji}\big) = \mathbf{0}$.

Now, we compute the FIM w.r.t. $\vM_{\text{diag}}$,  $\vM_{\text{asym}}$ and $\vM_{\text{up}}$ separately.

Like Eq \eqref{eq:gauss_prec_low_M} in  Appendix \ref{app:sym_m} , the FIM w.r.t. the upper-triangular block is
\begin{align*}
 - \Unmyexpect{q(\lat|\lparam)}\sqr{  \nabla_{{M_{\text{up}}}_{ij}} \nabla_{M_\text{up}} \log q(\vlat|\lparam)  }\Big|_{\lparam=\mathbf{0}} 
 = 4 \vI_{ij}
\end{align*}.

Like Eq \eqref{eq:gauss_prec_diag_Mv2} in  Appendix \ref{app:sym_m} , the FIM w.r.t. the diagonal block is
\begin{align*}
 - \Unmyexpect{q(\lat|\lparam)}\sqr{  \nabla_{{M_{\text{diag}}}_{ij}} \nabla_{M_\text{diag}} \log q(\vlat|\lparam)  }\Big|_{\lparam=\mathbf{0}} 
 = 2 \vI_{ij}
\end{align*}.

By the chain rule, 
  the FIM w.r.t. $\vM_{\text{asym}}$ can be computed as follows, where $(i,j)\in {\cal I}_{\text{asym}}$.
\begin{align*}
 &- \Unmyexpect{q(\lat|\lparam)}\sqr{  \nabla_{{M_{\text{asym}}}_{ij}} \nabla_{M_\text{asym}} \log q(\vlat|\lparam)  }\Big|_{\lparam=\mathbf{0}} \\
=&- \Unmyexpect{q(\lat|\lparam)}\sqr{  \nabla_{{M_{\text{asym}}}_{ij}}  \mathrm{Asym} \big( \nabla_{M}  \log q(\vlat|\lparam) \big) }\Big|_{\lparam=\mathbf{0}}  \\
=&- \Unmyexpect{q(\lat|\lparam)}\sqr{ \sum_{k,l} \big[ \nabla_{{M_{\text{asym}}}_{ij}} M_{kl} \big] \nabla_{M_{kl}} \mathrm{Asym} \big( \nabla_{M}  \log q(\vlat|\lparam) \big) }\Big|_{\lparam=\mathbf{0}} \\
=&- \Unmyexpect{q(\lat|\lparam)}\sqr{   \underbrace{\big[ \nabla_{{M_{\text{asym}}}_{ij}} M_{ij} \big]}_{=1} \nabla_{M_{ij}} \mathrm{Asym} \big( \nabla_{M}  \log q(\vlat|\lparam) \big) }\Big|_{\lparam=\mathbf{0}} \\
=&- \mathrm{Asym} \big(\Unmyexpect{q(\lat|\lparam)}\sqr{  \nabla_{M_{ij}}  \nabla_{M}  \log q(\vlat|\lparam)  } \big) \Big|_{\lparam=\mathbf{0}} \\
=& \mathrm{Asym} \big( \underbrace{ \nabla_{M_{ij}}\big[ \vM + \vM^T \big]}_{=\vI_{ij}+\vI_{ji}}   \big)  \,\,\, (\text{By Lemma \ref{lemma:fim_M_gauss_prec}}) \\
=&  \vI_{ij}
\end{align*} where we obtain the last step since
that $\mathrm{Asym}(\vI_{ji})=\mathbf{0}$ when $i<j$ since 
$(i,j)\in {\cal I}_{\text{asym}}$ and $(j,i) \not \in  {\cal I}_{\text{asym}}$.
Therefore, the FIM w.r.t. the asymmetric block is
\begin{align*}
 - \Unmyexpect{q(\lat|\lparam)}\sqr{  \nabla_{{M_{\text{asym}}}_{ij}} \nabla_{M_\text{asym}} \log q(\vlat|\lparam)  }\Big|_{\lparam=\mathbf{0}} 
 =  \vI_{ij}
\end{align*}.

Like the symmetric case (see Eq \eqref{eq:gauss_sym_prec_ng} Appendix \ref{app:sym_m})
when we  evaluate gradients at $\lparam_0=\{\vdelta_0,\vM_0\} =\mathbf{0}$, we have
\begin{align*}
 \nabla_{\delta_i} {\cal L} \big|_{\lparam=0} &= \big[ \nabla_{\delta_i} \vdelta \big]^T \vB_t^{-1} \nabla_{\mu} {\cal L }\\
 \nabla_{M_{ij}} {\cal L}\big|_{\lparam=0}  
&= - \mathrm{Tr}\big(  \big[\nabla_{M_{ij}} \big(\vM + \vM^T\big) \big]   \vB_t^{-1} \big[\nabla_{\Sigma} {\cal L }\big] \vB_t^{-T}   \big)
\end{align*}

Let's denote $\vG_M= -2  \vB_t^{-1} \big[\nabla_{\Sigma} {\cal L }\big] \vB_t^{-T} $.
Therefore, we can show that Euclidean gradients are
\begin{align*}
 \vG_{M_{\text{diag}}} = \mathrm{Diag} (\vG_M); \,\,\,\,\,\,
 \vG_{M_{\text{up}}} = \mathrm{Up} \big ( \vG_M + \vG_M^T \big) = 2 \mathrm{Up} (\vG_M); \,\,\,\,\,\,
 \vG_{M_{\text{asym}}} = \mathrm{Asym} (\vG_M) ;\,\,\,\,\,\,
\vg_\delta =  \vB_t^{-1} \nabla_{\mu} {\cal L }
\end{align*}
The natural gradients w.r.t. $\vM_{\text{diag}}$,  $\vM_{\text{up}}$, and $\vM_{\text{asym}}$ are
$\half \mathrm{Diag} (\vG)$, $\half \mathrm{Up} (\vG)$, and $ \mathrm{Asym} (\vG)$ respectively.
The natural gradient w.r.t. $\vdelta$ is $ \vB_t^{-1} \nabla_{\mu} {\cal L }  $.

Natural gradients can be expressed as in the following compact form:
\begin{align*}
 \vngrad_{\delta_0}^{(t)} &   = \vB_t^{-1} \nabla_{\mu} {\cal L } \\
 \vngrad_{M_0}^{(t)} & = \vC_{\text{up}} \odot \kappa_{\text{up}}\big( -2  \vB_t^{-1} \big[\nabla_{\Sigma} {\cal L } \big] \vB_t^{-T} \big)
\end{align*} where
\begin{align*}
 \vC_{\text{up}} = 
 \begin{bmatrix}
\half \vJ_A & \vJ_B   \\
 \mathbf{0} & \half \vI_D
      \end{bmatrix}  \in {\cal{M}_{\text{up}}}(k)
\end{align*}

Therefore, our update is
\begin{align}
\vmu_{t+1} & \leftarrow \vmu_{t} - \beta \vB_t^{-T} \vB_t^{-1}  \vg_{\mu_t} \nonumber\\
\vB_{t+1} & \leftarrow   \vB_t \vh \rnd{ \beta \vC_{\text{up}} \odot \kappa_{\text{up}}\big( 2 \vB_t^{-1} \vg_{\Sigma_t} \vB_t^{-T} \big) }
\label{eq:gauss_prec_triaup_ngd_aux}
\end{align}

\subsubsection{Induced Structures}
\label{app:matrix_structure}

When $\vB \in {\cal B}_{\text{up}}(k)$,   we can show that the covariance matrix $\vSigma = (\vB\vB^T)^{-1}$ has a low rank structure.
This structure is useful for posterior approximation

Notice that the precision matrix $\vS=\vB\vB^T$ is a block arrowhead matrix as shown below.
\begin{align*}
\vS &= \vB\vB^T \\
&=  \begin{bmatrix}
\vB_A  \vB_A^T + \vB_B \vB_B^T & \vB_B \vB_D \\
\\
\vB_D \vB_B^T & \vB_D^2
      \end{bmatrix}
\end{align*}

Now, we can show that the covariance matrix $\vSigma=\vP^{-1}$ admits a rank-$k$ structure.
\begin{align*}
\vSigma &=  
\begin{bmatrix}
  \vB_A^{-T} \vB_A^{-1} & - \vB_A^{-T} \vB_A^{-1} \vB_B \vB_D^{-1} \\
  \\
-\vB_D^{-1} \vB_B^T  \vB_A^{-T} \vB_A^{-1} & \vB_D^{-1} \vB_B^T  \vB_A^{-T} \vB_A^{-1} \vB_B \vB_D^{-1} + \vB_D^{-2}
      \end{bmatrix} \\
      &= \vU_k \vU_k^T +  
      \begin{bmatrix}
      \mathbf{0} &  \\
       &  \vB_D^{-2}
      \end{bmatrix}
\end{align*}
where $\vU_k$ is a $p$-by-$k$ matrix as shown below and $\vU_k$ is a rank-$k$ matrix since $\vB_A^{-T}$ is full $k$ rank (invertible).
\begin{align*}
\vU_k =  \begin{bmatrix}
-\vB_A^{-T} \\
\\
\vB_D^{-1} \vB_B^T \vB_A^{-T}
      \end{bmatrix} 
\end{align*}

\subsubsection{Singular FIMs}
\label{app:diag_zero}

In Appendix \ref{app:matrix_structure}, we know that when $\vB \in {\cal B}_{\text{up}}(k)$  takes the block upper triangular structure, the covariance is a low-rank matrix.
\begin{align*}
\vSigma   &= ( \vB \vB^T )^{-1} \\
      &= \vU_k \vU_k^T +  
      \begin{bmatrix}
     {\color{red} \mathbf{0} } & \mathbf{0} \\
      \mathbf{0} &  \vB_D^{-2}
      \end{bmatrix}
\end{align*}
As shown in Appendix \ref{app:ng_block_triangular_gauss}, the FIM $\vF_{\lparam}(\lparam_0)$ is non-singular.
Equivalently, we can use auxiliary parameterization $\vA \in {\cal B}_{\text{low}}(k)$ for the covariance $\vSigma=\vA\vA^T$ if we choose to use the covariance  as a global parameterization $\gparam=\{\vmu,\vSigma\}$.

In fact, the zero block (the $k$-by-$k$ matrix) highlighted in red ensures the FIM $\vF_{\lparam}(\lparam_0)$ is non-singular when $k>0$.
The group structure contains such a zero block so that the FIM is non-singular.
It is tempting to use a non-zero block to replace the zero block in the above expression to get a more flexible structure.
Unfortunately, the FIM $\vF_{\lparam}(\lparam_0)$ may become singular by doing so.

Th singularity issue also appears even when we use a common (global) parameterization $\gparam$ for a low-rank (e.g., rank-one) Gaussian \citep{tran2020bayesian,mishkin2018slang,sun2013linear} such as $\vSigma = \vv \vv^T + \mathrm{Diag}(\vd^2) $, where $\vv, \vd \in \real^{p}$ are both learnable parameters.
For illustration, let's consider a rank-one structure in the covariance matrix $\vSigma \in {\cal S}_{++}^{p \times p}$ of Gaussians, which is a case considered in \citet{tran2020bayesian}, where the global parameterization is chosen to be $\gparam=\{\vmu, \vv, \vd \}$ so that the covariance $\vSigma=\vv \vv^T+ \mathrm{Diag}(\vd^2  )$ has a rank-one structure.
We will give two examples to show that
the FIM $\vF_{\gparam}$ is singular when $\gparam=\{\vmu, \vv, \vd \}$, where $\vmu, \vv, \vd \in \real^{p} $ are all learnable
vectors.
To avoid the singularity issue, \citet{tran2020bayesian} have to use a block approximation of the FIM $\vF_{\gparam}$.
\citet{mishkin2018slang} also consider a rank-one matrix in the precision matrix $\vS$ of Gaussians, where an additional approximation is used to fix this singularity issue.
\citet{sun2013linear} reduce the degree of freedom in a p-dimensional low-rank Gaussians such as $\vSigma = \vv \vv^T + d^2 \vI$ to avoid this issue\footnote{When $p=1$, the FIM  of the low-rank Gaussian considered by \citet{sun2013linear} is still singular.}, where $d$ is chosen to be a learnable \emph{scalar} instead of a vector. However, the covariance used in 
\citet{sun2013linear} is less flexible than the covariance induced by our group structures since the degree of freedom for
the covariance used in \citet{sun2013linear} is $p+1$ while the degree of freedom for the covariance induced by the block triangular group with $k=1$ is $2p-1$.

Now, we give two examples to illustrate the singularity issue in a rank-one $p$-dimensional Gaussian with \emph{constant mean} and the covariance structure
$\vSigma = \vv \vv^T + \mathrm{Diag}(\vd^2) $, where
$\gparam=\{\vv, \vd \}$ and $\vv, \vd \in \real^{p} $ are all learnable vectors.

Example (1):
First of all, in 2-dimensional ($p=2$) Gaussian cases with constant mean, we know that the degree of freedom of the full covariance $\vSigma$ is 3 since $\vSigma \in {\cal S}_{++}^{2 \times 2}$ is symmetric.
It is easy to see when $\gparam=\{\vv, \vd \}$, the  degree of freedom in the rank-one Gaussian case with constant mean is 4, which implies the FIM is singular since the maximum degree of freedom is 3 obtained in the full Gaussian case.

Example (2):
This issue also appears in higher dimensional cases. We consider an example in a 3-dimensional ($p=3$) rank-one Gaussian with \emph{constant zero} mean.
Let's consider the following case where $\vv = \begin{bmatrix}
1 \\ 0 \\ 0\end{bmatrix}$, and $\vd=\begin{bmatrix}1 \\ 1 \\ 1\end{bmatrix}$ so that $\vSigma:=\vv \vv^T + \mathrm{Diag}(\vd^2)$.
Let $\valpha=\begin{bmatrix}
            \vd \\ \vv 
            \end{bmatrix}  \in \real^{6}$.
The FIM in this case is denoted by $\vF_{\gparam}(\valpha)$, where 
the global parameter is  $\gparam=\{\vv, \vd \}$.
In this case,
$\vF_{\gparam}(\valpha)$ computed by Auto-Diff (see \eqref{eq:fim_singular_st_gauss}) is given below.
\begin{align*}
 \vF_{\gparam}(\valpha) =
 \begin{bmatrix}
0.5 & 0 & 0 & 0.5 & 0 & 0 \\  
0 & 2 & 0 & 0 & 0 & 0 \\  
0 & 0 & 2 & 0 & 0 & 0 \\  
0.5 & 0 & 0 & 0.5 & 0 & 0 \\  
0 & 0 & 0 & 0 & 0.5 & 0 \\  
0 & 0 & 0 & 0 & 0 & 0.5   
 \end{bmatrix}
\end{align*} where $\valpha=\begin{bmatrix}
             1  & 1 & 1  & 1 & 0 & 0
            \end{bmatrix}^T$ when $\vd=\begin{bmatrix}
             1  & 1 & 1 
            \end{bmatrix}^T$ and $\vv=\begin{bmatrix}
             1  & 0 & 0
            \end{bmatrix}^T$.
            
It is easy to see that
$ \vF_{\gparam}(\valpha)$ is singular. Therefore, the FIM $\vF_{\gparam}$ under the global parameterization $\gparam=\{\vv,\vd\}$ for the rank-one Gaussian can be singular.

Even when we allow to learn the mean $\vmu$ in the rank-one Gaussian cases, the FIM $\vF_{\gparam}$ is still singular
where $\gparam=\{\vmu,\underbrace{\vv,\vd}_{\valpha}\}$ since $\vF_{\gparam}=\begin{bmatrix}
                                         \vF_{\gparam}(\vmu) &  \mathbf{0} \\               
                                          \mathbf{0} & \vF_{\gparam}(\valpha)
                                                        \end{bmatrix}$ is block-diagonal and $\vF_{\gparam}(\valpha)$ is singular at $\vmu=\mathbf{0}$.

\subsubsection{Complexity analysis and Efficient Computation}
\label{app:complexiity}
When $\vB \in {\cal B}_{\text{up}}(k)$ is a $p$-by-$p$ invertible matrix, it can be written as
\begin{align*}
\vB = \begin{bmatrix}
\vB_A & \vB_B \\
 \mathbf{0} & \vB_D
      \end{bmatrix}
\end{align*} where $\vB_A$ is a $k$-by-$k$ invertible matrix and $\vB_D$ is a diagonal and invertible matrix.

To generate samples, we first  compute the following matrix.
\begin{align*}
 \vB^{-T} = \begin{bmatrix}
\vB_A^{-T} & \mathbf{0} \\
 -\vB_D^{-T} \vB_B^T \vB_A^{-T}  & \vB_D^{-T}
      \end{bmatrix}
\end{align*}

Given $\vB^{-T}$ is known, for variational inference, we can easily generate a sample in $O(k^2 p)$ as $\vlat = \vmu + \vB^{-T} \vepsilon$, where $\vepsilon \sim \gauss(\mathbf{0},\vI)$.
Similarly, $\vS^{-1} \vg_{\mu} = \vB^{-T} \vB^{-1} \vg_\mu$ can be computed in $O(k^2p)$.

Since $\vM \in {\cal M}_{\text{up}}(k)$, it can be written as
\begin{align*}
\vM = \begin{bmatrix}
\vM_A &     \vM_B  \\
 \mathbf{0}  & \vM_D
      \end{bmatrix}  
\end{align*} where $\vM_A$ is a $k$-by-$k$ symmetric matrix and $\vM_D$ is a diagonal matrix.

We can compute $\vh(\vM)$ in $O(k^2 p)$ when 
$\vM \in {\cal M}_{\text{up}}(k)$
\begin{align*}
\vh(\vM) :=
\vI + \vM + \half \vM^2 =\begin{bmatrix}
\vI_A + \vM_A + \half \vM_A^2 &  \vM_B + \half \big( \vM_A \vM_B + \vM_B \vM_D \big) \\
 \mathbf{0} & \vI_D + \vM_D + \half \vM_D^2
                         \end{bmatrix}
\end{align*}

Similarly, we can compute the matrix product  $\vB \vh(\vM)$ in $O(k^2 p)$.

Now, we discuss how to compute
$\kappa_{\text{up}}\big( 2 \vB_t^{-1} \vg_{\Sigma} \vB_t^{-T} \big)$

We assume $\vg_\Sigma$ can be expressed as the following form.
\begin{align*}
\vg_\Sigma = \half \begin{bmatrix}
        \vH_{11} & \vH_{12} \\
        \vH_{21} & \vH_{22}
        \end{bmatrix}
\end{align*} where $\vH_{21}=\vH_{12}^T$.

\begin{align*}
2 \vB^{-1} \vg_\Sigma \vB^{-T} 
& =  \begin{bmatrix}
\vE - \vF^T \vB_B^T \vB_A^{-T} - \vB_A^{-1} \vB_B \vF  + \vB_A^{-1} \vB_B \vB_D^{-1} \vH_{22} \vB_D^{-T}  \vB_B^T \vB_A^{-T} & & \vF^T - \vB_A^{-1}\vB_B \vB_D^{-1} \vH_{22}\vB_D^{-T} \\
\vF - \vB_D^{-1} \vH_{22} \vB_D^{-T} \vB_B^{T} \vB_A^{-T} && \vB_D^{-1} \vH_{22} \vB_D^{-T}
                           \end{bmatrix}
\end{align*}  
where $\vE= \vB_A^{-1} \vH_{11} \vB_A^{-T}$ and $\vF = \vB_D^{-1} \vH_{21} \vB_A^{-T}  $

Therefore, we have
\begin{align*}
 \kappa_{\text{up}}\big( 2 \vB_t^{-1} \vg_{\Sigma} \vB_t^{-T} \big) = 
 \begin{bmatrix}
\vE - \vF^T \vB_B^T \vB_A^{-T} - \vB_A^{-1} \vB_B \vF  + \vB_A^{-1} \vB_B \vB_D^{-1} \vH_{22} \vB_D^{-T}  \vB_B^T \vB_A^{-T} & & \vF^T - \vB_A^{-1}\vB_B \vB_D^{-1} \vH_{22}\vB_D^{-T} \\
\mathbf{0} && \mathrm{Diag} \big( \vB_D^{-1} \vH_{22} \vB_D^{-T} \big)
                           \end{bmatrix}
\end{align*}

Notice that by Stein's identity , we have
\begin{align*}
 \vg_\Sigma  
 &= \half \Unmyexpect{q(\lat)} \big[ \nabla_{\lat}^2 f(\vlat) \big]
\end{align*} where $\vlat = \vmu + \vB^{-T} \vepsilon$ and $\vepsilon \sim \gauss(\mathbf{0},\vI)$.

For a $k$-rank approximation, if we can compute $O(k)$  Hessian-vector products,
let's consider  the following expression.
\begin{align*}
\begin{bmatrix}
\vv_1 \\
\vv_2
\end{bmatrix} &= 
\begin{bmatrix}
\vH_{11} \vB_A^{-T} - \vH_{12}  \vB_D^{-T}  \vB_B^T \vB_A^{-T} \\
\vH_{21}  \vB_A^{-T} -\vH_{22} \vB_D^{-T}  \vB_B^T \vB_A^{-T}
\end{bmatrix} \\ 
&= \Unmyexpect{q(\lat)} \big[ \begin{bmatrix}
\nabla_{\lat_1}^2 f(\vlat_1,\vlat_2) &\nabla_{\lat_1} \nabla_{\lat_2}f(\vlat_1,\vlat_2) \\
\nabla_{\lat_2} \nabla_{\lat_1}f(\vlat_1,\vlat_2) & \nabla_{\lat_2}^2 f(\vlat_1,\vlat_2)
\end{bmatrix}  \begin{bmatrix}
\vB_A^{-T} \\
 - \vB_D^{-T} \vB_B^T \vB_A^{-T}
\end{bmatrix} \big]
\end{align*}

Therefore, we have
\begin{align}
 \kappa_{\text{up}}\big( 2 \vB_t^{-1} \vg_{\Sigma} \vB_t^{-T} \big) = 
 \begin{bmatrix}
 \big( \vB_A^{-1}\vv_1  - \vB_A^{-1} \vB_B \vB_D^{-1} \vv_2 \big) & & \big( \vB_D^{-1} \vv_2 \big)^T \\
\mathbf{0} && \vB_D^{-1} \mathrm{Diag} \big( \vH_{22} \big) \vB_D^{-T} 
                           \end{bmatrix}
                           \label{eq:upper_hvp}
\end{align}
We can compute this in $O(k^2 p)$ since $\vB_D$ is diagonal, where we assume we can efficiently compute
$O(k)$  Hessian-vector products and compute/approximate  diagonal entries of the Hessian $\mathrm{Diag} \big( \vH_{22} \big)$.

\subsubsection{Block Lower-triangular Group}
\label{app:block_lower_tri}

Similarly, we can define a block lower-triangular group ${\cal{B}_{\text{low}}}(k)$ and a local parameter space
${\cal{M}_{\text{low}}}(k)$.

\begin{align*}
{\cal{B}_{\text{low}}}(k)  = \Big\{ 
\begin{bmatrix}
\vB_A &  \mathbf{0}  \\
 \vB_C & \vB_D
      \end{bmatrix} \Big| & \vB_A \in \mathrm{GL}^{k \times k},\,\,
 \vB_D  \in{\cal D}^{d_0 \times d_0}_{++}  \Big\}; \,\,\,\,\,
{\cal{M}_{\text{low}}}(k)  = \Big\{ 
\begin{bmatrix}
\vM_A &  \mathbf{0}\\
  \vM_C  & \vM_D
      \end{bmatrix} \Big| &  \vM_A \in{\cal S}^{k \times k}, \,\,
 \vM_D  \in{\cal D}^{d_0 \times d_0} \Big\}
\end{align*}

we consider the following parametrization for the Gaussian $\gauss(\vlat|\vmu,\vS^{-1})$, where the precision $\vS$ belongs to a sub-manifold of $\mathcal{S}_{++}^{p\times p}$, auxiliary parameter $\vB$ belongs to ${\cal{B}_{\text{low}}}(k)$, and local parameter $\vM$ belongs to ${\cal{M}_{\text{low}}}(k)$,
\begin{equation*}
    \begin{split}
       \gparam &:= \crl{\vmu \in\real^p, \,\,\, \vS=\vB \vB^T \in \mathcal{S}_{++}^{p\times p} \,\,\,|\,\,\, \vB \in {\cal{B}_{\text{low}}}(k) }, \,\,\,\\
        \aparam &:= \crl{ \vmu \in\real^p, \,\,\, \vB \in {\cal{B}_{\text{low}}}(k) }, \\
        \lparam &:= \crl{ \vdelta\in\real^p, \,\,\, \vM \in {\cal{M}_{\text{low}}}(k)   }.
    \end{split}
\end{equation*}
The map $\vpsi \circ \vphi_{\aparam_t}(\lparam)$ at $\aparam_t := \{\vmu_t, \vB_t\}$ is chosen as below, which is the same as     \eqref{eq:gauss_xnes_prec}
\begin{equation*}
    \begin{split}
        \crl{ \begin{array}{c} \vmu \\ \vS \end{array} } &= \vpsi(\aparam) :=  \crl{ \begin{array}{c} \vmu \\ \vB\vB^\top \end{array} } \\
        \crl{ \begin{array}{c} \vmu \\ \vB \end{array} } &= \vphi_{\aparam_t}(\lparam) :=  \crl{ \begin{array}{c} \vmu_t + \vB_t^{-T} \vdelta \\ \vB_t \vh (\vM) \end{array} }.
    \end{split}
\end{equation*}

We can show Assumption 1 and 2 are satisfied similar to Appendix  \ref{app:proof_up_group_lemma}.

Our update over the auxiliary parameters is
\begin{align}
\vmu_{t+1} & \leftarrow \vmu_{t} - \beta \vB_t^{-T} \vB_t^{-1} \vg_{\mu_t} \nonumber \\
\vB_{t+1} & \leftarrow   \vB_t \vh \rnd{ \beta \vC_{\text{low}} \odot \kappa_{\text{low}}\big( 2 \vB_t^{-1} \vg_{\Sigma_t} \vB_t^{-T} \big) }
\label{eq:low_sym_gauss_prec_ng}
\end{align} where 
\begin{align*}
 \vC_{\text{low}} = 
 \begin{bmatrix}
\half \vJ_A & \mathbf{0}   \\
 \vJ_C & \half \vI_D
      \end{bmatrix}  \in {\cal{M}_{\text{low}}}(k)
\end{align*} where $\vJ $ denotes a matrix of ones and factor $\half$ appears in the symmetric part of ${\vC}_{\text{low}}$.
$\odot$ denotes the element-wise product, $\kappa_{\text{low}}(\vX)$ extracts non-zero entries of   ${\cal{M}_{\text{low}}}(k)$ from $\vX$ so that $\kappa_{\text{low}}(\vX) \in {\cal{M}_{\text{low}}}(k)$. We can compute this update in $O(k^2 p)$.

When $\vB \in {\cal{B}_{\text{low}}}(k)$,  we  show that the precision matrix $\vS=\vB\vB^T$ has a low rank structure.
This structure is useful for optimization.

The precision matrix $\vS$ admits a rank-$k$ structure  as shown below.
\begin{align*}
\vS = \vB\vB^T 
=  \begin{bmatrix}
\vB_A \vB_A^T & \vB_A \vB_C^T \\
\\
\vB_C \vB_A^T & \vB_C \vB_C^T + \vB_D^2
      \end{bmatrix} 
       = \vV_k \vV_k^T +  
      \begin{bmatrix}
      \mathbf{0} &  \\
       &  \vB_D
      \end{bmatrix}; \,\,\,\,\,
      \vV_k =  \begin{bmatrix}
\vB_A \\
\\
\vB_C
      \end{bmatrix} 
\end{align*} 
where $\vV_k$ is a $d$-by-$k$ matrix and $\vV_k$ is a rank-$k$ matrix since $\vB_A$ is full $k$ rank.

Similarly, we can show that the covariance matrix $\vSigma=\vS^{-1}$ is a block arrowhead matrix.
\begin{align*}
 \vSigma &= 
 \begin{bmatrix}
 \vB_A^{-T} & -\vB_A^{-T} \vB_C^T \vB_D^{-1}\\
\\
\mathbf{0} & \vB_D^{-1}
      \end{bmatrix}
       \begin{bmatrix}
 \vB_A^{-1} & \mathbf{0}\\
\\
 - \vB_D^{-1} \vB_C \vB_A^{-1}  
 & \vB_D^{-1}
      \end{bmatrix}\\
      &= 
       \begin{bmatrix}
 \vB_A^{-T} \vB_A^{-1} + \vB_A^{-T} \vB_C^T \vB_D^{-2} \vB_C \vB_A^{-1} & - \vB_A^{-T}\vB_C^T \vB_D^{-2} \\
\\
-\vB_D^{-2} \vB_C \vB_A^{-1} & \vB_D^{-2}
      \end{bmatrix}
\end{align*}

Now, we discuss how to compute
$\kappa_{\text{low}}\big( 2 \vB_t^{-1} \vg_{\Sigma} \vB_t^{-T} \big)$.

Similarly,
we assume $\vg_\Sigma$ can be expressed as the following form.
\begin{align*}
\vg_\Sigma = \half \begin{bmatrix}
        \vH_{11} & \vH_{12} \\
        \vH_{21} & \vH_{22}
        \end{bmatrix}
\end{align*} where $\vH_{21}=\vH_{12}^T$.

Therefore, we have
\begin{align*}
 \kappa_{\text{low}} \big( 2\vB^{-1} \vg_{\Sigma}  \vB^{-T} ) 
              = &
       \begin{bmatrix}
 \vF  &  \mathbf{0} \\
\\
  -\vB_D^{-1} \vB_C  \vF  + \vB_D^{-1} \vE_{2}& 
   \vB_D^{-1} \mathrm{Diag} \big[ \vB_C \vF \vB_C^T  +  \vH_{22} - \vB_C \vE_2^T - \vE_2 \vB_C^T \big] \vB_D^{-1} 
      \end{bmatrix}
\end{align*} where

\begin{align*}
       \begin{bmatrix}
\vE_{1} \\
\\
 \vE_{2}
      \end{bmatrix}
&      :=
       \begin{bmatrix}
\vH_{11} &  \vH_{21}^T \\
\\
 \vH_{21}& \vH_{22}
      \end{bmatrix}
 \begin{bmatrix}
\vB_A^{-T}   \\
\\
 \mathbf{0}
      \end{bmatrix} 
     =
       \begin{bmatrix}
\vH_{11} \vB_A^{-T}\\
\\
 \vH_{21}\vB_A^{-T}
      \end{bmatrix} \\
      \vF &:= \vB_A^{-1} \vE_1 = \vB_A^{-1} \vH_{11} \vB_A^{-T} 
\end{align*}

Note that we have the following identity.
\begin{align*}
 \mathrm{Diag}(\vA \vB) =  \mathrm{Diag}(\vB^T \vA^T) = \mathrm{Sum} ( \vA \odot \vB^T, \text{column})
\end{align*} where $\mathrm{Sum} ( \vX, \text{column})$ returns a column vector by summing $\vX$ over its columns.

Using this identity, 
we can further simplify the term as
\begin{align*}
 \kappa_{\text{low}} \big( 2\vB^{-1} \vg_{\Sigma}  \vB^{-T} ) 
& =      \begin{bmatrix}
 \vF  &  \mathbf{0} \\
\\
  -\vB_D^{-1} \vB_C  \vF  + \vB_D^{-1} \vE_{2}& 
   \vB_D^{-1} \big[\mathrm{Diag}( \vH_{22} ) +  \mathrm{Sum} ( \vB_C \odot (\vB_C\vF -  2\vE_2) , \text{column})   \big] \vB_D^{-1} 
      \end{bmatrix}
\end{align*}

\subsection{Alternative Structures Inspired by the Heisenberg Group}
\label{app:Heisenberg_group}

First of all, the Heisenberg group is defined as follows.
\begin{align*}
\vB = \begin{bmatrix}
1 & \va^T & c \\
\mathbf{0} & \vI & \vb\\
0 & \mathbf{0} & 1
      \end{bmatrix}
\end{align*} where $\va$ and $\vb$ are column vectors while $c$ is a scalar.

We construct the following set inspired by the Heisenberg group, where $1<k_1+k_2<p$ and $d_0=p-k_1-k_2$.

\begin{align*}
  {\cal B}_{\text{up}} (k_1,k_2) = \{  \begin{bmatrix}
\smash[b]{ \overbrace{ \begin{matrix} \vB_A \end{matrix} }^{\text{$k_1$-by-$k_1$}} } & \smash[b]{ \overbrace{\begin{matrix} \vB_{B_1} & \vB_{B_2} \end{matrix}}^{\vB_B} } \\
\smash[b]{\begin{matrix} \mathbf{0} \end{matrix} }& \smash[b]{\begin{matrix}  \vB_{D_1} & \vB_{D_2} \end{matrix} } \\
\smash[b]{\begin{matrix} \mathbf{0}  \end{matrix} } &
\smash[b]{   \begin{matrix} \mathbf{0} & \smash[b]{  \underbrace{ \begin{matrix} & \vB_{D_4} \end{matrix} }_{\text{$k_2$-by-$k_2$}}  } \end{matrix} }   
      \end{bmatrix} |    \vB_A \in \mathrm{GL}^{k_1 \times k_1}, \vB_{D_1}  \in{\cal D}_{++}^{d_0 \times d_0}, \vB_{D_4} \in \mathrm{GL}^{k_2 \times k_2} \}
\end{align*}

We can re-express the structure  as follows
\begin{align*}
 {\cal B}_{\text{up}} (k_1,k_2)
  = \Big\{  \begin{bmatrix}
\vB_A & \vB_B \\
\mathbf{0} & \vB_D
      \end{bmatrix} \Big| 
      \vB_D = \begin{bmatrix}
      \vB_{D_1} & \vB_{D_2}\\
      \mathbf{0} & \vB_{D_4}
     \end{bmatrix}
 \Big\}
\end{align*} where 
$
     \vB_A \in \mathrm{GL}^{k_1 \times k_1}
     $, 
$\vB_{D_1}  \in{\cal D}_{++}^{d_0 \times d_0}$,
$\vB_{D_4} \in \mathrm{GL}^{k_2 \times k_2}$.

We can show that $  {\cal B}_{\text{up}} (k_1,k_2)$ is a matrix group, which is more flexible than the block triangular group.

Similarly, we define a local parameter space ${\cal{M}_{\text{up}}}(k_1,k_2)$ as
\begin{align*}
{\cal{M}_{\text{up}}}(k_1,k_2)  = \Big\{ 
\begin{bmatrix}
\vM_A &  \vM_{B_1} & \vM_{B_2}  \\
 \mathbf{0} & \vM_{D_1} & \vM_{D_2} \\
 \mathbf{0} & \mathbf{0} & \vM_{D_4} \\
      \end{bmatrix} \Big| &  \vM_A \in{\cal S}^{k_1 \times k_1}, \,\,
 \vM_{D_1} \in{\cal D}^{d_0 \times d_0},\,\,
 \vM_{D_4} \in{\cal S}^{k_2 \times k_2}
 \Big\}
\end{align*}

Likewise, we consider the following parametrization for the Gaussian $\gauss(\vlat|\vmu,\vS^{-1})$, where the precision $\vS$ belongs to a sub-manifold of $\mathcal{S}_{++}^{p\times p}$, auxiliary parameter $\vB$ belongs to ${\cal{B}_{\text{up}}}(k_1,k_2)$, and local parameter $\vM$ belongs to ${\cal{M}_{\text{up}}}(k_1,k_2)$,
\begin{equation*}
    \begin{split}
       \gparam &:= \crl{\vmu \in\real^p, \,\,\, \vS=\vB \vB^T \in \mathcal{S}_{++}^{p\times p} \,\,\,|\,\,\, \vB \in {\cal{B}_{\text{up}}}(k_1,k_2)  }, \,\,\,\\
        \aparam &:= \crl{ \vmu \in\real^p, \,\,\, \vB \in {\cal{B}_{\text{up}}}(k_1,k_2) }, \\
        \lparam &:= \crl{ \vdelta\in\real^p, \,\,\, \vM \in {\cal{M}_{\text{up}}}(k_1,k_2)   }.
    \end{split}
\end{equation*}
The map $\vpsi \circ \vphi_{\aparam_t}(\lparam)$ at $\aparam_t := \{\vmu_t, \vB_t\}$ is chosen as below, which is the same as     \eqref{eq:gauss_xnes_prec}
\begin{equation*}
    \begin{split}
        \crl{ \begin{array}{c} \vmu \\ \vS \end{array} } &= \vpsi(\aparam) :=  \crl{ \begin{array}{c} \vmu \\ \vB\vB^\top \end{array} } \\
        \crl{ \begin{array}{c} \vmu \\ \vB \end{array} } &= \vphi_{\aparam_t}(\lparam) :=  \crl{ \begin{array}{c} \vmu_t + \vB_t^{-T} \vdelta \\ \vB_t \vh (\vM) \end{array} }.
    \end{split}
\end{equation*}

We can show Assumption 1 and 2 are satisfied similar to Appendix \ref{app:proof_up_group_lemma}.
Our update over the auxiliary parameters is
\begin{align}
\vmu_{t+1} & \leftarrow \vmu_{t} - \beta \vB_t^{-T} \vB_t^{-1}  \vg_{\mu_t} \nonumber \\
\vB_{t+1} & \leftarrow   \vB_t \vh \rnd{ \beta \vC_{\text{up}} \odot \kappa_{\text{up}}\big( 2 \vB_t^{-1} \vg_{\Sigma_t} \vB_t^{-T} \big) }
\label{eq:gauss_prec_heisen_ngd_aux}
\end{align} where
$\odot$ denotes the element-wise product, $\kappa_{\text{up}}(\vX)$ extracts non-zero entries of   ${\cal{M}_{\text{up}}}(k_1,k_2)$ from $\vX$ so that $\kappa_{\text{up}}(\vX) \in {\cal{M}_{\text{up}}}(k_1,k_2)$,  $ \vC_{\text{up}}$ is a constant matrix defined below,
$\vJ $ denotes a matrix of ones and factor $\half$ appears in the symmetric part of ${\vC}_{\text{up}}$.
\begin{align*}
 \vC_{\text{up}} = 
 \begin{bmatrix}
\half \vJ_A &  \vJ_{B_1} & \vJ_{B_2}   \\
 \mathbf{0} & \half \vI_{D_1} & \vJ_{D_2} \\
  \mathbf{0} & \mathbf{0} & \half \vJ_{D_4}  \\
      \end{bmatrix}  \in {\cal{M}_{\text{up}}}(k_1,k_2)
\end{align*}

We can also efficiently implement  this update by using Hessian-vector products.

Similarly, we can define a lower version of this group  denoted by ${\cal{B}_{\text{low}}}(k_1,k_2) $ and derive our update for this structure.
\begin{align*}
 {\cal B}_{\text{low}} (k_1,k_2)
  = \Big\{  \begin{bmatrix}
\vB_A & \mathbf{0} \\
\vB_C  & \vB_D
      \end{bmatrix} \Big| 
      \vB_D = \begin{bmatrix}
      \vB_{D_1} & \mathbf{0}\\
      \vB_{D_3} & \vB_{D_4}
     \end{bmatrix}
 \Big\}
\end{align*} where 
$
 \vB_A \in \mathrm{GL}^{k_1 \times k_1}
$, 
$\vB_{D_1}  \in{\cal D}_{++}^{d_0 \times d_0}$,
$\vB_{D_4} \in \mathrm{GL}^{k_2 \times k_2}$.

\end{appendices}

\end{document}